%% file: my_thesis.tex
\documentclass[a4paper,11pt]{book}

\usepackage[T1]{fontenc}
\usepackage[utf8]{inputenc}
\usepackage[round]{natbib}

\usepackage[french,german,english]{babel}

\usepackage{lmodern} 

\usepackage{fourier} 
\usepackage{setspace} 

\usepackage{graphicx,xcolor}
\graphicspath{{images/}}
\newcommand\circledd[1]{\tikz[baseline=(char.base)]{
            \node[shape=circle,fill,inner sep=1pt] (char) {\small\textcolor{white}{#1}};}}

\usepackage{booktabs}
\usepackage{lipsum}
\usepackage{microtype}
\usepackage{url}
\usepackage{fancyhdr}

\fancypagestyle{plain}{
	\fancyhf{}

	\fancyfoot[EL,OR]{\thepage}}
\fancypagestyle{addpagenumbersforpdfimports}{
	\fancyhead{}
	
	\fancyfoot{}
	\fancyfoot[RO,LE]{\thepage}
}

\usepackage{listings}
\usepackage{hyperref}
\hypersetup{pdfborder={0 0 0},
	colorlinks=true,
	linkcolor=black,
	citecolor=black,
	urlcolor=black}
\ifpdf
\usepackage[final]{pdfpages}
\else
\usepackage{calc}
\usepackage{breakurl}
\usepackage[nlwarning=false]{hypdvips}
\usepackage{backref}
\renewcommand*{\backref}[1]{}
\fi
\usepackage{bookmark}

\makeatletter
\renewcommand\@pnumwidth{20pt}
\makeatother

\makeatletter
\def\cleardoublepage{\clearpage\if@twoside \ifodd\c@page\else
    \hbox{}
    \thispagestyle{empty}
    \newpage
    \if@twocolumn\hbox{}\newpage\fi\fi\fi}
\makeatother \clearpage{\pagestyle{plain}\cleardoublepage}

\usepackage{color}
\usepackage{tikz}
\usepackage[explicit]{titlesec}
\newcommand*\chapterlabel{}
\titleformat{\chapter}[display]  
	{\normalfont\bfseries\Huge} 
	{\gdef\chapterlabel{\thechapter\ }}     
 	{0pt} 
 	  {\begin{tikzpicture}[remember picture,overlay]
    \node[yshift=-8cm] at (current page.north west)
      {\begin{tikzpicture}[remember picture, overlay]
        \draw[fill=black] (0,0) rectangle(35.5mm,15mm);
        \node[anchor=north east,yshift=-7.2cm,xshift=34mm,minimum height=30mm,inner sep=0mm] at (current page.north west)
        {\parbox[top][30mm][t]{15mm}{\raggedleft \rule{0cm}{0.6cm}\color{white}\chapterlabel}};  
        \node[anchor=north west,yshift=-7.2cm,xshift=37mm,text width=\textwidth,minimum height=30mm,inner sep=0mm] at (current page.north west)
              {\parbox[top][30mm][t]{\textwidth}{\rule{0cm}{0.6cm}\color{black}#1}};
       \end{tikzpicture}
      };
   \end{tikzpicture}
   \gdef\chapterlabel{}
  } 
\titlespacing*{name=\chapter,numberless}{-3.7cm}{83.2pt-\parskip}{-3.2pt+\parskip}
\titlespacing*{\chapter}{-3.7cm}{50pt-\parskip-\parskip}{30pt+\parskip+\parskip}
\titlespacing*{\section}{0pt}{13.2pt}{1em-\parskip}  
\titlespacing*{\subsection}{0pt}{13.2pt}{1em-\parskip}
\titlespacing*{\subsubsection}{0pt}{13.2pt}{1em-\parskip}
\titlespacing*{\paragraph}{0pt}{13.2pt}{1em-\parskip}

\newcounter{myparts}
\newcommand*\partlabel{}
\titleformat{\part}[display]  
	{\normalfont\bfseries\Huge} 
	{\gdef\partlabel{\thepart\ }}     
 	{0pt} 
 	  {\ifpdf\setlength{\unitlength}{20mm}\else\setlength{\unitlength}{0mm}\fi
	  \addtocounter{myparts}{1}
	  \begin{tikzpicture}[remember picture,overlay]
    \node[anchor=north west,xshift=-65mm,yshift=-6.9cm-\value{myparts}*20mm] at (current page.north east) 
      {\begin{tikzpicture}[remember picture, overlay]
        \draw[fill=black] (0,0) rectangle(62mm,20mm);   
        \node[anchor=north west,yshift=-6.1cm-\value{myparts}*\unitlength,xshift=-60.5mm,minimum height=30mm,inner sep=0mm] at (current page.north east)
        {\parbox[top][30mm][t]{55mm}{\raggedright \color{white}Part \partlabel \rule{0cm}{0.6cm}}};  
        \node[anchor=north east,yshift=-6.1cm-\value{myparts}*\unitlength,xshift=-63.5mm,text width=0.97\textwidth,minimum height=30mm,inner sep=0mm] at (current page.north east)
              {\parbox[top][30mm][t]{\textwidth}{\raggedleft \rule{0cm}{0.6cm}\color{black}#1}};
       \end{tikzpicture}
      };
   \end{tikzpicture}
   \gdef\partlabel{}
  } 
\titlespacing*{\part}{11.06cm}{26.4pt-\parskip-\parskip}{0pt}

\usepackage{amsmath}
\usepackage{amsfonts}
\usepackage{amssymb}
\usepackage{mathtools}
\usepackage[figurename=Figure]{caption}
\usepackage{soul}
\usepackage{booktabs,arydshln}
\usepackage{algorithm}
\usepackage[export]{adjustbox}
\usepackage{multirow}
\usepackage{threeparttable}
\usepackage{tabularx}
\usepackage{tcolorbox}
\usepackage[inline]{enumitem}
\usepackage{xparse}
\usepackage{subcaption}
\usetikzlibrary{calc}
\usetikzlibrary{bayesnet}
\usepackage[noend]{algpseudocode}

\definecolor{orange2}{rgb}{0.77254902, 0.352941176, 0.066666667}
\definecolor{yellow2}{rgb}{0.749019608, 0.564705882, 0.0}
\definecolor{red}{rgb}{1.0,0.0,0.0}
\definecolor{orange}{rgb}{1.0,0.65,0.0}
\definecolor{green}{rgb}{0.0,0.75,0.0}
\definecolor{blue}{rgb}{0.0,0.0,1.0}
\definecolor{purple}{rgb}{0.5,0.0,0.5}
\definecolor{white}{rgb}{1.0,1.0,1.0}
\definecolor{redo}{rgb}{1.0,0.0,0.0}
\definecolor{orangeo}{rgb}{1.0,0.65,0.0}
\definecolor{greeno}{rgb}{0.0,1.0,0.0}
\definecolor{blueo}{rgb}{0.0,0.0,1.0}
\definecolor{purpleo}{rgb}{0.5,0.0,0.5}
\definecolor{white}{rgb}{1.0,1.0,1.0}

\newcommand{\bb}[1]{\mathbf{#1}}

\newcommand{\br}{\bb{r}}
\newcommand{\bhr}{\bb{\hat{r}}}
\newcommand{\bR}{\bb{R}}

\newcommand{\bhRzero}{\bb{\hat{R}}^0}
\newcommand{\bhRone}{\bb{\hat{R}}^1}
\newcommand{\bru}{\br_{u}}
\newcommand{\bhru}{\bhr_{u}}
\newcommand{\bhrut}{\bhr_{u}^{t}}

\newcommand{\bhruzero}{\bhr_{u}^{0}}
\newcommand{\brui}{\br_{u,i}}

\newcommand{\bk}{\bb{k}}
\newcommand{\bhk}{\bb{\hat{k}}}
\newcommand{\bK}{\bb{K}}
\newcommand{\bKI}{\bb{K^I}}
\newcommand{\bku}{\bk_{u}}

\newcommand{\bkIi}{\bk^{I}_{i}}
\newcommand{\bkIic}{\bk^{I}_{i,\bc}}
\newcommand{\bhku}{\bhk_{u}}

\newcommand{\bz}{\bb{z}}
\newcommand{\btz}{\bb{\tilde{z}}}

\newcommand{\bzu}{\bz_{u}}
\newcommand{\bzur}{\bz_{u}^{r}}
\newcommand{\bzuk}{\bz_{u}^{k}}
\newcommand{\bzuc}{\bz_{u}^{c}}
\newcommand{\bzuct}{\bz_{u}^{t}}
\newcommand{\btzuct}{\btz_{u}^{t}}

\newcommand{\bc}{\bb{c}}

\newcommand{\bcu}{\bc_{u}}
\newcommand{\bcut}{\bc_{u}^{t}}

\newcommand{\bx}{\bb{x}}
\newcommand{\bmu}{\bb{\mu}}
\newcommand{\bS}{\bb{\Sigma}}
\newcommand{\bmuu}{\bmu_{u}}
\newcommand{\bSu}{\bS_{u}}

\newcommand{\mystar}{{\fontfamily{lmr}\selectfont$\star$}}

\makeatletter

\def\adl@drawiv#1#2#3{%
        \hskip.5\tabcolsep
        \xleaders#3{#2.5\@tempdimb #1{1}#2.5\@tempdimb}%
                #2\z@ plus1fil minus1fil\relax
        \hskip.5\tabcolsep}
\newcommand{\cdashlinelr}[1]{%
  \noalign{\vskip\aboverulesep
           \global\let\@dashdrawstore\adl@draw
           \global\let\adl@draw\adl@drawiv}
  \cdashline{#1}
  \noalign{\global\let\adl@draw\@dashdrawstore
           \vskip\belowrulesep}}

\makeatletter
\def\resetMathstrut@{%
  \setbox\z@\hbox{%
    \mathchardef\@tempa\mathcode`\(\relax
      \def\@tempb##1"##2##3{\the\textfont"##3\char"}%
      \expandafter\@tempb\meaning\@tempa \relax
  }%
  \ht\Mathstrutbox@1.2\ht\z@ \dp\Mathstrutbox@1.2\dp\z@
}
\makeatother

\begin{document}
\setlength{\parindent}{0pt}
\setlength{\parskip}{0pt} 
\frontmatter
\input{titlepage}

\include{dedication}

\setcounter{page}{0}
\include{acknowledgements}
\include{abstracts}

\cleardoublepage
\pdfbookmark{\contentsname}{toc}
\tableofcontents

\cleardoublepage
\phantomsection
\addcontentsline{toc}{chapter}{List of Figures} 
\listoffigures
 
\cleardoublepage
\phantomsection
\addcontentsline{toc}{chapter}{List of Tables} 
\listoftables

\setlength{\parskip}{1em}

\mainmatter
\cleardoublepage
\include{main/introduction}
\cleardoublepage
\part{Extracting Multi-Faceted Explanations from Text Documents}
\label{p1}
\include{main/AAAI2021/aaai2021}
\include{main/ACL2021/acl2021}
\cleardoublepage
\part{Iterative Critiquing of Explanations}
\label{p2}
\include{main/IJCAI2021/main_final}
\include{main/RECSYS2021/main}
\cleardoublepage
\include{main/conclusion}
\cleardoublepage
\phantomsection
\part*{Appendices}
\addtocontents{toc}{\vspace{\normalbaselineskip}}
\cleardoublepage
\bookmarksetup{startatroot}

\include{appendix}
\backmatter
\include{biblio}


\end{document}

%% file: titlepage.tex
\begin{titlepage}
\begin{otherlanguage}{english}
\begin{center}
\sffamily

\null\vspace{2cm}
{\huge Textual Explanations and Critiques\\in Recommendation Systems} \\[24pt] 
    
\vfill

\begin{tabular} {cc}
\parbox{0.3\textwidth}{\includegraphics[width=4cm]{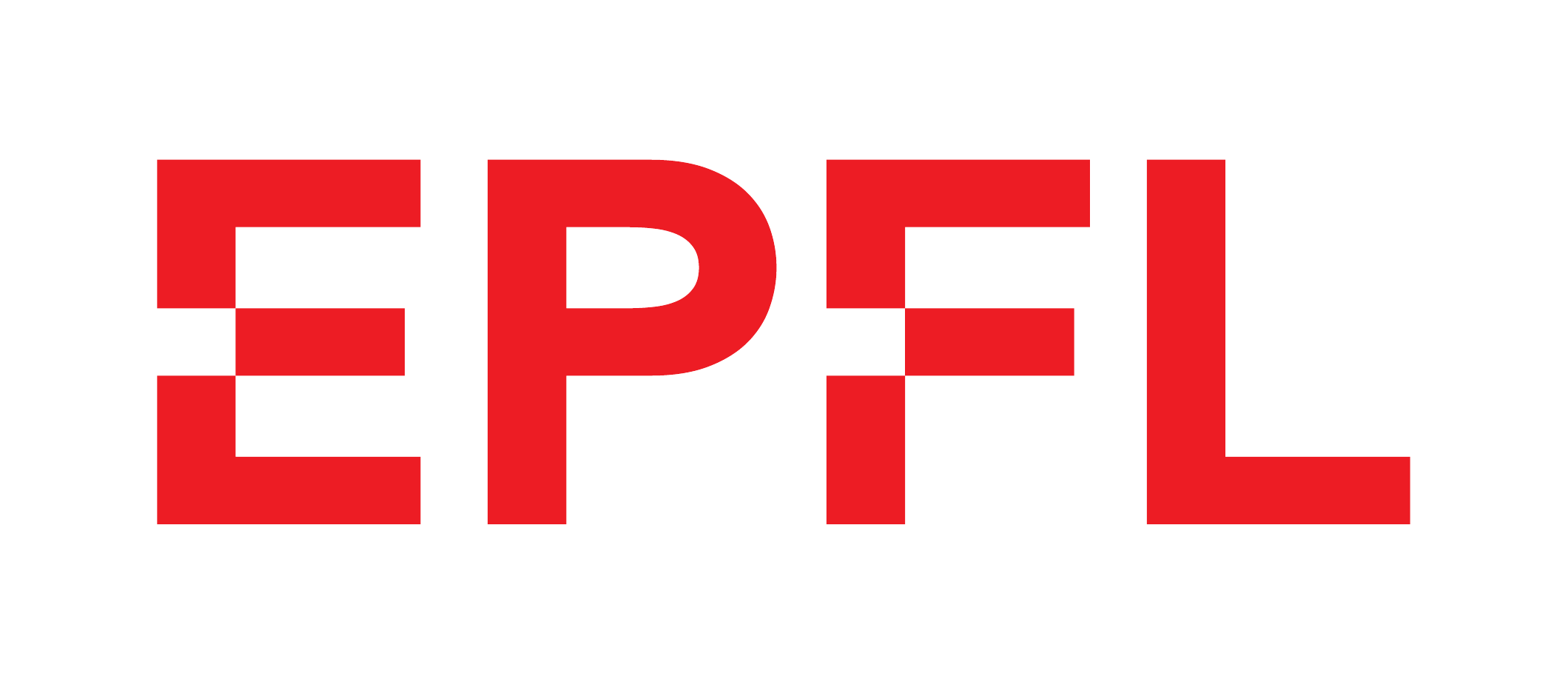}}
&
\parbox{0.7\textwidth}{%
	Thèse n. 9047\\
	Présentée le 25 février 2022\\\\
	à la Faculté informatique et communications\\
	Laboratoire d'intelligence artificielle\\
	Programme doctoral en informatique et communications\\

%
	École polytechnique fédérale de Lausanne\\[6pt]
	pour l'obtention du grade de Docteur ès Sciences\\
	\\par\\ [4pt]
	\null \hspace{3em} Diego Matteo ANTOGNINI\\[9pt]
\small
Acceptée sur proposition du jury:\\[4pt]
Prof. R. West, président du jury\\
Prof. B. Faltings, directeur de thèse\\
Prof. J. McAuley, rapporteur\\
Prof. S. Sanner, rapporteur\\
Prof. A. Bosselut, rapporteur\\[12pt]
Lausanne, EPFL, 2022}
\end{tabular}
\end{center}
\vspace{2cm}
\end{otherlanguage}
\end{titlepage}

%% file: dedication.tex
\cleardoublepage
\thispagestyle{empty}

\vspace*{3cm}

\begin{raggedleft}
Our greatest weakness lies in giving up. The most certain \\ way to succeed is always to try just one more time. \\
     --- Thomas A. Edison\\
\end{raggedleft}

\vspace{4cm}

\begin{center}
    To my family, for their love and support.\dots
\end{center}

%% file: acknowledgements.tex
\chapter*{Acknowledgements}
\markboth{Acknowledgements}{Acknowledgements}
\addcontentsline{toc}{chapter}{Acknowledgements}

First of all, I would like to express my deepest gratitude to my thesis advisor, Prof. Boi Faltings, for his insightful supervision, support, flexibility, and scientific freedom to pursue my research interests. I am also grateful to Dr. Claudiu Musat for always being so positive and encouraging me to always give my best and never give up. I would like to thank Prof. Julian McAuley for his guidance and care during my research stay at the University of California, San Diego. 

\vspace{0.5em}
I had the honor of having Prof. Robert West (EPFL), Prof. Julian McAuley (UCSD), Prof. Scott Sanner (University of Toronto), and Prof. Antoine Bosselut (EPFL) for my thesis committee. Thank you for taking the time to read my thesis, participate in my thesis defense and for your insightful feedback and helpful suggestions on the thesis.

\vspace{0.5em}
During my Ph.D. journey, I spent a great time with current and former members of the artificial intelligence laboratory (LIA): Panayiotis, Aleksei, Igor, Ljubomir, Fei, and also Naman, Zeki, and Adam. Special thanks to Aleksei for all the great times in San Diego. When we were not surfing, we were climbing one of the highest peaks in the U.S.A. in 13 hours (Mount Whitney, 4,421m), eating great food, or simply going through the Death Valley or cathedral rocks in Sedona. Thank you also to Panayiotis for the great times and discussions. Thank you to Damien Doy for making me discover hike challenges: around 40km in 14 hours from Champéry (1,050m) - Dents du Midi (3,258) - Evionnaz (469m). A special thanks to Alexandre Perez for bike tours, restaurants, and other great times. Also, I would like to thank our former secretaries Sylvie Thomet and Patricia Defferrard for their tireless help with all the administrative services. Also, special thanks to Nichole Loyd Janlov who helped me so much with all the administrative steps to come to UCSD. Finally, the IT support of EPFL made sure IT problems were minimal and promptly resolved. Thank you to Stéphane Ecuyer, Carlos Perez, and Yohan Moulin.

\vspace{0.5em}
A heartfelt thank you to the 30+ students that I had the privilege of supervising during my Ph.D.: Kirtan Padh, Saibo Geng, Shuangqi Li, Antoine Scardigli, Milena Filipovic, Blagojce Mitrevski, Martin Milenkoski, Nikola Milojkovic, and many more. Although a huge thank you to Prof. Hatem Ghorbel and Prof. Cédric Billat for great discussions and for asking me to assess many students' bachelor and master theses. Same for Pierre Ferrari and Renato Rota for CFC/EFZ final projects.

\vspace{0.5em}
I am very thankful to my parents, Luciano and Ana, and my siblings, Sergio and Laura, and my nephew Noa, for their support and encouragement. Finally, I wish to thank my wonderful fiancée, Michaela, for her support and patience, but most importantly for her love, her person, all the delicious meals and desserts she made, and all the great adventures we had together!

\enlargethispage{\baselineskip}
\vspace{0.3em}
\noindent\textit{Lugano, March 7, 2022}
\hfill D.~A.

%% file: abstracts.tex

\cleardoublepage
\chapter*{Abstract}
\markboth{Abstract}{Abstract}
\addcontentsline{toc}{chapter}{Abstract (English/Français)} 

Artificial intelligence and machine learning algorithms have become ubiquitous.
Although they offer a wide range of benefits, their adoption in decision-critical fields is limited by their lack of interpretability, particularly with textual data. Moreover, with more data available than ever before, it has become increasingly important to explain automated predictions.

Generally, users find it difficult to understand the underlying computational processes and interact with the models, especially when the models fail to generate the outcomes or explanations, or both, correctly. This problem highlights the growing need for users to better understand the models' inner workings and gain control over their actions. This dissertation focuses on two fundamental challenges of addressing this need. The first involves \textit{explanation generation}: inferring high-quality explanations from text documents in a scalable and data-driven manner. The second challenge consists in making explanations actionable, and we refer to it as \textit{critiquing}. This dissertation examines two important applications in natural language processing and recommendation tasks.

First, we present two models for extracting high-quality and personalized justifications from text documents under the paradigm of selective rationalization -- a binary selection of input features from which the model computes the outcome. 
We propose the concept of multi-dimensional rationales and emphasize that a single overall selection does not accurately capture the multi-faceted nature of useful rationales. The first method relies on multi-task learning and learns the rationales in an unsupervised manner, with no prior in the data. Inspired by the role of concept-based thinking in human reasoning, our second model is a generalization: it assumes that only one label is observed. We empirically demonstrate the efficiency of the models in terms of explanation and predictive performance as well as their scalability. It opens new doors for applications, such as recommendation or summarization.

Second, we use the high-level idea of multi-dimensional rationalization to mine massive corpora and construct large personalized justification datasets. We show that human users significantly prefer our explanations over those produced by state-of-the-art methods.

Last, we propose two conversational explainable recommendation systems. The first explains a user rating by inferring a set of keyphrases. Conditioned on them, it generates an~abstractive personal justification. We allow users to interact iteratively with the explanations~and refine the recommendation through an unsupervised critiquing method. Our second model is based on multimodal modeling and self-supervision. It enables fast and efficient multi-step critiquing. Using real-world datasets, we show that our models exhibit better performance than state-of-the-art models in terms of recommendation, explanation, and critiquing.

Overall, we demonstrate that interpretability does not come at the cost of reduced performance in two consequential applications. Our framework is applicable to other fields as well. This dissertation presents an effective means of closing the gap between promise and practice in artificial intelligence.

\textbf{Keywords}: explainability, interpretability, deep learning, selective rationalization, conversational explainable recommendation systems, multi-step critiquing, variational autoencoder

\begin{otherlanguage}{french}
\cleardoublepage
\chapter*{Résumé}
\markboth{Résumé}{Résumé}

L'intelligence artificielle et les algorithmes d'apprentissage automatique sont devenus omniprésents.
Bien qu'ils offrent un large éventail d'avantages, leur adoption dans les domaines critiques pour la décision est limitée par leur manque d'interprétabilité, en particulier avec les données textuelles. De plus, comme les données disponibles sont plus nombreuses que jamais, il est devenu de plus en plus important d'expliquer les prédictions automatisées.

En général, les utilisateurs ont du mal à comprendre les processus de calcul sous-jacents et à interagir avec les modèles, surtout lorsque ces derniers ne parviennent pas à générer correctement les résultats ou les explications, ou les deux. Ce problème met en évidence le besoin croissant des utilisateurs de mieux comprendre le fonctionnement interne des modèles et d'avoir un contrôle sur leurs actions. Cette thèse se concentre sur deux défis fondamentaux pour répondre à ce besoin. Le premier concerne la \textit{génération d'explications} : inférer des explications de haute qualité à partir de documents textuels à grande échelle et d'une manière orientée données. Le second défi consiste à rendre les explications interactives, ce que nous appelons \textit{critiquer}. Cette thèse examine deux applications importantes dans le traitement du langage naturel et la recommandation.

Tout d'abord, nous présentons deux modèles pour extraire des justifications personnalisées et de haute qualité à partir de documents textuels sous le paradigme de la rationalisation sélective - une sélection binaire de caractéristiques d'entrée à partir desquelles le modèle calcule le résultat. Nous proposons le concept de justifications multidimensionnelles et montrons qu'une seule sélection globale ne permet pas d'identifier avec précision la nature multidimensionnelle des justifications utiles. La première méthode repose sur l'apprentissage multi-tâches et apprend à extraire les justifications de manière non supervisée, sans assomptions sur les données. Inspiré par le rôle de la pensée basée sur les concepts dans le raisonnement humain, notre deuxième modèle en est une généralisation : il suppose qu'une seule étiquette est observée. Nous démontrons empiriquement l'efficacité des modèles en termes de performances d'explication et de prédiction, ainsi que leur évolutivité. Cela ouvre de nouvelles portes pour des applications telles que la recommandation ou le résumé.

Deuxièmement, nous utilisons l'idée de haut niveau de la rationalisation multidimensionnelle pour exploiter des corpus massifs et construire de grands ensembles de données de justifications personnalisées. Nous montrons que les utilisateurs humains préfèrent nettement nos explications à celles produites par les méthodes les plus récentes.

Enfin, nous proposons deux systèmes de recommendation conversationnels et explicables. Le premier explique une évaluation d'utilisateur en déduisant un ensemble de mots-clés. En fonction de ceux-ci, il génère une justification personnalisé. Nous permettons aux utilisateurs d'interagir itérativement avec les explications et d'affiner la recommandation par une méthode de critique non supervisée. Notre deuxième modèle est basé sur la modélisation multimodale et l'auto-supervision. Il permet une critique rapide et efficace en plusieurs étapes. En utilisant des ensembles de données réelles, nous montrons que nos modèles présentent de meilleures performances que les modèles de pointe en termes de recommandation, d'explication et de critique.

Dans l'ensemble, nous démontrons que l'interprétabilité ne se fait pas au prix d'une réduction des performances dans deux applications conséquentes. Notre cadre est également applicable à d'autres domaines. Cette thèse présente un moyen efficace de combler le fossé entre les promesses et la pratique en intelligence artificielle.

\textbf{Mots-clés} : explicabilité, interprétabilité, apprentissage profond, rationalisation sélective, systèmes de recommandation explicables et conversationnels, critiquer, auto-encodeur variationnel

\end{otherlanguage}

\cleardoublepage
\chapter*{Publications}
\markboth{Publications}{Publications}
\addcontentsline{toc}{chapter}{Publications} 

The contributions and the results presented in this thesis were published in the following publications:

\begin{enumerate}
	\item\textbf{ \cite{antognini2019multi}:} \underline{Antognini, D.}, Musat, C., and Faltings, B. (2021b). Multi-dimensional explanation of target variables from documents. In \textit{Proceedings of the AAAI Conference on Artificial Intelligence: AAAI 2021}, 35(14):12507–12515.
	\item\textbf{ \cite{antognini-2021-concept}:} \underline{Antognini, D.} and Faltings, B. (2021b). Rationalization through concepts. In \textit{Findings of the Association for Computational Linguistics: ACL 2021}, pages 761–775, Online. Association for Computational Linguistics.
	\item\textbf{ \cite{antognini2020interacting}:} \underline{Antognini, D.}, Musat, C., and Faltings, B. (2021a). Interacting with explanations through critiquing. In Zhou, Z.-H., editor, In \textit{Proceedings of the Thirtieth International Joint Conference on Artificial Intelligence: IJCAI 2021}, pages 515–521. International Joint Conferences on Artificial Intelligence Organization. Main Track.
	\item\textbf{ \cite{fast_critiquing}:} \underline{Antognini, D.} and Faltings, B. (2021a). Fast multi-step critiquing for vae-based recommender systems. In \textit{Fifteenth ACM Conference on Recommender Systems: RecSys 2021}, page 209–219, New York, NY, USA. Association for Computing Machinery.
	\item\textbf{ \cite{multi_step_demo}:} *Petrescu, D. A., *\underline{Antognini, D.}, and Faltings, B. (2021). Multi-step critiquing user interface for recommender systems. In \textit{Fifteenth ACM Conference on Recommender Systems: RecSys 2021}, page 760–763, New York, NY, USA. Association for Computing Machinery.
\end{enumerate}

The following publications were part of the research conducted during the Ph.D.. They present results that are supplementary to or build upon but are not covered in this dissertation:

\begin{enumerate}
	\setcounter{enumi}{5}
	\item\textbf{ \cite{editing_cooking}:} \underline{Antognini, D.}, Li, S., Faltings, B., and McAuley, J. (2023). Assistive recipe editing through critiquing. In Proceedings of the 17th Conference of the European Chapter of the Association for Computational Linguistics: Main Volume, Online. Association for Computational Linguistics.
	\item\textbf{ \cite{fast_critiquing_pos}:} \underline{Antognini, D.} and Faltings, B. (2022). Positive \& Negative Critiquing for VAE-based Recommenders. arXiv preprint arXiv:2204.02162.
	\item\textbf{ \cite{fairness_kirtan}:} Padh, K., \underline{Antognini, D.}, Lejal Glaude, E., Faltings, B., and Musat, C. (2021). Addressing fairness in classification with a model-agnostic multi-objective algorithm. In Peters, J. and Sontag, D., editors, In \textit{Proceedings of the 37th Conference on Uncertainty in Artificial Intelligence (UAI): UAI 2021}, Proceedings of Machine Learning Research. PMLR.
	\item\textbf{ \cite{antognini-faltings-2020-hotelrec}:} \underline{Antognini, D.} and Faltings, B. (2020b). HotelRec: a novel very large-scale hotel recommen- dation dataset. In \textit{Proceedings of the 12th Language Resources and Evaluation Conference: LREC 2021}, pages 4917–4923, Marseille, France. European Language Resources Association.
	\item\textbf{ \cite{antognini-faltings-2020-gamewikisum}:} \underline{Antognini, D.} and Faltings, B. (2020a). GameWikiSum: a novel large multi-document summarization dataset. In \textit{Proceedings of the 12th Language Resources and Evaluation Conference: LREC 2020}, pages 6645–6650, Marseille, France. European Language Resources Association.
	\item\textbf{ \cite{antognini-faltings-2019-learning}:} \underline{Antognini, D.} and Faltings, B. (2019). Learning to create sentence semantic relation graphs for multi-document summarization. In \textit{Proceedings of the 2nd Workshop on New Frontiers in Summarization}, pages 32–41, Hong Kong, China. Association for Computational Linguistics.
	\item\textbf{ \cite{milojkovic2019multi}:} Milojkovic, N., \underline{Antognini, D.}, Bergamin, G., Faltings, B., and Musat, C. (2020). Multi-gradient descent for multi-objective recommender systems. In Proceedings of the AAAI (2020) - Workshop on Interactive and Conversational Recommendation Systems (WICRS).
	\item\textbf{ \cite{model_evaluation}:} Filipovic, M., Mitrevski, B., \underline{Antognini, D.}, Lejal Glaude, E., Faltings, B., and Musat, C. (2021). Modeling online behavior in recommender systems: The importance of temporal context. In Proceedings of the Perspectives on the Evaluation of Recommender Systems Workshop at RecSys 2021 (PERSPECTIVE 2021).
	\item\textbf{ \cite{cross_recommendation}:} Milenkoski, M., \underline{Antognini, D.}, and Musat, C. (2021). Recommending burgers based on pizza preferences: Addressing data sparsity with a product of experts. In Proceedings of the First Workshop of Cross-Market Recommendation at RecSys 2021 (XMRec 2021).
	\item\textbf{ \cite{momentum_gradient}:} Mitrevski, B., Filipovic, M., \underline{Antognini, D.}, Lejal Glaude, E., Faltings, B., and Musat, C. (2021). Momentum-based gradient methods in multi-objective recommendation. In Proceedings of the First Workshop on Multi-Objective Recommender Systems at RecSys 2021 (MORS 2021).
	\item\textbf{ \cite{giannakopoulos2017dataset}:} *Giannakopoulos, A., *\underline{Antognini, D.}, Musat, C., Hossmann, A., and Baeriswyl, M. (2017). Dataset construction via attention for aspect term extraction with distant supervision. In \textit{2017 IEEE International Conference on Data Mining Workshops (ICDMW): ICDMW 2017}, pages 373–380. IEEE.
\end{enumerate}

%% file: main/introduction.tex
\chapter{Introduction}

Thanks to the era of big data and modern technology, the past decade has witnessed unprecedented progress in artificial intelligence (AI) and machine learning (ML). AI-based decisions now pervade our daily lives and offer a wide range of benefits. In many domains, including healthcare, criminal justice, job hiring, and education, automated predictions have a tangible impact on the final decision \citep{citron2014scored,10.1145/2783258.2788613,barocas-hardt-narayanan,vyas2021hidden}. As research has shown, however, modern systems sometimes have unintentional and undesirable biases \citep[e.g.,][]{pmlr-v81-buolamwini18a,propublica16,pariser11}. Therefore, it is crucial to provide the underlying reasons for automated decisions.

In recent years, new initiatives and regulations have been proposed. In 2016, for example, the European Parliament adopted a set of comprehensive regulations for the collection, storage, and use of personal information: the European General Data Protection Regulation (GDPR)\footnote{\url{https://eur-lex.europa.eu/eli/reg/2016/679/2016-05-04}.} \citep{Goodman_Flaxman_2017}. Around the same time, the U.S. Defense Advanced Research Projects Agency (DARPA) created an explainable artificial intelligence program that also emphasizes the importance of human-computer interaction for ML systems \citep{Gunning2016}. Both of these initiative stress a social ``right to explanation'': users can ask for an explanation of any algorithmic decision or instance of profiling associated with them. Furthermore, prior studies have shown that explanations improve overall system transparency \citep[e.g.,][]{ExplainingRecommendation,sinha2002role} and trustworthiness \citep[e.g.,][]{zhang2018exploring,Kunkel2018TrustrelatedEO}, increasing at an accelerating pace the variety of applications \citep{MILLER20191}.


Nevertheless, current AI and ML systems are far from perfect and do make mistakes. Explainable systems are also concerned, because they can generate faulty, unfaithful, or fragile explanations independently of the correctness of the predictions. \cite{dietvorst2015algorithm,castelo2019task,madoc56152,noexplainability2020} showed that people have less confidence in the models when seeing them err for some tasks, even when the machine performs better than humans. 
This algorithmic aversion can be overcome either by\begin{enumerate}
	\item letting users modify (even slightly) the models \citep{dietvorst2018algorithm} or
	\item explaining how the models induce their predictions \citep{yeomans2019making,noexplainability2020}.
\end{enumerate}
These strategies align with  the view of \cite{shneiderman2016}: highlighting the need for users to understand better underlying computational processes in search, recommender, and other algorithms and the need to give to users the potential to control their actions better.

Interpretability for natural language processing tasks is a large, fast-growing area of research. It includes text classification (e.g., sentiment analysis, risk prediction for diabetes, anemia, and breast cancer), natural language inference, reading comprehension, and question answering \citep{bastings-etal-2019-interpretable,deyoung-etal-2020-eraser,wiegreffe2021teach}. Because more data are available than ever before, it has become increasingly important to explain automated predictions. In recent years, selective rationalization has become popular \citep{lei-etal-2016-rationalizing,bastings-etal-2019-interpretable,deyoung-etal-2020-eraser,yu2021}. The key idea  of selective rationalization is to explain complex neural networks by selecting relevant text snippets from the input text -- called a rationale -- that suffice on their own to yield the same prediction as the original outcome. Rationalization can be naturally extended to different types of inputs, like images. Moreover, in practice, a rationale can potentially be understood and verified against domain knowledge \citep{lei-etal-2016-rationalizing,yala2019deep,chang2019game}. However, rationales are limited by their single dimensionality: current rationalization models strive for one overall selection to explain the outcome. However, useful rationales can be multi-faceted, involving support for different outcomes with different degrees \citep{musat2015personalizing,chang2019game}.

Therefore, it is important to provide fine-grained explanations to users. Recommendation systems represent an impactful real-life application in which interactions are naturally present. There is not only a practical need to better model users and improve the recommendations themselves but also a growing demand for explainability\footnote{For example, since July 2021, Google Search has partly explained to users why it found the results it shows them. \href{https://www.reuters.com/technology/google-is-starting-tell-you-how-it-found-search-results-2021-07-22/}{www.reuters.com/technology/google-is-starting-tell-you-how-it-found-search-results-2021-07-22/}.}. Finally, many data are publicly available, and they contain user feedback and often user-written reviews; this facilitates research and development that targets new, data-driven techniques for explainable recommender systems. 

In an increasingly digital world, recommender systems are an inevitable part of virtually everyone’s daily digital routine. They influence how we perceive our environment, from media content to human relationships. In response to the continuously growing quantity of information, products, and choices, modern recommender systems suggest content tailored to users' interests. Most importantly, users deliver feedback and often write reviews about the products they consume. Research has shown that \begin{enumerate*}
 	\item users offer written opinions about the topics they care about \citep{zhang2014explicit,musat2015personalizing} and
 	\item human explanations for high-level decisions are often based on key concepts \citep{ARMSTRONG1983263,tenenbaum1999bayesian}.
 \end{enumerate*}
 These observations underline that rationales are naturally present in user-written texts, which will play a key role in \begin{enumerate*}
\item the understanding of users' preferences and 
\item better building their profiles to improve subsequent personalized recommendations.
\end{enumerate*} Finally, several studies \citep{KonstanRiedl12umuai,knijnenburg2012explaining} have emphasized that personalization reduces search effort and makes the user experience more comfortable and engaging. Furthermore, it plays a crucial role in many e-commerce websites, such as Amazon, Twitter, YouTube, and Netflix \citep{10.1145/3370082}. For example, in 2012 35\% of purchases on Amazon and 75\% of what users watch on Netflix were influenced by their recommender systems \citep{mackenzie2013retailers}.

Although modern recommender systems accurately capture users' preferences and achieve high performance, they offer little transparency regarding their inner workings. Thus, their performance comes at the cost of an increased complexity that makes them seem like black boxes to end users, which may result in distrust or rejection of the recommendations themselves~\citep{herlocker2000explaining,ExplainingRecommendation}. It has been shown that providing explanations along with item recommendations \begin{enumerate}
 \item enables users to understand why a particular item has been suggested and hence to make better decisions~\citep{chang2016crowd,bellini2018knowledge},
 \item increases the system's overall transparency for users~\citep{sinha2002role,ExplainingRecommendation}, and 
 \item improves users' perceived trustworthiness~\citep{putrust2006,zhang2018exploring,Kunkel2018TrustrelatedEO}.
 \end{enumerate}

Although it is important to provide explanations for recommendation systems, one obstacle to doing so involves \textbf{explanation generation}: how do we generate useful explanations? Multiple approaches to producing simple explanations have been proposed: aspect words \citep{McAuley:2013:HFH:2507157.2507163,keyphraseExtractionDeep}, template sentences \citep{zhang2014explicit}, and user-item similarity \citep{herlocker2000explaining}. However, \cite{kunkel2019let,chang2016crowd,wilkinson2021} showed that not all explanations are equivalent: highly personalized justifications\footnote{We use the terms ``explanations'', ``rationales'', and ``justifications'' interchangeably.} using natural language lead to substantial improvements in perceived recommendation quality and trustworthiness compared to more straightforward explanations. Nevertheless, generating useful and meaningful explanations remains a difficult task due to the lack of ground-truth datasets specifying what ``good'' justifications are and indicating whether they are multi-faceted.

\begin{figure}[!t]
\centering
\includegraphics[width=1.0\textwidth]{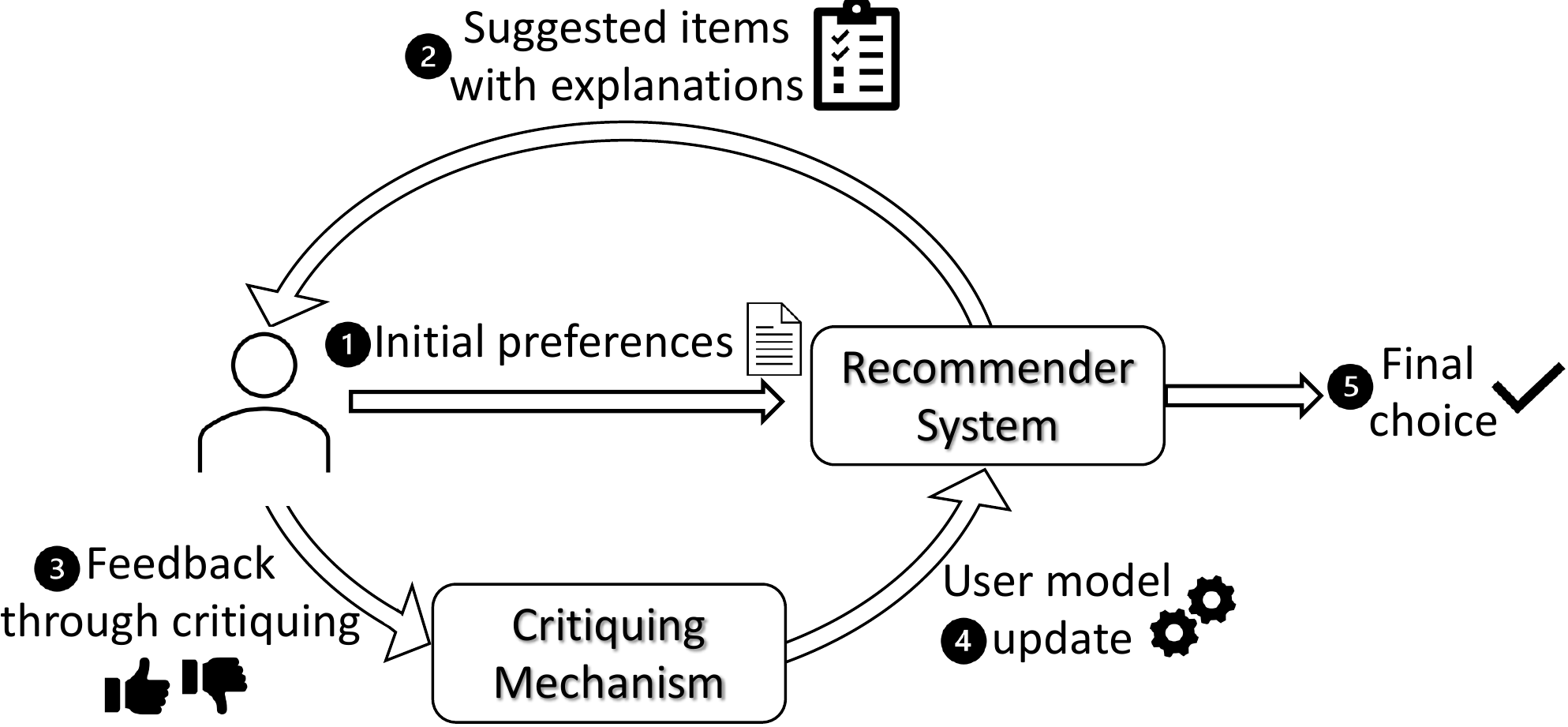}	
\caption{\label{fig_intr_crit}The cycle of conversational critiquing. First, the user informs the recommender system of his or her initial preferences. Then, the model returns a list of recommended items with their respective explanations. The user can interact with the explanations and critique them. This process (\protect\circledd{2} - \protect\circledd{4}) repeats until the user accepts the recommendation and ceases to provide critiques.}
\end{figure}

An essential advantage of explanations is that they provide a basis for feedback. If users understand what has generated the suggestions, they can refine the recommendations by interacting directly with the explanations~\citep{10.1145/3442381.3450123}. Therefore, the second challenge involves \textbf{critiquing} -- a conversational recommendation method that incrementally adapts recommendations in response to user preferences~\citep{chen2012critiquing}. This iterative (i.e., multi-step) process continues until the user accepts the recommendation and ceases to provide critiques. Figure~\ref{fig_intr_crit} illustrates the overall interaction cycle. It has been shown that critiquing has a substantive impact on the user's experience and improves the user's decision accuracy \citep{pu2005integrating,Viappiani06preference-basedsearch,viappiani2008preference}. Another benefit of critiquing is that it enables the system to correct and improve an incomplete or inaccurate model of the user's preferences \citep{10.1007/978-3-540-73078-1_11}. 

Historically, most work on critiquing \citep[e.g.,][]{Faltings2004b,Viappiani06preference-basedsearch,viappiani2008preference,HybridChen,10.1007/978-3-540-73078-1_11} has assumed a fixed set of discrete and domain-dependent attributes along with explicit constraint- and utility-based methods. However, a major limitation of direct filtering-based interactions is assuming a fixed set of known attributes. Another drawback is their lack of effectiveness for domains with expressive feature sets: critiques might not generalize to other items with correlated (hidden) attributes \citep{keyphraseExtractionDeep}. Recently, few studies \citep{keyphraseExtractionDeep,luo2020b,luo2020,hanze2020} have assumed that item properties are represented as keyphrases sourced from subjective user reviews. However, it is not trivial to co-embed item critiques with the latent user preferences. These limitations hinder the pace at which companies are deploying explainable recommender systems. 

\section{Dissertation Structure and Contributions}

This dissertation addresses the two aforementioned challenges: explanation generation and multi-step critiquing. More specifically, our approach first extracts textual explanations from natural language texts using novel multi-aspect rationales. Then, our method tailors and leverages textual explanations for \textit{explainable} recommendation systems. Our critiquing methods enable explanations to be actionable, which in turn enables the recommenders to become conversational. Our framework is generalizable beyond recommendation systems and other domains, such as medicine, law, finance, or cooking. Examples include medical-report generation from radiography, diagnostic and procedural coding of clinical notes, the removal of biases and sensitive information from documents, and the rewriting of cooking recipes to satisfy dietary restrictions. Finally, the input data are not restricted to text.

Part~\ref{p1} of this dissertation presents two approaches to extracting high-quality, personalized, multi-faceted justifications from text documents under the paradigm of selective rationalization. A unique advantage of rational models over many other interpretable methods \citep[e.g.,][]{iclr2015,ribeiro2016should,serrano-smith-2019-attention,li2016understanding,pmlr-v37-xuc15,chen2018shapley} is that unselected words are guaranteed not to contribute to the final prediction. Rationales are faithful (i.e., meaningfully influencing the model's predictions) and can also be considered high-quality compressed information of a data sample with respect to the downstream task. This opens new doors for applications, such as user profiling, summarization, recommendation, and many others. In the two proposed methods, we infer multi-dimensional rationales and emphasize that a single overall selection does not accurately capture the multi-faceted nature of useful rationales. Moreover, we demonstrate the efficiency and scalability of our two approaches on massive corpora.

Based on the high-level idea of rationalization, Part~\ref{p2} uses the induced multi-dimensional rationales -- which reflect users' ratings and subratings -- to construct large personalized recommendation justification datasets and train models to infer personalized explanations alongside recommendations. The justifications are generated in an abstractive fashion and are intended to reflect the user's preferences for each recommended item. We show that human users significantly prefer our explanations over those produced by state-of-the-art techniques. Moreover, using language-based feedback, the user can then interact with these explanations to refine the recommendation, enabling a type of reasoning from the machine side. By explicitly critiquing certain aspects of the items, the recommender can integrate these constraints for subsequent recommendations. We propose two approaches with different properties that achieve good performance in adapting to the preferences expressed in multi-step critiquing and generate consistent explanations without the cost of reduced recommendation performance. Finally, we introduce multiple model-agnostic interfaces to test such systems on an actual use case, and we highlight some limitations of these approaches in the hope that the research community will find ways to overcome them.

The two major parts of this dissertation address the challenges outlined above: selective rationalization (i.e., explanation generation) and developing a conversational explainable recommendation system (i.e., explanation generation and  critiquing). Both parts propose several innovative techniques along with their applications. Our dissertation's main contributions are as follows:

\begin{enumerate}
	\item \textbf{From single-dimensional to multi-dimensional rationales.}\hspace{\parindent} We first focus on selective rationalization for explaining the outcomes corresponding to \textit{multiple} labels. We take the best of the attention and rationale methods and propose the Multi-Target Masker (MTM), which addresses their limitations by replacing the hard binary mask with a soft multi-dimensional mask (one for each target); it is learned in an unsupervised and multi-task learning manner while jointly predicting all the target variables. To the best of our knowledge, this study is the first to neither assume any prior on the data distribution nor train, tune, or maintain multiple models, one for each label to explain. Moreover, our experiments show that our proposed model significantly outperforms attention and strong baselines in terms~of mask precision, coherence, and aspect rating performance. Thus, interpretability no longer comes at the cost of reduced performance.
	\item \textbf{Generalizing multi-dimensional rationalization through latent concepts.}\hspace{\parindent} Our second contribution involves a more challenging and realistic setting than the first, but one that has the same advantages. We aim to learn the same multi-dimensional rationales while observing only \textit{one} label. We present the Concept-based Rationalizer (ConRAT), a novel rationalization scheme that first identifies a set of concepts in a document and then decides which ones are currently described (binary selection). The model explains the prediction with a linear aggregation of concepts. In addition,~we~design two regularizers that guide the model to induce interpretable concepts and propose two optional techniques, knowledge distillation and concept pruning, in order to boost the performance further. Our method significantly outperforms strong \textit{supervised} baseline models in terms of predictive and explanation performance. Finally, the concepts generated align with human rationalization and are preferred over rationales inferred by state-of-the-art methods trained on each aspect label independently.
	\item \textbf{Generating large-scale personalized justification datasets for recommendation.}\hspace{\parindent} We introduce a data-driven and scalable pipeline, based on the previous contributions, that extracts high-quality and personalized justifications from users' reviews but that is also applicable to other types of documents. The extracted text snippets reflect user ratings and  subratings. Applied in the area of recommendation, we show that human users prefer our justifications more than 80\% of the time over those produced by state-of-the-art techniques.
	\item \textbf{Interacting with explanations through critiquing.}\hspace{\parindent} We examine the problem of recommendation with actionable fine-grained explanations. We propose a transformer-based model (T-RECS) that explains a rating by first inferring a set of keyphrases describing the intersection between the profiles of a user and an item. Conditioned on the keyphrases, the model generates a synthetic personalized justification. We then leverage these explanations in a new unsupervised method for single-step and multi-step critiquing. Our system achieves good performance in adapting to the preferences expressed in multi-step critiquing and generates consistent explanations. In the continuation, we introduce four different web interfaces that help users make decisions and find their ideal item. Because our application is model-agnostic (for both recommender systems and critiquing mechanisms), it allows great flexibility and further extensions. Our interfaces are, above all, a useful tool for research on recommendation with critiquing and for highlighting some limitations of these approaches.
	\item \textbf{Making multi-step critiquing faster and more efficient with self-supervision.}\hspace{\parindent} Finally, we address the current recommender-system models' inherent trade-off between the efficiency, recommendation, explanation, and critiquing performance. We present a novel multimodal variational autoencoder (M\&Ms-VAE), trained under a weak supervision scheme to simulate both fully and partially observed variables. It is computationally efficient and handles critiques up to 25.6x faster than existing models. Last but not least, it does not suffer from the cold-start problem, thanks to our multimodal modeling and training scheme.	
\end{enumerate}

The five contributions outlined above are described in detail in the next four chapters (contributions 3 and 4 are presented together). Each chapter corresponds to one or more publications. All the chapters begin with a preface, consisting of a summary of the work and, when applicable, a detailed presentation of their individual contributions using the CRediT framework \citep{brand2015beyond}. The final chapter offers concluding remarks and discusses possible directions for future research.

%% file: main/AAAI2021/aaai2021.tex
\chapter{Multi-Dimensional Rationalization}
\label{chapter_aaai2021}

\section{Preface}

\textbf{Contribution and Sources.}\hspace{\parindent} This chapter is largely based on \cite{antognini2019multi}. The detailed individual contributions are listed below using the CRediT taxonomy \citep{brand2015beyond} (terms are selected as applicable).

\begin{table}[!h]
\begin{tabular}{@{}l@{\hspace{1mm}}l@{}}
Diego Antognini (author): & Conceptualization, Methodology, Software, Validation,\\
& Investigation, Formal Analysis, Writing -- Original Draft,\\
&Writing -- Review \& Editing.\\ \\
Claudiu Musat: & Writing -- Review \& Editing (supporting), Supervision (supporting). \\ \\
Boi Faltings: & Writing -- Review \& Editing, Administration, Supervision.
\end{tabular}
\end{table}

\textbf{Summary.}\hspace{\parindent} Automated predictions require explanations to be interpretable by humans.
Past work used attention and rationale mechanisms to find words that predict the target variable of~a document. Often though, they result in a tradeoff between noisy explanations or a drop in accuracy. 
Furthermore, rationale methods cannot capture the multi-faceted nature of justifications for multiple targets, because of the non-probabilistic nature of the mask.
In this chapter, we propose the Multi-Target Masker (MTM) to address these shortcomings.
The novelty lies in the soft multi-dimensional mask that models a relevance probability distribution over the set of target variables to handle ambiguities and remove the low-correlation assumption in the data. 
Additionally, two regularizers guide MTM to induce long, meaningful explanations.
We evaluate MTM on two datasets and show, using standard metrics and human annotations, that the resulting masks are more accurate and coherent than those generated by the state-of-the-art methods. 
Moreover, MTM is the first to also achieve the highest F1 scores for all the target variables simultaneously while training a single model.

\section{Introduction}

Neural models have become the standard for natural language processing tasks. Despite the large performance gains achieved by these complex models, they offer little transparency about their inner workings. Thus, their performance comes at the cost of interpretability, limiting their practical utility. Integrating interpretability into a model would supply reasoning for the prediction, increasing its utility.

Perhaps the simplest means of explaining predictions of complex models is by selecting relevant input features. Prior work includes various methods to find relevant words in the text input to predict the target variable of a document. Attention mechanisms \citep{iclr2015,luong-etal-2015-effective} model the word selection by a conditional importance distribution over the inputs, used as explanations to produce a weighted context vector for downstream modules. However, their reliability has been questioned \citep{jain2019attention,pruthi-etal-2020-learning}.
Another line of research includes rationale generation methods \citep{NIPS2017_7062,li2016understanding,lei-etal-2016-rationalizing}. 
If the selected text input features are short and concise -- called a rationale or mask -- and suffice on their own to yield the prediction, it can potentially be understood and verified against domain knowledge \citep{lei-etal-2016-rationalizing,chang2019game}. Specifically, these rationale generation methods have been recently proposed to provide such explanations alongside the prediction.
Ideally, a good rationale should yield the same or higher performance as using the full input.

\begin{figure}[t]
\centering
\begin{tabular}{@{}c@{\hspace{1mm}}c@{}}
     \underline{Attention Model} & \underline{Multi-Target Masker (Ours)} \\
    Trained on $\ell_{pred}$ & Trained on $\ell_{pred}$ \\
    and no constraint & with $\lambda_p$, $\ell_{sel}$, and $\ell_{cont}$ \\ 
    \multicolumn{1}{c}{\includegraphics[width=0.3\textwidth,height=.65cm]{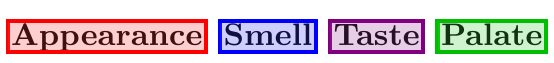}} & \multicolumn{1}{c}{\includegraphics[width=0.3\textwidth,height=.65cm]{main/AAAI2021/Figures/example_legend.pdf}}\\
     \includegraphics[width=0.4\textwidth,height=7.25cm]{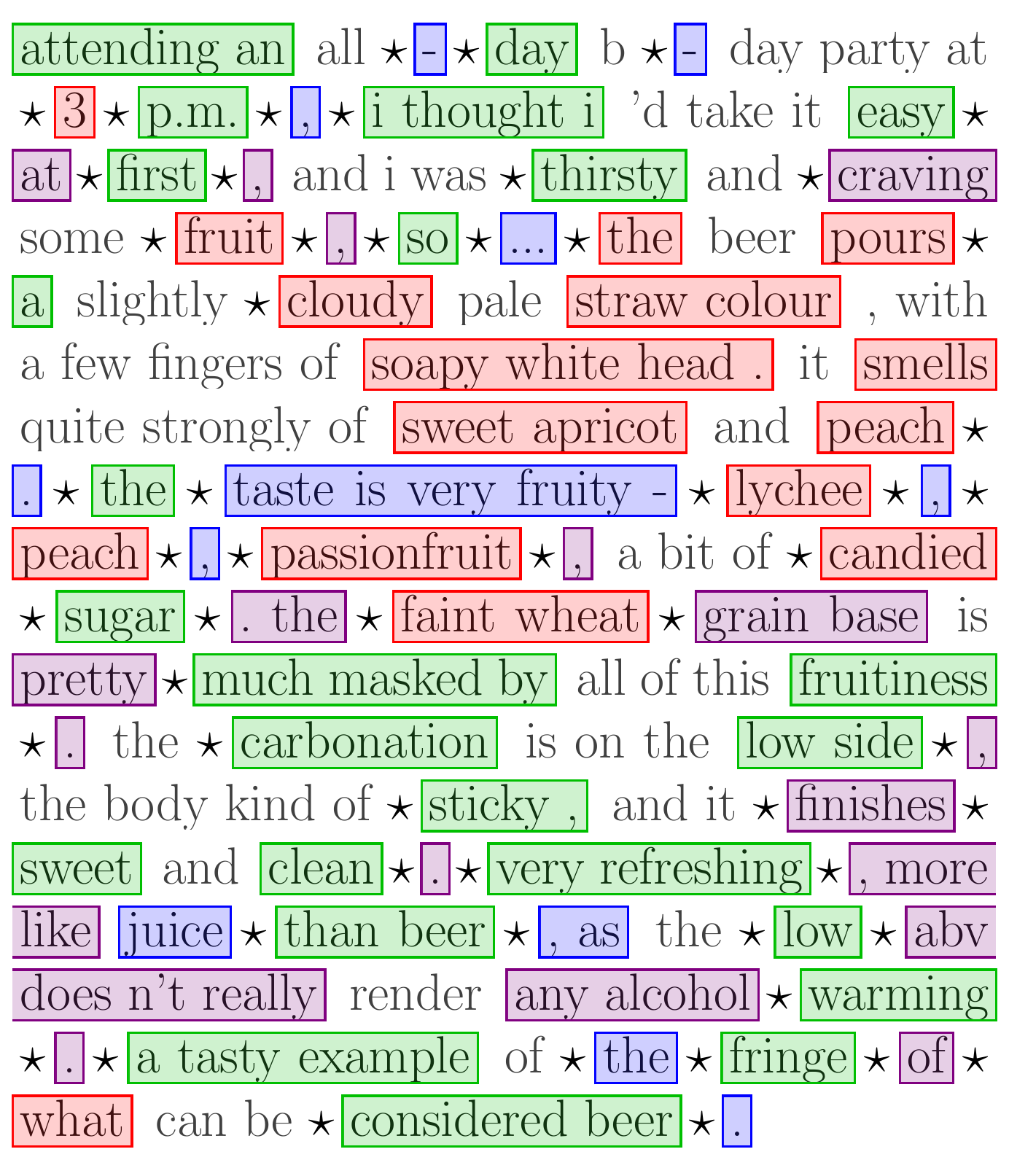} &
     \includegraphics[width=0.45\textwidth,height=7.25cm]{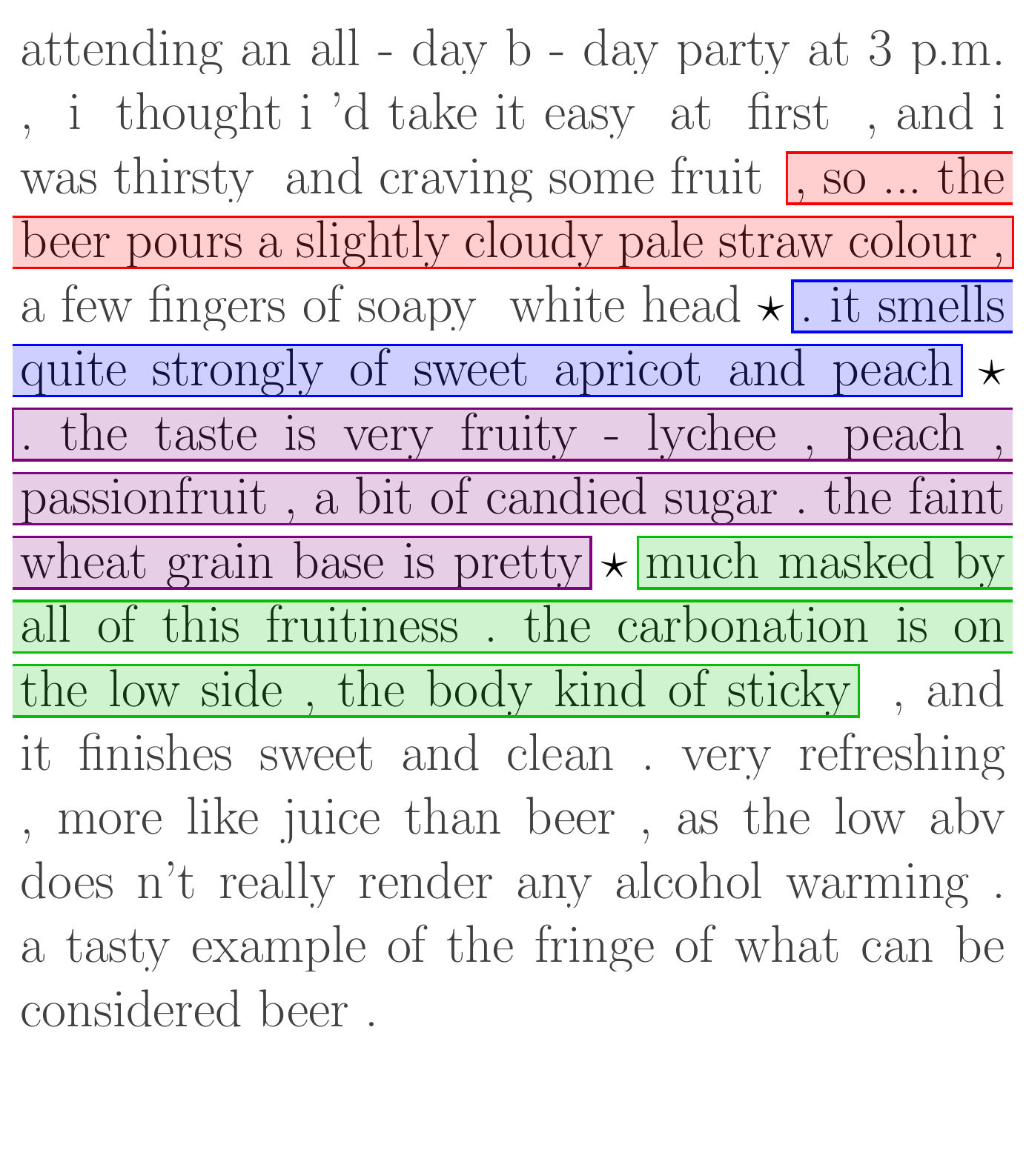}\\
     Aspect Changes \mystar : $56$ & Aspect Changes \mystar : $3$
     \end{tabular}

\caption{\label{sample_lambda}A beer review with explanations produced by an attention model and our Multi-Target Masker model. The colors depict produced rationales (i.e., justifications) of the rated aspects: Appearance, Smell, Taste, and Palate. 
	The induced rationales mostly lead to long sequences that clearly describe each aspect (one switch \mystar\ per aspect), while the attention model has many short, noisy interleaving sequences.
}
\end{figure}

The key motivation of this chapter arises from the limitations of existing methods. First, the attention mechanisms induce an importance distribution over the inputs, but the resulting explanation consists of many short and noisy word sequences (Figure~\ref{sample_lambda}). In addition, the rationale generation methods produce coherent explanations, but the rationales are based on a binary selection of words, leading to the following shortcomings: 
\begin{enumerate}[topsep=0pt]
	\item they explain only one target variable,
	\item they make a priori assumptions about the data, and 
	\item they make it difficult to capture ambiguities in the text.
\end{enumerate}
Regarding the first shortcoming, rationales can be multi-faceted by definition and involve support for different outcomes. If that is the case, one has to train, tune, and maintain one model per target variable, which is impractical. For the second, current models are prone to pick up spurious correlations between the input features and the output. Therefore, one has to ensure that the data have low correlations among the target variables, although this may not reflect the real distribution of the data. Finally, regarding the last shortcoming, a strict assignment of words as rationales might lead to ambiguities that are difficult to capture. For example, in an hotel review that states \textit{``The room was large, clean, and close to the beach.''},  the word \textit{``room''} refers to the aspects \textit{Room}, \textit{Cleanliness}, and \textit{Location}. All these limitations are implicitly related due to the non-probabilistic nature of the~mask. For further illustrations, see Figure~\ref{sample_new_hotel} and the Appendix~\ref{samples_with_visualization}.

In this chapter, we take the best of the attention and rationale methods and propose the Multi-Target Masker to address their limitations by replacing the hard binary mask with a soft multi-dimensional mask (one for each target), in an unsupervised and multi-task learning manner, while jointly predicting all the target variables. We are the first to use a probabilistic multi-dimensional mask to explain multiple target variables jointly without any assumptions on the data, unlike previous rationale generation methods. More specifically, for each word, we model a relevance probability distribution over the set of target variables plus the irrelevant case, because many words can be discarded for every target. Finally, we can control the level of interpretability by two regularizers that guide the model in producing long, meaningful rationales.
Compared to existing attention mechanisms, we derive a target importance distribution for each word instead of one over the entire sequence length.

Traditionally, interpretability came at the cost of reduced performance. In contrast, our evaluation shows that on two datasets, in beer and hotel review domains, with up to five correlated targets, our model outperforms strong attention and rationale baselines approaches and generates masks that are strong feature predictors and have a meaningful interpretation. We show that it can be a benefit to:\begin{enumerate}[topsep=0pt]
	\item guide the model to focus on different parts of the input text, 
	\item capture ambiguities of words belonging to multiple aspects, and 
	\item further improve the sentiment prediction for all the aspects.
\end{enumerate}
Therefore, interpretability does not come at a cost in our paradigm. 

\section{Related Work}
\subsection{Interpretability}
Developing interpretable models is of considerable interest to the broader research community; this is even more pronounced with neural models \citep{kim2015mind,doshi2017towards}. 
There has been much work with a multitude of approaches in the areas of analyzing and visualizing state activation \citep{KarpathyJL15,li2016visualizing,montavon2018methods}, attention weights \citep{jain2019attention,serrano-smith-2019-attention,pruthi-etal-2020-learning,clark-etal-2019-bert}, and learned sparse and interpretable word vectors \citep{faruqui2015sparse,faruqui-etal-2015-retrofitting,herbelot2015building}.
Other works interpret black box models by locally fitting interpretable models \citep{ribeiro2016should,NIPS2017_7062}.
\cite{li2016understanding} proposed erasing various parts of the input text using reinforcement learning to interpret the decisions. 
However, this line of research aims at providing post-hoc explanations of an already-trained model.
Our work differs from these approaches in terms of what is meant by an explanation and its computation. We defined an explanation as one or multiple text snippets that -- as a substitute of the input text -- are sufficient for the predictions.

\subsection{Multi-Aspect Sentiment Classification}
Multi-aspect sentiment classification is sometimes seen as a sub-problem of sentiment analysis. \cite{beer,pappas2014explaining} employed heuristic-based methods or topic models. Neural models achieve significant improvements with less feature engineering. \cite{yin2017document} built a hierarchical attention model with aspect representations by using a set of manually defined topics. \cite{li2018document} extended this chapter with user attention and additional features such as overall rating, aspect, and user embeddings. The disadvantage of these methods is their limited interpretability because they rely on many features in addition to the review text.

\subsection{Attention-based Models}
Attention models \citep{vaswani2017attention,yang2016hierarchical,LinFSYXZB17} have been shown to improve prediction accuracy, visualization, and interpretability. The most popular and widely used attention mechanism is soft attention~\citep{iclr2015}, rather than hard attention~\citep{luong-etal-2015-effective} or sparse ones~\citep{martins2016softmax}.
According to various studies \citep{jain2019attention,serrano-smith-2019-attention,pruthi-etal-2020-learning}, standard attention modules noisily predict input importance; the weights cannot provide safe and meaningful explanations.
Moreover, \cite{pruthi-etal-2020-learning} showed that standard attention modules can fool people into thinking that predictions from a model biased against gender minorities do not rely on the gender.
Our approach differs in two ways from attention mechanisms. First, the loss includes two regularizers to favor long word sequences for interpretability. Second, the normalization is not done over the sequence length but over the target set for each word; each has a relevance probability distribution over the set of target variables.

\subsection{Rationale Models}
The idea of using human rationales during training has been explored in \cite{zhang-etal-2016-rationale,bao-etal-2018-deriving,deyoung-etal-2020-eraser}. Although they have been shown to be beneficial, they are costly to collect and might vary across annotators. In our work, no annotation is~needed.

One of the first rationale generation methods was introduced by \cite{lei-etal-2016-rationalizing} in which a generator masks the input text fed to the classifier.
This framework is a cooperative game that selects rationales to accurately predict the label by maximizing the mutual information \citep{chen2018learning}.
\cite{yu-etal-2019-rethinking} proposed conditioning the generator based on the predicted label from a classifier reading the whole input, although it slightly underperformed when compared to the original model \citep{chang2020invariant}.
\cite{chang2019game} presented a variant that generated rationales to perform counterfactual reasoning. Finally, \cite{chang2020invariant} proposed a generator that can decrease spurious correlations in which the selective rationale consists of an extracted chunk of a pre-specified length, an easier variant than the original one that generated the rationale.
In all, these models are trained to generate a hard binary mask as a rationale to explain the prediction of a target variable, and the method requires as many models to train as variables to explain. Moreover, they rely on the assumption that the data have low internal correlations.

In contrast, our model addresses these drawbacks by jointly predicting the rationales of all the target variables (even in the case of highly correlated data) by generating a soft multi-dimensional mask. The probabilistic nature of the masks can handle ambiguities in the induced rationales. In Chapter~\ref{chapter_ijcai_2021}, we show~how~to use the induced rationales to generate personalized explanations for recommendation and how human users significantly prefer these over those produced by state-of-the-art models.

\section{MTM: A Multi-Target Masker}

\begin{figure}[!t]
\centering
\includegraphics[width=0.8\textwidth]{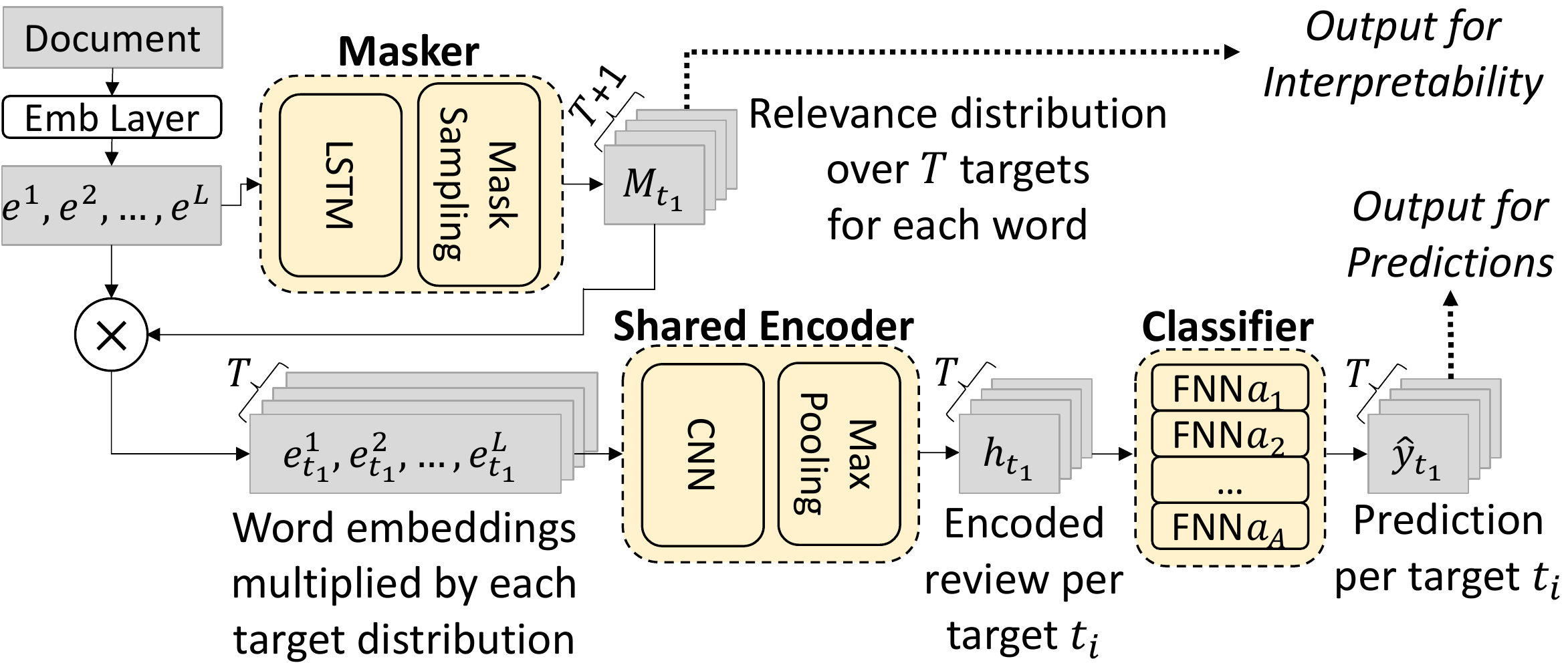}
\caption{\label{model_architecture}The proposed Multi-Target Masker (\textit{MTM}) model architecture to predict and explain $T$ target variables.}
\end{figure}

Let $X$ be a random variable representing a document composed of $L$ words ($x^1, x^2, ..., x^L$), and $Y$ the target $T$-dimensional vector\footnote{Our method is easily adapted for regression problems.}. Our proposed model, called the Multi-Target Masker (MTM), is composed of three components:~1)~a~\textbf{masker} module that computes a probability distribution over the target set for each word, resulting in $T+1$ masks (including one for the irrelevant case); 2)~an \textbf{encoder} that learns a representation of a document $X$ conditioned on the induced masks; 3)~a \textbf{classifier} that predicts the target variables. The overall model architecture is shown in Figure~\ref{model_architecture}. Each module is interchangeable with other models.
\subsection{Model Overview}

\subsubsection{Masker.} The masker first computes a hidden representation $h^\ell$ for each word $x^\ell$ in the input sequence, using their word embeddings $e^1, e^2, ..., e^L$. Many sequence models could realize~this task, such as recurrent, attention, or convolution networks. In our case, we chose a recurrent model to learn the dependencies between the words.
Let $t_i$ be the $i^{th}$ target for $i=1, ..., T$, and~$t_0$ the irrelevant case, because many words are irrelevant to every target. We define the multi-dimensional mask $\mathbf{M} \in \mathbb{R}^{(T+1) \times L}$ as the target relevance distribution $M^\ell \in \mathbb{R}^{(T+1)}$ of each word~$x^\ell$ as follows:
\begin{equation}
    P(\mathbf{M}|X) = \prod_{\ell=1}^L P(M^\ell|x^\ell) = \prod_{\ell=1}^L \prod_{i=0}^T P(m_{t_i}^\ell|x^\ell).
\end{equation}
Because we have categorical distributions, we cannot directly sample $P(M^\ell|x^\ell)$ and backpropagate the gradient through this discrete generation process. Instead, we model the variable $m^\ell_{t_i}$ using the straight through gumbel-softmax~\citep{JangGP17,MaddisonMT17} to approximate sampling from a categorical distribution\footnote{We also experimented with the implicit reparameterization trick using a Dirichlet distribution \citep{figurnov2018implicit} instead, but we did not obtain a significant improvement.}. We model the parameters of each Gumbel-Softmax distribution~$M^\ell$ with a single-layer feed-forward neural network followed by applying a log softmax, which induces the log-probabilities of the $\ell^{th}$  distribution: $\omega_\ell = \text{log}(\text{softmax}(W h^\ell + b))$. $W$ and $b$ are shared across all tokens so that the number of parameters stays constant with respect to the sequence length. We control the sharpness of the distributions with the~temperature~parameter~$\tau$, which dictates the peakiness of the relevance distributions. In our case, we keep the temperature low to enforce the assumption that each word is relevant about one or two targets. Note that compared to attention mechanisms, the word importance is a~probability distribution over~the~targets $\sum_{i=0}^T P(m_{t_i}^\ell|x^\ell) = 1$ instead of a normalization over the sequence length $\sum_{\ell=1}^L P(t^\ell | x^\ell) = 1$.

Given a soft multi-dimensional mask $\mathbf{M} \in \mathbb{R}^{(T+1) \times L}$ and the input $X \in \mathbb{R}^L$, we define each sub-mask $M_{t_i} \in \mathbb{R}^L$ as follows:
\begin{equation}
    M_{t_i} = P(m_{t_i}^1|x^1) , P(m_{t_i}^2|x^2) , ... , P(m_{t_i}^L|x^L).
\end{equation}
To integrate the word importance of the induced sub-masks $M_{t_i}$ within the model, we weight the word embeddings by their importance towards a target variable $t_i$, such that $E_{t_i} = E \odot M_{t_i} =$ $e_1 \cdot P(m^1_{t_i}|x^1) , e_2 \cdot P(m^2_{t_i}|x^2), ... , e_L \cdot P(m^L_{t_i}|x^L)$. Thereafter, each modified embedding $E_{t_i}$ is fed into the encoder block. Note that $E_{t_0}$ is ignored because $M_{t_0}$ only serves to absorb probabilities of words that are insignificant\footnote{if $P(m_{t_0}^\ell|x^\ell) \approx 1.0$, it implies $\sum_{i=1}^T P(m_{t_i}^\ell|x^\ell) \approx 0$ and consequently, $e_{t_i}^\ell \approx \vec{0}$ for  $i=0, ..., T$.}.

\subsubsection{Encoder and Classifier.} The encoder includes a convolutional network, followed by max-over-time pooling to obtain a fixed-length feature vector. We chose a convolutional network because it led to a smaller model, faster training, and performed empirically similarly to recurrent and attention models. It produces the fixed-size hidden representation $h_{t_i}$ for each target $t_i$. To exploit commonalities and differences among the targets, we share the weights of the encoder for all $E_{t_i}$. Finally, the classifier block contains for each target variable $t_i$ a two-layer feedforward neural network, followed by a softmax layer to predict the outcome~$\hat{y}_{t_i}$.

\subsubsection{Extracting Rationales.}
To explain the prediction $\hat{y}_{t_i}$ of one target $Y_{t_i}$, we generate its rationale by selecting each word $x^\ell$, whose relevance towards $t_i$ is the most likely: $P(m_{t_i}^\ell|x^\ell) = \max_{j=0, ..., T}{P(m_{t_j}^\ell|x^\ell)}$. Then, we can interpret $P(m_{t_i}^\ell|x^\ell)$ as the model confidence of $x^\ell$ relevant to~$Y_{t_i}$.

\subsection{Enabling the Interpretability of Masks}

The first objective to optimize is the prediction loss, represented as the cross-entropy between the true target label $y_{t_i}$ and the prediction $\hat{y}_{t_i}$ as follows:~\begin{equation}
    \ell_{pred} = \sum_{i=1}^{T} \ell_{cross\_entropy}(y_{t_i}, \hat{y}_{t_i}).
\end{equation}
However, training MTM to optimize $\ell_{pred}$ will lead to meaningless sub-masks $M_{t_i}$ because the~model tends to focus on certain words. Consequently, we guide the model to produce long, meaningful word sequences, as shown in Figure~\ref{sample_lambda}.
We propose two regularizers to control the number of selected words and encourage consecutive words to be relevant to the same target. For the first term, we calculate the~probability $p_{sel}$ of tagging a word as relevant to any target using the non-aspect mask $m_{t_0}^\ell$ for each token $x^\ell$ as follows:~\begin{equation}
        p_{sel} = \frac{1}{L} \sum_{\ell=1}^L \big( 1 - P(m_{t_0}^\ell|x^\ell) \big).
\end{equation}
We then compute the cross-entropy with a prior hyperparameter $\lambda_p$ to control the expected number of selected words among all target variables, which corresponds to the expectation of a binomial distribution $(p_{sel})$. We minimize the difference between $p_{sel}$ and~$\lambda_p$ as follows:~\begin{equation}
        \ell_{sel} = \ell_{binary\_cross\_entropy}(p_{sel}, \lambda_p).
\end{equation}
The second regularizer discourages the target transition of two consecutive words by minimizing the mean variation of their target distributions, $M^\ell$ and $M^{\ell-1}$. We generalize the formulation of a hard binary selection as suggested by \cite{lei-etal-2016-rationalizing} to a soft probabilistic multi-target selection as follows\footnote{Early experiments with other distance functions, such as the Kullback–Leibler divergence, produced inferior results.}:~\begin{equation}
    \begin{split}
        p_{dis} &= \frac{1}{L} \sum_{\ell=1}^L \frac{\big|\big| M^\ell - M^{\ell-1} \big|\big|_1 }{A + 1},\\
        \ell_{cont} &= \ell_{binary\_cross\_entropy}(p_{dis}, 0).
    \end{split}
\end{equation}
We train our Multi-Target Masker end-to-end and optimize the loss $\ell_{MTM} = \ell_{pred} + \lambda_{sel} \cdot \ell_{sel} + \lambda_{cont} \cdot \ell_{cont}$, where $\lambda_{sel}$ and $\lambda_{cont}$ control the impact of each~constraint.

\section{Experiments}

We assess our model in two dimensions: the quality of the explanations and the predictive performance. Following previous work \citep{lei-etal-2016-rationalizing,chang2020invariant}, we use sentiment analysis as a demonstration use case, but we extend it to the multi-aspect case. However, we are interested in learning rationales for every aspect at the same time without any prior assumption on the data, where aspect ratings can be highly~correlated.
We first measure the quality of the induced rationales using human aspect sentence-level annotations and an automatic topic model evaluation method. In the second set of experiments, we evaluate MTM on the multi-aspect sentiment classification task in two different domains.
\subsection{Experimental Details}
\label{experimental_details}

The review encoder was either a bi-directional recurrent neural network using LSTM~\citep{hochreiter1997long} with $50$ hidden units or a multi-channel text convolutional neural network, similar to~\cite{kim2015mind}, with $3$-, $5$-, and $7$-width filters and $50$ feature maps per filter. Each aspect classifier is a two-layer feedforward neural network with a rectified linear unit activation function~\citep{nair2010rectified}. We used the $200$-dimensional pretrained word embeddings of~\cite{lei-etal-2016-rationalizing} for beer reviews. For the hotel domain, we trained word2vec~\citep{mikolov2013distributed} on a large collection of hotel reviews~\citep{antognini-faltings-2020-hotelrec} with an embedding size~of~$300$.
We used a dropout~\citep{srivastava2014dropout} of~$0.1$, clipped the gradient norm at $1.0$, added a L2-norm regularizer with a factor of~$10^{-6}$, and trained using early stopping with a patience of three iterations. 
We used~Adam \citep{KingmaB14} with a learning rate of $0.001$. The~temperature $\tau$ for the Gumbel-Softmax distributions was fixed at $0.8$. The two regularizers and the prior of our model were~$\lambda_{sel} = 0.03$, $\lambda_{cont} = 0.03$, and $\lambda_p = 0.15$ for the \textit{Beer} dataset and $\lambda_{sel} = 0.02$, $\lambda_{cont} = 0.02$, and $\lambda_p = 0.10$ for the \textit{Hotel}~one. We ran all experiments for a maximum of $50$ epochs with a batch-size of $256$. We tuned all models on the dev set with 10 random search trials. For \cite{lei-etal-2016-rationalizing}, we used the code from the~authors. For reproducibility purposes, we include additional details in Appendix~\ref{app:additional_training}.

\begin{table}[!t]
\caption{\label{app_dataset_description_main}Statistics of the multi-aspect review datasets. Both datasets have high correlations between aspects.}
\centering
\begin{tabular}{@{}lcc@{}}
\multicolumn{1}{c}{\bf Dataset} & \multicolumn{1}{c}{\bf Beer}  & \multicolumn{1}{c}{\bf Hotel}\\
\toprule
Number of reviews& $1,586,259$ & $140,000$\\
Average words per review& $147.1 \pm 79.7$  & $188.3 \pm 50.0$\\
Average sentences per review& $10.3 \pm 5.4$  & $10.4 \pm 4.4$\\
Number of Aspects & $4$ & $5$ \\
Avg./Max corr. between aspects & $71.8\% / 73.4\%$ & $63.0\% / 86.5\%$\\\end{tabular}
\end{table}

\subsection{Datasets}
\label{sec_datasets_aaai}

\cite{beer} provided $1.5$~million English beer reviews from BeerAdvocate. Each contains multiple sentences describing various beer aspects: \textit{Appearance}, \textit{Smell}, \textit{Palate}, and \textit{Taste}; users also provided a five-star rating for each aspect. 
To evaluate the robustness of the models across domains, we sampled $140,000$ hotel reviews from~\cite{antognini-faltings-2020-hotelrec}, that contains $50$~million reviews from TripAdvisor. Each review contains a five-star rating for each aspect: \textit{Service}, \textit{Cleanliness}, \textit{Value}, \textit{Location}, and \textit{Room}. The descriptive statistics are shown in Table~\ref{app_dataset_description_main}.

There are high correlations among the rating scores of different aspects in the same review ($71.8\%$ and $63.0\%$ on average for the beer and hotel datasets, respectively). This makes it difficult to directly learn textual justifications for single-target rationale generation models \citep{chang2020invariant,chang2019game,lei-etal-2016-rationalizing}. Prior work used~separate decorrelated train sets for each aspect and excluded~aspects with a high correlation, such as \textit{Taste}, \textit{Room}, and \textit{Value}. However, these assumptions do not reflect the real data distribution. Therefore, we keep the original data (and thus can show that our model does not suffer from the high correlations). We binarize the problem as in previous work \citep{bao-etal-2018-deriving,chang2020invariant}: ratings at three and above are labeled as positive and the rest as negative. We split the data into $80/10/10$ for the train, validation, and test sets.
Compared to the beer reviews, the hotel ones were longer, noisier, and less structured, as shown in Appendices~\ref{samples_with_visualization}. Both datasets do not contain annotated rationales.

\subsection{Baselines}
\label{sec_baselines}

We compare our Multi-Target Masker~(\textit{MTM}) with various baselines. We group them in three levels of interpretability:
\begin{itemize}
    \item \textit{None}. We cannot extract the input features the model used to make the predictions;
    \item \textit{Coarse-grained}. We can observe what parts of the input a model used to discriminate all aspect sentiments without knowing what part corresponded to what aspect;
    \item \textit{Fine-grained}. For each aspect, a model selects input features to make the prediction.
\end{itemize}

We first use a simple baseline, \textit{SENT}, that reports the majority sentiment across the aspects, as the aspect ratings are highly correlated. Because this information is not available at testing, we trained a model to predict the majority sentiment of a review as suggested by \cite{wang-manning-2012-baselines}. The second baseline we used is a shared encoder followed by $T$~classifiers that we denote \textit{BASE}. These models do not offer any interpretability. We extend it with~a shared attention mechanism~\citep{iclr2015} after the encoder, noted as \textit{SAA} in our study, that provides a coarse-grained interpretability; for all aspects, \textit{SAA} focuses on the same words in the~input.

Our final goal is to achieve the best performance and provide fine-grained interpretability in order to visualize what sequences of words a model focuses on to predict the aspect sentiments.
To this end, we include other baselines: two trained \textit{separately} for each aspect (e.g., current rationale models) and two trained with a \textit{multi-aspect} sentiment loss. 
For the first ones, we employ the well-known \textit{NB-SVM} \citep{wang-manning-2012-baselines} for sentiment analysis tasks, and we then use the Single-Aspect Masker (\textit{SAM}) \citep{lei-etal-2016-rationalizing}, each trained separately for each aspect.

\begin{figure}[t]
\centering
\begin{subfigure}[t]{\textwidth}
    \centering
\includegraphics[width=\textwidth]{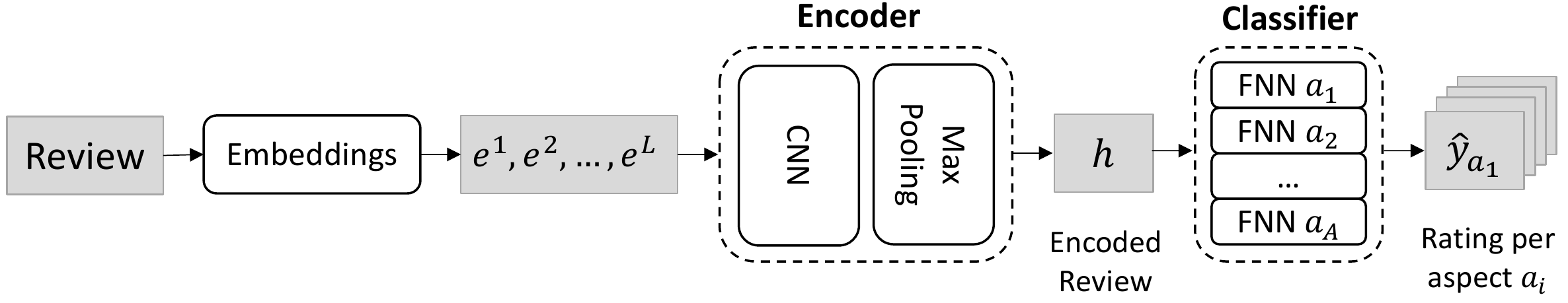} 
    \caption{Baseline model $\text{Emb}$ + $\text{Enc}_\textsubscript{CNN}$ + \text{Clf} (\textit{BASE}).}
\end{subfigure}
\\
\begin{subfigure}[t]{\textwidth}
    \centering
\includegraphics[width=\textwidth]{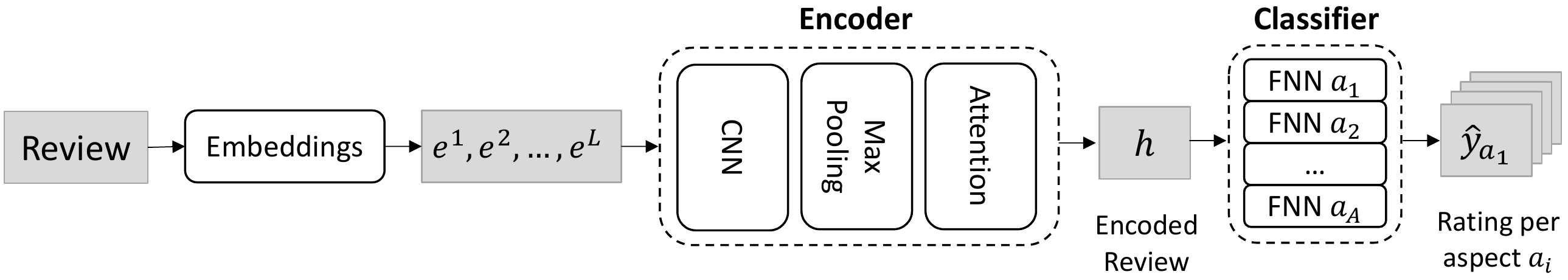}
\caption{Baseline model $\text{Emb}$ + $\text{Enc}_\textsubscript{CNN}$ + $\text{A}_\textsubscript{Shared}$ + \text{Clf} (\textit{SAA}, CNN variant).}
\end{subfigure}
\\
\begin{subfigure}[t]{\textwidth}
    \centering
\includegraphics[width=\textwidth]{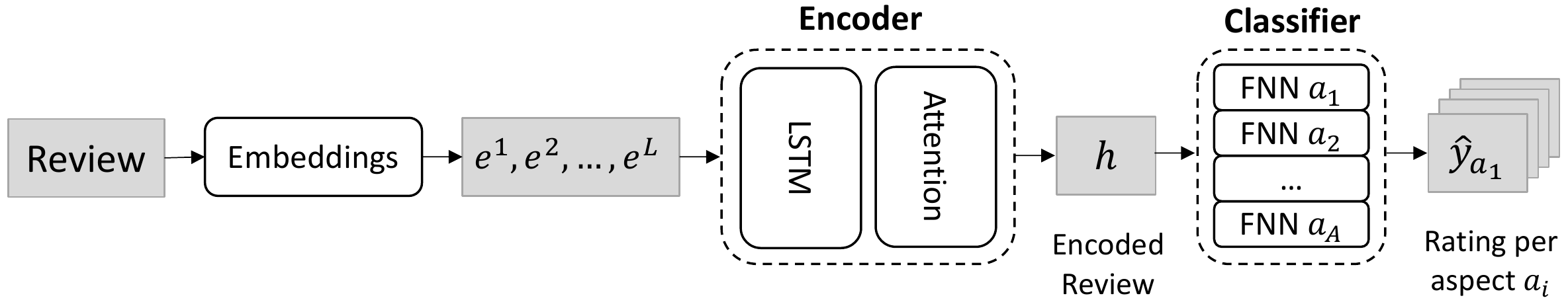}
\caption{Baseline model $\text{Emb}$ + $\text{Enc}_\textsubscript{LSTM}$ + $\text{A}_\textsubscript{Shared}$ + \text{Clf} (\textit{SAA}, LSTM variant).}
\end{subfigure}
\\
\begin{subfigure}[t]{\textwidth}
    \centering
\includegraphics[width=\textwidth]{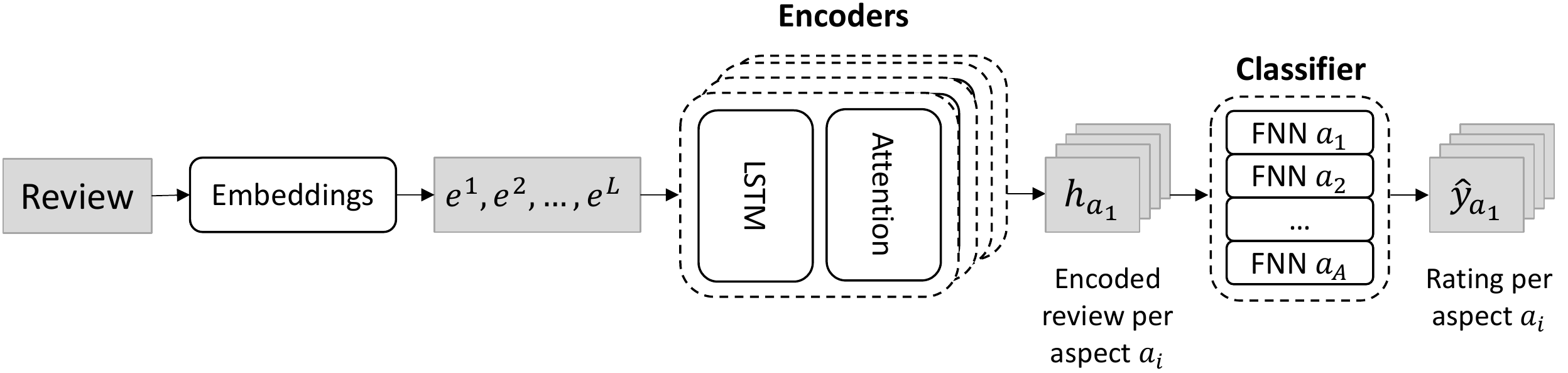} 
\caption{Baselines $\text{Emb}$ + $\text{Enc}_\textsubscript{LSTM}$ + $\text{A}^\textsubscript{[Sparse]}_\textsubscript{Aspect-wise}$ + \text{Clf}. Attention is either additive (\textit{MAA}) or sparse (\textit{MASA}).}
\end{subfigure}
\caption{\label{baseline_architecture}Baseline architectures.}
\end{figure}

The two last methods contain a separate encoder, attention mechanism, and classifier~for each aspect. We utilize two types of attention mechanisms, additive~\citep{iclr2015} and sparse~\citep{martins2016softmax}, as sparsity in the attention has been shown to induce useful, interpretable representations. We call them Multi-Aspect Attentions~(\textit{MAA}) and Sparse-Attentions~(\textit{MASA}), respectively. Diagrams of the baselines can be found in Figure~\ref{baseline_architecture}.

We demonstrate that the induced sub-masks $M_{t_1}, ..., M_{t_T}$ computed from \textit{MTM} bring fine-grained interpretability and are meaningful for other models to improve performance. To do so, we extract and concatenate the masks to the word embeddings, resulting in contextualized embeddings~\citep{peters-etal-2018-deep}, and train \textit{BASE} with those. We call this variant \textit{MTM\textsuperscript{C}}, that is smaller and has faster inference than~\textit{MTM}.

\section{Results}

\subsection{Multi-Rationale Interpretability}

We first verify whether the inferred rationales of \textit{MTM} are meaningful and interpretable, compared to the other models.

\subsubsection{Precision.}
\label{subsub_prec}
Evaluating explanations that consist of coherent pieces of text is challenging because there is no gold standard for reviews. \cite{beer} have provided $994$ beer reviews with sentence-level aspect annotations (although our model computes masks at a finer level). Each sentence was annotated with one aspect label, indicating what aspect that sentence covered. We evaluate the precision of the words selected by~each model, as in \cite{lei-etal-2016-rationalizing}.
We use trained models on the \textit{Beer} dataset and extracted a similar number of selected words for a fair comparison. We also report the results of the models from \cite{lei-etal-2016-rationalizing}: \textit{NB-SVM}, the Single-Aspect Attention and Masker (\textit{SAA} and \textit{SAM}, respectively); they use the separate decorrelated train sets for each aspect because they compute hard masks\footnote{When trained on the original data, they performed significantly worse, showing the limitation in handling correlated variables.}.

\begin{table}[t]
\caption{\label{perfs_precision_beer}Performance related to human evaluation, showing the precision of the selected words for each aspect of the \textit{Beer} dataset. The percentage indicates the number~of~selected words of the reviews. MTM infers rationales that better correlate with human judgements.}
\centering
\begin{threeparttable}
    \centering
\begin{tabular}{@{}lccc@{}}
 & \multicolumn{3}{c}{\textbf{Precision / \% Highlighted Words}}\\
\cmidrule(lr){2-4}
\bf Model & \textit{Smell} & \textit{Palate} & \textit{Appearance}\\
\toprule
NB-SVM\tnote{*} & $21.6$ / $7\%$ & $24.9$ / $7\%$ & $38.3$ / $13\%$\\
SAA\tnote{*} & $88.4$ / $7\%$ & $65.3$ / $7\%$ & $80.6$ / $13\%$\\
SAM\tnote{*} & $95.1$ / $7\%$ & $80.2$ / $7\%$ & $96.3$ / $14\%$\\
MASA & $87.0$ / $4\%$ & $42.8$ / $5\%$ & $74.5$ / $\ \ 4\%$\\
MAA & $51.3$ / $7\%$ & $32.9$ / $7\%$ & $44.9$ / $14\%$\\
\textbf{MTM (Ours)} & $\mathbf{96.6}$ / $7\%$ & $\mathbf{81.7}$ / $7\%$ & $\mathbf{96.7}$ / $14\%$\\
\end{tabular}
\begin{tablenotes}
     \item[*] Model trained separately for each aspect.
   \end{tablenotes}
\end{threeparttable}
\end{table}

Table~\ref{perfs_precision_beer} presents the precision of the masks and attentions computed on the sentence-level aspect annotations. We show that the generated sub-masks obtained with our Multi-Target Masker (\textit{MTM}) correlates best with the human judgment. In comparison to \textit{SAM}, the \textit{MTM} model obtains significantly~higher precision with an average of $+1.13$. Interestingly, \textit{NB-SVM} and attention models (\textit{SAA}, \textit{MASA}, and \textit{MAA}) perform  poorly compared with the mask models, especially \textit{MASA}, which focuses only on a couple of words due to the sparseness of the attention. In Section~\ref{app:impact_length}, we also analyze the impact of the length of the explanations.

\subsubsection{Semantic Coherence.}
\label{sec_mask_coh}

In addition to evaluating the rationales with human annotations, we compute their semantic interpretability. According to \cite{aletras2013evaluating,lau2014machine}, normalized point mutual information (NPMI) is a good metric for the qualitative evaluation of topics because it matches human judgment most closely. However, the top-$N$ topic words used for evaluation are often selected arbitrarily. To alleviate this problem, we followed \cite{lau2016sensitivity}. We compute the topic coherence over several cardinalities and report the results and average (see Appendix~\ref{sec_topic_words}); those authors claimed that the mean leads to a more stable and robust evaluation.

\begin{table}[t]
\centering
\caption{\label{topics_perfs}Performance on automatic evaluation, showing the average topic coherence (NPMI) across different top-$N$ words for each dataset. We treat each aspect $a_i$ as a topic and use the masks/attentions to compute~$P(w|a_i)$. MTM produces more coherent~rationales.}
\begin{threeparttable}[t]
    \centering
\begin{tabular}{
@{}lccccccc@{}}
& \multicolumn{7}{c}{\textbf{NPMI}}\\
\cmidrule(lr){2-8}
\textbf{Model} & $N=5$ & $10$ & $15$ & $20$ & $25$ & $30$ & \ Mean\tnote{\textsuperscript{\textdagger}}\\
\toprule
\multicolumn{8}{c}{\textit{Beer}} \\
SAM\tnote{*} & $0.046$ & $0.120$ & $0.129$ & $0.243$ & $0.308$ & $0.396$ & \ $0.207$ \\
MASA & $0.020$ & $0.082$ & $0.130$ & $0.168$ & $0.234$ & $0.263$ & \ $0.150$\\
MAA & $0.064$ & $\mathbf{0.189}$ & $0.255$ & $0.273$ & $0.332$ & $0.401$ & \ $0.252$ \\
\textbf{MTM} & $\mathbf{0.083}$ & $0.187$ & $\mathbf{0.264}$ & $\mathbf{0.348}$ & $\mathbf{0.477}$ & $\mathbf{0.410}$ & \ $\mathbf{0.295}$ \\
\midrule
\multicolumn{8}{c}{\textit{Hotel}} \\
SAM\tnote{*} & $0.041$ & $0.103$ & $0.152$ & $0.180$ & $0.233$ & $0.281$ & \ $0.165$ \\
MASA & $0.043$ & $0.127$ & $0.166$ & $0.295$ & $0.323$ & $0.458$ & \ $0.235$\\
MAA & $0.128$ & $0.218$ & $\mathbf{0.352}$ & $0.415$ & $0.494$ & $0.553$ & \ $0.360$\\
\textbf{MTM} & $\mathbf{0.134}$ & $\mathbf{0.251}$ & $0.349$ & $\mathbf{0.496}$ & $\mathbf{0.641}$ & $\mathbf{0.724}$ & \ $\mathbf{0.432}$\\
\end{tabular}
\begin{tablenotes}
     \item[*] Model trained separately for each aspect.
     \item[\textsuperscript{\textdagger}] The metric that correlates best with human judgment \citep{lau2016sensitivity}.
   \end{tablenotes}
\end{threeparttable}
\begin{tikzpicture}[overlay]

\node [coordinate, shift={(0.2cm, 2.65cm)},inner sep=0pt](c1){};
\node [coordinate,shift={(0.2cm, -3.15cm)},inner sep=0pt](c2){};
\node [coordinate,shift={(-1.25cm, -3.15cm)},inner sep=0pt](c3){};
\node [coordinate,shift={(-1.25cm, 2.65cm)},inner sep=0pt](c4){};
\draw [fill=gray, fill opacity=0.1, draw=gray,rounded corners = 1ex] (c1) -- (c2) -- (c3) -- (c4) -- cycle; 

\end{tikzpicture}
\end{table}

The results are shown in~Table~\ref{topics_perfs}. We show that the computed masks by \textit{MTM} lead to the highest mean NPMI and, on average, $20\%$  superior results in both datasets, while only needing a single training. Our \textit{MTM} model significantly outperforms \textit{SAM} and the attention models (\textit{MASA} and \textit{MAA}) for $N \ge 20$ and $N=5$. For $N=10$ and $N=15$, \textit{MTM} obtains~higher scores in two out of four cases ($+.033$ and $+.009$). For the other two, the difference was below $.003$. SAM obtains poor results in all cases and must be trained as many times as the number of~aspects.

We analyzed the top words for each aspect by conducting a human evaluation to identify intruder words (i.e., words not matching the corresponding aspect). Generally, our model found better topic words: approximately $1.9$ times fewer intruders than other methods for each aspect and each dataset. More details are available in Appendix~\ref{sec_topic_words}.

\subsection{Multi-Aspect Sentiment Classification}
\label{masc}

\begin{table}[t]
\caption{\label{perfs_full_beer}Performance of the multi-aspect sentiment classification task for the \textit{Beer} (top) and \textit{Hotel} (bottom) datasets. MTM obtains the best F1 scores and requires fewer parameters.}
\centering
\hspace*{-1.5cm}
\begin{tabular}
{@{}cl>{}ll@{}>{}c@{}>{}c@{}>{}c@{\hspace*{2mm}}>{}c@{\hspace*{2mm}}>{}c@{\hspace*{2mm}}>{}c@{\hspace*{2mm}}>{}c@{\hspace*{2mm}}>{}c@{}}
& & & & & \multicolumn{5}{c}{\textbf{F1 Scores}}\\
\cmidrule(lr){6-10}
\multirow{13}{*}{\rotatebox[origin=c]{90}{\centering \textit{Beer Reviews}}} &  \multicolumn{1}{c}{\bf Interp.} & \multicolumn{2}{c}{\bf Model} & \multicolumn{1}{c}{\bf Params} & \multicolumn{1}{c}{\bf Macro} & \multicolumn{1}{c}{\boldmath$A_1$} & \multicolumn{1}{c}{\boldmath$A_2$} & \multicolumn{1}{c}{\boldmath$A_3$} & \multicolumn{1}{c}{\boldmath$A_4$}\\
\cmidrule[0.08em]{2-10}
& \multicolumn{1}{c}{\multirow{2}{*}{\parbox{1.cm}{\centering None}}}
& SENT & Sentiment Majority & $560k$ & $73.01$ & $71.83$ & $75.65$ & $71.26$ & $73.31$\\
& & BASE & $\text{Emb}_{200}$ + $\text{Enc}_\textsubscript{CNN}$ + \text{Clf} & $188k$ & $76.45$ & $71.44$ & $78.64$ & $74.88$ & $80.83$\\
\cmidrule{3-10}
& \multicolumn{1}{c}{\multirow{2}{*}{\parbox{1.cm}{\centering Coarse-grained}}}
& \multirow{2}{*}{SAA} & $\text{Emb}_{200}$ + $\text{Enc}_\textsubscript{CNN}$ + $\text{A}_\textsubscript{Shared}$ + \text{Clf} & $226k$ & $77.06$ & $73.44$ & $78.68$ & $75.79$ & $80.32$\\
& & & $\text{Emb}_{200}$ + $\text{Enc}_\textsubscript{LSTM}$ + $\text{A}_\textsubscript{Shared}$ + \text{Clf} & $219k$ & $78.03$ & $74.25$ & $79.53$ & $75.76$ & $82.57$\\
\cmidrule{3-10}
& \multicolumn{1}{c}{\multirow{6}{*}{\parbox{1.cm}{\centering Fine-grained}}}
& NB-SVM & \citep{wang-manning-2012-baselines} & $4 \cdot 560k$ & $72.11$ & $72.03$ & $74.95$ & $68.11$ & $73.35$\\
& & SAM & \citep{lei-etal-2016-rationalizing} & $4 \cdot 644k$ & $76.62$ & $72.93$ & $77.94$ & $75.70$ & $79.91$\\
& & MASA & $\text{Emb}_{200}$ + $\text{Enc}_\textsubscript{LSTM}$ + $\text{A}^\textsubscript{Sparse}_\textsubscript{Aspect-wise}$ + \text{Clf} & $611k$ & $77.62$ & $72.75$ & $79.62$ & $75.81$ & $82.28$\\
& & MAA & $\text{Emb}_{200}$ + $\text{Enc}_\textsubscript{LSTM}$ + $\text{A}_\textsubscript{Aspect-wise}$ + \text{Clf} & $611k$ & $78.50$ & $74.58$ & $79.84$ & $77.06$ & $82.53$\\
\cmidrule{3-10}
& & MTM\textsuperscript{} & $\text{Emb}_{200}$ + \text{Masker} + $\text{Enc}_\textsubscript{CNN}$ + \text{Clf} (Ours) & $289k$ & $78.55$ & $74.87$ & $79.93$ & $77.39$ & $82.02$\\
& & \textbf{MTM\textsuperscript{C}} & \textbf{$\text{Emb}_{200+4}$ +  $\text{Enc}_\textsubscript{CNN}$ + \text{Clf} (Ours)} & $191k$ & $\mathbf{78.94}$ & $\mathbf{75.02}$ & $\mathbf{80.17}$ & $\mathbf{77.86}$ & $\mathbf{82.71}$\\
\end{tabular}
\hspace*{-2.25cm}
\begin{tabular}
{@{}cl>{}ll@{}>{}c@{}>{}c@{}>{}c@{\hspace*{2mm}}>{}c@{\hspace*{2mm}}>{}c@{\hspace*{2mm}}>{}c@{\hspace*{2mm}}>{}c@{}}
 & & & & & \multicolumn{6}{c}{\textbf{F1 Scores}}\\
\cmidrule(lr){6-11}
\multirow{14}{*}{\rotatebox[origin=c]{90}{\centering \textit{Hotel Reviews}}} & \multicolumn{1}{c}{\bf Interp.} & \multicolumn{2}{c}{\bf Model} & \multicolumn{1}{c}{\bf Params} & \multicolumn{1}{c}{\bf Macro} & \multicolumn{1}{c}{\boldmath$A_1$} & \multicolumn{1}{c}{\boldmath$A_2$} & \multicolumn{1}{c}{\boldmath$A_3$} & \multicolumn{1}{c}{\boldmath$A_4$} & \multicolumn{1}{c}{\boldmath$A_5$}\\
\cmidrule[0.08em]{2-11}
& \multicolumn{1}{c}{\multirow{2}{*}{\parbox{1.cm}{\centering None}}}
& SENT & Sentiment Majority& $309k$ & $85.91$ & $89.98$ & $90.70$ & $92.12$ & $65.09$ & $91.67$\\
& & BASE & $\text{Emb}_{300}$ + $\text{Enc}_\textsubscript{CNN}$ + \text{Clf} & $263k$ & $90.30$ & $92.91$ & $93.55$ & $94.12$ & $76.65$ & $94.29$\\
\cmidrule(lr){3-11}
& \multicolumn{1}{c}{\multirow{2}{*}{\parbox{1.cm}{\centering Coarse-grained}}}
& \multirow{2}{*}{SAA} & $\text{Emb}_{300}$ + $\text{Enc}_\textsubscript{CNN}$ + $\text{A}_\textsubscript{Shared}$ + \text{Clf} & $301k$ & $90.12$ & $92.73$ & $93.55$ & $93.76$ & $76.40$ & $94.17$\\
& & & $\text{Emb}_{300}$ + $\text{Enc}_\textsubscript{LSTM}$ + $\text{A}_\textsubscript{Shared}$ + \text{Clf} & $270k$ & $88.22$ & $91.13$ & $92.19$ & $93.33$ & $71.40$ & $93.06$\\
\cmidrule(lr){3-11}
& \multicolumn{1}{c}{\multirow{6}{*}{\parbox{1.cm}{\centering Fine-grained}}}

& NB-SVM & \citep{wang-manning-2012-baselines} & $5 \cdot 309k$ & $87.17$ & $90.04$ & $90.77$ & $92.30$ & $71.27$ & $91.46$\\

& & SAM & \citep{lei-etal-2016-rationalizing} & $5\cdot824k$ & $87.52$ & $91.48$ & $91.45$ & $92.04$ & $70.80$ & $91.85$\\
& & MASA & $\text{Emb}_{200}$ + $\text{Enc}_\textsubscript{LSTM}$ + $\text{A}^\textsubscript{Sparse}_\textsubscript{Aspect-wise}$ + \text{Clf} & $1010k$ & $90.23$ & $93.11$ &    $93.32$ & $93.58$ & $77.21$ & $93.92$ \\
& & MAA & $\text{Emb}_{300}$ + $\text{Enc}_\textsubscript{LSTM}$ + $\text{A}_\textsubscript{Aspect-wise}$ + \text{Clf} & $1010k$ & $90.21$ & $92.84$ & $93.34$ & $93.78$ & $76.87$ & $94.21$\\
\cdashlinelr{3-11}
& & \multirow{1}{*}{MTM} & $\text{Emb}_{300}$ + \text{Masker} + $\text{Enc}_\textsubscript{CNN}$ + \text{Clf} (Ours)& $404k$ & $89.94$ & $92.84$ & $92.95$ & $93.91$ & $76.27$ & $93.71$\\
& & \textbf{MTM\textsuperscript{C}} & \textbf{$\text{Emb}_{300+5}$ + $\text{Enc}_\textsubscript{CNN}$ + \text{Clf} (Ours)} & $267k$ & $\mathbf{90.79}$ & $\mathbf{93.38}$ & $\mathbf{93.82}$ & $\mathbf{94.55}$ & $\mathbf{77.47}$ & $\mathbf{94.71}$ \\
\end{tabular}
\end{table}

We showed that the inferred rationales of \textit{MTM} were significantly more accurate and semantically coherent than those produced by the other models. Now, we inquire as to whether the masks could become a benefit rather than a cost in performance for the multi-aspect sentiment classification.

\subsubsection{Beer Reviews.}
\label{sec_beer_reviews}

We report the macro F1 and individual score for each aspect $A_i$. Table~\ref{perfs_full_beer} (top) presents the results for the \textit{Beer} dataset. The Multi-Target Masker (\textit{MTM}) performs better on average than all the baselines and provided fine-grained interpretability. Moreover, \textit{MTM} has two times fewer parameters than the aspect-wise attention models.

The contextualized variant \textit{MTM\textsuperscript{C}} achieves a macro F1 score absolute improvement of $0.44$~and $2.49$ compared~to \textit{MTM} and \textit{BASE}, respectively. These results highlight that the inferred masks are meaningful to improve the performance while bringing fine-grained interpretability to \textit{BASE}. It is $1.5$ times smaller than \textit{MTM} and has a faster inference.

\textit{NB-SVM}, which offers fine-grained interpretability and was trained separately for each aspect, significantly underperforms when compared to \textit{BASE} and, surprisingly, to \textit{SENT}. As shown in Table~\ref{app_dataset_description_main}, the sentiment correlation between any pair of aspects of the \textit{Beer} dataset is on average $71.8\%$. Therefore, by predicting the sentiment of one~aspect correctly, it is likely that other aspects share the same~polarity. We suspect that the linear model \textit{NB-SVM} cannot~capture the correlated relationships between aspects, unlike the non-linear (neural) models that have a higher capacity. The shared attention models perform better than \textit{BASE} but provide only coarse-grained interpretability. \textit{SAM} is outperformed by all the models except \textit{SENT}, \textit{BASE}, and \textit{NB-SVM}. 

\begin{figure}[!t]
\centering

\begin{tabular}{@{}l@{}}
	\includegraphics[width=0.8\textwidth]{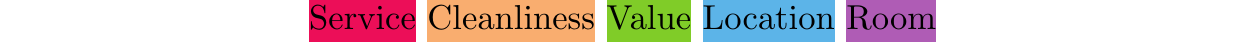} \\
	 \multicolumn{1}{c}{\underline{Multi-Target Masker (Ours)}}\\
     \includegraphics[width=0.8\textwidth]{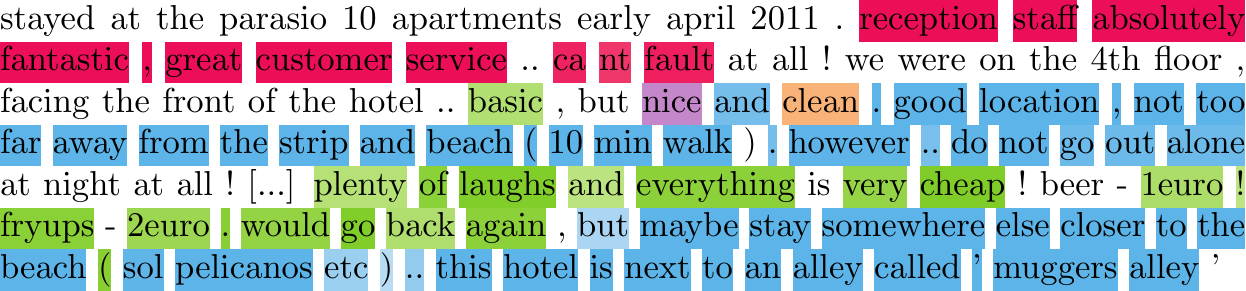} \\
     \multicolumn{1}{c}{\underline{Single-Aspect Masker}}\\
     \includegraphics[width=0.8\textwidth]{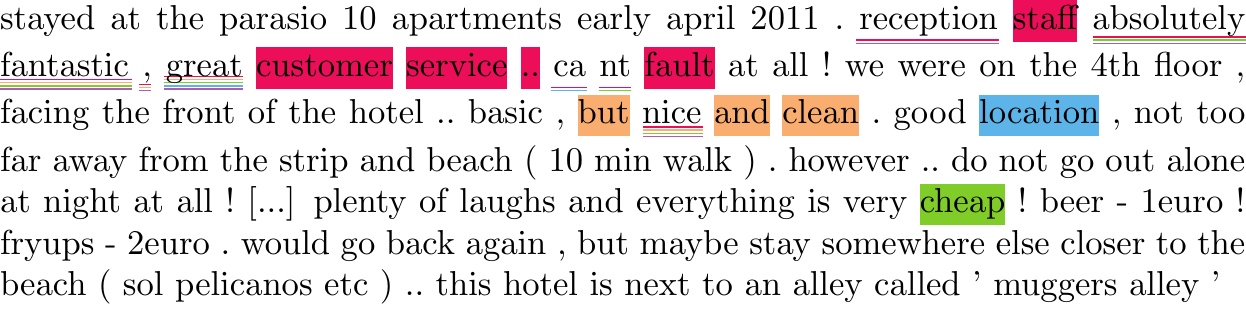} \\
     \multicolumn{1}{c}{\underline{Multi-Aspect Attentions}}\\
     \includegraphics[width=0.8\textwidth]{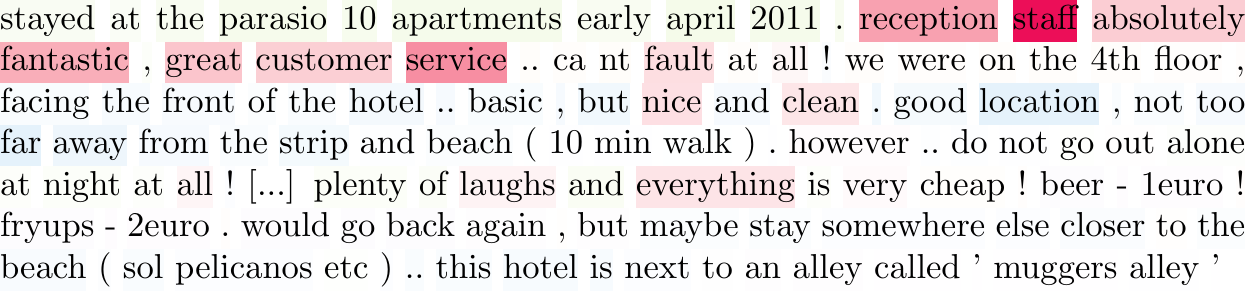} \\
     \multicolumn{1}{c}{\underline{Multi-Aspect Sparse-Attentions}}\\
     \includegraphics[width=0.8\textwidth]{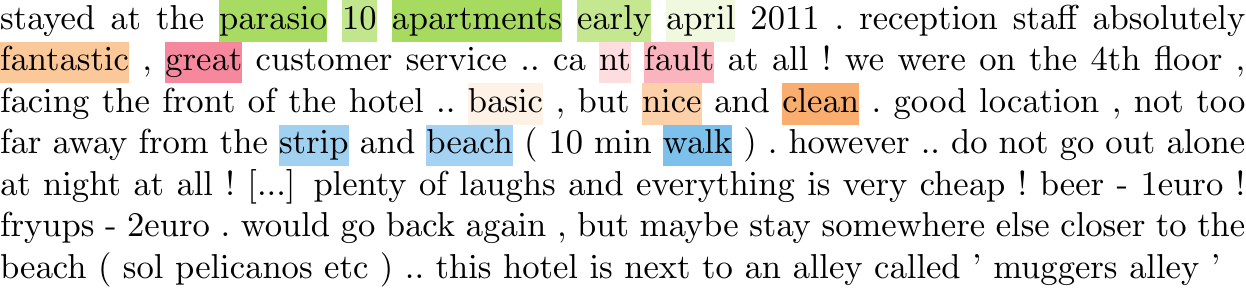}\\
\end{tabular}

\caption{\label{sample_new_hotel}Induced rationales on a truncated hotel review, where shade colors represent the model confidence towards the aspects. \textit{MTM} finds most of the crucial spans of words with a small amount of noise. \textit{SAM} lacks coverage but identifies words where half are correct and the others ambiguous (represented with colored underlines).}
\end{figure}

\subsubsection{Model Robustness - Hotel Reviews.}
We check the robustness of our model on another domain. Table~\ref{perfs_full_beer} (bottom) presents the results of the \textit{Hotel} dataset. The contextualized variant \textit{MTM\textsuperscript{C}}~outperforms all other models significantly with a macro F1 score improvement of $0.49$. Moreover, it achieves the best individual F1 score for each aspect~$A_i$. This shows that the learned mask $\mathbf{M}$ of \textit{MTM} is again meaningful because it increases the performance and adds interpretability to \textit{BASE}. Regarding \textit{MTM}, we see that it performs slightly worse than the aspect-wise attention models  \textit{MASA} and \textit{MAA} but has $2.5$ times fewer~parameters.

A visualization of a truncated hotel review with the extracted rationales and attentions is available in Figure~\ref{sample_new_hotel}. Not only do probabilistic masks enable higher performance, they better capture parts of reviews related to each aspect compared to other methods. More samples of beer and hotel reviews can be found in Appendix~\ref{full_beer_sample}.

To summarize, we have shown that the regularizers in \textit{MTM} guide the model to produce high-quality masks as explanations while performing slightly better than the strong attention models in terms of prediction performance. However, we demonstrated that including the inferred masks into word embeddings and training a simpler model achieved the best performance across two datasets and at the same time, brought fine-grained interpretability. Finally, \textit{MTM} supported high correlation among multiple target variables.

\subsubsection{Hard Mask versus Soft Masks.}
\label{sec_decorr}

\textit{SAM} is the neural model that obtained the lowest relative macro F1 score in the two datasets compared with \textit{MTM\textsuperscript{C}}: a difference of $-2.32$ and $-3.27$ for the \textit{Beer} and \textit{Hotel} datasets, respectively. Both datasets have a high average correlation between the aspect ratings: $71.8\%$ and $63.0\%$, respectively (see Table~\ref{app_dataset_description_main}).
Therefore, it makes it challenging for rationale models to learn the justifications of the aspect ratings directly. Following the observations of \cite{lei-etal-2016-rationalizing,chang2019game,chang2020invariant}, this highlights that single-target rationale models suffer from high correlations and require data to satisfy certain constraints, such as low correlations. In contrast, \textit{MTM} does not require any particular assumption on the data.

We compare \textit{MTM} in a setting where the aspect ratings were less correlated, although it does not reflect the real distribution of the aspect ratings. We employ the decorrelated subsets of the \textit{Beer} reviews from \cite{lei-etal-2016-rationalizing,chang2020invariant}. As shown in Table~\ref{app_dataset_description}, it has an average correlation of $27.2\%$ and the aspect \textit{Taste} is removed. 

\begin{table}[!t]
\centering
\caption{\label{app_dataset_description}Statistics of the multi-aspect review datasets. \textit{Beer} and \textit{Hotel} represent real-world beer and hotel reviews, respectively. \textit{Decorrelated Beer} is a subset of the \textit{Beer} dataset with a low-correlation assumption between aspect ratings, leading to a more straightforward and unrealistic dataset.}
\begin{tabular}{@{}lccc@{}}
& & & \textbf{Decorrelated}\\
\multicolumn{1}{c}{\bf Dataset} & \multicolumn{1}{c}{\bf Beer}  & \multicolumn{1}{c}{\bf Hotel} & \multicolumn{1}{c}{\bf Beer} \\
\toprule
Number of reviews& $1,586,259$ & $140,000$& $280,000$   \\
Average word-length of review& $147.1 \pm 79.7$  & $188.3 \pm 50.0$ & $157.5 \pm 84.3$ \\
Average sentence-length of review& $10.3 \pm 5.4$  & $10.4 \pm 4.4$ & $11.0 \pm 5.7$ \\
Number of aspects& $4$ & $5$ & $3$ \\
Average correlation between aspects& $\mathbf{71.8\%}$ & $\mathbf{63.0\%}$ & $\mathbf{27.2\%}$ \\
Max correlation between two aspects& $\mathbf{73.4\%}$ & $\mathbf{86.5\%}$ & $\mathbf{29.8\%}$ \\
\end{tabular}
\end{table}

\begin{table}[!t]
\centering
\caption{\label{app_perfs_small_beer}Performance of the multi-aspect sentiment classification task for the \textbf{\textit{decorrelated}} \textit{Beer} dataset. MTM still outperforms baselines in the unrealistic decorrelated settings.}
\hspace*{-0.75cm}
\begin{tabular}{@{}l>{}ll@{}>{}c@{}>{}c@{}>{}c@{\hspace*{2mm}}>{}c@{\hspace*{2mm}}>{}c@{\hspace*{2mm}}>{}c@{}}
& & & & \multicolumn{4}{c}{\textbf{F1 Score}}\\
\cmidrule(lr){5-8}
\multicolumn{1}{c}{\bf Interp.} & \multicolumn{2}{c}{\bf Model} & \multicolumn{1}{c}{\bf Params} & \multicolumn{1}{c}{\bf Macro} & \multicolumn{1}{c}{\boldmath$A_1$} & \multicolumn{1}{c}{\boldmath$A_2$} & \multicolumn{1}{c}{\boldmath$A_3$}\\
\toprule
\multicolumn{1}{c}{\multirow{2}{*}{\parbox{1.cm}{\centering None}}}
& SENT & Sentiment Majority & $426k$ & $68.89$ & $67.48$ & $73.49$ & $65.69$\\
& BASE & $\text{Emb}_{200}$ + $\text{Enc}_\textsubscript{CNN}$ + \text{Clf} & $173k$ & $78.23$ & $78.38$ & $80.86$ & $75.47$\\
\midrule
\multicolumn{1}{c}{\multirow{2}{*}{\parbox{1.cm}{\centering Coarse-grained}}}
& \multirow{2}{*}{SAA} & $\text{Emb}_{200}$ + $\text{Enc}_\textsubscript{CNN}$ + $\text{A}_\textsubscript{Shared}$ + \text{Clf} & $196k$ & $78.19$ & $77.43$ & $80.96$ & $76.16$\\
& & $\text{Emb}_{200}$ + $\text{Enc}_\textsubscript{LSTM}$ + $\text{A}_\textsubscript{Shared}$ + \text{Clf} & $186k$ & $78.16$ & $75.88$ & $81.25$ & $77.36$\\
\midrule
\multicolumn{1}{c}{\multirow{6}{*}{\parbox{1.cm}{\centering Fine-grained}}}
& NB-SVM & \citep{wang-manning-2012-baselines} & $3 \cdot 426k$ & $74.60$ & $73.50$ & $77.32$ & $72.99$\\
& SAM & \citep{lei-etal-2016-rationalizing} & $3 \cdot 644k$ & $77.06$ & $77.36$ & $78.99$ & $74.83$\\
& MASA & $\text{Emb}_{200}$ + $\text{Enc}_\textsubscript{LSTM}$ + $\text{A}^\textsubscript{Sparse}_\textsubscript{Aspect-wise}$ + \text{Clf} & $458k$ & $78.82$ & $77.35$ & $81.65$ & $77.47$\\
& MAA & $\text{Emb}_{200}$ + $\text{Enc}_\textsubscript{LSTM}$ + $\text{A}_\textsubscript{Aspect-wise}$ + \text{Clf} & $458k$ & $78.96$ & $78.54$ & $81.56$ & $76.79$\\
\cdashlinelr{2-8}
& MTM\textsuperscript{} & $\text{Emb}_{200}$ + \text{Masker} + $\text{Enc}_\textsubscript{CNN}$ + \text{Clf} (Ours) & $274k$ & $79.32$ & $78.58$ & $81.71$ & $77.66$\\
& \textbf{MTM\textsuperscript{C}} & \textbf{$\text{Emb}_{200+4}$ +  $\text{Enc}_\textsubscript{CNN}$ + \text{Clf} (Ours)} & $175k$ & $\mathbf{79.66}$ & $\mathbf{78.74}$ & $\mathbf{82.02}$ & $\mathbf{78.22}$ \\
\end{tabular}
\end{table}

We find similar trends but stronger results (see Table~\ref{app_perfs_small_beer}): \textit{MTM} significantly generates better rationales and achieves higher F1 scores than \textit{SAM} and the attention models. The contextualized variant \textit{MTM\textsuperscript{C}} further improves the performance. Some visualizations are available in~Appendix~\ref{samples_with_visualization}.

\subsection{Impact of the Length of Rationales}
\label{app:impact_length}

In this section, we study the distribution of the rationales' lengths and their impact on the rationale performance in terms of the precision, recall, and F1 score. In Section~\ref{subsub_prec}, and for a fair comparison with prior work, we report only the precision on three aspects at a similar number of selected words on the human sentence-level aspect annotations. 
 
We use the \textit{Beer} dataset and also consider the extra aspect \textit{Overall} (which leads to five aspects in total). We train \textit{MTM} on it as in Section~\ref{experimental_details}. We compute the distribution of the explanation length (in percent) on the validation set for each aspect $a_i, i=1,...,5$. We calculate all the $100$ percentiles $P^{a_i}_p$. Then, we infer the sub-masks $M_{a_i}$ and generate the rationale by selecting each word $x^\ell$, whose relevance towards $a_i$ satisfies $P(m_{a_i}^\ell|x^\ell) \ge P^{a_i}_p$. Finally, we compute the precision, recall, and F1 score on the human annotation for each percentile~$P^{a_i}_p$.

\begin{figure}[!t]
\centering
\includegraphics[width=\textwidth,height=0.7cm]{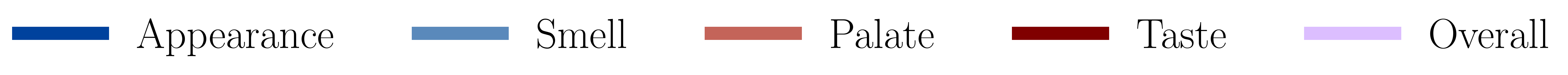}\\
\includegraphics[width=0.33\textwidth,height=4.9cm]{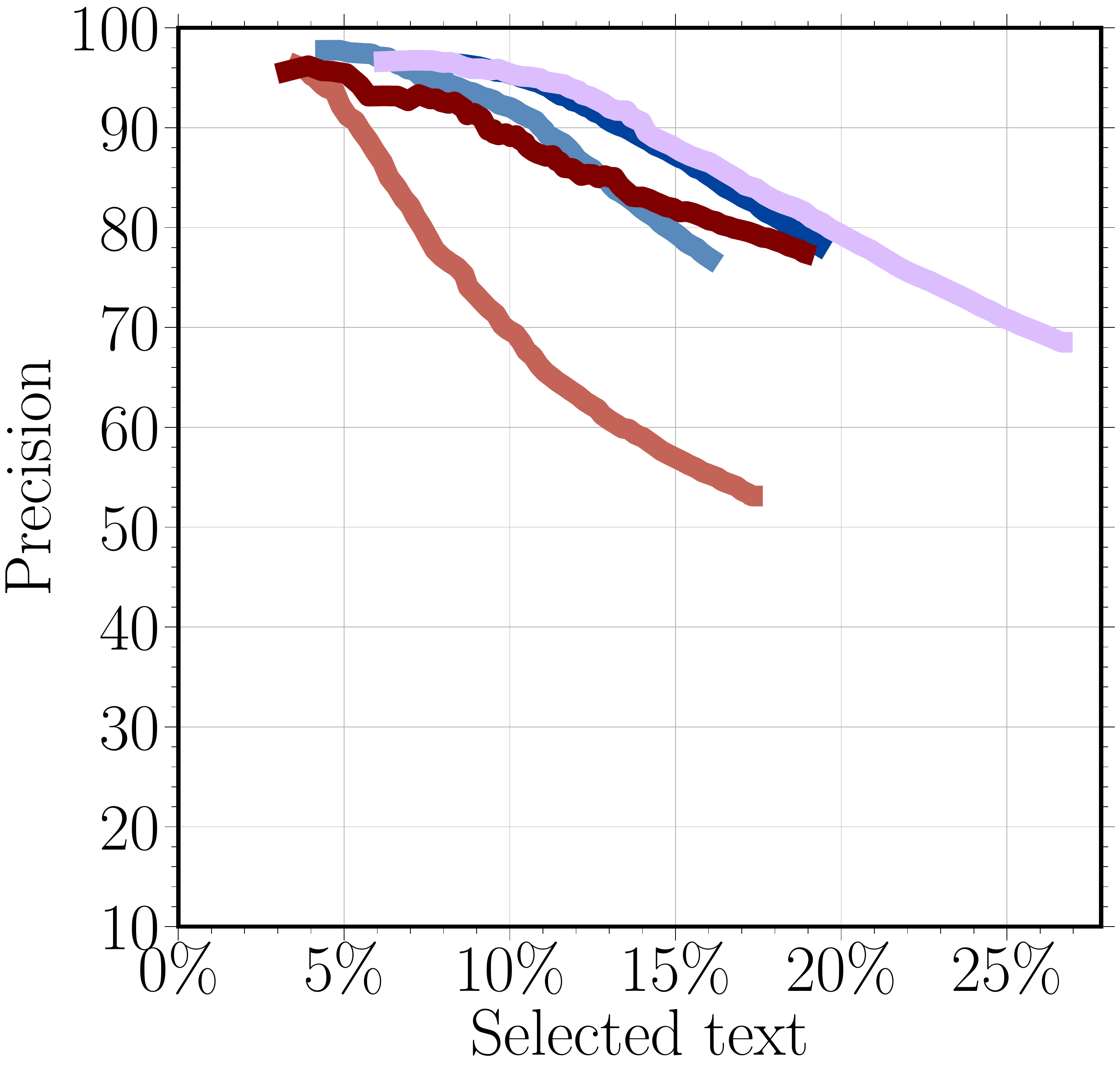}
\includegraphics[width=0.33\textwidth,height=4.9cm]{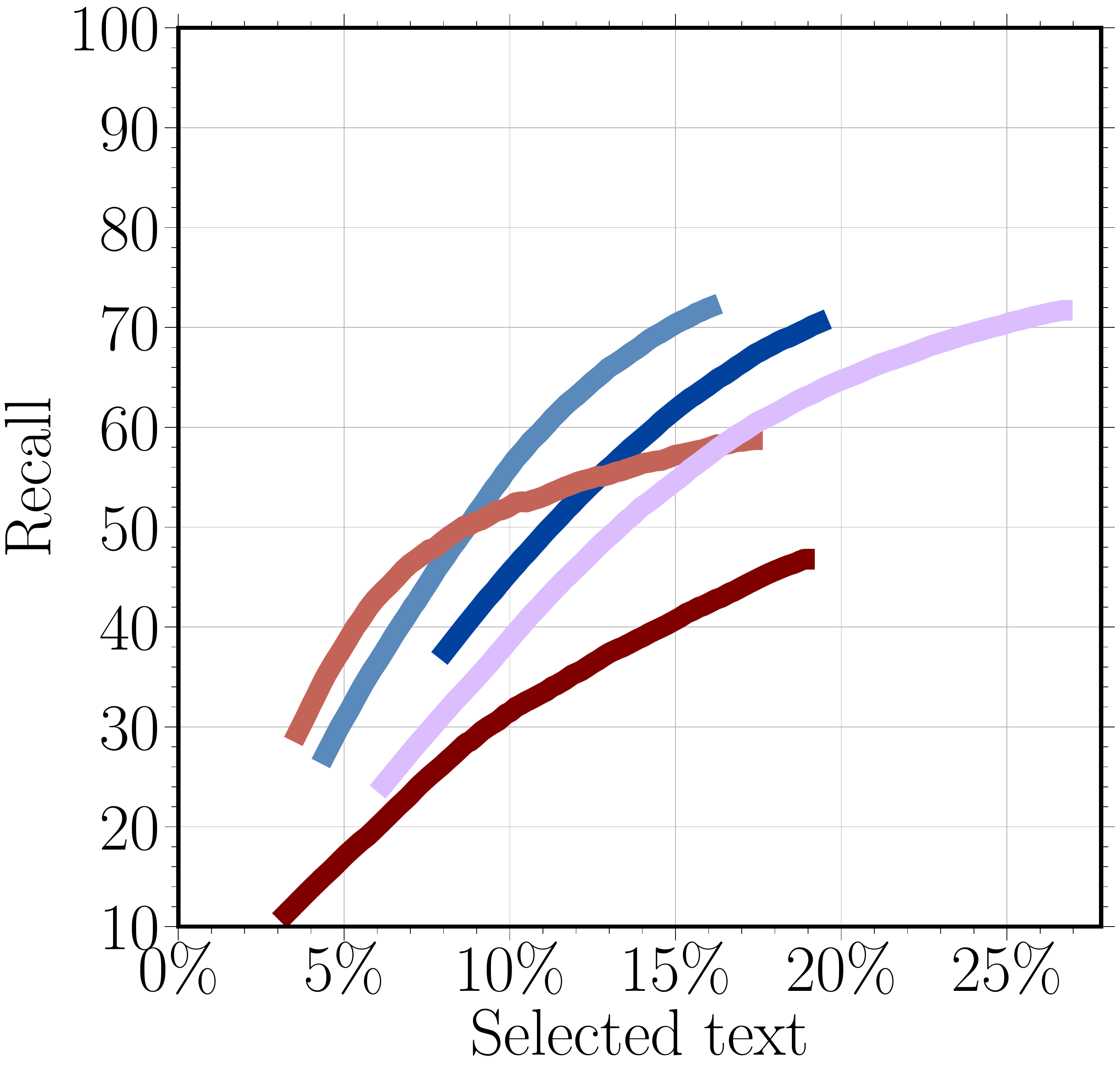}
\includegraphics[width=0.33\textwidth]{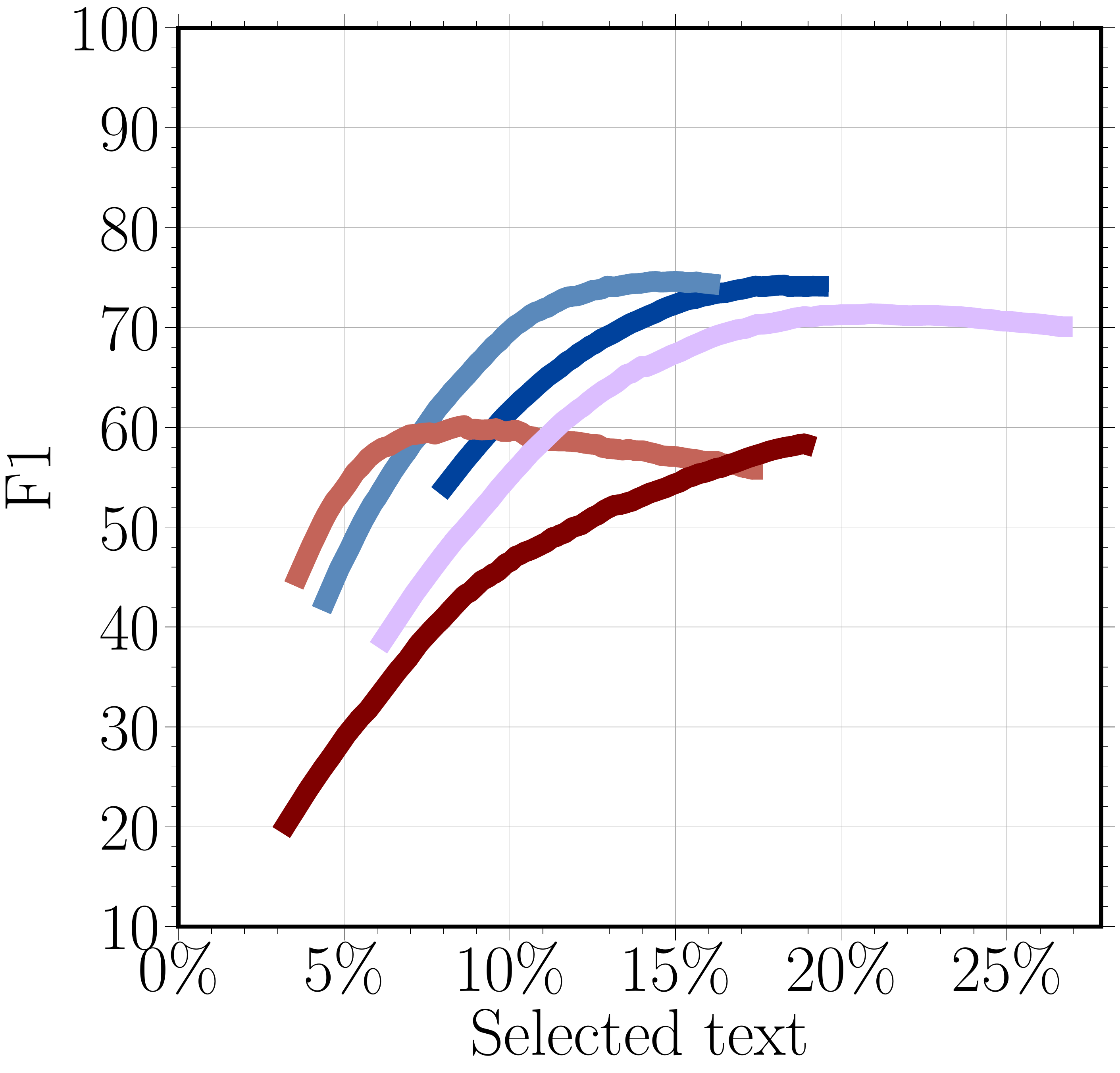}
\caption{\label{app_he_ann}Performance related to human evaluation, showing the precision, recall, and F1 scores of the rationales for each aspect (including the overall rating) of the \textit{Beer} dataset. The percentage of words indicates the number of highlighted words of the full~review.}
\end{figure}

Figure~\ref{app_he_ann} shows the result of the five aspects. First, we observe that the length distribution of the rationales for the aspect \textit{Overall} is the most spread compared to the other aspects, and the one of the aspect \textit{Smell} the least. The other three aspects share a similar distribution.

In terms of precision, we notice that the aspect \textit{Palate} drops quickly compared to the other ones, that decrease linearly when we augment the portion of selected text. According to RateBeer\footnote{\url{https://www.ratebeer.com/Story.asp?StoryID=103}.}, the aspect \textit{Palate} is the most difficult one to rate, which confirms our findings. For most aspects the precision remains high after selecting $75\%$ of the text ($5 \cdot 15\%$), showing the effectiveness of our approach. In terms of recall, they all increase unsurprisingly when more words are highlighted.

Finally, we show that \textit{MTM} generally achieves a very high precision~($>90\%$) by highlighting only $5-15\%$ words of the input text, which reduces the cognitive load of users to identify the important parts of documents.

\section{Conclusion}

Providing explanations for automated predictions carries much more impact, increases transparency, and might even be necessary.
Past work has proposed using attention mechanisms or rationale methods to explain the prediction of a target variable. The former produce noisy explanations, while the latter do not properly capture the multi-faceted nature of useful rationales. Because of the non-probabilistic assignment of words as justifications, rationale methods are prone to suffer from ambiguities and spurious correlations and thus, rely on unrealistic assumptions about the data.

The Multi-Target Masker (MTM) addresses these drawbacks by replacing the binary mask with a probabilistic multi-dimensional mask (one dimension per target), learned in an unsupervised and multi-task learning manner, while jointly predicting all the target variables.
According to comparison with human annotations and automatic evaluation on two real-world datasets, the inferred masks were more accurate and coherent than those that were produced by the state-of-the-art methods.
It is the first technique that delivers both the best explanations and highest accuracy for multiple targets simultaneously.

In Chapter~\ref{chapter_ijcai_2021}, we use inferred masks for recommendation: to generate personalized and relevant explanations to support recommendations and build better user and item profiles.
As future work, because masks embed opinions in reviews, they could be used to produce structured summaries across multiple aspects. Also, our method can be applied to different types of inputs like images. In that case, the words in the text data are replaced~by regions or patches in an image, obtained by a classical pretrained convolutional neural network. The sparsity constraint stays similar, and the continuity is based on two-dimensional distances instead.

%% file: main/ACL2021/acl2021.tex
\chapter{Rationalization through Concepts}
\label{chapter_acl_2021}

\section{Preface}
\textbf{Contribution and Sources.}\hspace{\parindent} This chapter is largely based on \cite{antognini-2021-concept}. The detailed individual contributions are listed below using the CRediT taxonomy \citep{brand2015beyond} (terms are selected as applicable).

\begin{table}[!h]
\begin{tabular}{@{}l@{\hspace{1mm}}l@{}}
Diego Antognini (author): & Conceptualization, Methodology, Software, Validation,\\
& Investigation, Formal Analysis, Writing -- Original Draft,\\
&Writing -- Review \& Editing.\\ \\
Boi Faltings: & Writing -- Review \& Editing, Administration, Supervision.
\end{tabular}
\end{table}

\textbf{Summary.}\hspace{\parindent} Automated predictions require explanations to be interpretable by humans.
One type of explanation is a rationale, i.e., a selection of input features such as relevant text snippets from which the model computes the outcome. However, a single overall selection does not provide a complete explanation, e.g., weighing several aspects for decisions. In Chapter~\ref{chapter_aaai2021}, we use several aspect ratings to infer multi-dimensional rationales. However, in some cases, the availability of those labels is not guaranteed.
To this end, we present in this chapter a novel self-interpretable model called ConRAT. Inspired by how human explanations for high-level decisions are often based on key concepts, ConRAT extracts a set of text snippets as concepts and infers which ones are described in the document. Then, it explains the one outcome with a linear aggregation of concepts. Two regularizers drive ConRAT to build interpretable concepts. In addition, we propose two techniques to boost the rationale and predictive performance further. Experiments on both single- and multi-aspect sentiment classification tasks show that ConRAT is the first to generate concepts that align with human rationalization while using only the overall label. Further, it outperforms state-of-the-art methods trained on each aspect label independently.

\section{Introduction}

Neural models have become the standard for many tasks, owing to their large performance gains. However, their adoption in decision-critical fields is more limited because of their lack of interpretability, particularly with textual data.

One of the simplest means of explaining predictions of complex models is by selecting relevant input features. Attention mechanisms \citep{iclr2015} model the selection using a conditional importance distribution over the inputs, but the resulting explanations are noisy \citep{jain2019attention,pruthi-etal-2020-learning}. Multi-head attention \citep{vaswani2017attention} extends attention mechanisms to attend information from different perspectives jointly. However, no explicit mechanisms guarantee a logical connection between different views \citep{voita-etal-2019-analyzing,kovaleva-etal-2019-revealing}.
Another line of research includes rationale generation methods \citep{lei-etal-2016-rationalizing,chang2020invariant,antognini2019multi}. If the selected~text input features are short and concise -- called a rationale -- and suffice on their own to yield the prediction, it can potentially be understood and verified against domain knowledge \citep{chang2019game}. 

\begin{figure}[!t]
\centering
\includegraphics[width=0.7\linewidth]{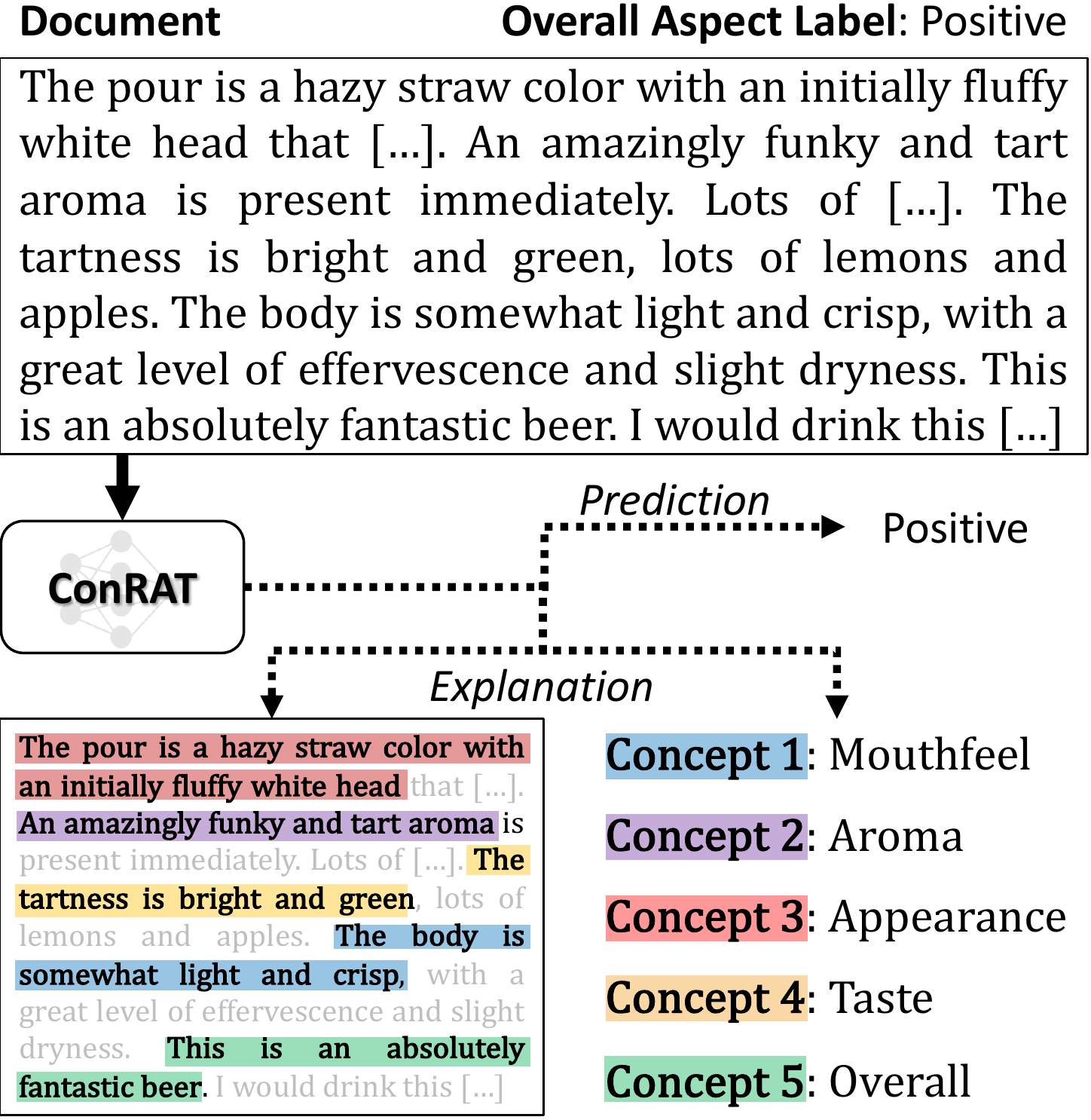}
\caption{\label{front}An illustration of ConRAT. Given a beer~review, ConRAT identifies five excerpts that relate to~particular concepts of beers (i.e., the explanation), depicted in color, from which it computes the outcome.}
\end{figure}

The key motivation for this chapter arises from the limitations of rationales. Rationalization models strive for one overall selection to explain the outcome by maximizing the mutual information between the rationale and the label. However, useful rationales can be multi-faceted, where each facet relates to a particular ``concept'' (see Figure~\ref{front}). For example, users typically justify their opinions of a product by weighing explanations: one for each aspect they care about \citep{musat2013,musat2015personalizing}.

Inspired by how human reasoning comprises concept-based thinking \citep{ARMSTRONG1983263,tenenbaum1999bayesian}, we aim to discover, in an unsupervised manner, a set of concepts to explain the outcome with a weighted average, similar to multi-head attention. In this chapter, we relate concepts to semantically meaningful and consistent excerpts across multiple texts. Unlike topic modeling, where documents are described by a set of latent topics comprising word distributions, our latent concepts relate to text snippets that are relevant for the prediction.

Another motivation for this chapter is to generate interpretable concepts. The explanation~of~an outcome should rely on concepts that~satisfy the desiderata introduced in \cite{NEURIPS2018_3e9f0fc9}. They should
\begin{enumerate}[topsep=0pt]
  \item preserve relevant information,
  \item not overlap with each other and be diverse, and
  \item be human-understandable.
\end{enumerate} Figure~\ref{front} shows an example of concepts in the beer domain.

In this chapter, we present a novel self-explaining neural model: the concept-based rationalizer (ConRAT) (see Figure~\ref{front} and Figure~\ref{architecture}). Our new rationalization scheme first identifies a set of concepts in a document and then decides which ones are currently described (binary selection). ConRAT explains the prediction with a linear aggregation of concepts. The model is trained end-to-end, and the concepts are learned in an unsupervised manner. In addition,~we~design two regularizers that guide ConRAT to induce interpretable concepts and propose two optional techniques, knowledge distillation and concept pruning, in order to boost the performance further.

We evaluate ConRAT on both single- and multi-aspect sentiment classification with up to five target labels. Upon training ConRAT only on the overall aspect, the results show that ConRAT generates concepts that are relevant, diverse, and non-overlapping, and they also recover human-defined concepts. Furthermore, our model significantly outperforms strong supervised baseline models in terms of predictive and explanation performance.

\section{Related Work}

Developing interpretable models is of considerable interest to the broader research community. Researchers have investigated many approaches to improve the interpretability of neural networks. 

\subsection{Interpretability}
The first line of research aims at providing post-hoc explanations of an already trained model. For example, gradient and perturbation-based methods attribute the decision to important input features \citep{ribeiro2016should,pmlr-v70-sundararajan17a,NIPS2017_7062,pmlr-v70-shrikumar17a}.
Other studies identified the causal relationships between input-output pairs \citep{alvarez-melis-jaakkola-2017-causal,goyal2019explaining}. In contrast, our model is inherently interpretable as it directly produces the prediction with an explanation.

Another line of research has developed interpretable models. \cite{quint2018interpretable} extended a variational autoencoder with a differentiable decision tree. \cite{alaniz2019explainable} proposed an explainable observer-classifier framework whose predictions can be exposed as a binary tree. However, these methods have been designed for images only, while our work focuses on text input.

The works most relevant to ours relate to interpretable models from the rationalization field~\citep{lei-etal-2016-rationalizing,bastings-etal-2019-interpretable,yu-etal-2019-rethinking,chang2020invariant,jain-etal-2020-learning,paranjape-etal-2020-information}. These methods justify their predictions by selecting rationales (i.e., relevant tokens in the input text). However, they are limited to explain~only~the prediction with mostly one text span and rely on the assumption that the data have low internal correlations \citep{antognini2019multi}. \cite{chang2019game} extended previous methods to extract an additional rationale in order to counter the prediction. In our work, ConRAT produces multi-faceted rationales and explains the prediction through a linear aggregation of the extracted concepts. However, if we set the number of concepts to one, ConRAT reduces to a special case of a rationale model.

\subsection{Explanations through Concepts}

Researchers have proposed multiple approaches for concept-based explanations. \cite{conf/icml/KimWGCWVS18} designed a post-hoc technique to learn concept activation vectors by relying on human annotations that characterize concepts of interest. Similarly, \cite{bau2017network,10.1007/978-3-030-01237-3_8} generated visual explanations for a classifier. Our concepts are learned in an unsupervised manner and not defined a~priori. 

Few studies have learned concepts on images in an unsupervised fashion. \cite{li2018deep} explained predictions based on the similarity of the input to ``prototypes'' learned during training. \cite{NEURIPS2018_3e9f0fc9} used an autoencoder to extract relevant concepts and explain the prediction. \cite{ghorbani2019towards} designed an unsupervised concept discovery method to explain trained models. \cite{koh2020concept} employed the discovered concepts to predict the target label. Our work's key difference is that we focus on text data, while all these methods treat only image inputs.

To the best of our knowledge, \cite{bouchacourt2019educe} is the only study that has proposed a self-interpretable concept-based model for text data using reinforcement learning. It computes the predictions and provides an explanation in terms of the presence or~absence of concepts in the input (i.e., text excerpts of variable lengths). However, their method achieves poor overall performance. In addition, it is unclear whether the discovered concepts are interpretable. Conversely, ConRAT is differentiable, clearly outperforms strong models in terms of predictive and explanation performance, and it infers relevant, diverse, non-overlapping, and human-understandable concepts. 

\subsection{Topic Modeling}
Topic models, such as latent Dirichlet allocation \citep{blei2003latent}, describe documents with a mixture of latent topics. Each topic represents a word distribution. Some studies combined topic models with recurrent neural~models \citep{dieng2016topicrnn,zaheer2017latent}. However, the goal of these generative models and the topics remains different than this chapter's.~We aim to build a self-interpretable model that predicts and explains the outcome with latent concepts.

\section{ConRAT: A Concept-based Rationalizer}
\begin{figure}[!t]
\centering
\includegraphics[width=0.7\linewidth]{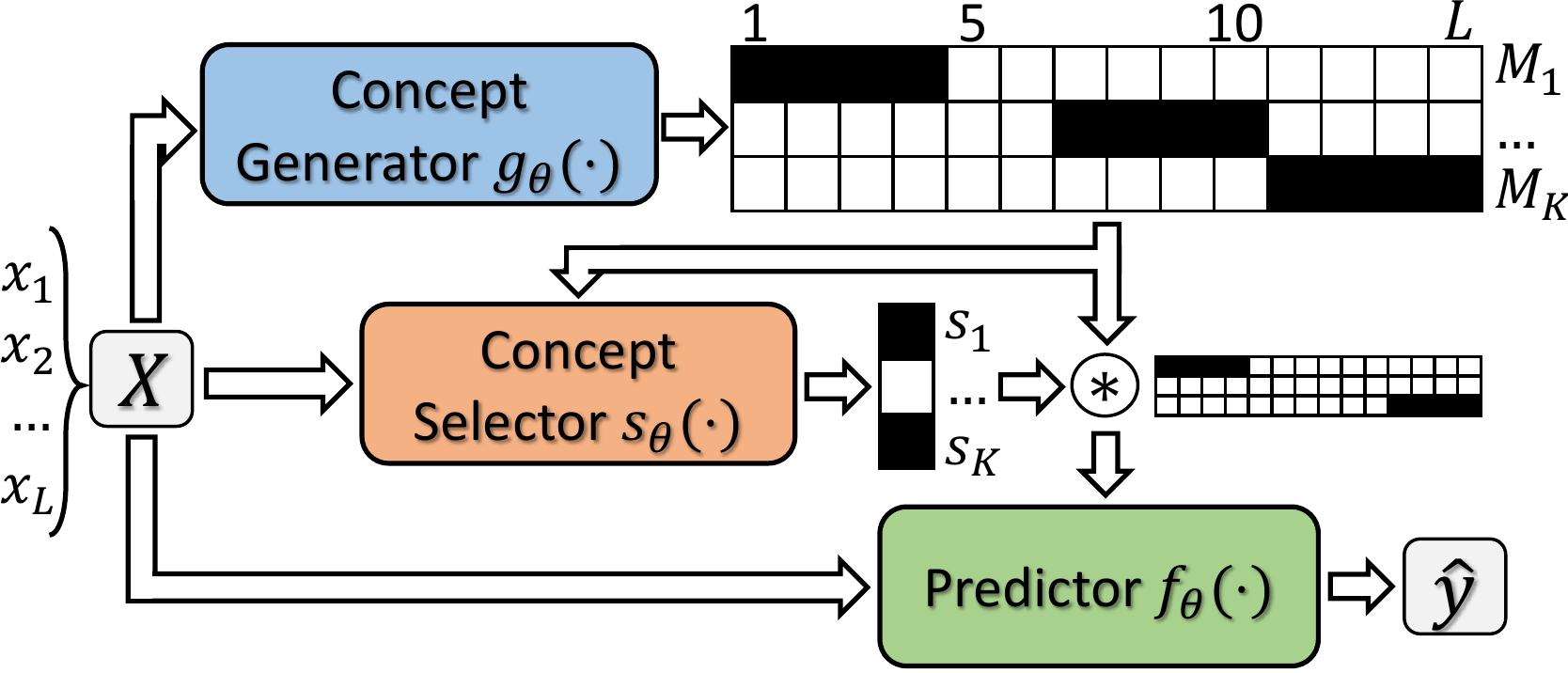}
\caption{\label{architecture}The proposed self-explaining model ConRAT. The model predicts and explains~$\hat{y}$. Given a document~$X$, the concept generator produces one binary mask per concept. The concept selector decides which concepts are present in the input. The predictor~aggregates each selected concept's prediction to~compute~$\hat{y}$.}
\end{figure}

Figure~\ref{architecture} depicts the architecture of our proposed self-explaining model: the Concept-based Rationalizer (ConRAT). Let $X$ be a random variable representing a document composed of $T$ words $(x_1, x_2, \dots, x_T)$, $y$ the ground-truth label, and $K$ the desired numbers of concepts\footnote{Our method is easily adapted for regression problems.}.
 Given a document $X$ and a label $y$, our goal is to explain the prediction $\hat{y}$ by finding a set of $K$ concepts $C_1,$ $\dots, C_K$ that are masked versions of $X$. ConRAT learns concepts by maximizing the mutual information between $\bb{C}$ and $y$. We guide ConRAT to create separable and consistent concepts via two regularizers to make them human-understandable.

\subsection{Model Overview}

ConRAT is divided into three submodels: a \textbf{Concept Generator} $g_\theta (\cdot)$, which finds the concepts~$C_1, \dots, C_K$; a \textbf{Concept Selector} $s_\theta(\cdot)$, which detects whether a concept $C_k$ is present or absent (i.e., $s_k \in \{1,0\}$) in the input $X$; and a \textbf{Predictor}~$f_\theta(\cdot)$, which predicts the outcome $\hat{y}$ based on the concepts $\bb{C}$ and their presence scores $S$.

\subsubsection{Concept Generation.}
\label{sec_concept_generation}Inspired by the selective rationalization field \citep{lei-etal-2016-rationalizing}, we define ``concept'' as a sequence of consecutive words in the input text. Previous studies extracted only one concept $C_1$ that is sufficient to explain the target variable $y$. In our work, a major difference is that we aim to find $K$ concepts~$C_1, \cdots, C_K$ that represent different topics or aspects and altogether explain the target variable $y$. We interpret the model as being linear in the concepts rather than depending on one overall selection of word.
More formally, we define a concept as follows:\begin{equation}
  C_k = M_k \odot X,
  \label{eq:concept}
\end{equation}where $M_k \in \mathbb{S}$ denotes a binary mask, $\mathbb{S}$ is a subset of $\mathbb{Z}_2^T$ with some constraints (introduced in Section~\ref{discovery_concepts}), and $\odot$ is the element-wise multiplication of two vectors. 

We parametrize the binary masks $\bb{M} \in \mathbb{Z}_2^{K \times T}$ with the concept generator model $g_\theta(\cdot)$, based on~a bi-directional recurrent neural network. Following previous rationalization research \citep{yu-etal-2019-rethinking,chang2020invariant}, we force $g_\theta(\cdot)$ to select one chunk of text per concept with a pre-specified length $\ell \in [1, T]$\footnote{In early experiments, we relaxed the length constraint and generated instead $K$ differentiable masks with continuity regularizers. However, this variant produced majorly inferior results. We hypothesize that there are too many constraints to optimize with only the target label as a strong signal.}.
 Instead of predicting the mask~$M_k$ directly, $g_\theta(\cdot)$ produces a score for each position $t$. Then, it samples the start position~$t^*_k$ of the chunk for each $C_k$ using the straight-through Gumbel-Softmax \citep{MaddisonMT17,JangGP17}. Finally, we compute $M_k$ as~follows:\begin{equation}
\begin{split}
  T^* &\sim Gumbel(g_\theta(X)),\\
    M_{k,t} &= \mathbb{1}[t\in [t^*_k, min(t^*_k + \ell-1, T)]],
    \end{split} 
    \label{eq:hard_constraint}
\end{equation}
where $\mathbb{1}$ denotes the indicator function. Although the equation is not differentiable, we can employ the straight-through technique \citep{bengio2013estimating} and approximate it with the gradient of a causal convolution and a convolution kernel of an all-one vector of length $\ell$.

\subsubsection{Concept Selection.}
\label{sec_concept_selection}
A key objective of ConRAT is to produce semantically consistent and separable concepts. So far, the generator $g_\theta(\cdot)$ generates $K$ concepts for any input document. However, some documents might mention only a subset of those. Thus, the goal of the concept selector model $s_\theta(\cdot)$ is to enable ConRAT to ignore absent concepts.

Specifically, for each concept $C_k$, the model first computes a concept representation $H_{C_k}$ using a standard attention mechanism \citep{iclr2015} (the tokens whose $M_{k,t}=0$ are masked out). Then, we take the dot product of $H_{C_k}$ with a weight vector, followed by a sigmoid activation function to induce the log-probabilities of a relaxed Bernoulli distribution \citep{JangGP17}. Finally, we sample the presence score $s_k \in \{0,1\}$ of each concept independently:\begin{equation}
  S \sim RelaxedBernoulli(s_\theta(X, \bb{M})),
\end{equation}where $\bb{M}$ denote the binary masks and $X$ the input tokens.

\subsubsection{Prediction.}

As inputs, the predictor $f_\theta (\cdot)$ takes the document~$X$, the masks $\bb{M}$, and the presence scores~$S$ for all concepts. First, we extract the concepts, which are masked versions of $X$. Differently than in Equation~\ref{eq:concept}, the concepts are ignored if $s_k=0$:\begin{equation}
  C_k = (M_k * s_k) \odot X.
\end{equation}
Second, the model produces the hidden representation $h'_{C_k}$ with another recurrent neural network, followed by a LeakyReLU activation function \citep{xu2015empirical}. Then, it computes the logits of $y$ by applying a linear projection for each concept: \begin{equation}
  P_k = Wh'_{C_k} + b,
\end{equation} where $W$ and $b$ are the projection parameters. Finally, $f_\theta (\cdot)$ computes the final outcome as follows:\begin{equation}
  p(y | C, \bb{M}, X) = softmax(\sum_{k=1}^K \alpha_k P_k s_k),
\end{equation}where $\alpha_k$ are model parameters that can be interpreted as the degree to which a particular concept contributes to the final prediction.

\subsection{Unsupervised Discovery of Concepts}
\label{discovery_concepts}

The above formulations integrate the explanation into the outcome computation. However, $M_k$ is by definition faithful to the model's inner workings but not comprehensible for the end user~\citep{jacovi-goldberg-2020-towards}. Following \cite{NEURIPS2018_3e9f0fc9}, we~aim the concepts to follow three desiderata:
\begin{enumerate}[topsep=0pt]
\item \textbf{Fidelity}: they should preserve relevant information,
\item \textbf{Diversity}: they should be non-overlapping and diverse, and
\item \textbf{Grounding}: they should have an immediate human-understandable interpretations.
\end{enumerate}

The hard constraint in Equation~\ref{eq:hard_constraint} naturally enforces the grounding by forcing the concept to be a sequence of $\ell$ words. For the fidelity, it~is partly integrated in ConRAT by the prediction loss, which is the cross-entropy between the ground-truth label $y$ and the prediction $\hat{y}$: $\mathcal{L}_{pred} = CE(\hat{y},y)$. Recall that the concepts are substitutes of the input that are sufficient for the prediction. We emphasize the word ``partly'' because nothing prevents ConRAT from picking up spurious correlations.

We propose two regularizers to encourage ConRAT in finding non-overlapping, relevant, and dissimilar concepts. The first favors the orthogonality of concepts by penalizing redundant rows in $\bb{M}$:
\begin{equation}
  \mathcal{L}_{overlap} = || \bb{M}\bb{M}^T - \ell \cdot \mathbb{1}||_F^2,
\end{equation} where $|| \cdot ||_F$ stands for the Frobenius norm of a matrix, $\mathbb{1}$ denotes the identity matrix, and $\ell$ the pre-specified concept length. 
However, $\mathcal{L}_{overlap}$ alone does not prevent ConRAT from learning little relevant concepts. Therefore, we propose a second regularizer to encourage fidelity and diversity by minimizing the cosine similarity between the concept representations $H_{C_k}$ (see Section~\ref{sec_concept_selection}):\begin{equation}
  \mathcal{L}_{div} = \frac{1}{K}\frac{1}{K-1}\sum_{\substack{k_1,k_2=1\\k_1 \ne k_2}}^K cos(H_{C_{k_1}}, H_{C_{k_2}}).
\end{equation}

In both regularizers, we do not consider the presence scores $S$ because a model could always select only one concept; this strategy is suboptimal and reduces to a special case of rationale models (i.e., $S$ would become a one-hot vector).

To summarize, the concepts are learned in an unsupervised manner and align with the three desiderata mentioned above: diversity is achieved with $\mathcal{L}_{overlap}$ and $\mathcal{L}_{div}$; fidelity is enforced by $\mathcal{L}_{pred}$ and $\mathcal{L}_{div}$, and the hard constraint in Equation~\ref{eq:hard_constraint} ensures the grounding. Finally, we train ConRAT end-to-end and minimize the loss jointly $\mathcal{\bb{L}}=\mathcal{L}_{pred} + \lambda_O \mathcal{L}_{overlap} + \lambda_D \mathcal{L}_{div}$, where $\lambda_O$ and~$\lambda_D$ control the impact of each regularizer.
\subsection{Improving  Overall Performance Further}
\label{sec_improv_perfs}
The purpose of self-explaining models is to compute outcomes while being more interpretable. However, one key point is to achieve predictive performance comparable to that of black-box models. We propose two techniques to further improve both interpretability and performance; however, ConRAT does not require these techniques to outperform other methods, as we will see later.

\textbf{Knowledge Distillation.}\hspace{\parindent} We can train ConRAT not only via the information provided by the true labels but also by observing how a teacher model behaves \citep{44873}. In that case, we introduce the teacher model $T_\theta ( \cdot )$, which is a simple recurrent neural network similar to the predictor $f_\theta (\cdot)$. It is trained one the same data, but it uses the whole input $X$ instead of subsets selected by each~$C_k$. The overall training loss becomes $\mathcal{\bb{L}}=\mathcal{L}_{pred} + \lambda_O \mathcal{L}_{overlap} + \lambda_D \mathcal{L}_{div} + \lambda_T(\hat{y}_{T_\theta} - \hat{y}_{f_\theta})^2$.

\textbf{Pruning Concepts.}\hspace{\parindent} Depending on the number of concepts and the pre-specified length, the total number of selected words can be close to or higher than the document length\footnote{e.g., if a document contains $200$ tokens and we aim to extract $10$ concepts of $20$ tokens, all words should be selected.}. In practice, it is hard to extract meaningful concepts in such settings. To alleviate this problem, we propose to prune concepts at inference and select the top-k~concepts that overlap the least with the others. More specifically, we compute the overlap as follows: for each sample in the validation set, we measure the average overlap ratio between $M_{k_1}$ and $M_{k_2}$ for each concept-pair $(C_{k_1}, C_{k_2}), k_1 \ne k_2$. Then, we select the top-k concepts whose scores are the lowest. Finally, to compute the new prediction $\hat{y}$, we update $s_k=1$ if $C_k$ is in the top-k or $s_k=0$ otherwise.

\section{Experiments}

\subsection{Datasets}
\label{datasets}

We evaluate the quantitative performance of ConRAT using two binary classification datasets. The first one is the single-aspect Amazon Electronics dataset \citep{ni-etal-2019-justifying}. We followed the filtering process in \cite{chang2019game} to keep only the reviews that contain evidence for both positive and negative sentiments. Specifically, we considered the first 50 tokens after the words ``pros:'' and ``cons:'' as the rationale annotations for the positive and negative labels, respectively. We randomly picked 24,000 balanced samples with ratings of four and above or two and below.

The second dataset comprises the multi-aspect beer reviews \citep{beer} used in the field of rationalization \citep{lei-etal-2016-rationalizing,yu-etal-2019-rethinking}. Each review describes various beer aspects: Appearance, Aroma, Palate, Taste, and~Overall; users also provided a five-star rating for each aspect. However, we only use the overall rating for ConRAT. The dataset includes 994 beer reviews with sentence-level aspect annotations. Following the evaluation protocol in \cite{bao-etal-2018-deriving,chang2020invariant}, we binarized the ratings $\le2$ as negative and $\ge3$ as positive. We sampled 60,000 balanced examples. Our setting is more challenging than those in previous studies because we assess the performance on all aspects (instead of three) and consider all examples for the sampling (instead of de-correlated subsets), reflecting the real data distribution. Table~\ref{dataset_description} shows the data statistics.  
\begin{table}[t]
\centering
\caption{\label{dataset_description}Statistics of the review datasets.}
\begin{tabular}{@{}lcc@{}}
\multicolumn{1}{c}{\bf Dataset} & \multicolumn{1}{c}{\bf Amazon}  & \multicolumn{1}{c}{\bf Beer}\\
\toprule
\# Reviews & $24,000$ & $60,000$\\
Split Train/Val/Test & 20k/2k/2k & 50k/5k/5k\\
\# Annotations & $471$ & $994$\\
\# Human Aspects & $1$ & $5$\\
\# Words per review& $224 \pm 125$  & $184 \pm 58$\\
\# Sentences per review & $12.0 \pm 7.6$  & $12.6 \pm 4.9$\\
\end{tabular}
\end{table}
\subsection{Baselines}
We consider the following baselines. \textbf{RNP} is a generator-predictor framework proposed by \cite{lei-etal-2016-rationalizing} for rationalizing neural prediction. The generator selects text spans as rationales, which are then fed to the classifier for the final prediction. \cite{yu-etal-2019-rethinking} introduced \textbf{RNP-3P}, which extends RNP to include the complement predictor as the third player. It maximizes the predictive accuracy from unselected words. The training consists of an adversarial game with the three players. \textbf{Intro-3P} \citep{yu-etal-2019-rethinking} improves RNP-3P by conditioning the generator on~the predicted outcome of a teacher model. \textbf{InvRAT} is a game-theoretic method that competitively rules out spurious words with strong correlations to the output. The game-theoretic approach \textbf{CAR} aims to infer a rationale and a counterfactual rationale that counters the true label. We follow \cite{chang2020invariant} and consider for all methods their hard constraint variant (i.e., selecting one chunk of text) with different lengths for generating~rationales.

RNP-3P and Intro-3P are trained with the policy gradient \citep{williams1992simple}. The others estimate the gradients of the rationale selections using the straight-through technique \citep{bengio2013estimating}.

All rationalization methods, except CAR, strive for a single overall selection ($K=1$) to explain the outcome. For the multi-aspect dataset, we train and tune each baseline independently for each aspect. The key difference with ConRAT is that the model is only trained on the overall aspect label and infers one rationale of $K$ concepts; the baselines are trained $K$ times to infer one rationale of one concept.
 
\subsection{Experimental Details}

To seek fair comparisons, we try to keep a similar number of parameters across all models, and we employ the same architecture for each player (generators, predictors, and discriminators/teachers) in all models: bi-directional gated recurrent units \citep{chung2014empirical} with a hidden dimension 256. We use the 100-dimensional GloVe word embeddings \citep{pennington2014glove}, Adam \citep{KingmaB14} as optimization method with a learning rate of 0.001. We set the convolutional neural network in the concept selector similarly to \cite{kim2015mind} with 3-, 5-, and 7-width filters and 50 feature maps per filter. For ConRAT, we set the regularizer factors as follow: $\lambda_O=0.05$, $\lambda_D=0.05$, and $\lambda_T=0.5$.
We use the open-source implementation for all models, and we tune them by maximizing the prediction accuracy on the dev set with $16$ random searches. For reproducibility purposes, we include additional details in Appendix~\ref{app_training}.

\begin{table}[t]
    \centering
   \caption{\label{exp_rq1_amazon}Accuracy and objective performance of rationales in automatic evaluation for the Amazon dataset. ConRAT infers rationales that align with factual and counter factual ones.}
\begin{threeparttable}
\begin{tabular}{@{}l@{\hspace{0.75cm}}c@{\hspace{0.75cm}}ccc@{}c@{\hspace{0.75cm}}ccc@{}}
& & \multicolumn{3}{c}{\textit{Factual}} & & \multicolumn{3}{c}{\textit{Counter Factual}}\\
\cmidrule{3-5}\cmidrule{7-9}
\textbf{Model} & \textbf{Acc.} & \textbf{P} & \textbf{R} & \textbf{F} & &  \textbf{P} & \textbf{R} & \textbf{F}\\
\toprule
RNP & $\mathbf{75.5}$ &$32.6$ & $18.8$ & $23.8$ & & \multicolumn{3}{c}{$-$}\\
RNP-3P & $70.0$ & $49.4$ & $28.4$ & $36.0$ & & \multicolumn{3}{c}{$-$}\\
Intro-3P  & $75.2$ & $22.1$ & $12.8$ & $16.2$ & & \multicolumn{3}{c}{$-$}\\
InvRAT & $71.5$ & $44.3$ & $25.5$ & $32.4$ & & \multicolumn{3}{c}{$-$}\\
ConRAT-1 (Ours) & $\mathbf{75.5}$ & $\mathbf{56.4}$ & $\mathbf{32.5}$ & $\mathbf{41.3}$ & & \multicolumn{3}{c}{$-$}\\
\cdashlinelr{1-9}
CAR & $73.6$& $33.0$ & $19.1$ & $24.2$ & & $\mathbf{44.1}$ & $\mathbf{25.4}$ & $\mathbf{32.2}$\\
ConRAT-6 (Ours) & $\mathbf{75.4}$ & $\mathbf{50.0}$ & $\mathbf{28.8}$ & $\mathbf{36.6}$ & & $32.3$ & $18.6$ & $23.6$\\
ConRAT-4 (Ours) & $75.3$ & $46.4$ & $26.7$ & $33.9$ & & $29.6$ & $17.1$ & $21.6$\\
ConRAT-2 (Ours) & $75.3$ & $33.7$ & $19.4$ & $24.6$ & & $8.9$  & $5.1$  & $6.5$\\
\end{tabular}
\end{threeparttable}
\end{table}

\subsection{RQ 1: Can ConRAT find evidence for factual and counterfactual rationales?}
\label{rq1}
We aim to validate whether ConRAT can identify the two evidences for positive and negative sentiments. We set the concept length $\ell=30$, we compare the generated rationales with the annotations, and we report the precision, recall, and F1 score. In this experiment, no teacher is used in ConRAT.

Table~\ref{exp_rq1_amazon} contains the results. The top rows contain the results when only the factual rationales are considered for the evaluation, and ConRAT-1 uses only one concept. We see that ConRAT surpasses the baselines in finding rationales that align with human annotations, and it also matches the test accuracy with the baselines. Interestingly, we note that the baselines achieving the highest accuracy underperform in finding the correct rationales.

For the factual and counterfactual rationales, CAR finds one rationale to support the outcome and another one to counter it, in an adversarial game. However, the concepts inferred by ConRAT are not guaranteed to align with the rationales~as there is no explicit signal to infer counterfactual concepts. Thus, we increase the number of concepts up to six and prune ConRAT to consider only the two most dissimilar concepts (see Section~\ref{sec_improv_perfs}).

The bottom of Table~\ref{exp_rq1_amazon} shows the results. With only two concepts, ConRAT-2 outperforms CAR in terms of test accuracy and  matches~the performance for the factual rationales, but it poorly identifies counterfactual rationales. However, there is~a major improvement when we increase the number of concepts and use pruning. Indeed, the word~distribution of the factual and counterfactual rationales are different, hence captured with pruning. ConRAT's factual rationales are better than those of all models. The counterfactual ones~get~closer to those produced by CAR. We show later~in Section~\ref{rq3} that pruning helps in achieving better correlation with human judgments but is not required.

\subsection{RQ 2: Are concepts inferred by ConRAT consistent with human rationalization?}
\label{seq_rq2_concepts}

We investigate whether ConRAT can recover all beer aspects by using only the overall ratings. Because beer reviews are smaller in length than Amazon ones, we set the concept length $\ell$ to 10 and 20. We fix the number of concepts to ten and prune ConRAT to keep five. We manually map them to the closest aspect for comparison. We trained the teacher model, used in Intro-3P and ConRAT, and obtained 91.4\% accuracy. More results and illustrations are available in Appendix~\ref{app_rq2_conrad} and Appendix~\ref{app_samples}, respectively.

\subsubsection{Objective Evaluation.}
\label{rq2_exp_obj}

\begin{table}[!t]
    \centering
   \caption{\label{exp_rq2_beer_objective}Objective performance of rationales for the multi-aspect beer reviews. ConRAT only uses the overall label and ignores the other aspect labels. All baselines are trained separately on each aspect rating. \textbf{Bold} and \underline{underline} denote the best and second-best results, respectively. ConRAT is competitive with the fully-supervised baselines.}
\hspace*{-1.5cm}
\begin{threeparttable}
\begin{tabular}{@{}c@{\hspace{1.5mm}}l@{\hspace{2.0mm}}
c@{\hspace{2.0mm}}
c@{\hspace{1mm}}c@{\hspace{1mm}}c@{}c@{\hspace{2.0mm}}
c@{\hspace{1mm}}c@{\hspace{1mm}}c@{}c@{\hspace{2.0mm}}
c@{\hspace{1mm}}c@{\hspace{1mm}}c@{}c@{\hspace{2.0mm}}
c@{\hspace{1mm}}c@{\hspace{1mm}}c@{}c@{\hspace{2.0mm}}
c@{\hspace{1mm}}c@{\hspace{1mm}}c@{}c@{\hspace{2.0mm}}
c@{\hspace{1mm}}c@{\hspace{1mm}}c@{}c
@{}}
& & & \multicolumn{3}{c}{\textit{Average}} & & \multicolumn{3}{c}{\textit{Appearance}} & & \multicolumn{3}{c}{\textit{Aroma}} & & \multicolumn{3}{c}{\textit{Palate}}& & \multicolumn{3}{c}{\textit{Taste}}& & \multicolumn{3}{c}{\textit{Overall}}\\
\cmidrule{4-6}\cmidrule{8-10}\cmidrule{12-14}\cmidrule{16-18}\cmidrule{20-22}\cmidrule{24-26}
& \textbf{Model} & \textbf{Acc.} & \textbf{P} & \textbf{R} &  \textbf{F} & & \textbf{P} & \textbf{R} & \textbf{F}  & & \textbf{P} & \textbf{R} & \textbf{F} & &  \textbf{P} & \textbf{R} & \textbf{F} & &  \textbf{P} & \textbf{R} & \textbf{F} & &  \textbf{P} & \textbf{R} & \textbf{F}\\
\toprule
\multirow{5}{*}{\rotatebox{90}{\textit{$\ell=20$}}}
& RNP & $81.1$ & $30.7$ & $22.1$ & $24.9$ & & $30.8$ & $23.2$ & $26.5$ & & $22.1$ & $21.0$ & $21.5$ & & $17.7$ & $24.1$ & $20.4$ & & $28.1$ & $16.7$ & $20.9$ & & $54.9$ & $25.8$ & $35.1$\\
& RNP-3P & $80.5$ & $29.1$ & $22.5$ & $25.0$ & & $30.4$ & $25.6$ & $27.8$ & & $19.3$ & $20.4$ & $19.8$ & & $10.3$ & $12.0$ & $11.1$ & & $43.9$ & $\underline{28.4}$ & $\underline{34.5}$ & & $41.6$ & $26.0$ & $32.0$\\
& Intro-3P & $\underline{85.6}$ & $24.2$ & $19.6$ & $21.3$ & & $28.7$ & $24.8$ & $26.6$ & & $14.3$ & $14.4$ & $14.3$ & & $16.6$ & $19.3$ & $17.9$ & & $24.2$ & $13.6$ & $17.4$ & & $37.0$ & $25.9$ & $30.5$\\
& InvRAT & $82.9$ & $\underline{41.8}$ & $\underline{31.1}$ & $\underline{34.8}$ & & $\underline{54.5}$ & $\underline{45.5}$ & $\underline{49.6}$ & & $\underline{26.1}$ & $\underline{27.6}$ & $\underline{26.9}$ & & $\underline{22.6}$ & $\underline{25.9}$ & $\underline{24.1}$ & & $\underline{46.6}$ & $27.4$ & $34.5$ & & $\underline{59.0}$ & $\underline{29.3}$ & $\underline{39.2}$\\
& ConRAT\tnote{*} (Ours) & $\mathbf{91.4}$ & $\mathbf{50.0}$ & $\mathbf{42.0}$ & $\mathbf{44.9}$ & & $\mathbf{57.8}$ & $\mathbf{53.0}$ & $\mathbf{55.3}$ & &$\mathbf{31.9}$ & $\mathbf{35.5}$ & $\mathbf{33.6}$ & & $\mathbf{29.0}$ & $\mathbf{36.3}$ & $\mathbf{32.3}$ & & $\mathbf{56.5}$ & $\mathbf{33.9}$ & $\mathbf{42.4}$ & & $\mathbf{74.9}$ & $\mathbf{51.0}$ & $\mathbf{60.7}$\\
\midrule
\multirow{5}{*}{\rotatebox{90}{\textit{$\ell=10$}}}
& RNP & $\underline{84.4}$ & $41.3$ & $16.6$ & $23.2$ & & $40.1$ & $12.0$ & $18.5$ & & $\mathbf{33.3}$ & $\mathbf{18.7}$ & $\mathbf{24.0}$ & & $\mathbf{25.1}$ & $\mathbf{17.4}$ & $\mathbf{20.6}$ & & $32.3$ & $9.8$ & $15.07$ & & $76.0$ & $25.1$ & $37.8$\\
& RNP-3P & $83.1$ & $31.1$ & $13.5$ & $18.6$ & & $41.8$ & $19.2$ & $26.3$ & & $22.2$ & $12.4$ & $15.9$ & & $16.5$ & $10.4$ & $12.7$ & & $33.2$ & $10.6$ & $16.1$ & & $41.9$ & $14.7$ & $21.8$\\
& Intro-3P & $80.9$ & $21.8$ & $10.8$ & $14.3$ & & $51.0$ & $26.0$ & $34.4$ & & $18.8$ & $9.7$ & $12.8$ & & $16.5$ & $10.6$ & $12.9$ & & $9.7$ & $2.6$ & $4.1$ & & $13.1$ & $5.2$ & $7.4$\\
& InvRAT & $81.9$ & $\underline{47.1}$ & $\underline{17.8}$ & $\underline{25.5}$ & & $\mathbf{59.4}$ & $\underline{26.1}$ & $\mathbf{36.3}$ & & $31.3$ & $15.5$ & $20.8$ & & $16.4$ & $9.6$ & $12.1$ & & $\underline{39.1}$ & $\underline{11.6}$ & $\underline{17.9}$ & & $\mathbf{89.1}$ & $\underline{26.4}$ & $\underline{40.7}$\\
& ConRAT\tnote{*} (Ours) & $\mathbf{91.3}$ & $\mathbf{48.1}$ & $\mathbf{20.1}$ & $\mathbf{28.0}$ & & $\underline{51.7}$ & $\mathbf{26.2}$ & $\underline{34.8}$ & & $\underline{32.6}$ & $\underline{17.4}$ & $\underline{22.7}$ & & $\underline{23.0}$ & $\underline{13.8}$ & $\underline{17.3}$ & & $\mathbf{45.3}$ & $\mathbf{13.1}$ & $\mathbf{20.3}$ & & $\underline{88.0}$ & $\mathbf{30.1}$ & $\mathbf{44.9}$\\
\end{tabular}
\begin{tablenotes}
     \item[*] The model is only trained on the overall label and does not have access to the other ground-truth labels.
  \end{tablenotes}
\end{threeparttable}
\end{table}

Similar to Section~\ref{rq1}, we compare the generated rationales with the human annotations on the five aspects and the average performance. The main results are shown in Table~\ref{exp_rq2_beer_objective}. On average, ConRAT achieves the best performance while trained only on the overall ratings. This shows that the generated concepts, learned in an unsupervised manner, are separable, consistent, and correlated with human judgments to a certain extent. For the concept length $\ell=20$, ConRAT produces significant superior results for all aspects, whereas the difference with InvRAT is less pronounced for $\ell=10$. Finally, ConRAT's concepts lead to the highest accuracy and respect the grounding desideratum, thanks to the teacher.

We hypothesize that the baselines underperform due to the high correlations among the aspect ratings. Thus, they are more prone to pick up spurious correlations between the input features and the output. By considering multiple concepts simultaneously, ConRAT reduces the impact of spurious correlations. Regarding Intro-3P and RNP-3P, both suffer from instability issues due to the policy gradient \citep{chang2020invariant,yu-etal-2019-rethinking}.

\begin{figure}[!t]
\centering
\begin{tabular}{@{}c@{}}
\includegraphics[width=0.45\textwidth]{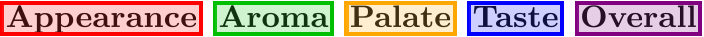}\\
     ConRAT (Ours)\\
     \includegraphics[width=0.6\textwidth,height=3.25cm]{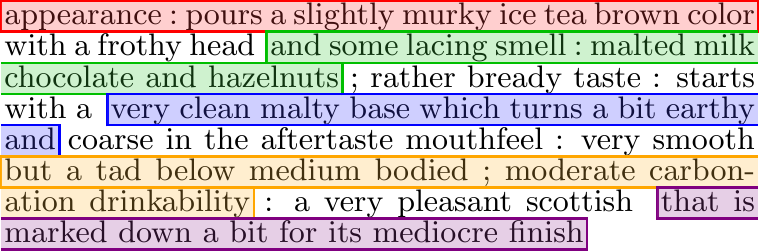}\\
  RNP \citep{lei-etal-2016-rationalizing}\\
  \includegraphics[width=0.6\textwidth,height=3.25cm]{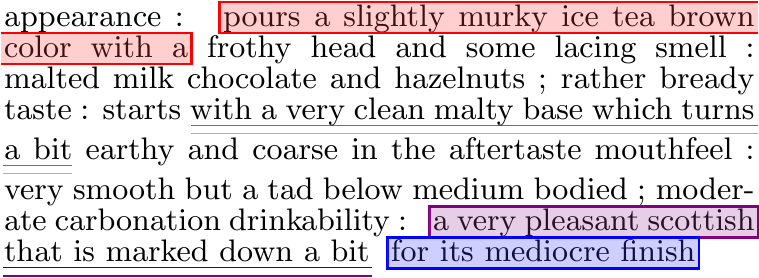}\\
  InvRAT \citep{chang2020invariant}\\
  \includegraphics[width=0.6\textwidth,height=3.25cm]{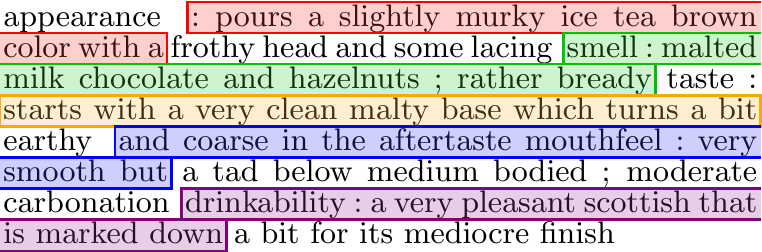}
\end{tabular}
\caption{\label{sample}Concepts generated (with $\ell$=10)~for~a beer review. \underline{Underline} highlights ambiguities. The color depicts the aspects: Appearance, Aroma, Palate, Taste, and~Overall. \textbf{ConRAT is trained only on the overall label}. RNP produces the same rationale for the aspect Aroma and Palate. InvRAT infers wrong rationales for the aspects Palate and Taste. In contrast, ConRAT reduces the impact of spurious correlations.} 
\end{figure}

We visualize an example in Figure~\ref{sample}. We observe that ConRAT induces interpretable~concepts, while the best baselines suffer from spurious correlations. By reading our concepts alone, humans will easily predict the aspect label and its polarity.


\subsubsection{Subjective Evaluation.}
\label{rq2_exp_subj}

We conduct a human evaluation using Amazon’s Mechanical Turk (details in Appendix~\ref{app_rq2_hum}) to judge the understandability of the concepts. Following \cite{chang2019game}, we sampled 100 balanced reviews from the hold-out set for each aspect, model, and concept length, resulting in 5,000 samples. We showed the examples in random order. An evaluator is presented with the concept generated by one of the five methods (unselected words are not visible). We credit a success when the evaluator guesses the true aspect label and its sentiment. We report the success rate as the performance metric. A random guess has a 10\% success rate.

\begin{figure}[!t]
\centering
\includegraphics[width=0.6\textwidth]{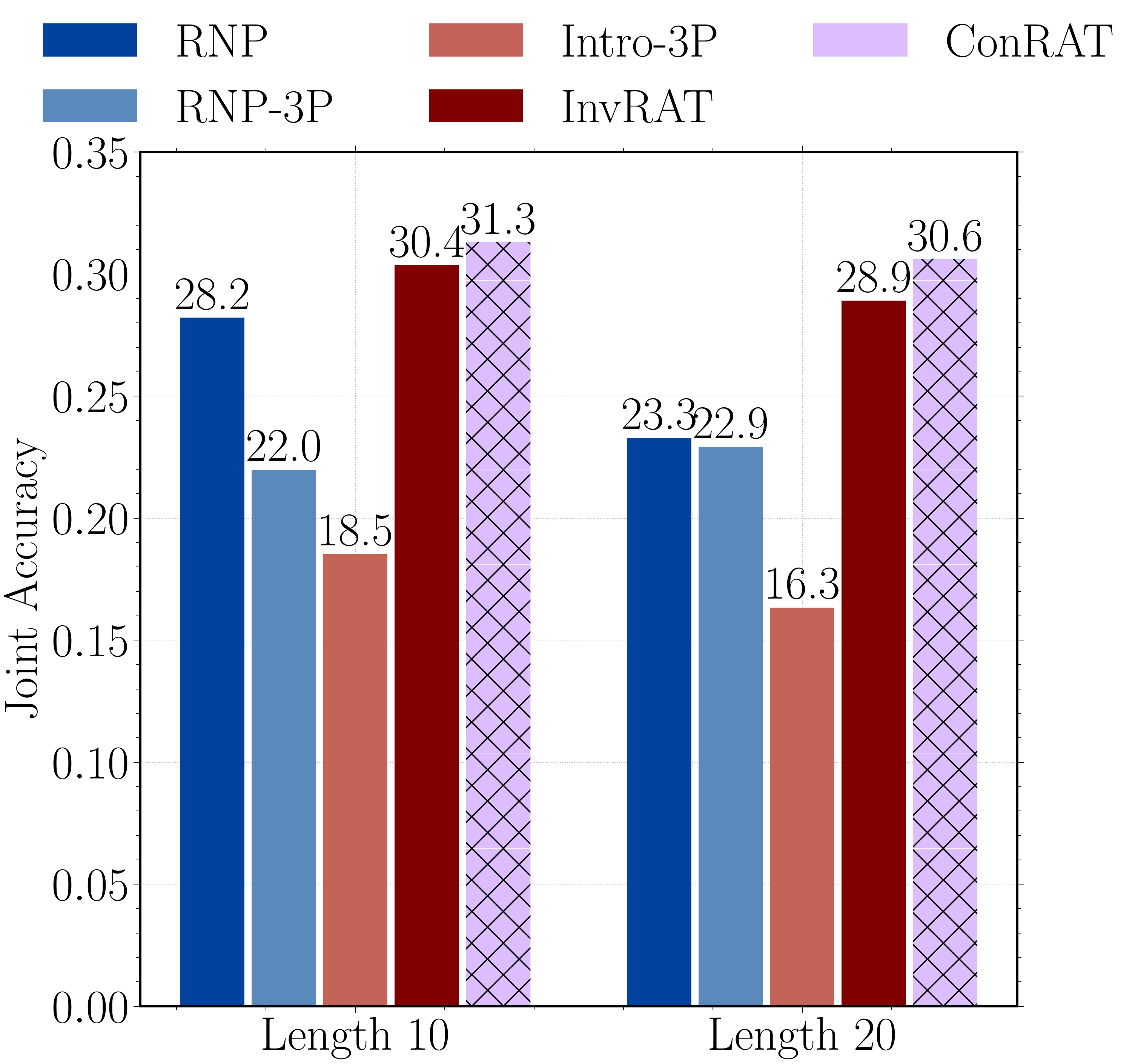}
\caption{\label{exp_rq2_beer_sub}Subjective performance of rationales for the multi-aspect beer reviews. Evaluators need to guess both the sentiment and what aspect the concept is about, which makes random guess only 10\%. ConRAT infers meaningful rationales according to human annotators.}
\end{figure}

Figure~\ref{exp_rq2_beer_sub} shows the main results. Similar to the objective evaluation, ConRAT reaches the best performance, followed by InvRAT. Moreover, ConRAT only requires a single training on the overall aspect. It emphasizes that the discovered concepts satisfy the fidelity and diversity desiderata and better correlate with human judgments compared with supervised baselines. 
We study the error rates on each aspect. The Aroma and Palate aspects cause the highest for all models. One possible reason is that users confuse these with the aspect Taste, hence their high correlations in rating scores \citep{antognini2019multi}.

\subsection{RQ 3: How does the number of concepts $K$ in ConRAT affect the performance?}
\label{rq3}

We study the impact of the number of concepts $K$ in ConRAT on the performance, as discussed in Section~\ref{rq2_exp_obj}. We set the number of concepts to~the number of aspects ($K$=5) and then increase it to $K$=10 and $K$=20. We prune ConRAT to keep only the five most dissimilar concepts (see Section~\ref{sec_improv_perfs}). 

Results are shown in Table~\ref{exp_rq3}. First, we observe that the performance is already better than the baselines in Table~\ref{exp_rq2_beer_objective} with $K$=5. Second, when increasing $K$ and pruning ConRAT, the performance is boosted further. However, we remark that the interpretability of the concepts follows a bell curve and significantly decreases when $K$=20. One potential reason is that we expect overlaps between the discriminative concepts that relate to beer aspects\footnote{As shown in Table~\ref{dataset_description}, the mean length of beer reviews is 184 words. With $\ell$=20 and $C$=20, 400 words are highlighted.}.
Thus, the five most dissimilar concepts might align less with human-defined concepts.

\begin{table}[t]
    \centering
   \caption{\label{exp_rq3}Impact of the number of concepts in ConRAT on the objective performance for the beer reviews.}
\begin{threeparttable}
\begin{tabular}{@{}
cc
cccc@{}}
& & & \multicolumn{3}{c}{\textit{Average}} \\
\cmidrule{4-6}
\multicolumn{2}{c}{\textbf{\#Concepts}} & \textbf{Acc.} & \textbf{P} & \textbf{R} & \textbf{F}\\
\toprule
\multirow{3}{*}{\rotatebox{90}{\textit{$\ell=20$}}}
& $K=5\ \ $ & $90.95$ & $48.96$ & $37.59$ & $41.37$\\
& $K=10$ & $\mathbf{91.35}$ & $\mathbf{50.02}$ & $\mathbf{41.96}$ & $\mathbf{44.86}$\\
& $K=20$ & $90.24$ & $37.78$ & $31.19$ & $32.84$\\
\midrule
\multirow{3}{*}{\rotatebox{90}{\textit{$\ell=10$}}}
& $K=5\ \ $ & $89.64$ & $47.60$ & $19.23$ & $26.90$\\
& $K=10$ & $\mathbf{91.25}$ & $\mathbf{48.12}$ & $\mathbf{20.11}$ & $\mathbf{27.97}$\\
& $K=20$ & $91.05$ & $35.71$ & $14.84$ & $20.71$\\
\end{tabular}
\end{threeparttable}
\end{table}

\begin{table}[!t]
    \centering
   \caption{\label{exp_rq4}Ablation study of ConRAT with five concepts. The results confirm our intuition and that our regularizers enforce the interpretable desiderata. }
\begin{threeparttable}
\begin{tabular}{@{}
lcccc@{}}
& & \multicolumn{3}{c}{\textit{Average}} \\
\cmidrule{3-5}
\textbf{Model}& \textbf{Acc.} & \textbf{P} & \textbf{R} & \textbf{F}\\
\toprule
ConRAT  & $89.64$ & $47.60$ & $19.23$ & $26.90$\\
\ \ - No $\mathcal{L}_{overlap}$ & $91.05$ & $31.50$ & $13.16$ & $18.37$\\
\ \ - No $\mathcal{L}_{div}$ & $90.85$ & $34.49$ & $11.69$ & $16.95$\\
\ \ - No $s_\theta(\cdot):$$s_k=1 \forall k$ & $89.74$ & $43.13$ & $14.95$ & $21.42$\\
\ \ - No Teacher & $86.52$ & $45.31$ & $19.65$ & $26.99$\\
\end{tabular}
\end{threeparttable}
\end{table}

\subsection{RQ 4: How does each module of ConRAT contribute to the overall performance?}

Finally, we analyze the importance of each module in an ablation study. To avoid any bias from pruning, we set the number of concepts to five\footnote{We obtain similar results with $K$=10 and $K$=20.}.

Table~\ref{exp_rq4} shows the results. When ConRAT ignores the overlapping or the diversity regularizer, we observe a large drop in the rationale performance. This is expected as the diversity desideratum is not encouraged anymore. However, we remark that the sentiment prediction accuracy increases, which is possibly caused by spurious correlation with the ground-truth label.
When all concepts are considered ($s_k=1$ $\forall k$), we note that the sentiment accuracy stays similar. However, the objective performance decreases by 10\% for the precision and more than 20\% for the recall and F1 score. These results align with prior work: users write opinions about the topics they care about \citep{musat2015personalizing,antognini2020interacting}. ConRAT reduces the noise at training by selecting concepts described in the current document. Finally, the teacher model helps ConRAT to boost the sentiment accuracy by more than 3\% absolute score, without affecting the rationale quality.

\section{Conclusion}

Providing explanations for automated predictions carries much more impact, increases transparency, and might even be vital. Previous works have~proposed using rationale methods to explain the prediction of a target variable. However, they do not properly capture the multi-faceted nature of useful rationales. In this chapter, we proposed ConRAT, a novel self-explaining model that extracts a set of concepts and explains the outcome with a linear aggregation of concepts, similar to how humans~reason.
Our second contribution is two novel regularizers that guide ConRAT to generate interpretable concepts. Experiments on both single- and multi-aspect sentiment classification datasets show that ConRAT, by using only the overall label, is the first to provide superior rationale and predictive performance compared with supervised state-of-the-art methods trained for each aspect label. Moreover, ConRAT produces concepts considered superior in interpretability when evaluated by humans.

%% file: main/IJCAI2021/main_final.tex
\chapter{Generating Explanations and Interacting with Them through Critiquing}
\label{chapter_ijcai_2021}


\section{Preface}

\textbf{Contribution and Sources.}\hspace{\parindent} This chapter is largely based on \cite{antognini2020interacting} and \cite{multi_step_demo}. The detailed individual contributions are listed below using the CRediT taxonomy \citep{brand2015beyond} (terms are selected as applicable).

Authors' contributions in \cite{antognini2020interacting}:
\begin{table}[!h]
\begin{tabular}{@{}l@{\hspace{1mm}}l@{}}
Diego Antognini (author): & Conceptualization, Methodology, Software, Validation,\\
& Investigation, Formal Analysis, Writing -- Original Draft,\\
&Writing -- Review \& Editing.\\ \\
Claudiu Musat: & Writing -- Review \& Editing (supporting), Supervision (supporting). \\ \\
Boi Faltings: & Writing -- Review \& Editing, Administration, Supervision.
\end{tabular}
\end{table}

Authors' contributions in \cite{multi_step_demo} (* denotes equal contributions):
\begin{table}[!h]
\begin{tabular}{@{}l@{\hspace{1mm}}l@{}}
Diego Antognini* (author): & Conceptualization (lead), Methodology (lead), Software (equal), \\
& Writing -- Review \& Editing (lead).\\ \\
Diana Petrescu*: & Software (equal), Writing -- Original Draft, \\
& Writing -- Review \& Editing (supporting). \\ \\
Boi Faltings: & Conceptualization (supporting), Methodology (supporting),\\
& Administration, Supervision.
\end{tabular}
\end{table}

\textbf{Summary.}\hspace{\parindent} Using personalized explanations to support recommendations has been shown to increase trust and perceived quality. However, to actually obtain better recommendations, there needs to be a means~for users to modify the recommendation criteria by interacting with the explanation. In this chapter, we present~a novel explanation technique using aspect markers that learns to generate personalized explanations of recommendations from review texts, and we show that human users significantly prefer~these explanations over those produced by state-of-the-art techniques.
Our work's most important innovation is that it allows users to react to a recommendation by critiquing the textual explanation: removing (symmetrically adding) certain aspects they dislike or that are no longer relevant (symmetrically that are of interest). The system updates its user model~and~the resulting recommendations according to the critique. This is based on a novel unsupervised critiquing method for single- and multi-step critiquing with textual explanations. Empirical results show that our system achieves good performance in adapting to the preferences expressed in multi-step critiquing and generates consistent explanations. Finally, we introduce four web interfaces helping users make decisions and find their ideal item. Because our system is model-agnostic (for both recommender and critiquing models), it allows great flexibility and further extensions. Our interfaces are, above all, a useful tool for research on recommendation with critiquing. They allow to test such systems on a real use case and for highlighting some limitations of these approaches.

\section{Introduction}
\textbf{Explanations of recommendations are beneficial}. Modern recommender systems accurately capture users' preferences and achieve high performance. But, their performance comes at the cost of increased complexity, which makes them seem like black boxes to users. This may result in distrust or rejection of the recommendations~\citep{herlocker2000explaining,ExplainingRecommendation}.

There is thus value in providing \textit{textual explanations} of the recommendations, especially on e-commerce websites, because such explanations enable users to understand why~a particular item has been suggested and hence to make better decisions \citep{chang2016crowd,bellini2018knowledge,ExplainingRecommendation}. Furthermore, explanations increase overall system transparency \citep{ExplainingRecommendation,sinha2002role} and trustworthiness \citep{zhang2018exploring,Kunkel2018TrustrelatedEO}. 

However, not all explanations are equivalent. \cite{kunkel2019let,chang2016crowd} showed~that highly personalized justifications using \textit{natural language} lead to substantial increases in perceived recommendation quality and trustworthiness compared to simpler explanations, such as aspect \citep{keyphraseExtractionDeep}, template \citep{zhang2014explicit}, or similarity \citep{herlocker2000explaining}.

\begin{figure}[!t]
\centering
\includegraphics[width=0.9\linewidth]{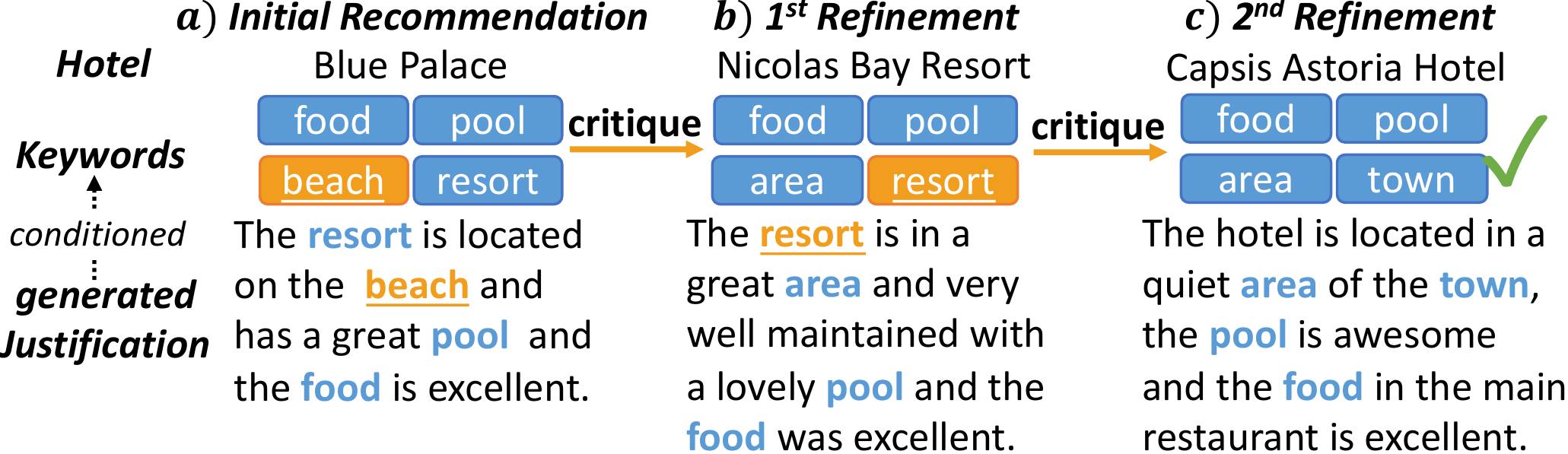}
\caption{\label{example_critiquing}A flow of conversational critiquing over two time steps.~a) The system proposes to the user a recommendation with a keyphrase explanation and a justification. The user can interact with the explanation and critique phrases. b) A new recommendation is produced from the user's profile and the critique. c) This process repeats until the user accepts the recommendation and ceases to provide critiques.}
\end{figure}

A second, and more important, benefit of explanations is that they provide a basis for feedback~\citep{10.1145/3442381.3450123}: if a~user~is unsatisfied~with~a recommendation, understanding what generated it allows them to \textit{critique} it (Figure~\ref{example_critiquing}). Critiquing -- a~conversational method of incorporating user preference feedback regarding item attributes into the recommended list of items -- has several advantages. First, it allows the system to correct and improve an incomplete or inaccurate~model of the user's preferences \citep{faltings2004solution}, which improves the user's decision accuracy \citep{pu2005integrating,chen2012critiquing}. Compared to preference elicitation, critiquing is more flexible: users can express preferences in any order and on any criteria~\citep{reilly2005explaining}.

\textbf{Useful explanations are hard to generate}. Prior research has employed users' reviews to capture their preferences and writing styles \citep[e.g.,][]{ni2018personalized,li2019towards,dong2017learning}. From past reviews, they generate \textit{synthetic} ones that serve as personalized \textit{explanations} of ratings given by users. However, many reviews are noisy, because they partly describe experiences or endorsements. It is thus nontrivial to identify meaningful justifications inside reviews. \cite{ni-etal-2019-justifying} proposed a pipeline for identifying justifications from reviews and asked humans to annotate them. \cite{chen2019co,chen2020towards} set the justification as the first sentence. However, these notions of justification were ambiguous,~and they assumed that a review contains only one justification.

\begin{figure}[!t]
\centering
\includegraphics[width=0.8\linewidth]{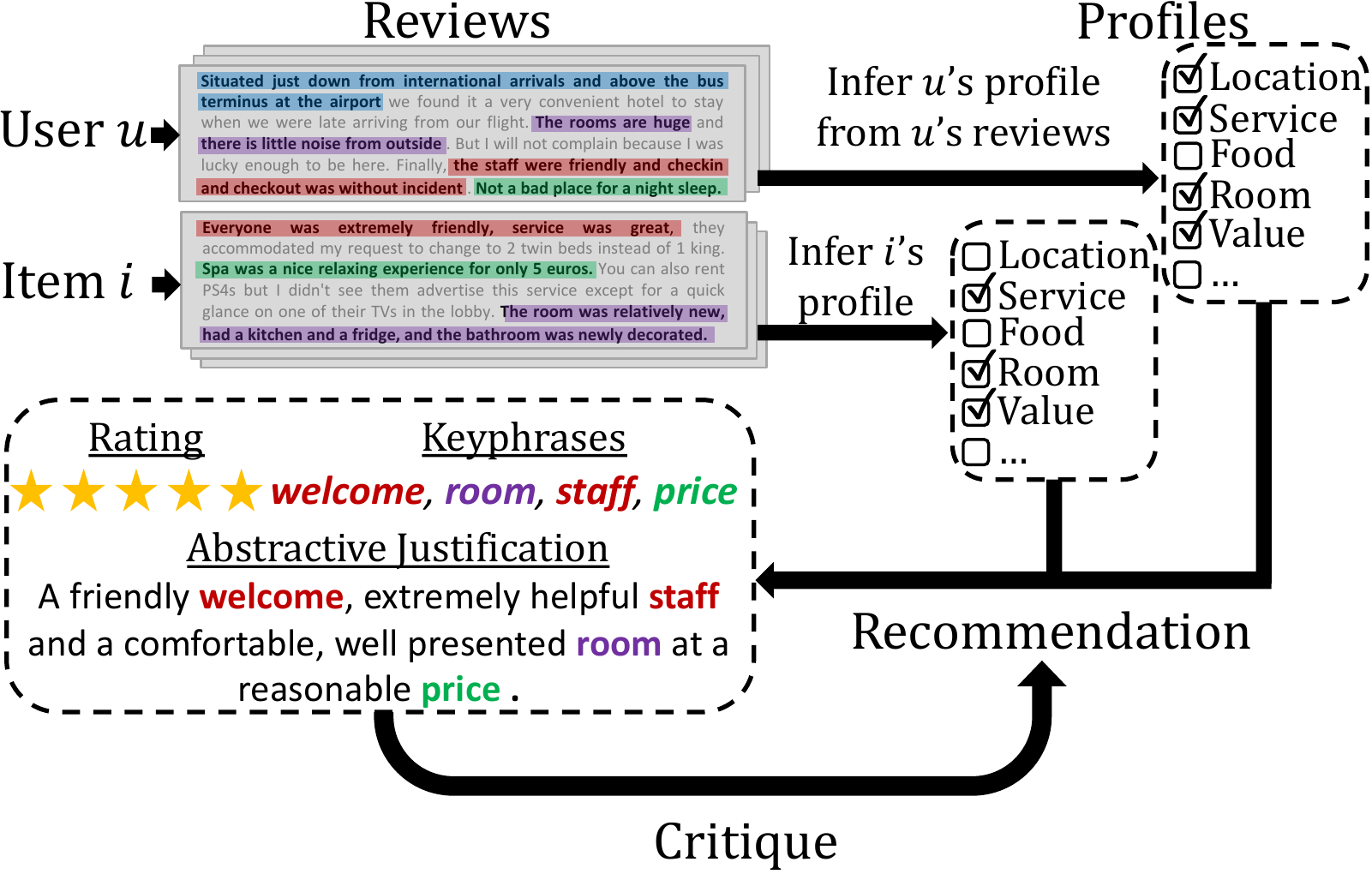}
\caption{\label{figure_pipeline}For reviews written by a user $u$ and a set of~reviews about an item $i$, we extract the justifications for each~aspect rating and implicitly build an interest profile. T-RECS outputs a personalized recommendation with~two~explanations: the keyphrases reflecting the overlap between the two profiles, and a synthetic justification conditioned on the~latter.}
\end{figure}

Recently, \cite{antognini2019multi} solved these shortcomings by introducing a justification extraction system with no prior limits imposed on their number or structure. This is important because a user typically justifies his overall rating with multiple explanations: one for each aspect the user cares about \citep{musat2013,zhang2014explicit,musat2015personalizing}. 
The authors showed that there is a connection between faceted ratings and snippets within the reviews: for each subrating, there exists at least one text fragment that alone suffices to make the prediction. They employed a sophisticated attention mechanism to favor long, meaningful word sequences; we call these \textbf{\textit{markers}}. Building upon their study, we show that these \textit{markers} serve to create better user and item profiles and can inform better user-item pair justifications. Figure~\ref{figure_pipeline} illustrates the pipeline.

\textbf{From explanations to critiquing.} To reflect the overlap between the profiles of a user and an item, one can produce~a set of keyphrases and then a synthetic justification. The user can correct his profile, captured by the system, by \textit{critiquing} certain aspects he does not like or that are missing or not relevant anymore and obtain a new justification (Figure~\ref{example_critiquing}). \cite{keyphraseExtractionDeep} introduced a keyphrase-based critiquing method in which attributes are mined~from reviews, and users interact with them. However, their models need~an~extra autoencoder to project the critique back into the latent space, and it is unclear how the models behave in multi-step~critiquing.

We overcome these drawbacks by casting the critiquing as an unsupervised attribute transfer task: altering a keyphrase explanation of a user-item pair representation to the critique. To this end, we entangle the user-item pair with the explanation in the same latent space. At inference, the keyphrase classifier modulates the latent representation until the classifier identifies it as the critique vector.

In this chapter, we address the problem recommendation with fine-grained explanations. We first demonstrate how~to extract multiple relevant and personalized justifications from the user's reviews to build a profile that reflects his preferences and writing style (Figure~\ref{figure_pipeline}). Second, we propose T-RECS, a recommender with explanations. T-RECS explains a rating by first inferring a set of keyphrases describing the intersection between the profiles of a user and an item. Conditioned on the keyphrases, the model generates a synthetic personalized justification. We then leverage these explanations in an unsupervised critiquing method for single- and multi-step critiquing. We evaluate our model using two real-world recommendation datasets. T-RECS outperforms strong baselines in explanation generation, effectively re-ranks recommended items in single-step critiquing. Finally, T-RECS also better models the user's preferences in multi-step critiquing while generating consistent textual justifications.

\section{Related Work}
\subsection{Textual Explainable Recommendation}
Researchers have investigated many approaches to generating textual explanations of recommended items for users. \cite{McAuley:2013:HFH:2507157.2507163} proposed a topic model to discover latent factors from reviews~and explain recommended items. \cite{zhang2014explicit} improved the understandability of topic words and aspects by filling template sentences. 

Another line of research has generated synthetic reviews as explanations. Prior studies have employed users' reviews and tips to capture their preferences and writing styles. \cite{catherine2017transnets} predicted and explained ratings by encoding the user's review and identifying similar~reviews. \cite{lu2018like,chen2019co} extended the previous work~to~generate short synthetic reviews. \cite{dual2020} optimized both tasks in dual forms. \cite{dong2017learning,costa2018automatic} proposed an attribute-to-sequence model to learn how to generate reviews given categorical attributes. \cite{ni2018personalized} improved review generation by leveraging aspect information using a sequence-to-sequence model with attention. Instead~of reviews, others have generated tips \citep{li2017neural,li2019persona}.
However, the tips are scarce and uninformative \citep{chen2019co}; many reviews are noisy because they describe partially general experiences or endorsements \citep{ni-etal-2019-justifying}. 

\cite{ni-etal-2019-justifying} built a sequence-to-sequence model conditioned on the aspects to generate relevant explanations for an existing recommender system;~the~fine-grained aspects are provided by the user in the inference. They identified justifications from reviews~by~segmenting them into elementary discourse units (EDU) \citep{mann1988rhetorical} and asking annotators to label them as ``good'' or ``bad'' justifications. \cite{chen2019co} set the justification as the first sentence. All assumed that a review contains only one justification. Whereas their notions of justification were ambiguous, we extract multiple justifications from reviews using \textit{markers} that justify subratings. Unlike their models, ours predicts keyphrases on which the justifications are conditioned and integrates critiquing.

\subsection{Critiquing}
Refining recommended items allows users~to interact with the system until they are satisfied. Some methods are example critiquing \citep{williams1982rabbit}, in which users critique a set of items; unit critiquing~\citep{unitcritiquing}, in which users critique an item's attribute and request another one instead; and compound critiquing \citep{reilly2005explaining} for more~aspects. \cite{mccarthy2010experience} collaboratively utilized critiques from users. The major drawback of these approaches is the assumption of a fixed set of known attributes.

\cite{keyphraseExtractionDeep} circumvented this limitation by extending the neural collaborative filtering model \citep{he2017neural}. First, the model explains a recommendation by predicting a set of keywords (mined from users' reviews). In \cite{chen2020towards}, based on \cite{chen2019co}, the model samples only one keyword via the Gumbel-Softmax function. Our work applies a deterministic strategy similar to \cite{keyphraseExtractionDeep}.

 Second, \cite{keyphraseExtractionDeep} project the critiqued keyphrase explanations back into the latent space, via an autoencoder that perturbs the training, from which the rating and the explanation are predicted. In this manner, the user's critique~modulates his latent representation. The model of \cite{chen2020towards} is trained in a two-stage manner: one to perform recommendation and predict one keyword and another to learn critiquing from online feedback, which requires additional data. By contrast, our model is simpler and learns critiquing in an~unsupervised fashion: it iteratively edits the latent representation until the new explanation matches the critique. 
 Finally, \cite{luo2020b}~examined various linear aggregation methods on latent representations for multi-step critiquing. In comparison, our gradient-based critiquing iteratively updates the latent representation for each~critique.

\begin{figure}[!t]
\centering
\includegraphics[width=.6\linewidth]{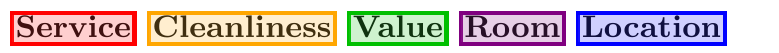}
\includegraphics[width=0.85\linewidth]{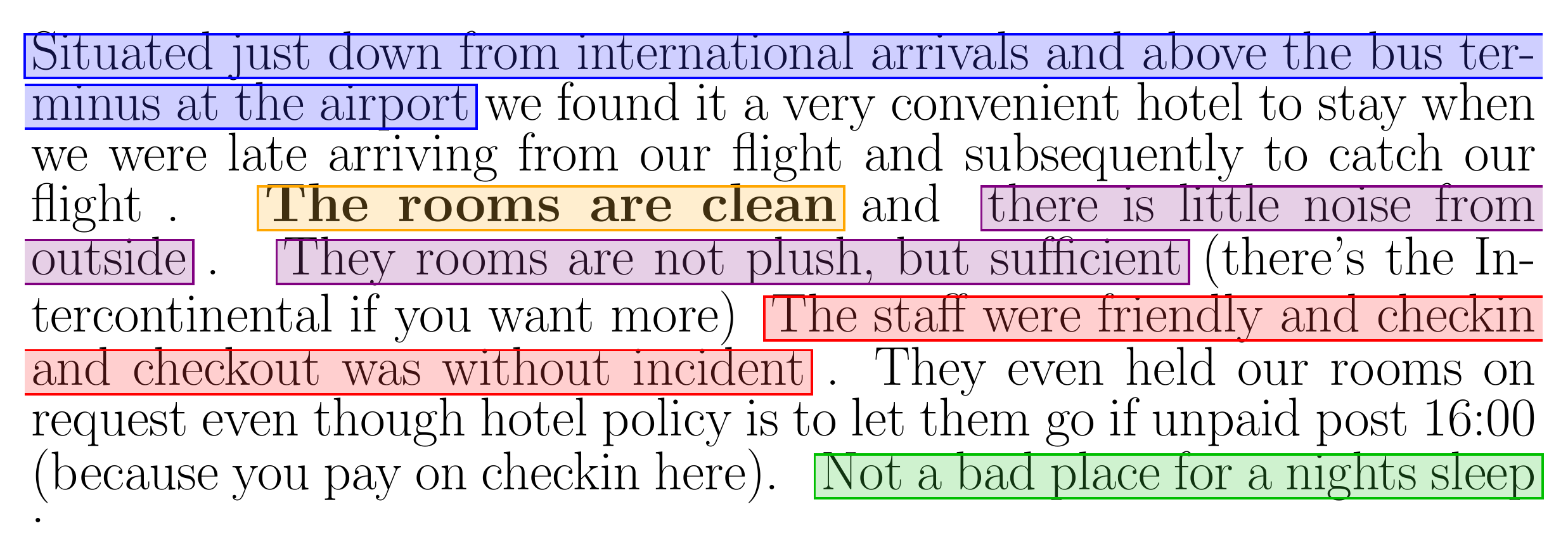}
\includegraphics[width=.45\linewidth]{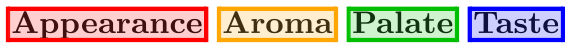}
\includegraphics[width=0.85\linewidth,height=4.0cm]{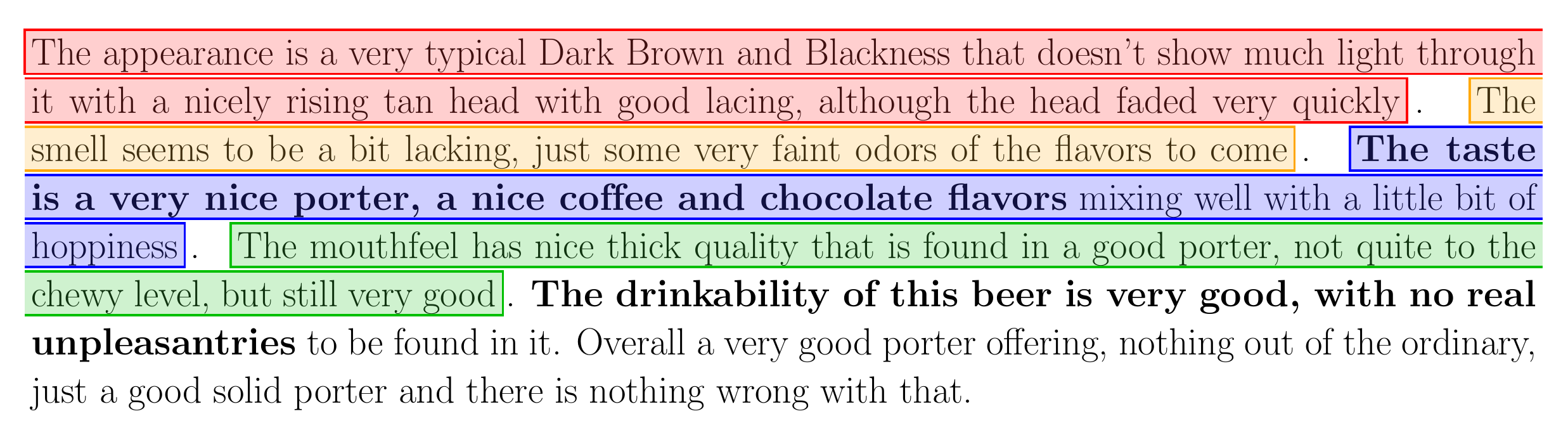}
\caption{\label{examples_masks_no_edus}Extracted justifications from a hotel review and a beer review. The inferred \textit{markers} depict the excerpts that explain the ratings of the aspects: Service, Cleanliness, Value, Room, and Location. We denote~in \textbf{bold} the EDU-based justification from the model of~\protect\citep{ni-etal-2019-justifying}.}
\end{figure}

\section{Extracting Justifications from Reviews}
\label{sec_extr_just}
In this section, we introduce the pipeline for extracting high-quality and personalized justifications from users' reviews. We claim that a user justifies his overall experience with multiple explanations: one for each aspect he cares about. Indeed, it has been shown that users write opinions about the topics they care about \citep{zhang2014explicit,musat2015personalizing}. Thus, the pipeline must satisfy two requirements:
\begin{enumerate}[noitemsep,topsep=0pt]
    \item extract text snippets that reflect a rating or subrating, and
    \item be data~driven and scalable to mine massive review corpora and to construct a large personalized recommendation justification dataset.
\end{enumerate}

In Chapter~\ref{chapter_aaai2021}, we proposed the multi-target masker (MTM) to find text fragments that explain faceted ratings in an unsupervised manner. MTM fulfills the two requirements. For each word, the model computes a distribution over the aspect set, which corresponds to the aspect ratings (e.g., service, location) and ``not aspect.'' In parallel, the model minimizes the number of selected words and discourages aspect transition between consecutive words. These two constraints guide the model to produce long, meaningful sequences of words called \textbf{\textit{markers}}. The model updates its parameters by using the inferred \textit{markers} to predict the aspect sentiments jointly and improves the quality of the \textit{markers} until convergence. 

Given a review, MTM extracts the \textit{markers} of each aspect. A sample is shown in Figure~\ref{examples_masks_no_edus}. Similarly to \cite{ni-etal-2019-justifying}, we filter out \textit{markers}~that are unlikely to be suitable justifications: including third-person pronouns or being too short.~We use the constituency parse tree to ensure that \textit{markers} are verb phrases. The detailed processing is available in Appendix~\ref{app_process_filtering}.

\section{T-RECS: A Multi-Task Transformer with Explanations and Critiquing}
\label{sec_model}
\begin{figure}[!t]
\hspace*{-0.75cm}
\centering
   \includegraphics[width=1.1\linewidth]{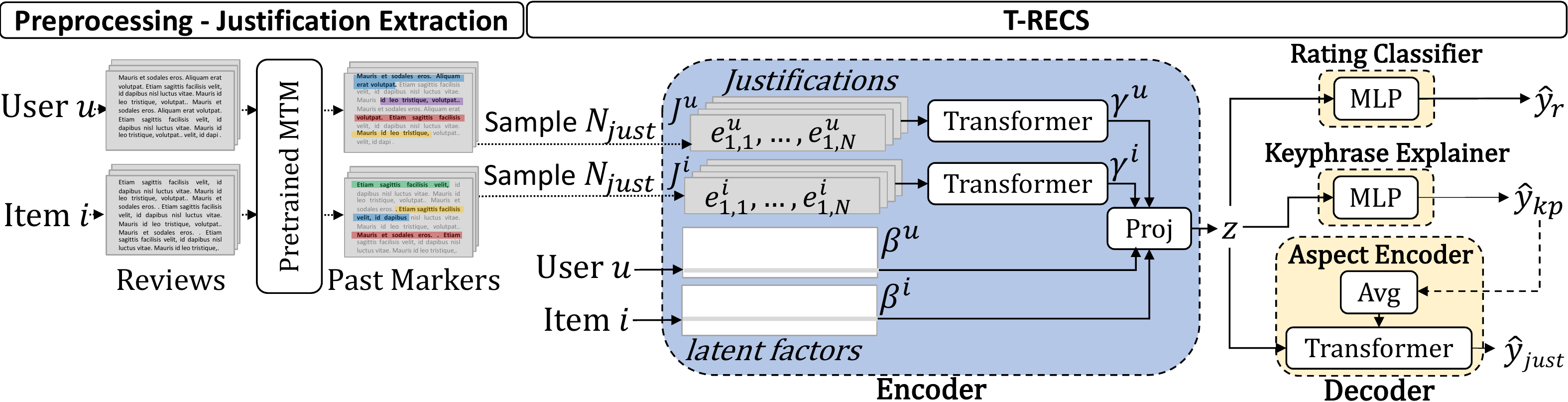}
   \caption{\label{general_architecture}(Left) Preprocessing for the users and the items.  
   For each user $u$ and item $i$, we first extract \textit{markers} from their past reviews (highlighted~in color), using the pretrained multi-target masker (see Section \ref{sec_extr_just}), that become their respective~justifications. Then, we sample $N_{just}$ of them and build the justification references $J^u$ and~$J^i$, respectively.
   (Right) T-RECS architecture. Given a user $u$ and an item $i$ with their justification references $J^u$, $J^i$ and latent factors $\beta^u$, $\beta^i$, T-RECS produces a joint embedding $\bb{z}$ from which it predicts a rating $\bb{\hat{y}}_r$, a keyphrase explanation $\bb{\hat{y}}_{kp}$, and a natural language justification $\bb{\hat{y}}_{just}$ conditioned on $\bb{\hat{y}}_{kp}$.}
\end{figure}

Figure~\ref{general_architecture} depicts the pipeline and our proposed T-RECS model. Let $U$ and $I$ be the user and item sets. For each user $u \in U$ (respectively an item $i \in I$), we extract \textit{markers} from the user's reviews on the training set, randomly select $N_{just}$, and build a justification~reference~$J^u$ (symmetrically~$J^i$). 

Given a user $u$, an item $i$, and their justification history $J^u$ and $J^i$, our goal is to~predict
\begin{enumerate*}
	\item a~rating~$\bb{y}_r$,
	\item a keyphrase explanation $\bb{y}_{kp}$ describing the relationship between $u$ and $i$, and 
	\item a natural language justification $\bb{y}_{just}=\{w_1, ..., w_N\}$
\end{enumerate*}, where $N$ is the length of the justification. $\bb{y}_{just}$ explains the rating $\bb{y}_r$ conditioned~on~$\bb{y}_{kp}$.

\subsection{Model Overview}
For each user and item, we extract \textit{markers} from their past reviews (in the train set) and build their justification history $J^u$ and $J^i$, respectively (see Section \ref{sec_extr_just}). T-RECS~is divided into four submodels: an \textbf{Encoder} $E$, which produces the latent representation $\bb{z}$ from the historical justifications and latent factors~of~the~user~$u$ and the item $i$; a \textbf{Rating Classifier} $C^r$, which classifies the rating $\bb{\hat{y}_r}$; a \textbf{Keyphrase Explainer} $C^{kp}$, which predicts~the keyphrase explanation $\bb{\hat{y}}_{kp}$ of the latent representation~$\bb{z}$; and a \textbf{Decoder}~$D$, which decodes the justification $\bb{\hat{y}}_{just}$ from~$\bb{z}$ conditioned on $\bb{\hat{y}}_{kp}$, encoded via the \textbf{Aspect Encoder} $A$. T-RECS involves four~functions:
\begin{equation}
	\bb{z}=E(u,i), \quad\quad \bb{\hat{y}}_r=C^r(\bb{z}), \quad\quad \bb{\hat{y}}_{kp} = C^{kp}(\bb{z}), \quad\quad \bb{\hat{y}}_{just} = D(\bb{z}, A(\bb{\hat{y}}_{kp})).
\end{equation}

The above formulation contains two types of personalized explanations: a list of keyphrases $\bb{\hat{y}}_{kp}$ that reflects the different aspects of item $i$ that the user $u$ cares about (i.e., the overlap between their profiles) and a natural language explanation~$\bb{\hat{y}}_{just}$ that justifies the rating, conditioned on $\bb{\hat{y}}_{kp}$. The set of keyphrases is mined from the reviews and reflects the different~aspects deemed important by the users. The keyphrases enable an interaction mechanism: users can express agreement or disagreement with respect to one or multiple aspects and hence critique the recommendation.

\subsubsection{Entangling user-item.} A key objective of T-RECS is to build a powerful latent representation. It accurately captures user and item profiles with their writing styles and entangles the rating, keyphrases, and~a natural language justification. Inspired by the superiority of the Transformer~\citep{vaswani2017attention} for text generation~tasks~\citep{abs-1912-02164,devlin-etal-2019-bert,radford2019language,abs-1905-12926}, we propose a Transformer-based encoder that learns latent personalized features from users' and items' justifications. We first pass each justification $J^u_j$ (respectively $J^i_j$) through the Transformer to compute the intermediate representations $\bb{h}^u_j$ (respectively~$\bb{h}^i_j$)\footnote{We propose a solution to address the case where users do not have written reviews in Appendix~\ref{app_cold_start_trecs}.}. We apply a sigmoid function on the representations and average them to get $\bb{\gamma}^u$ and $\bb{\gamma}^i$:\begin{equation}
    \bb{\gamma}^u = \frac{1}{|J^u|}\sum\nolimits_{j \in J^u} \sigma(\bb{h}^u_j), \quad\quad \bb{\gamma}^i = \frac{1}{|J^i|}\sum\nolimits_{j \in J^i} \sigma(\bb{h}^i_j).
\end{equation}In parallel, the encoder maps the user $u$ (item $i$) to the latent factors $\bb{\beta}^u$ ($\bb{\beta}^i$) via an embedding layer. We compute~the~latent representation $\bb{z}$ by concatenating the latent personalized features and factors and applying a linear projection:
~$\bb{z}=$ $E(u,i)= W[\bb{\gamma}^u \mathbin\Vert \bb{\gamma}^i \mathbin\Vert \bb{\beta}^u \mathbin\Vert \bb{\beta}^i] + \bb{b},$
where~$\mathbin\Vert$~is the concatenation operator, and $W$ and $\bb{b}$ the projection parameters.

\subsubsection{Rating Classifier \& Keyphrase Explainer.} Our framework classifies the interaction between the user~$u$ and item $i$ as positive or negative. Moreover, we predict the keyphrases that describe the overlap of their profiles.~Both models are a two-layer feedforward neural network with LeakyRelu activation function. Their respective losses~are:\begin{equation}
\begin{split}
    \mathcal{L}_{r}(C^r(\bb{z}),\bb{y}_r) &= (\bb{\hat{y}}_r - \bb{y}_r)^2,\\
    \mathcal{L}_{kp}(C^{kp}(\bb{z}),\bb{y}_{kp}) &= -\sum\nolimits_{k=1}^{|K|} y_{kp}^k \log \hat{y}_{kp}^k,
\end{split}
\end{equation}where $\mathcal{L}_{r}$ is the mean square error, $\mathcal{L}_{kp}$ the binary cross-entropy, and $K$ the whole set of keyphrases.

\subsubsection{Justification Generation.} The last component consists of generating the justification. Inspired by ``plan-and-write''~\citep{yao2019plan}, we advance the personalization of the justification by incorporating the keyphrases $\bb{\hat{y}}_{kp}$. In other words, T-RECS generates a natural language justification conditioned on the
\begin{enumerate*}
    \item user,
    \item item, and
    \item aspects of the item that the user would consider important.
\end{enumerate*}
We encode these via the Aspect Encoder $A$ that takes the average of their word embeddings from the embedding layer in the Transformer. The aspect embedding is denoted by $\bb{a}_{kp}$ and added to the latent representation: $\bb{\tilde{z}} = \bb{z} + \bb{a}_{kp}$. Based on $\bb{\tilde{z}}$, the Transformer decoding block computes the output probability $\hat{y}^{t,w}_{just}$ for the word $w$ at time-step $t$. We train using teacher-forcing \citep{teacherforcing} and cross-entropy with label smoothing \citep{szegedy2016rethinking}:\begin{equation}
 \mathcal{L}_{just}(D(\bb{z}, \bb{a}_{kp}), \bb{y}_{just}) = - \sum\nolimits_{t=1}^{|\bb{y}_{just}|} CE(y^{t,w}_{just}, \hat{y}^{t,w}_{just}).
\end{equation}We train T-RECS end-to-end and minimize jointly~the~loss $\bb{\mathcal{L}} = \lambda_{r} \mathcal{L}_{r} + \lambda_{kp} \mathcal{L}_{kp} + \lambda_{just} \mathcal{L}_{just}$, where $\lambda_{r}$, $\lambda_{kp}$, and $\lambda_{just}$ control the impact of each loss. All objectives share the latent representation~$\bb{z}$ and are thus mutually regularized by the function $E(u,i)$ to limit overfitting by any~objective.

\subsection{Unsupervised Critiquing}
\label{sec_critiquing}
The purpose of critiquing is to refine the recommendation based on the user's interaction with the explanation, the keyphrases $\bb{\hat{y}}_{kp}$, represented with a binary vector. The user critiques either a keyphrase $k$ by setting $\hat{y}_{kp}^k=0$ (i.e., disagreement) or symmetrically adding a new one (i.e.,$\hat{y}_{kp}^k=1$). We denote the critiqued keyphrase explanation as~$\bb{\tilde{y}}^*_{kp}$.\begin{figure}[t]
\centering
   \includegraphics[width=0.8\linewidth]{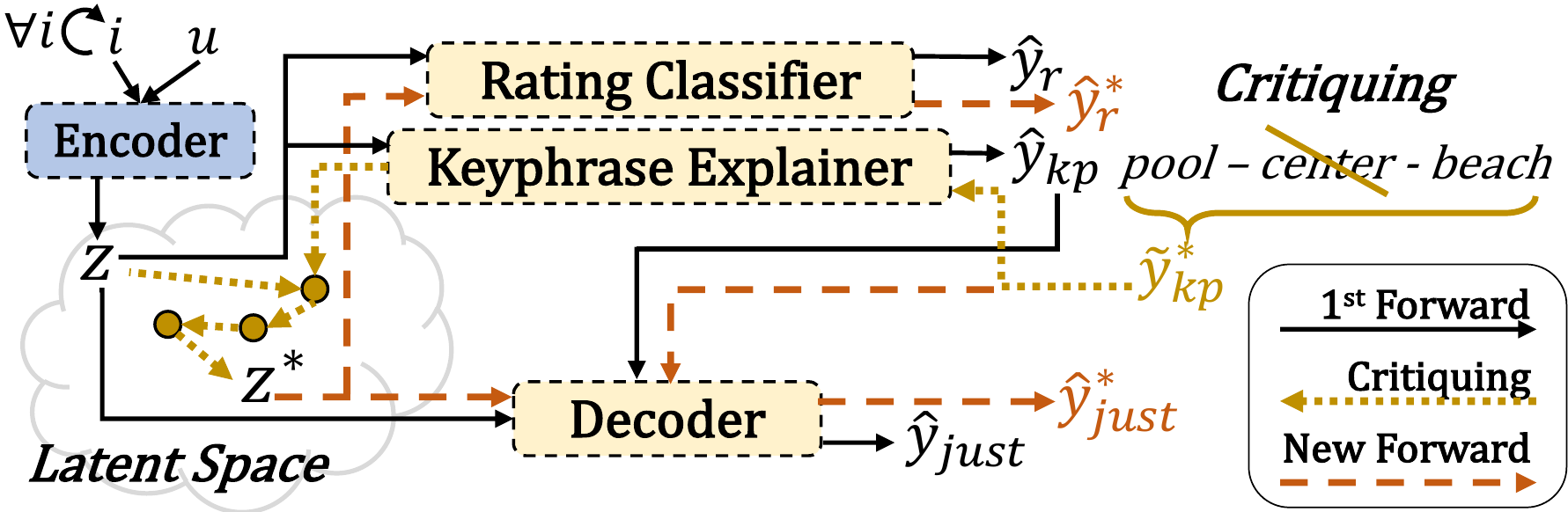}
   \caption{\label{general_critiquing}Workflow of considering to recommend items to~a user~$u$. We illustrate it for a given item $i$. \textbf{Black} denotes~the forward pass to infer the rating $\bb{\hat{y}}_r$ with the explanations $\bb{\hat{y}}_{kp}$ and $\bb{\hat{y}}_{just}$.~\textcolor{yellow2}{\textbf{Yellow}} indicates the critiquing: the user critiques the binary-vector keyphrase explanation $\bb{\hat{y}}_{kp}$ (e.g., \textit{center}) to $\bb{\tilde{y}^{*}}_{kp}$, which modulates the latent space into $\bb{z}^*$ for each item. \textcolor{orange2}{\textbf{Orange}} shows the new forward pass for the subsequent recommendation $\bb{\hat{y}}^*_r$ and explanations $\bb{\hat{y}}^*_{kp}$, $\bb{\hat{y}}^*_{just}$.
   }
\end{figure} 

\begin{algorithm}[H]
\caption{\label{alg_critiquing}Iterative Critiquing Gradient Update.}
\begin{algorithmic}[1]
\Function{Critique}{latent vector $\bb{z}$, target keyphrases $\bb{\tilde{y}}^*_{kp}$, trained keyphrases Explainer~$C^{kp}$, decay coefficient~$\zeta$, threshold $T$, a maximum number of iterations $I$}
    \State Set $\bb{z}^*_0 = \bb{z}, \quad \alpha_0 = 1.0,\quad $ and $t = 1$;
    \While {$|C^{kp}(\bb{z}^*) - \bb{\tilde{y}^*}_{kp}| > T$ and $t \le I$}
        \State $\bb{g}_{t-1} = \nabla_{\bb{z}^*_{t-1}} \mathcal{L}_{kp}(C^{kp}(\bb{z}^*_{t-1}), \bb{\tilde{y}^*}_{kp})$;
        \State {$\bb{z}^*_t = \bb{z}^*_{t-1} - \alpha_{t-1} \frac{\bb{g}_{t-1}}{||\bb{g}_{t-1}||_2}$;}
        \State {$\alpha_t = \zeta\alpha_{t-1}$;}
        \State {$t = t + 1$;}
    \EndWhile		
    \\
	\Return $\bb{z}^*_t$;
\EndFunction
\end{algorithmic}
\end{algorithm}

The overall critiquing process is depicted in Figure~\ref{general_critiquing} and Algorithm~\ref{alg_critiquing}. Inspired by the recent success in editing the latent space on the unsupervised text attribute transfer task~\citep{abs-1905-12926,abs-1912-02164}, we employ the trained Keyphrase Explainer $C^{kp}$ and the critiqued explanation $\bb{\tilde{y}}^*_{kp}$ to provide the gradient from which we update the latent representation $\bb{z}$ (depicted in \textcolor{yellow2}{\textbf{yellow}}). More formally, given a latent representation $\bb{z}$ and a binary critique vector $\bb{\tilde{y}}^*_{kp}$, we want to find a new latent representation $\bb{z}^*$ that will produce a new keyphrase explanation close to the critique, such that $|C^{kp}(\bb{z}^*) - \bb{\tilde{y}}^*_{kp}| \le T$, where $T$ is a threshold. In order to achieve this goal, we iteratively compute the gradient \underline{with respect to $\bb{z}$ instead of the model} \underline{parameters $C^{kp}_\theta$}. We then modify $\bb{z}$ in the direction of the gradient until we get a new latent representation $\bb{z}^*$ that $C^{kp}$ considers close enough to~$\bb{\tilde{y}}^*_{kp}$ (shown in \textcolor{orange2}{\textbf{orange}}). We emphasize that we use the gradient to modulate $\bb{z}$ rather than the parameters~$C^{kp}$. 

Let denote the gradient as $\bb{g}_t$ and a decay coefficient as $\zeta$. For each iteration~$t$ and $\bb{z}^*_0 = \bb{z}$, the modified latent representation $\bb{z}^*_t$ at the $t$\textsuperscript{th} iteration can be formulated as follows:\begin{equation}
    \bb{g}_t = \nabla_{\bb{z}^*_t} \mathcal{L}_{kp}(C^{kp}(\bb{z}^*_t), \bb{\tilde{y}}^*_{kp}); \quad\quad \bb{z}^*_t = \bb{z}^*_{t-1} - \zeta^{t-1} \bb{g}_t/||\bb{g}_t||_2.
\end{equation}Because this optimization is nonconvex, there is no guarantee that the difference between the critique vector and the inferred explanation will differ by only $T$. In our experiments in Section~\ref{sec_rq4_ijcai}, we found that a limit of $50$ iterations works well in practice, and that the newly induced explanations remain consistent.

\section{Experiments}

In this section, we answer the following research questions:

\begin{itemize}
    \item \textbf{RQ 1}: Are \textit{markers} appropriate justifications for recommendation?
    \item \textbf{RQ 2}: Does T-RECS generate high-quality, relevant, and personalized explanations?
    \item \textbf{RQ 3}: Do T-RECS justifications benefit the overall recommendation quality?
    \item \textbf{RQ 4}: Can T-RECS enable critiquing and effectively re-rank recommended items by critiquing explanations?
    \end{itemize}

\subsection{Experimental Settings}

\begin{table}[!t]
    \centering
   \caption{\label{stats_datasets_ijcai}Descriptive statistics of the datasets.}
\begin{threeparttable}
\begin{tabular}{@{}lcccccccc@{}}
& & & & & & \multicolumn{3}{c}{Avg. \#KP per}\\
\cmidrule{7-9}
\textbf{Dataset} & \textbf{\#Users} & \textbf{\#Items} & \textbf{\#Inter.} & \textbf{Dens.} & \textbf{KP Cov.} & \textbf{Just.} & \textbf{Rev.} & \textbf{User}\\
\toprule
Hotel & 72,603 & 38\,896& 2.2M & 0.08\% & 97.66\% & 2.15& 3.79 & 115\\
Beer & 7,304 & 8,702 & 1.2M & 2.02\% & 96.87\% & 3.72 & 6.97 & 1,210\\
\end{tabular}
\end{threeparttable}
\end{table}

\subsubsection{Datasets.}
We evaluate the quantitative performance of T-RECS using two real-world, publicly available datasets: BeerAdvocate~\citep{beer} and HotelRec~\citep{antognini-faltings-2020-hotelrec}. They contain 1.5 and~50 million reviews from BeerAdvocate and TripAdvisor. In addition to the overall rating, users also provided five-star aspect ratings\footnote{In cases where faceted ratings are not available at large cases, \cite{antognini-2021-concept,mukherjee2020uncertainty,niu-etal-2020-self} proposed elegant solutions to infer them from 20 or fewer samples.}.
We binarize the ratings with a threshold $t$: $t > 4$~for hotel reviews and $t > 3.5$ for beer reviews. We further filter out all users with fewer than 20 interactions and sort them chronologically. We keep the first~80\%~of interactions per user as the training data, leaving the remaining 20\% for validation and testing. We sample two justifications per review. Table~\ref{stats_datasets_ijcai} shows the statistics of the datasets.

We need to select keyphrases for explanations and critiquing. Hence, we follow the processing in~\cite{keyphraseExtractionDeep}~to~extract 200 keyphrases (distributed uniformly over the aspect categories) from the \textit{markers} on each dataset. Some examples are shown in Appendix \ref{app_kws_datasets}.

\subsubsection{Implementation Details.}
To extract \textit{markers}, we trained MTM with the hyperparameters following Chapter~\ref{chapter_aaai2021}. We obtained approximately 79\% and 91\% macro F1 Scores as shown in Table~\ref{perfs_full_beer}. Because the multi-aspect classification is not the focus, here we ignore the training procedure and performance evaluation.

We build the justification history $J^u$,$J^i$, with $N_{just}=32$. We set the embedding and attention dimension to 256 and to 1024 for the~feed-forward network. The encoder and decoder consist of two layers of Transformer with $4$~attention heads. We use a batch size of 128, dropout of 0.1, and Adam with learning rate $0.001$.
For critiquing, we choose a threshold and decay coefficient $T=0.015,\zeta=0.9$ and $T=0.01,\zeta=0.975$~for hotel and beer reviews. We tune all models on the dev set. For reproducibility purposes, we provide details~in~Appendix~\ref{app_reproducibility}.

\subsection{RQ 1: Are \textit{markers} appropriate justifications?}
\label{sec_rq1_ijcai}
We first verify whether \textit{markers} can serve as justifications for recommendation. We derive baselines from \cite{ni-etal-2019-justifying}: we split a review into elementary discourse~units (EDUs) and apply their classifier to get justifications; it is trained on~a manually~annotated dataset and generalizes well to other domains.~We employ two variants: EDU One and EDU All. The latter~includes all justifications, whereas the former includes only one.

We perform a human evaluation using Amazon's Mechanical Turk (see Appendix~\ref{app_hes} for more details) to judge the quality of the justifications extracted from the Markers, EDU One, and EDU All on both datasets. We employ three~setups: an evaluator is presented with 1. the three types of justifications; 2. only those from Markers and EDU All; and~3. EDU One instead of EDU All. We sampled 300 reviews (100 per setup) with generated justifications presented in random order. The annotators judged the justifications by choosing the most convincing in the pairwise setups and otherwise using best-worst scaling~\citep{louviere_flynn_marley_2015}, which has been shown to give more reliable results than Likert scales~\citep{van-der-lee-etal-2019-best,kiritchenko2016capturing}. We report the win rates for the pairwise comparisons and a normalized score ranging from -1 to +1.

Table~\ref{he_which_justification} shows that justifications extracted from Markers are preferred, on both datasets, more than 80\% of the time. Moreover, when compared to EDU All and EDU One, Markers achieve a score of $0.74$, three times higher than EDU All. Therefore, justifications extracted from the Markers are significantly better than EDUs, and a single justification cannot explain a review. Figure~\ref{examples_masks_no_edus} shows a sample for comparison.

\subsection{RQ 2: Does T-RECS infer high-quality and personalized explanations?}
We investigate whether T-RECS can generate \textit{personalized} explanations for a given user-item pair:
\begin{enumerate*}
    \item a relevant natural language justifications and
    \item a list of keyphrases, that best describes the intersection of the user and item profiles.
\end{enumerate*}
\begin{table}[t]
    \centering
   \caption{\label{he_which_justification}Human evaluation of explanations in terms of the win rate and the \textbf{b}est-\textbf{w}orst scaling. A score significantly different than Markers (post hoc Tukey HSD test) is denoted by {*}{*} for $p < 0.001$. Human annotators significantly prefer Markers as explanations.}
\begin{threeparttable}
\begin{tabular}{@{}lccccccc@{}}
& \multicolumn{3}{c}{\textit{Hotel}} & & \multicolumn{3}{c}{\textit{Beer}}\\
\cmidrule{2-4}\cmidrule{6-8}
\textbf{Winner}\ \ \ \textbf{Loser} & \multicolumn{3}{c}{\textbf{Win Rate}} & &  \multicolumn{3}{c}{\textbf{Win Rate}}\\
\cmidrule[0.08em]{1-8}
Markers\ \ \ EDU All & & 81\%{*}{*} & & & & 77\%{*}{*} & \\
Markers\ \ \ EDU One & & 93\%{*}{*} & & & & 90\%{*}{*} & \\
\\
& \multicolumn{3}{c}{\textit{Hotel}} & & \multicolumn{3}{c}{\textit{Beer}}\\
\cmidrule{2-4}\cmidrule{6-8}
\textbf{Model} & \textbf{Score} & \textbf{\#B} & \textbf{\#W} & & \textbf{Score} & \textbf{\#B} & \textbf{\#W}\\
\cmidrule[0.08em]{1-8}
EDU One & -0.95{*}{*} & \ 1 & 96 & & -0.93{*}{*} & \ 2 & 95\\
EDU All & \ 0.21{*}{*} & 24 & \ 3 & & \ 0.20{*}{*} & 23 & \ 3\\
Markers (Ours) & \textbf{0.74\ \ } & 75 & \ 1 & & \textbf{0.73}\ \  & 75 & \ 2\\
\end{tabular}
\end{threeparttable}
\end{table} 
\label{sec_rq2_ijcai}

\subsubsection{Natural Language Explanations.}
\label{sec_nle_justifications}We consider five baselines: 
ExpansionNet~\citep{ni2018personalized} is a sequence-to-sequence model with a user, item, aspect, and fusion attention mechanism that generates personalized reviews. DualPC~\citep{dual2020} and CAML~\citep{chen2019co} generate an explanation based on a rating and the user-item pair. Ref2Seq improves upon ExpansionNet by learning only from historical justifications of a user and an item. AP-Ref2Seq~\citep{ni-etal-2019-justifying} extends Ref2Seq with aspect planning~\citep{yao2019plan}, in~which aspects are given during the generation. All models use~beam search ($k=3$) during testing and the same keyphrases~as~aspects.
For automatic evaluation, we employ BLEU~\citep{papineni2002bleu}, ROUGE-L~\citep{lin2002manual}, and BertScore~\citep{bert-score}, a similarity metric based on BERT embeddings that has been shown to correlate better with human judgments. We also report the perplexity for evaluating the fluency and R\textsubscript{KW} for the explanation consistency as in \cite{chen2020towards}: the ratio of the target keyphrases present in the generated justifications.

The main results are presented in Table~\ref{just_auto_perfs} (more in Appendix~\ref{app_all_results}).~T-RECS achieves the highest scores on both datasets. 
We note that
\begin{enumerate*}
    \item sequence-to-sequence models better capture user and item information to produce more relevant justifications, and 
    \item using a keyphrase plan doubles the performance on average and improving explanation consistency.
\end{enumerate*}

\begin{table}[!t]
    \centering
\caption{\label{just_auto_perfs}Generated justifications on automatic evaluation using language generation metrics.}
\begin{threeparttable}
\begin{tabular}{@{}clccccc@{}}
& \textbf{Model} &  \textbf{BLEU} & \textbf{R-L} & \textbf{BERT\textsubscript{Score}} & \textbf{PPL}$\downarrow$ & \textbf{R\textsubscript{KW}}\\
\toprule
\multirow{6}{*}{\rotatebox{90}{\textit{Hotel}}}
& ExpansionNet & 0.53 & 6.91 & 74.81 & 28.87 & 60.09\\
& DualPC & 1.53 & 16.73 & 86.76 & 28.99 & 13.12\\
& CAML & 1.13 & 16.67 & 87.77 & 29.10 & 9.38\\
& Ref2Seq & 1.77 & 16.45 & 86.74 & 29.07 & 13.19\\
& AP-Ref2Seq & 7.28 & 33.71 & 88.31 & 21.31 & 90.20\\
& T-RECS (Ours) & \textbf{7.47} & \textbf{34.10} & \textbf{90.23} & \textbf{17.80} & \textbf{93.57}\\
\midrule
\multirow{6}{*}{\rotatebox{90}{\textit{Beer}}}
& ExpansionNet & 1.22 & 9.68 & 72.32 & 22.28 & 82.49\\
& DualPC & 2.08 & 14.68 & 85.49 &21.15 & 10.60\\
& CAML & 2.43 & 14.99 & 85.96 & 21.29& 10.18\\
& Ref2Seq & 3.51 & 15.96 & 85.27 & 22.34 & 12.10\\
& AP-Ref2Seq & 15.89 & 46.50 & 91.35 & 12.07 & 91.52\\
& T-RECS (Ours) & \textbf{16.54} & \textbf{47.20} & \textbf{91.50} & \textbf{10.24} & \textbf{94.96}\\
\end{tabular}
\end{threeparttable}
\end{table}

Because the justifications are intended for end users, we thus conduct a human evaluation, with the best models according to $R_{KW}$, on four dimensions using best-worst scaling:
\begin{enumerate}[noitemsep,topsep=0pt]
    \item \textbf{Overall} measures the overall subjective quality;
    \item \textbf{Fluency} represents the structure, grammar, and readability;
    \item \textbf{Informativeness} indicates whether the justification contains information pertinent to the user;
    \item \textbf{Relevance} measures how relevant the information is to an item.
\end{enumerate} We sample 300 explanations and showed them in random order.

\begin{table}[!t]
    \centering
   \caption{\label{just_human_perfs}Human evaluation of justifications in terms of best-worst scaling for \textbf{O}verall, \textbf{F}luency, \textbf{I}nformativenss, and \textbf{R}elevance. Most scores are significantly different than T-RECS (post hoc Tukey HSD test) with $p < 0.002$. $\dagger$ denotes a nonsignificant score. Explanations generated by T-RECS are largely preferred.}
\begin{threeparttable}
\begin{tabular}{@{}lccccccccc@{}}
& \multicolumn{4}{c}{\textit{Hotel}} & & \multicolumn{4}{c}{\textit{Beer}}\\
\cmidrule{2-5}\cmidrule{7-10}
\textbf{Model} & \textbf{O} & \textbf{F} & \textbf{I} & \textbf{R} & & \textbf{O} & \textbf{F} & \textbf{I} & \textbf{R}\\
\toprule
ExpansionNet & -0.58 & -0.67 & -0.52 & -0.56 & & -0.03 & -0.31 & 0.10 & -0.01\\
Ref2Seq & -0.27 & -0.19 & -0.30 & -0.26 & & -0.69 & -0.34 & -0.71 & -0.69\\
AP-Ref2Seq & 0.30 & 0.32 & 0.29 & 0.29 & & 0.22 & 0.25$\dagger$ & 0.21$\dagger$ & 0.25\\
T-RECS (Ours) & \textbf{0.55} & \textbf{0.54} & \textbf{0.53} & \textbf{0.53} & & \textbf{0.49} & \textbf{0.39} & \textbf{0.39} & \textbf{0.45}\\
\end{tabular}
\end{threeparttable}
\end{table}

Table~\ref{just_human_perfs} shows that our explanations are largely preferred on all criteria. T-RECS outperforms all methods on all criteria and performs $1.8$ times better than AP-Ref2Seq. Interestingly, we observe that ExpansionNet achieves significantly better results on the beer dataset with scores around $0.00$ for \textit{Overall} and \textit{Relevance}, and $0.10$ for \textit{Informativeness}; 
the information contained in beer reviews is easier to capture, as observed in \cite{bao-etal-2018-deriving}.

We provide samples of justifications generated by all models in Table~\ref{jus_comparison_hotel} and Table~\ref{jus_comparison_beer}. We observe that the \textit{markers} justify the subratings. Although they might be some overlaps between EDU All and Markers, justifications from EDU All often are incomplete or not relevant.
\begin{table}
\footnotesize
    \centering
   \caption{\label{jus_comparison_hotel}Comparisons of the extracted justifications from different models for two hotels on the hotel dataset. Colors denote aspects while underline denotes EDUs classified as ``good'' justifications.}
\begin{threeparttable}
\begin{tabular}{@{}lm{6.5cm}@{\hspace{6mm}}m{6cm}@{}}
\textbf{Model} & \multicolumn{1}{c}{\textbf{Casa del Sol Machupicchu}} & \multicolumn{1}{c}{\textbf{Southern Sun Waterfront Cape Town}}\\
\toprule
Review & \underline{\textcolor{red}{the hotel was decent}} \textcolor{red}{ the staff was very friendly.} \underline{the free pisco} \underline{sour class with kevin was a nice bonus!} however, \textcolor{purple}{the rooms were lacking.} \textcolor{green}{the wifi was incredibly slow and there was no air conditioning, so it got very hot at night.} \textcolor{purple}{we couldn't open the windows either because there were so many bugs, birds, and noise.} \textcolor{blue}{overall, the location is convenient, but was is not worth the price.}\newline
& this is my second year visiting cape town and staying here. \underline{\textcolor{blue}{excellent location to business district, conven-}} \underline{\textcolor{blue}{tion center, v\&a waterfront and access short}} \underline{\textcolor{blue}{distance to table mountain.}} \underline{\textcolor{red}{very nice}} \underline{\textcolor{red}{hotel, very friendly staff.} \textcolor{red}{breakfast is very good.}} \underline{\textcolor{purple}{rooms are nice}}~\textcolor{purple}{but bed mattress could be improved as bed is somewhat hard.} \underline{overall a very nice hotel.}\newline
\\
Rating & Overall: 3.0, \textcolor{red}{Service: 3.0}, \textcolor{yellow2}{Cleanliness: 4.0}, \textcolor{green}{Value: 2.0}, \textcolor{blue}{Location: 4.0}, \textcolor{purple}{Room: 3.0} & Overall: 4.0, \textcolor{red}{Service: 5.0}, \textcolor{yellow2}{Cleanliness: 5.0}, \textcolor{green}{Value: 4.0}, \textcolor{blue}{Location: 5.0}, \textcolor{purple}{Room: 3.0}\\
\midrule
Markers &
	-  \textcolor{purple}{the rooms were lacking.}\newline
	-  \textcolor{red}{the hotel was decent and the staff was very friendly.}\newline
	-  \textcolor{blue}{overall , the location is convenient , but was is not worth the price.}\newline
	-  \textcolor{purple}{we could n't open the windows either because there were so many bugs , birds , and noise.}\newline
	-  \textcolor{green}{the wifi was incredibly slow and there was no air conditioning , so it got very hot at night.}
	&
	-  \textcolor{red}{breakfast is very good.}\newline
	-  \textcolor{red}{very nice hotel , very friendly staff.}\newline
	-  \textcolor{purple}{rooms are nice but bed mattress could be improved as bed is somewhat hard.}\newline
	-  \textcolor{blue}{excellent location to business district , convention center , v\&a waterfront and access short distance to table mountain.}\\
\cdashlinelr{2-3}
EDU All &
	-  \underline{the hotel was decent}\newline
	-  \underline{the free pisco sour class with kevin was a nice bonus.}
&
	-  \underline{excellent location to business district , conven-}\newline
	\  \underline{tion center , v\&a waterfront and access short}\newline
	\  \underline{distance to table mountain.}\newline
	-  \underline{very nice hotel , very friendly staff. breakfast is}\newline
	\  \underline{very good.}\newline
	-  \underline{rooms are nice.}\newline
	-  \underline{overall a very nice hotel.}\\
\cdashlinelr{2-3}
EDU One & -  the hotel was decent & -  very nice hotel , very friendly staff . breakfast is very good\\
\bottomrule
\end{tabular}
\end{threeparttable}
\end{table}
\begin{table}
\footnotesize
    \centering
   \caption{\label{jus_comparison_beer}Comparisons of the extracted justifications from different models for two beers on the beer dataset. Colors denote aspects while underline denotes EDUs classified as ``good'' justifications.}
\begin{threeparttable}
\begin{tabular}{@{}lm{6.7cm}@{\hspace{6mm}}m{6cm}@{}}
\textbf{Model} & \multicolumn{1}{c}{\textbf{Saison De Lente}} & \multicolumn{1}{c}{\textbf{Bell's Porter}}\\
\toprule
Review
& poured from a 750ml bottle into a chimay branded chalice. a: \textcolor{red}{cloudy and unfiltered with a nice head that lasts and leaves good amounts of lacing in its tracks.} s: \textcolor{yellow2}{sour and bready with apple and yeast hints in there as well.} t: \textcolor{blue}{dry and hoppy with a nice crisp sour finish.} m: \underline{\textcolor{green}{medium bodied, high carbonation with big bubbles.}} d: \underline{easy to drink}, but i didn't really want more after splitting a 750ml with a buddy of mine.\newline
& \textcolor{red}{this beer pours black with a nice big frothy offwhite head.} \underline{\textcolor{yellow2}{smells or roasted malts, and chocolate.}} \underline{\textcolor{blue}{tastes of roasted malt with some chocolate}} \underline{\textcolor{blue}{and a hint of coffee.}} \textcolor{green}{the mouthfeel has medium body and is semi-smooth with some nice carbination.} \underline{drinkability is decent} i could drink a couple. \underline{overall a good choice} from bell's.\newline
\\
Rating & Overall: 3.0, \textcolor{red}{Appearance: 3.5}, \textcolor{yellow2}{Smell: 4.0}, \textcolor{green}{Mouthfeel: 3.5}, \textcolor{blue}{Taste: 3.5} & Overall: 3.5, \textcolor{red}{Appearance: 4.0}, \textcolor{yellow2}{Smell: 3.5}, \textcolor{green}{Mouthfeel: 3.5}, \textcolor{blue}{Taste: 4.0}\\
\midrule
Markers &
	- \textcolor{blue}{dry and hoppy with a nice crisp sour finish.}\newline
	- \textcolor{green}{medium bodied , high carbonation with big bubbles.}\newline
	- \textcolor{yellow2}{sour and bready with apple and yeast hints in there as~well.}\newline
	- \textcolor{red}{cloudy and unfiltered with a nice head that lasts and leaves good amounts of lacing in its tracks.}
&
	-  \textcolor{yellow2}{smells or roasted malts , and chocolate.}\newline
	-  \textcolor{red}{this beer pours black with a nice big frothy offwhite head.}\newline
	-  \textcolor{blue}{tastes of roasted malt with some chocolate and a hint of coffee.}\newline
	-  \textcolor{green}{the mouthfeel has medium body and is semi smooth with some nice carbination.}\\
\cdashlinelr{2-3}
EDU All &
- \underline{medium bodied , high carbonation with big bubbles.}\newline
- \underline{easy to drink} & 
	-  \underline{smells or roasted malts , and chocolate.}\newline
	-  \underline{tastes of roasted malt with some chocolate}\newline
	\  \underline{and a hint of~coffee.}\newline
	-  \underline{drinkability is decent}\newline
	-  \underline{overall a good choice}\\
\cdashlinelr{2-3}
EDU One & - easy to drink & -  drinkability is decent\\
\bottomrule
\end{tabular}
\end{threeparttable}
\end{table}

\subsubsection{Keyphrase Explanations.}
\label{sec_keyphrases_explanations}

Predicting keyphrases from the user-item latent representation is a natural~way to entangle them with and enable critiquing (see Section~\ref{sec_model}). We compare T-RECS with the popularity baseline and the models proposed in~\cite{keyphraseExtractionDeep}, which are extended versions~of the NCF model~\citep{he2017neural}. E-NCF and CE-NCF augment the NCF method with an \underline{e}xplanation and a~\underline{c}ritiquing neural component. Also, the authors provide \underline{v}ariational variants: VNCF, E-VNCF, and CE-VNCF. Here, we omit NCF and VNCF because they are trained only to predict ratings. We report the following metrics: NDCG, MAP, Precision, and Recall at~10 (see Appendix~\ref{app_kw_perfs2} for more metrics).\begin{table}[t]
    \centering
\caption{\label{kw_perfs}Keyphrase explanation quality using raking metrics at $N=10$.}
\begin{threeparttable}
\begin{tabular}{@{}lccccccccc@{}}
& \multicolumn{4}{c}{\textit{Hotel}} & & \multicolumn{4}{c}{\textit{Beer}}\\
\cmidrule{2-5}\cmidrule{7-10}
\textbf{Model} & \textbf{NDCG} & \textbf{MAP} & \textbf{P} & \textbf{R} &  & \textbf{NDCG} & \textbf{MAP} & \textbf{P} & \textbf{R}\\
\toprule
Pop & 0.333 & 0.208 & 0.143 & 0.396 &  & 0.250 & 0.229 & 0.176 & 0.253\\
\cdashlinelr{1-10}
E-NCF & 0.341 & 0.215 & 0.137 & 0.380 &  & 0.249 & 0.220 & 0.179 & 0.262\\
CE-NCF & 0.229 & 0.143 & 0.092 & 0.255 &  & 0.192 & 0.172 & 0.136 & 0.197\\
\cdashlinelr{1-10}
E-VNCF & 0.344 & 0.216 & 0.139 & 0.386 &  & 0.236 & 0.210 & 0.170 & 0.248\\
CE-VNCF & 0.229 & 0.134 & 0.107 & 0.297 &  & 0.203 & 0.178 & 0.148 & 0.215\\
\cdashlinelr{1-10}
T-RECS (Ours) & \textbf{0.376} & \textbf{0.236} & \textbf{0.158} & \textbf{0.436} &  & \textbf{0.316} & \textbf{0.280} & \textbf{0.228} & \textbf{0.332}\\
\end{tabular}
\end{threeparttable}
\end{table}

Table~\ref{kw_perfs} shows that T-RECS outperforms the CE-(V)NCF models by 60\%, Pop by 20\%, and E-(V)NCF models by 10\% to 30\% on all datasets. Interestingly, Pop performs better than~CE-(V)NCF, showing that many keywords are recurrent in reviews. Although the~task is thus not trivial, T-RECS retrieves up to 60\% of relevant keyphrases within the top 20. Thus, predicting keyphrases from the user-item latent space is a natural~way to entangle them with (and enable~critiquing). 

\subsection{RQ 3: Do T-RECS justifications benefit the overall recommendation quality?}
\label{sec_rq3_ijcai}In this section, we investigate whether justifications are beneficial to T-RECS and improve overall performance. We assess the performance on three different axes: rating prediction, preference prediction, and Top-N recommendation.

\subsubsection{Rating \& Preference Prediction.}
\label{sec_rating}
\begin{table}[t]
    \centering
   \caption{\label{rat_perfs}Performance of the rating prediction. Kendall's $\tau$ denotes the correlation between the users'own rankings and the models' ones. T-RECS consistently produces better recommendation.}
\begin{threeparttable}
\begin{tabular}{@{}lccclccc@{}}
& \multicolumn{3}{c}{\textit{Hotel}} & & \multicolumn{3}{c}{\textit{Beer}}\\
\cmidrule{2-4}\cmidrule{6-8}
\textbf{Model} & \textbf{MAE} & \textbf{RMSE} & \textbf{$\tau \uparrow$} & & \textbf{MAE} & \textbf{RMSE} & \textbf{$\tau \uparrow$}\\
\toprule
NMF & 0.3825 & 0.6171 & 0.2026 & & 0.3885 & 0.4459 & 0.4152\\
PMF & 0.3860 & 0.5855 & 0.0761 & & 0.3922 & 0.4512 & 0.4023\\
HFT & 0.3659 & 0.4515 & 0.4584 & & 0.3616 & 0.4358 & 0.4773\\
NARRE & 0.3564 & 0.4431 & 0.4476 & & 0.3620 & 0.4377 & 0.4506\\
\cdashlinelr{1-8}
NCF & 0.3619 & 0.4358 & 0.4200 & & 0.3638 & 0.4341 & 0.4696\\
E-NCF & 0.3579 & 0.4382 & 0.4145 & & 0.3691 & 0.4326 & 0.4685\\
CE-NCF & 0.3552 & 0.4389 & 0.4165 & & 0.3663 & 0.4390 & 0.4527\\
\cdashlinelr{1-8}
VNCF & 0.3502 & 0.4313 & 0.4408 & & 0.3666 & 0.4300 & 0.4706\\
E-VNCF & 0.3494 & 0.4365 & 0.4072 & & 0.3627 & 0.4457 & 0.4651\\
CE-VNCF & 0.3566 & 0.4545 & 0.3502 & & \textbf{0.3614} & 0.4330 & 0.4619\\
\cdashlinelr{1-8}
T-RECS (Ours) & \textbf{0.3306} & \textbf{0.4305} & \textbf{0.4702} & & \textbf{0.3614} & \textbf{0.4295} & \textbf{0.4909}\\
\bottomrule
\end{tabular}
\end{threeparttable}
\end{table}
We first analyze recommendation performance by the mean of rating prediction. We utilize the common Mean Squared Error (MSE) and Root Mean Squared Error (RMSE) metrics. However, the rating prediction performance alone does not best reflect the quality of recommendations, because users mainly see the relative ranking of different items~\citep{ricci2011introduction,musat2015personalizing}. Consequently, we measure also how well the item rankings computed by T-RECS agree with the user's own rankings as given by his own review ratings. We measure this quality by leveraging the standard metric Kendall's $\tau$ rank correlation~\citep{kendall1938measure}, computed on all pairs of rated-items by a user in the testing set. Overall, there are $153,954$ and $1,769,421$ pairs for the hotel and beer datasets, respectively.

We examine the following baseline methods together with T-RECS: NMF~\citep{hoyer2004non} is a non-negative matrix factorization model for rating prediction. PMF~\citep{mnih2008probabilistic} is a probabilistic matrix factorization method using ratings for collaborative filtering. HFT~\citep{McAuley:2013:HFH:2507157.2507163} is a strong latent-factor baseline, combined with a topic model aiming to find topics in the review text that correlate with the users' and items' latent factors. NARRE~\citep{10.1145/3178876.3186070} is a state-of-the-art model that predicts ratings and reviews' usefulness jointly. Finally, we include the six methods of~\cite{keyphraseExtractionDeep} described in Section~\ref{sec_keyphrases_explanations}.

The results are shown in Table~\ref{rat_perfs}. T-RECS consistently outperforms all the baselines, by a wide margin on the hotel dataset, including models based on collaborative filtering with/without review information or models extended with an explanation component and/or a critiquing component. Interestingly, the improvement in the hotel dataset in terms of MAE and RMSE is significantly higher than on the beer dataset. We hypothesize that this behavior is due to the~sparsity (see Table~\ref{stats_datasets_ijcai}), which has also been observed in the hotel domain in prior work~\citep{musat2015personalizing,antognini-faltings-2020-hotelrec}. On the beer dataset, we~note that reviews contain strong indicators and considerably improve the performance of NARRE and HFT compared to collaborative filtering methods. The extended (V)NCF models with either an explanation and/or a critiquing component improve MAE performance. Therefore, explanations can benefit the recommender systems to improve rating~prediction.

Table~\ref{rat_perfs} also contains the results in terms of preference prediction. T-RECS achieves up to $0.0136$ ($+2.85\%$) higher Kendall~correlation compared to the best baseline. Surprisingly, we note that CE-VNCF, NMF, and PMF show much worse results on the hotel datasets than on the beer dataset. This highlights that hotel reviews are noisier than~beer reviews and emphasizes the importance of capturing users' profiles, where T-RECS does best in comparison to other~models.

\subsubsection{Preference Prediction.}

\begin{figure}[!t]
\centering
\includegraphics[width=0.7\textwidth]{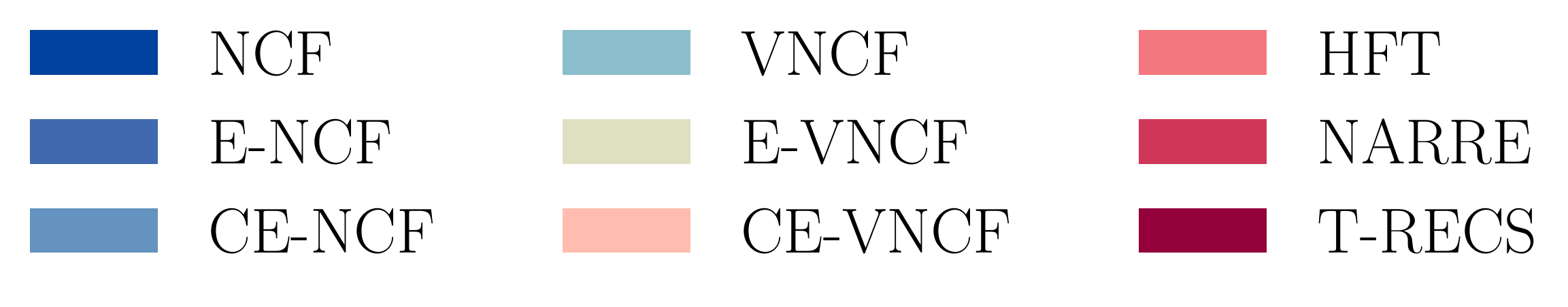}\\
\begin{subfigure}[t]{0.49\textwidth}
    \centering
    \includegraphics[width=\textwidth]{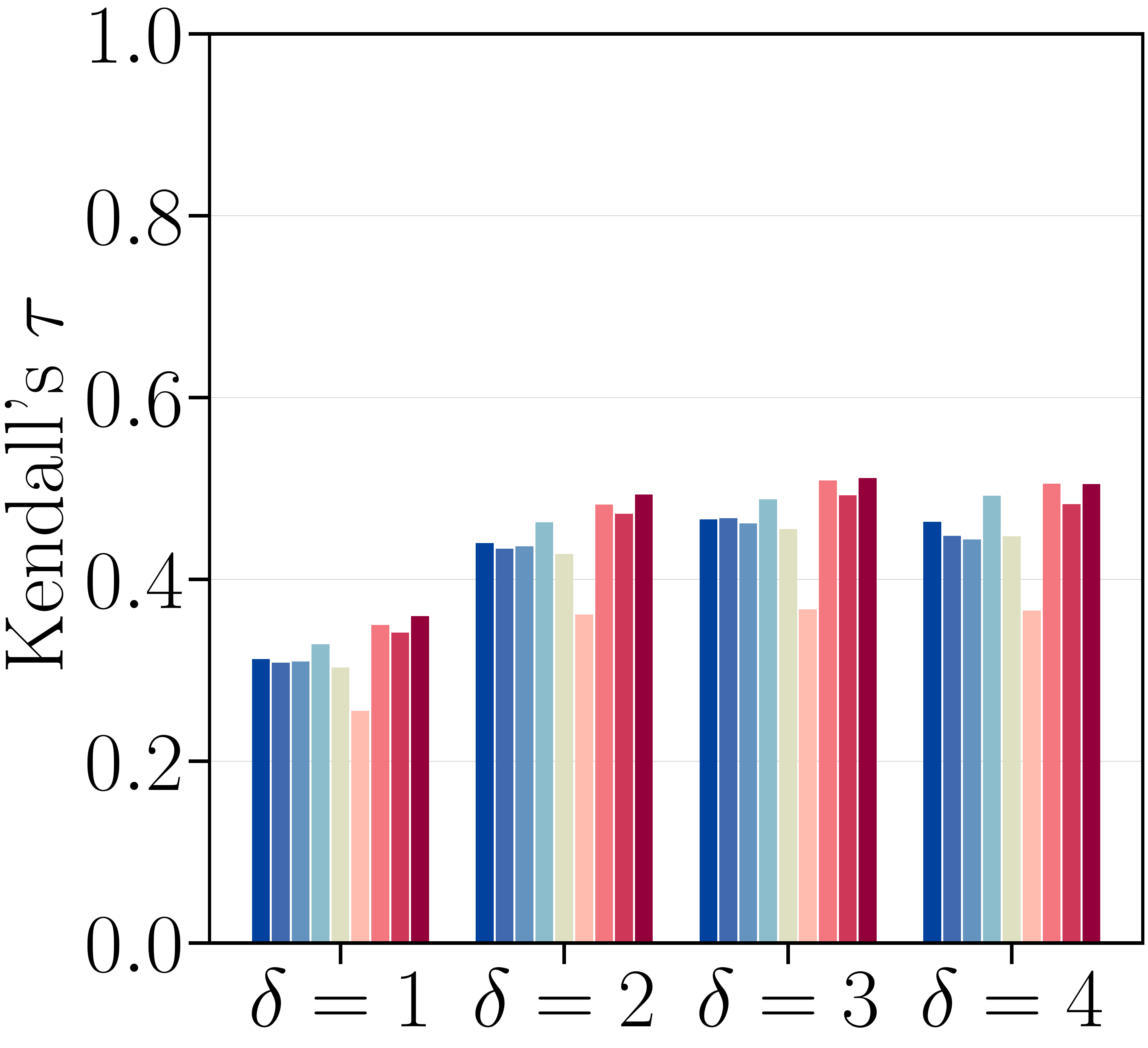}
    \caption{\label{kendall_hotel}Kendall's $\tau$ correlation on the \textbf{hotel} dataset.}
\end{subfigure}%
\hfill
\begin{subfigure}[t]{0.49\textwidth}
    \centering
    \includegraphics[width=\textwidth]{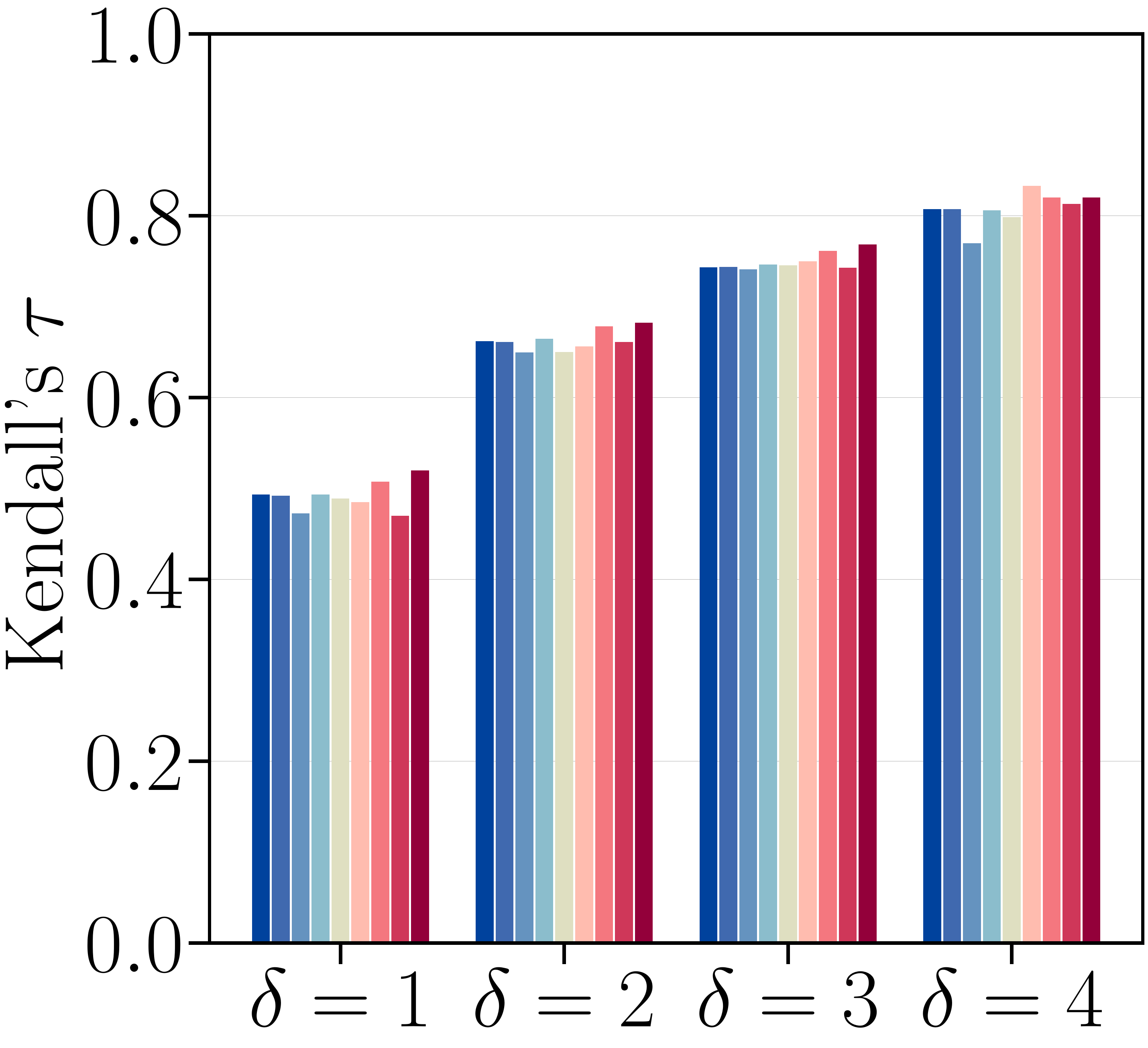}
    \caption{\label{kendall_beer}Kendall's $\tau$ correlation on the \textbf{beer} dataset.}
\end{subfigure}
\caption{\label{kendal_both}Performance of the preference prediction using Kendall's $\tau$ and $\delta = |y^i_r  - y^j_r|$.}
\end{figure}
In this experiment, we study a more fine-grained rank correlation. Following \cite{musat2013,musat2015personalizing}, we analyze the pairwise ranking of rated items by a user, and we impose a minimum value for the rating difference between two items $i$ and $j$, such that $\delta = |y^i_r  - y^j_r|$; the rating difference $\delta$ symbolizes the minimum preference strength. 

Figure~\ref{kendal_both} contains the Kendall's $\tau$ evaluation for multiple $\delta$ on both datasets. Overall, T-RECS increases the Kendall correlation similarly to other models but performs better on average. We observe that HFT's performance is similar to T-RECS, although slightly lower for most cases. On the beer dataset, we surprisingly note that CE-VNCF obtains a negligible higher score for $\delta = 4$, while significantly underperforming for $\delta < 4$, and especially on the hotel dataset. Finally, the Kendall's $\tau$ correlation increases majorly with the strength of preference pairs on the beer dataset and plateaus over $\delta \ge 2$ on the hotel dataset. It highlights that hotel reviews are noisier than beer reviews, and it emphasizes the importance of capturing users' profiles, where T-RECS does best in comparison to other models.
\begin{table}[!t]
    \centering
\caption{\label{rec_perfs}Performance of the Top-$N$ recommendation.}
\begin{threeparttable}[ht]
\begin{tabular}{@{}llcclcclcc@{}}
& & \multicolumn{2}{c}{\textbf{NDCG@N}} & & \multicolumn{2}{c}{\textbf{Precision@N}} & & \multicolumn{2}{c}{\textbf{Recall@N}}\\
\cmidrule{3-4}\cmidrule{6-7}\cmidrule{9-10}
& \textbf{Model} & N=10 & N=20 & & N=10 & N=20 & & N=10 & N=20\\
\toprule
\multirow{8}{*}{\rotatebox{90}{\textit{Hotel}}} & NCF & 0.1590 & 0.2461 & & 0.0231 & 0.0200 & & 0.2310 & 0.3991\\
& E-NCF & 0.1579 & 0.2432 & & 0.0234 & 0.0200 & & 0.2336 & 0.4004\\
& CE-NCF & 0.1585 & 0.2431 & & 0.0235 & 0.0201 & & 0.2352 & 0.4028\\
\cdashlinelr{2-10}
& VNCF & 0.1492 & 0.2431 & & 0.0220 & 0.0197 & & 0.2204 & 0.3932\\
& E-VNCF & 0.1505 & 0.2395 & & 0.0219 & 0.0192 & & 0.2188 & 0.3842\\
& CE-VNCF & \textbf{0.1738} & \textbf{0.2662} & &0.0221 & 0.0190 & & 0.2210 & 0.3809\\
\cdashlinelr{2-10}
& T-RECS (Ours) & 0.1674 & \textbf{0.2662} & & \textbf{0.0236} & \textbf{0.0207} & & \textbf{0.2358} & \textbf{0.4144} \\
\cmidrule[0.125em]{1-10}
\multirow{8}{*}{\rotatebox{90}{\textit{Beer}}} & NCF & 0.2172 & 0.3509 & & 0.0250 & 0.0212 & &  0.2499 & 0.4231 \\
& E-NCF & 0.2087 & 0.3363 & & 0.0243 & 0.0205 & & 0.2426 & 0.4103\\
& CE-NCF & 0.2226 & 0.3456 & & 0.0252 & 0.0205 & & 0.2517 & 0.4105\\
\cdashlinelr{2-10}
& VNCF & 0.1943 & 0.3329 & & 0.0235 & 0.0211 & & 0.2345 & 0.4213\\
& E-VNCF & 0.1387 & 0.2813 & & 0.0158 & 0.0168 & & 0.1579 & 0.3362\\
& CE-VNCF & 0.2295 & 0.3598 & & 0.0263 & 0.0218 & & 0.2630 & 0.4352\\
\cdashlinelr{2-10}
& T-RECS (Ours) & \textbf{0.2372} & \textbf{0.3674} & & \textbf{0.0269} & \textbf{0.0219} & &  \textbf{0.2693} & \textbf{0.4390}\\
\bottomrule
\end{tabular}
\end{threeparttable}
\end{table}

\subsubsection{Recommendation Performance.}

We evaluate the performance of T-RECS on the last dimension: Top-N recommendation. We adopt the widely used leave-one-out evaluation protocol~\citep{PERDIF2019,zhao2018categorical}; in particular, for each user, we randomly select one liked item in the test set alongside 99 randomly selected unseen items. We compare T-RECS with the state-of-the-art methods in \cite{keyphraseExtractionDeep}. Finally, we rank the item lists based on the recommendation scores produced by each method, and report the NDCG, Precision, and Recall at different~N.

Table~\ref{rec_perfs} presents the main results. Comparing to CE-(V)NCF models, which contain an explanation and critiquing components, our proposed model shows better recommendation performance for almost all metrics on the two datasets. On average, the variants (V)NCF reach higher results than the original method, which was not the case in the rating prediction and relative rankings tasks (see Section~\ref{sec_rating}), unlike T-RECS that shows consistent results.

\subsection{RQ 4: Can T-RECS enable critiquing?}
\label{sec_rq4_ijcai}We now investigate whether T-RECS can fill the gap between justifications and recommendation by enabling critiquing and effectively re-ranking recommended items.

\subsubsection{Single-Step Critiquing.}
\begin{figure}[t]
\centering
\includegraphics[width=0.8\textwidth]{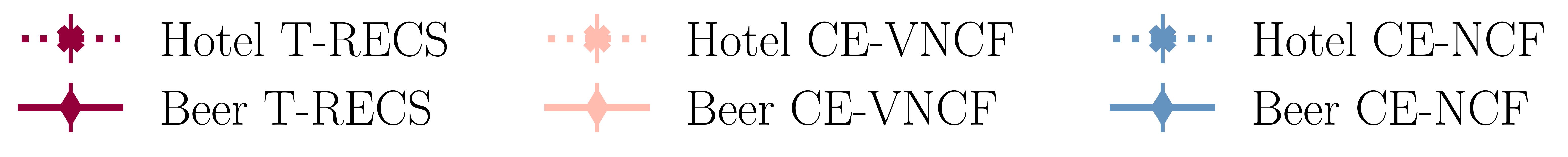}
\begin{subfigure}[t]{0.45\textwidth}
    \centering
    \includegraphics[width=\textwidth,height=6.35cm]{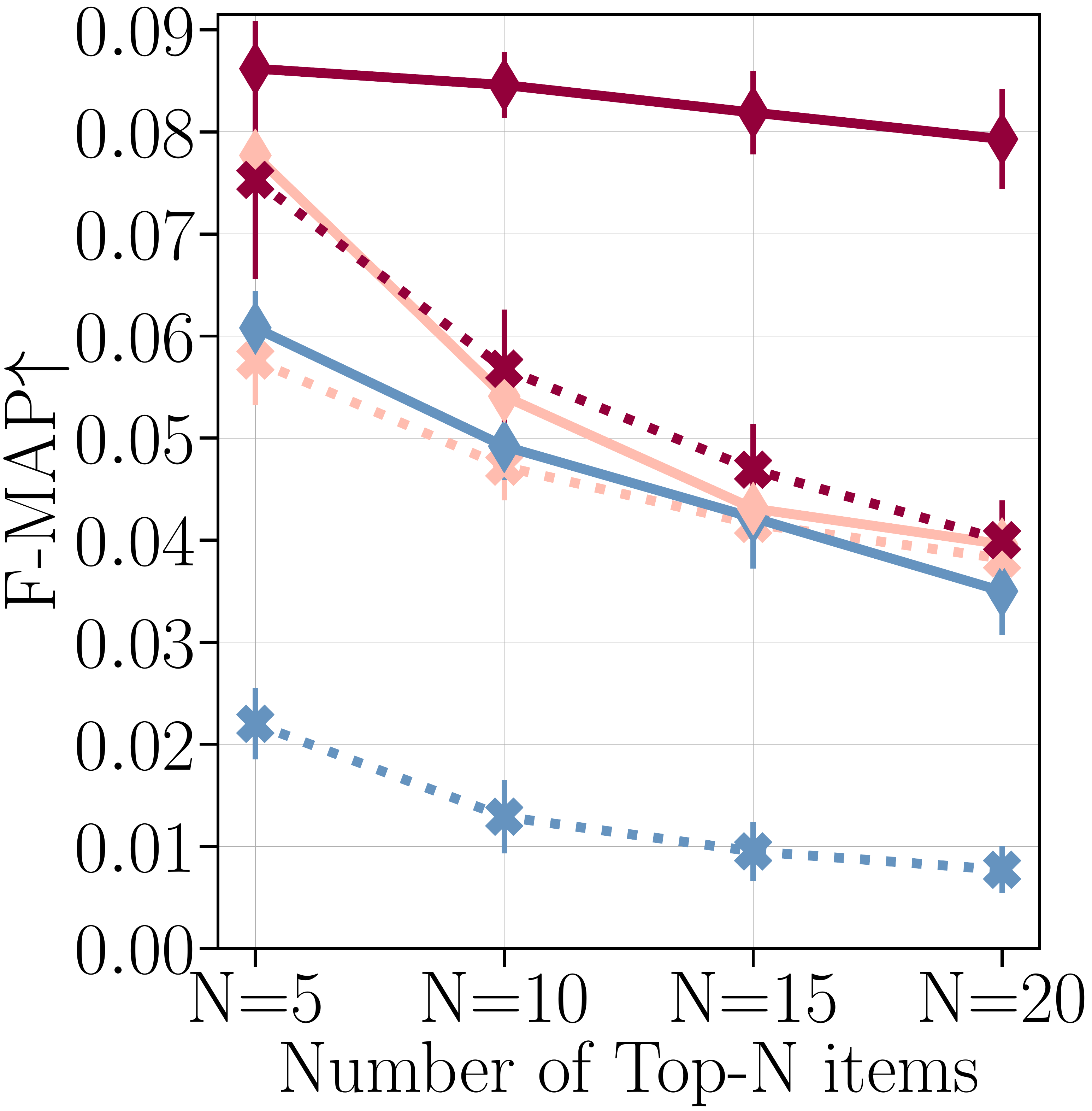}
    \caption{\label{single_crit_hotel}Falling MAP for different top-$N$. Error bars show the standard deviation.}
\end{subfigure}
\hfill
\begin{subfigure}[t]{0.5\textwidth}
    \centering
    \includegraphics[width=\textwidth,height=6.5cm]{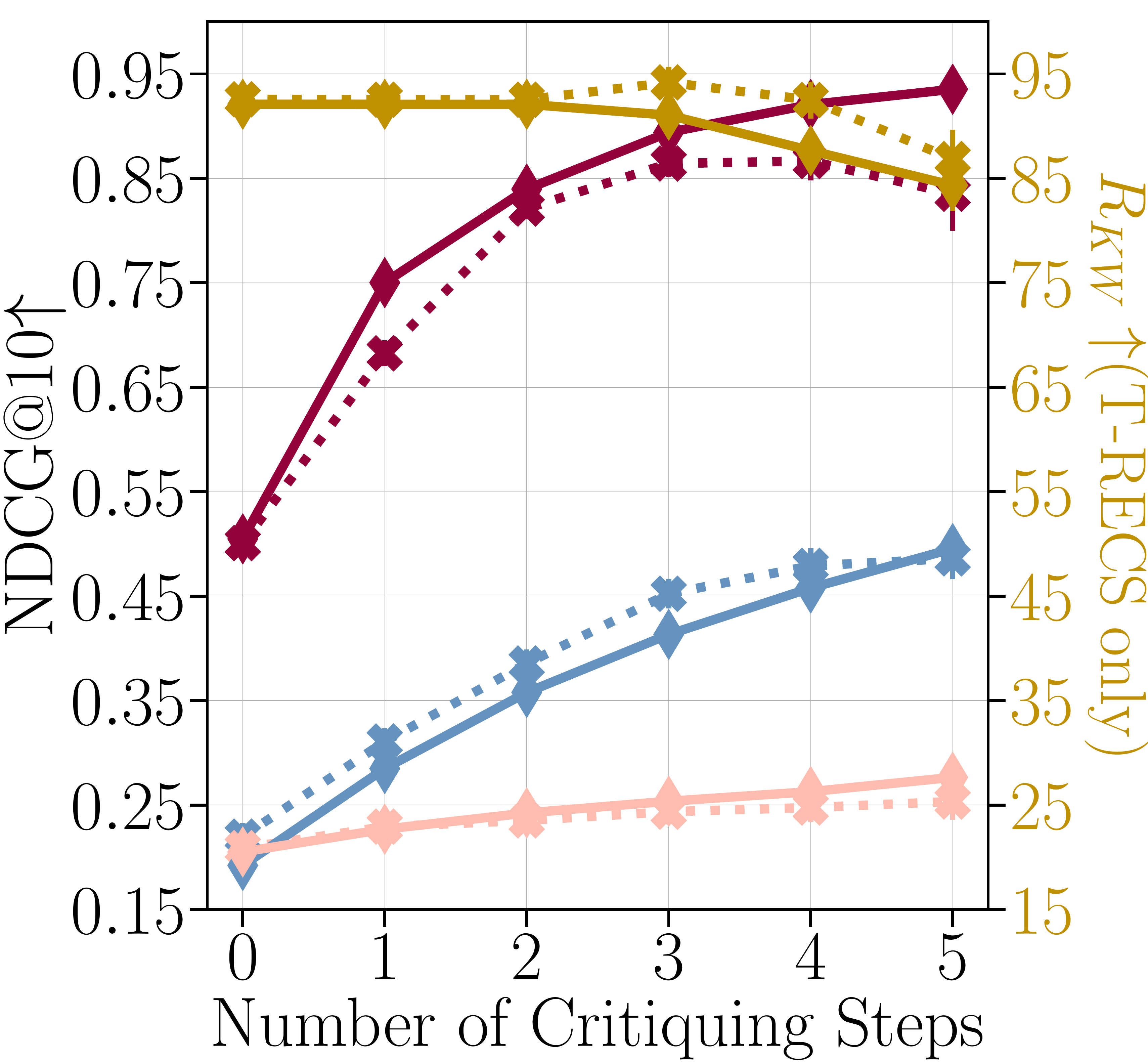}
    \caption{\label{mul_crit_hotel}Keyphrase prediction over multi-step critiquing with 95\% confidence interval. We also report the explanation consistency \textcolor{yellow2}{$R_{KW}$} for T-RECS.}
\end{subfigure}%
\caption{\label{single_mul_crit_both}Single- (top) and multi-step (bottom) critiquing.}
\end{figure}
For a given user, T-RECS recommends an item and generates personalized explanations, where the user can interact by critiquing one or multiple keyphrases. However, no explicit ground truth exists to evaluate the critiquing. We use F-MAP \citep{keyphraseExtractionDeep} to measure the effect of a critique. Given a user, a set of recommended items $\mathbb{S}$, and a critique~$k$, let $\mathbb{S}_k$ be the set of items containing $k$ in the explanation. The F-MAP measures the ranking difference of the affected items $\mathbb{S}_k$ before and after~critiquing $k$, using the Mean Average Precision at $N$: 
\begin{equation}
\textrm{F-MAP}=\textrm{MAP}@N_{\mathbb{S}_k}^{before} - \textrm{MAP}@N_{\mathbb{S}_k}^{after}. 
\end{equation}
A positive F-MAP indicates that the rank of items in $\mathbb{S}_k$ fell after $k$ is critiqued. We compare T-RECS with CE-(V)NCF and average the F-MAP over $5,000$ user-keyphrase pairs.

Figure~\ref{single_crit_hotel} presents the F-MAP performance on both datasets. All models show an anticipated positive F-MAP. The performance of T-RECS improves considerably on the beer dataset and is significantly higher for $N\le10$ on the hotel dataset. 
T-RECS and CE-(V)NCF handle the critique in different ways. CE-(V)NCF use an autoencoder learned jointly during training that projects the keyphrase explanation with the critiqued keyphrase removed back into the latent space. In contrast, T-RECS edits the entangled latent representation conforming to the new target explanation in the direction of the Keyphrase Explainer gradient.
The gap in performance may be caused by the extra loss~of the autoencoder, which brings noise during training. T-RECS only iteratively edits the latent representation at test time.

\subsubsection{Multi-Step Critiquing.}
\label{sec_multi_crit}Evaluating multi-step critiquing via ranking is difficult because many items can have the keyphrases of the desired item. Instead, we evaluate whether a system obtains a complete model of the user's preferences following \cite{pu2006increasing}. To do so, a simulated user expresses his keyphrase preferences iteratively according to a~randomly selected liked item. After each newly state preference, we evaluate the keyphrase explanations in the same manner~as in Section~\ref{sec_keyphrases_explanations} (we report complementary metrics in Appendix~\ref{app_multistep_add_metrics}). For T-RECS, we also report the explanation consistency $R_{KW}$. We expect a model to improve its performance over critiquing.

We run up to five-steps critiques over $1,000$ random selected users and up to $5,000$ random keyphrases for each dataset. 
Figure~\ref{mul_crit_hotel} shows that T-RECS builds through the critiques more accurate user profiles and consistent explanations. CE-NCF's top performance is significantly lower than T-RECS, and CE-VNCF plateaus, surely because of the KL divergence regularization, which limits the amount of information stored in the latent space. The explanation quality in T-RECS depends on the accuracy of the user's profile and may become saturated once we find it after four~steps.

\section{Multi-Step Critiquing User Interface}

In this section, we introduce four different web interfaces\footnote{Available under \url{https://lia.epfl.ch/critiquing/}.} helping users make decisions and find their ideal item. We have chosen the hotel recommendation domain as a use case even though our approach is trivially adaptable for other domains. Our system is model-agnostic (for both the recommender and critiquing models), allowing great flexibility and further extensions. 
Our interfaces are, above all, a useful tool to help research in recommendation with standard critiquing and critiquing based on neural generated-text techniques on a real use case. Finally, they can help to highlight the limitations of these systems to find ways to overcome them; they are not intended to be a finished usable product for end consumers.

\subsection{System Overview}
\label{sysoverview}
\begin{figure}[!t]
  \centering
  \begin{subfigure}[t]{\textwidth}
    \centering
  \includegraphics[width=\textwidth]{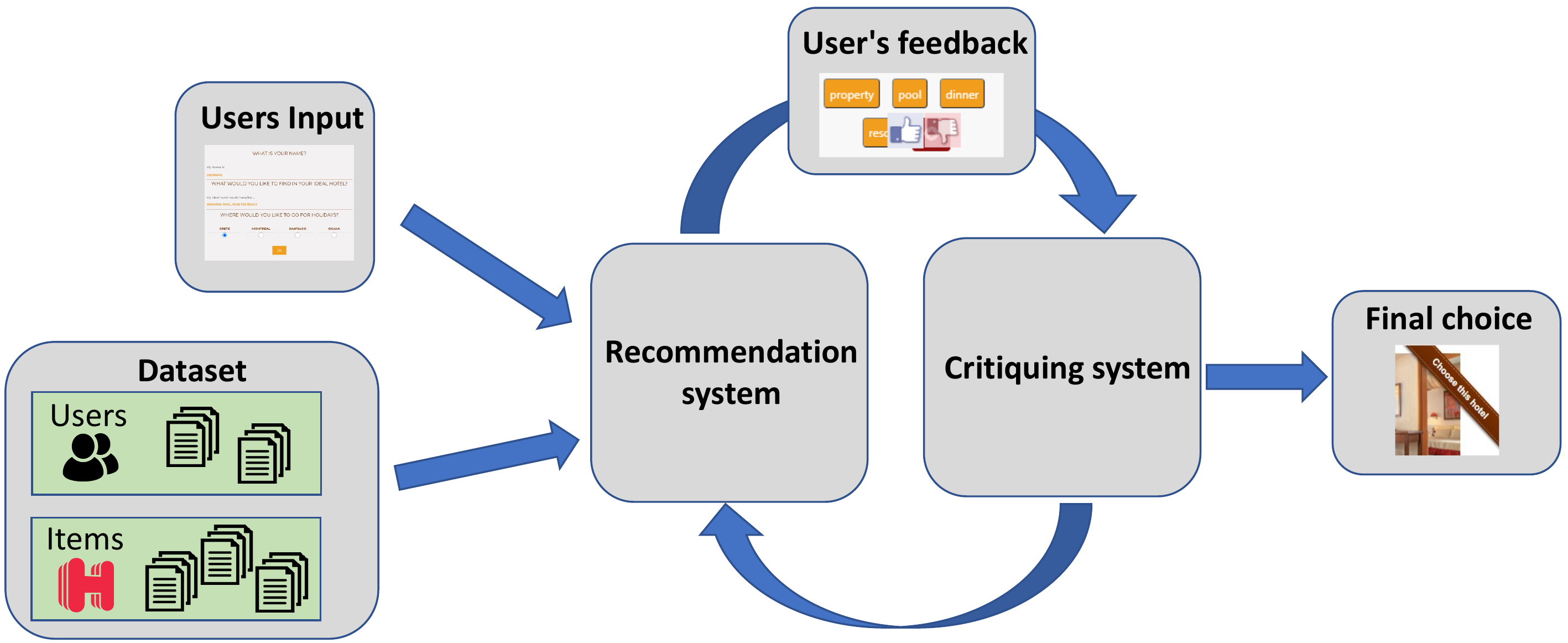}
  \caption{\label{fig:systemoverview}System interaction.}
\end{subfigure}
\\
\begin{subfigure}[!t]{\textwidth}
	  \centering
  \includegraphics[width=\textwidth]{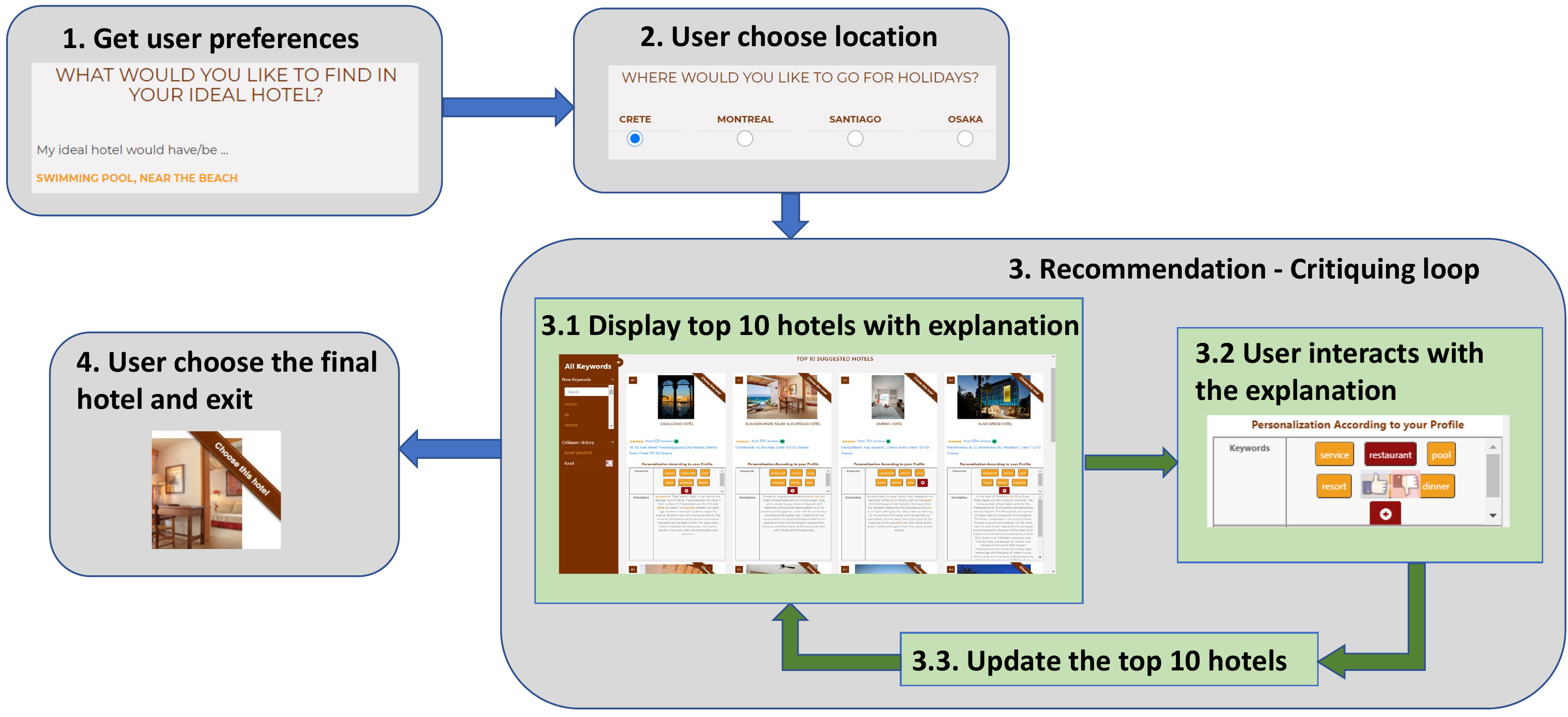}
  \caption{\label{fig:useroverview}User interaction.}
\end{subfigure}
\caption{Interactions overview.}
\end{figure}

The system overview is shown in Figure~\ref{fig:systemoverview}. To illustrate the flexibility of our system, we employ the HotelRec dataset~\citep{antognini-faltings-2020-hotelrec}, which contains 50 million reviews from \textit{TripAdvisor}, and automatically extracted 90 keyphrases to reflect hotel features (similar to those available on \textit{TripAdvisor} and \textit{Booking}).

Our system is model-agnostic both for the recommendation model and the critiquing mechanism. In case the recommender model does not handle the cold-start problem, we ask the users to provide an input to get their preferences and to define their initial profiles. We then match the query with an existing user in the database whose profile is most similar to the preferences formulated (according to a TF-IDF vector space model computed on user reviews). This enables more flexibility in the used recommender model.

We experiment with two recommender systems with critiquing mechanisms: T-RECS (Section~\ref{sec_model}) and CE-VAE~\citep{luo2020}. The latter improves the training performance of CE-VNCF. Both predict a set of keywords from the 90 keyphrases extracted during preprocessing, that best describes~the user profile, alongside the recommended items. These keywords are therefore used as an explanation for the recommendation and can then be critiqued. However, these two models have their own specificities that we detail in~Section~\ref{sec_int}.

The interaction with the interface from the user's point of view is shown in Figure~\ref{fig:useroverview}. Users are first asked to enter, in text form, what they would like for their ideal hotel. Then, users choose one of the four proposed destinations, where each contains approximately 25 to 45 hotels. Next, users enter the \textit{Recommendation - Critiquing loop}. The interface initially displays only the top 10 items recommended based on initial user input, along with the generated explanation. The users have the possibility to interact with the explanations by critiquing them. After each critique, the system updates the recommendation of the users. This process continues until users are satisfied and cease to provide additional critiques. At this point, this iterative process terminates as we consider that users have found their ideal hotel.

\subsection{Interface Details}
\label{sec_int}
\begin{figure}[!t]
  \centering
  \begin{subfigure}[t]{0.2\textwidth}
  \centering
  \includegraphics[height=6cm]{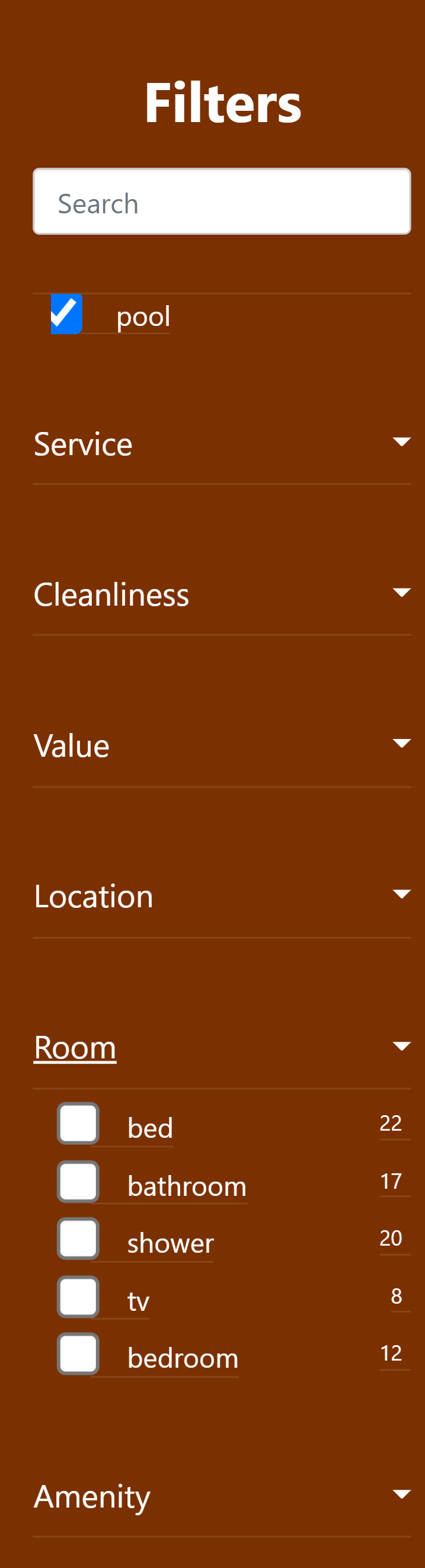}
  \caption{\label{fig:menuA}Menu for Int. A.}
  \end{subfigure}
\hspace{1mm}
  \begin{subfigure}[t]{0.245\textwidth}
  \centering
  \includegraphics[height=6cm]{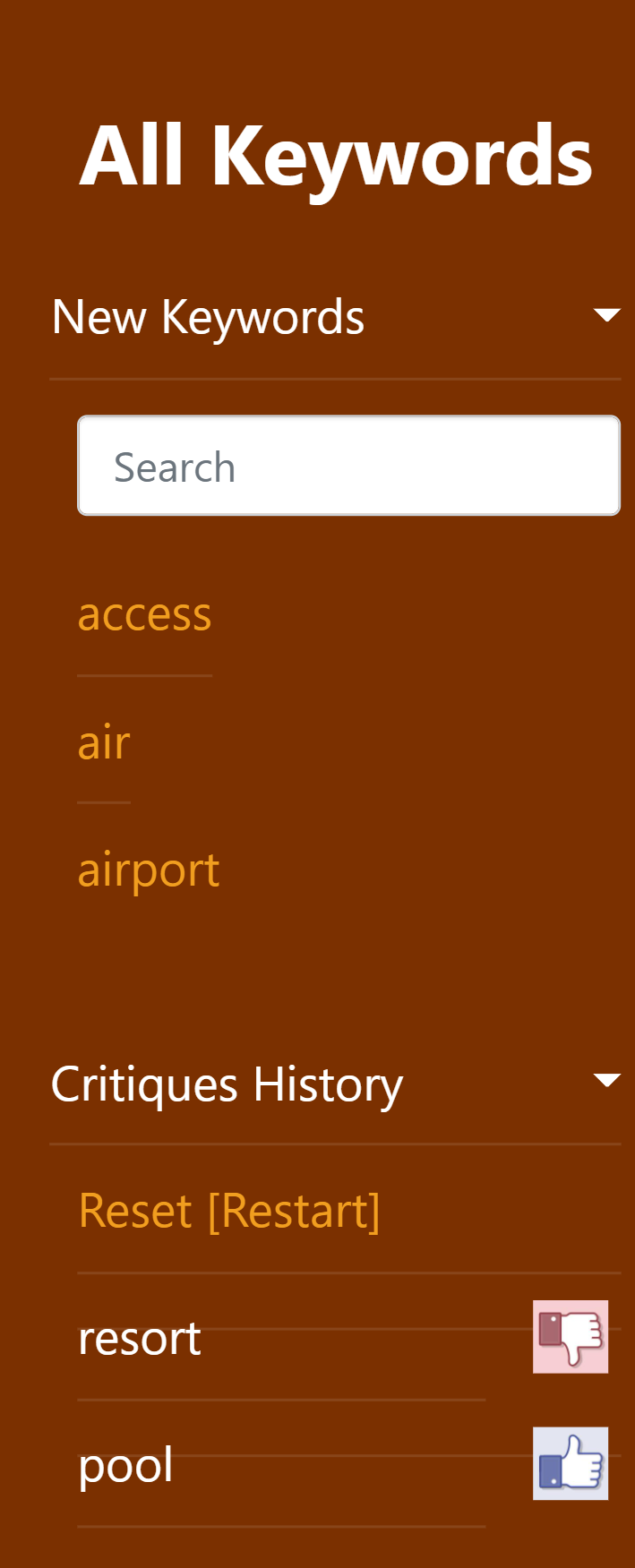}
  \caption{\label{fig:menuBCD}Menu for Int. B/C/D.}
  \end{subfigure}
\hspace{1mm}
  \begin{subfigure}[t]{0.45\textwidth}
  \centering
  \includegraphics[height=6cm]{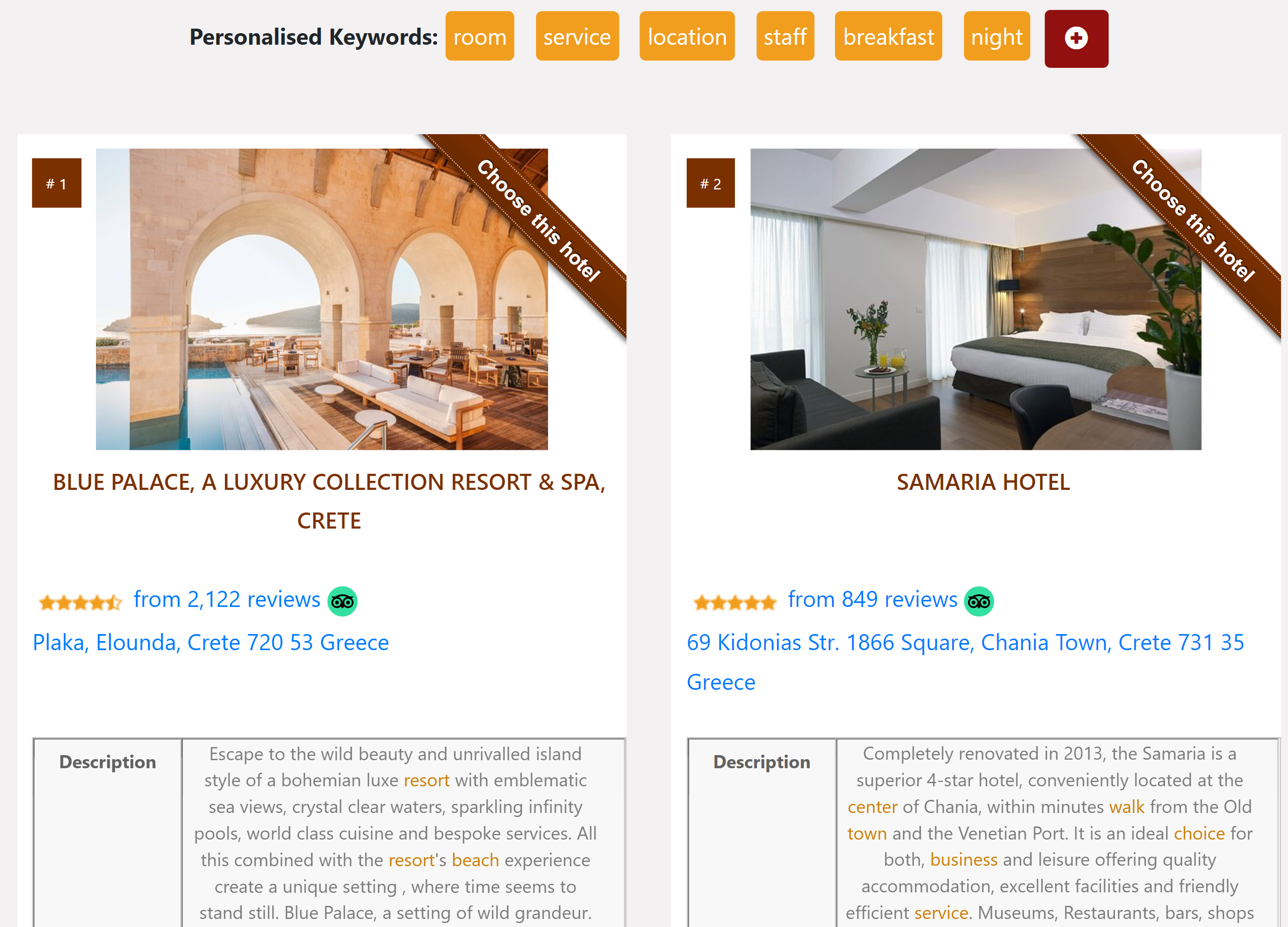}
  \caption{\label{fig:interfaceB}Int. B: Explanation shared by all items for CE-VAE with descriptions.}
  \end{subfigure}
  \\
    \begin{subfigure}[t]{0.475\textwidth}
  \centering
  \includegraphics[height=6cm]{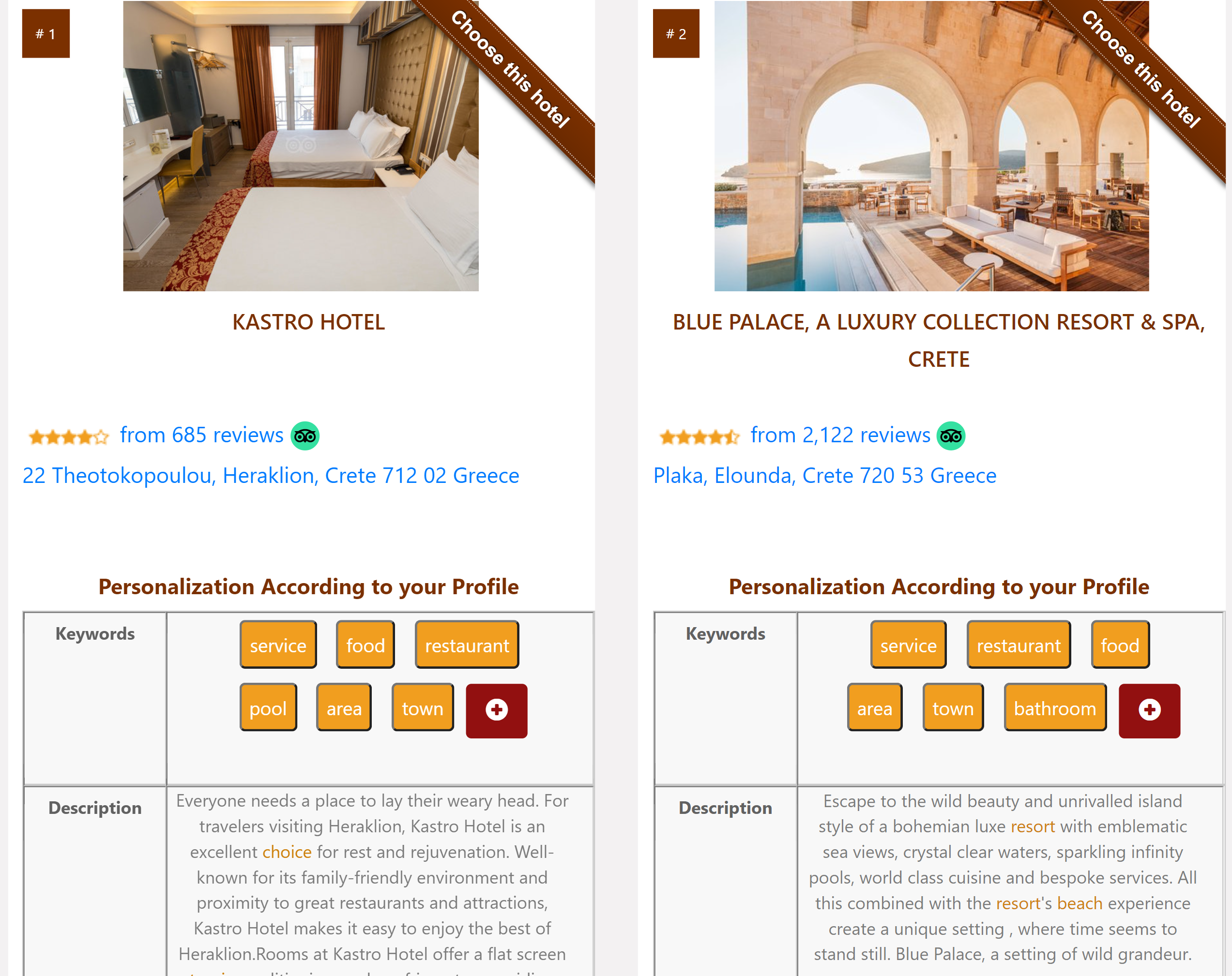}
  \caption{\label{fig:interfaceC}Int. C: One explanation per item for TRECS with descriptions.}
  \end{subfigure}
  \hfill
      \begin{subfigure}[t]{0.475\textwidth}
  \centering
  \includegraphics[height=6cm]{main/IJCAI2021/Images/InterfaceC_TRECS+Description.png}
  \caption{\label{fig:interfaceD}Int. D: One explanation per item for TRECS with justifications.}
  \end{subfigure}
  \caption{\label{fig:interfaces}Interfaces.}
\end{figure}
Users can interact with four interfaces, detailed below and shown in Figure~\ref{fig:interfaces}.

\textbf{Interface A: Static.}\hspace{\parindent} 
We use a static interface inspired by current hotel recommendation platforms as a reference point for user satisfaction. We allow users to filter hotels according to the keyword features (the same as the extracted ones during preprocessing). To make interaction easier, the keywords are grouped in 6~categories (see Figure~\ref{fig:menuA}). The interface also displays the hotel's description scraped from \textit{Tripadvisor}.

\textbf{Interface B: CE-VAE with shared keyphrases.}\hspace{\parindent} 
The recommender system predicts a set of keywords that best describe the user profile alongside the recommended items. For \textit{CE-VAE}, users can only interact with the keywords on a global level to express an explicit disagreement: all items share the same explanation. The top 6~keywords associated with the user profile are displayed at the top of the web page (see Figure~\ref{fig:interfaceB}).
Similar to \textit{Interface~A}, the hotel descriptions are also displayed. However, the system's keywords are highlighted in the descriptions. It allows the users to also interact with these descriptions by critiquing them.

\textbf{Interface C: T-RECS with fine-grained keyphrases.}\hspace{\parindent} In contrast to CE-VAE, T-RECS generates more fine-grained explanations by inferring personalized keywords for each user-item pair (see Figure~\ref{fig:interfaceC}). Moreover, users can critique a keyword to reflect a disagreement and highlight it as an important desired feature. As for \textit{Interface B}, the hotel descriptions are also displayed and enable critiquing interactions.

\textbf{Interface D: T-RECS with fine-grained keyphrases and generated justifications.}\hspace{\parindent} This interface is similar to \textit{Interface C} but, instead of displaying the general scrapped description for each hotel, it generates a personalized natural language justification conditioned on the keyword explanation. This justification, which is therefore based on user profiles, allows them to understand why those hotels are recommended. An example is shown in Figure~\ref{fig:interfaceD}.

In addition, for \textit{Interfaces B, C}, and \textit{D}, we give the possibility to search for additional keywords to critique (see Figure~\ref{fig:menuBCD}) than those predicted as relevant by the system. Finally, after experimenting with the four interfaces, users access the entire hotel catalog to determine if the different approaches helped them find the hotel that best suits~them.

\subsection{Discussion}
Some limitations of the natural language processing methods used can be observed by experimenting with the interfaces. We can take as an example the generated keywords. They are obtained by automated systems that generate them based on data mining techniques or machine learning models. However, meaningful keywords for the recommendation do not necessarily imply that they are helpful for the end user. On the one hand, the meanings of individual keywords are not always clear for the end user, especially if they are not familiar with the recommendation domain. On the other hand, it is not always obvious what critiquing a keyword means. Indeed, the keyword “pool” indicates the presence or the absence of a pool. However, for the keyword “breakfast”, it is not clear how the system interprets breakfast as most hotels propose breakfast options. Post-processing to filter vague keywords or the usage of n-grams could improve the understanding of the keywords but partially solve the problem.

Moreover, with the recent progress of natural language generation, models are now capable of generating syntactically and grammatically correct sentences~\citep{radford2019language}. However, we observe that the generated personalized justifications quite often lack diversity~\citep{ni-etal-2019-justifying}. Indeed, the justifications of the best items contain repeating parts of sentences. This makes it more difficult for the users to make a choice because the differences between the justifications are not emphasized. In addition, as is often the case in natural language applications, hallucinations can occur~\citep{rohrbach-etal-2018-object,holtzman2019curious}.

\section{Conclusion}
Recommendations can carry much more impact if they are supported by explanations. 
In this chapter, we presented T-RECS, a multi-task learning Transformer-based recommender, that produces explanations considered significantly superior when evaluated by humans.
The second contribution of T-RECS is the user's ability to react to a recommendation by \textit{critiquing} the explanation. 
We designed an unsupervised method for multi-step critiquing with explanations. Experiments show that T-RECS obtains stable and significant improvement in adapting to the preferences expressed in multi-step critiquing.

Finally, we designed several web interfaces that can be easily connected to different datasets, recommendation models, and critiquing mechanisms. In addition to the usual recommendations, they can provide users with keyword explanations and personalized natural language justifications. Users can interact with these keywords in different ways by critiquing them. It allows experimenting recommendations with critiquing based on neural natural language processing on a real use case. The interfaces are also a useful tool to highlight these systems' limitations and find ways to overcome them.

%% file: main/RECSYS2021/main.tex
\chapter{Faster and Better Multi-Step Critiquing via Self-Supervision}
\label{chapter_recsys_2021}

\section{Preface}
\textbf{Contribution and Sources.}\hspace{\parindent} This chapter is largely based on \cite{fast_critiquing}. The detailed individual contributions are listed below using the CRediT taxonomy \citep{brand2015beyond} (terms are selected as applicable).

\begin{table}[!h]
\begin{tabular}{@{}ll@{}}
Diego Antognini (author): & Conceptualization, Methodology, Software, Validation,\\
& Investigation, Formal Analysis, Writing -- Original Draft,\\
&Writing -- Review \& Editing.\\ \\
Boi Faltings: & Writing -- Review \& Editing, Administration, Supervision.
\end{tabular}
\end{table}

\textbf{Summary.}\hspace{\parindent} 
Recent studies have shown that providing personalized explanations alongside recommendations increases trust and perceived quality. Furthermore, it gives users an opportunity to refine the recommendations by \textit{critiquing} parts of the explanations.
On one hand,~current recommender systems model the recommendation, explanation, and critiquing objectives jointly, but this creates an inherent trade-off between their respective performance, and it yields poor results in multi-step critiquing settings.
On the other hand, although recent latent linear critiquing approaches are built~upon an existing recommender system, they suffer from computational inefficiency at inference due to the objective optimized at each~conversation's turn.
In this chapter, we address these deficiencies with M\&Ms-VAE, a novel variational autoencoder for recommendation and explanation that is based on multimodal modeling assumptions. We train the model under a weak supervision scheme to simulate both~fully~and partially observed variables. Then, we leverage the generalization ability of a trained M\&Ms-VAE model to embed the user preference and the critique separately. Our work's most important innovation is our critiquing module, which is built upon and trained in~a~self-supervised manner with a simple ranking objective.
Experiments on four real-world datasets demonstrate that among state-of-the-art models, our system is the first to dominate or match the performance in terms of recommendation, explanation, and multi-step critiquing. Moreover, M\&Ms-VAE processes the critiques up to 25.6x faster than the best baselines. Finally, we show that our model~infers coherent joint and cross generation, even under weak supervision, thanks to our multimodal-based modeling and training~scheme.

\section{Introduction}

Recommender systems accurately capture user preferences and achieve high performance. However, they offer little transparency regarding their inner workings. It has been shown that providing explanations along with item recommendations enables users to understand why a particular item has been suggested and hence to make better decision~\citep{chang2016crowd,bellini2018knowledge}. Additionally, explanations increase the system's overall transparency and trustworthiness~\citep{ExplainingRecommendation,zhang2018exploring,Kunkel2018TrustrelatedEO}.

An important advantage of explanations is that they provide a basis for feedback. If users understand what has generated the suggestions, they can refine the recommendations by interacting directly with the explanations. Critiquing~is a conversational recommendation method that incrementally adapts recommendations in response to user preferences~\citep{chen2012critiquing}. Example critiquing was introduced in information retrieval~\citep{williams1982rabbit} and first applied to recommender systems in~\cite{unitcritiquing}. Recognizing that critiquing is most useful when applied in multiple steps,~\cite{multistepcritiquingNeal} and~\cite{10.1145/332040.332446} introduced mechanisms based on constraint programming~\citep{Torrens2002abcd} with an application to travel planning. Multi-step critiquing with constraint programming was recognized as a form of preference elicitation, which enabled the analysis and optimization of its performance~\citep{Faltings2004b} and the addition of suggestions for active preference elicitation~\citep{Viappiani06preference-basedsearch}, which yielded dramatic improvements in decision accuracy in user studies. Multi-step critiquing was also shown to be superior to compound critiquing, which groups multiple attributes in a single step~\citep{10.1145/1250910.1250929}. A major limitation of all these approaches is that items have to be characterized by a set of discrete attributes.

After nearly a decade in which critiquing approaches received little attention,~\cite{keyphraseExtractionDeep} introduced a collaborative filtering recommender with explanations and an embedding-based critiquing method. This method allows users to critique the recommendation using arbitrary languages; a set of attributes is mined from reviews, and the users can interact~with them. Other works built upon the same paradigm~\citep{antognini2020interacting,chen2020towards}.~\cite{luo2020} showed that those models suffer from unstable training and high computational complexity, and they proposed a framework based on a variational autoencoder~\citep{kingmamw2014,liang2018variational}. However, these models learn a bidirectional mapping between the critique and the user latent space. This creates~an inherent trade-off between the recommendation and explanation performance, and it yields poor results in~multi-step~critiquing.
 
Recently,~\cite{luo2020b} proposed a latent linear critiquing (LLC) method built upon the  recommendation model PLRec~\citep{sedhain2016practical}. LLC co-embeds keyphrase attributes in the same embedding space as the recommender. The critiquing process consists~of~a weighted average between the user-preference embedding and the critique embeddings obtained through the conversation. The weights are optimized in a linear programming formulation using a max-margin scoring-based objective~(i.e., the pairwise difference of scores of items affected by the critique and the others). Following the same methodology,~\cite{hanze2020} changed the objective into a ranking-based one. While those models obtain good performance in multi-step critiquing, they suffer from computational inefficiency due to the objective function optimized at each turn.

To address both issues, we present M\&Ms-VAE, a novel variational autoencoder for recommendation and explanation with a separate critiquing module. Inspired by multimodal generative models~\citep{NEURIPS2018_1102a326,NEURIPS2019_0ae775a8,tsai2018learning,NEURIPS2020_43bb733c}, we treat the user's past interactions and keyphrase usage as different partially observed variables, and more importantly, we assume conditional independence between them. We can then approximate the variational joint posterior using a mixture of experts. We propose a training scheme that mimics weakly supervised learning to train the inference networks jointly but also independently. This is essential to our modeling, because M\&Ms-VAE is robust to a missing unobserved variable and can thus embed separately and efficiently the user interactions, the keyphrases, and the critique,~respectively. 

In a second step, we leverage the generalization ability of M\&Ms-VAE and design a novel blending module to re-rank recommended items according to a critique. The latter is trained only once on a synthetic dataset with a self-supervision objective. This generalizes into multi-step critiquing and enables fast critiquing.

To the best of our knowledge, this is the first work to revisit deep critiquing from the perspective of multimodal generative models and to propose a blending module trained in a simple self-supervised fashion. 
We evaluate our method using four real-world, publicly available datasets. The results demonstrate that the proposed M\&Ms-VAE model
\begin{enumerate}[topsep=0pt]
 	\item achieves superior or competitive performance in terms of recommendation, explanation, and multi-step critiquing in comparison to the state-of-the-art recommendation and critiquing methods, 
 	\item processes the critiques up to 25.6x faster than the best baselines and up to 9.2x faster using only the CPU, and 
 	\item induces coherent joint generation and cross generation, even under weak supervision.
 \end{enumerate}
 
\section{Preliminaries}

This section introduces the notation used in the chapter and the variational autoencoder for recommendation~\citep{liang2018variational}. Then, we review a recent study~\citep{luo2020} that built upon~\cite{liang2018variational} and revisited critiquing by proposing the critiquable-explainable VAE (CE-VAE) model. Finally, we highlight the key deficiencies that significantly limit its performance in practice.
\subsection{Notation}

Before proceeding, we define the following notation used throughout this chapter:
\begin{itemize}
	\item $U$, $I$, and $K$: The user, the item, and the keyphrase sets, respectively.
	\item $\bR \in \mathbb{R}^{|U|\times|I|}$: The user-by-item interaction matrix obtained with implicit feedback. Entries $\brui$ of~1~(respectively~0) denote a positive (respectively negative or unobserved) interaction between the user $u$ and item~$i$.
	\item $\bK \in \mathbb{R}^{|U|\times|K|}$: The binary user-keyphrase matrix that reflects the user $u$'s keyphrase-usage preference. Given user reviews from a corpus, we extract keyphrases that describe item attributes from all reviews (see Section~\ref{sec_datasets}).
	\item $\bKI \in \mathbb{R}^{|I|\times|K|}$: The binary item-keyphrase matrix. The process is similar to $\bK$ with the aggregation per item.
	\item $\bhru \in \mathbb{R}^{|I|}$ and $\bhku \in \mathbb{R}^{|K|}$: The predicted feedback and keyphrase explanation, respectively.
	\item $\bzu \in \mathbb{R}^{|H|}$: The user $u$'s latent embedding of dimension $H$ from the observed interaction $\bru$ and keyphrase-usage preference $\bku$.
	\item $\bzur \in \mathbb{R}^{|H|}$ and $\bzuk \in \mathbb{R}^{|H|}$: The user $u$'s latent embedding of dimension $H$ \textbf{only} from the observed interaction $\bru$ (symmetrically the keyphrase-usage preference $\bku$).
	\item $\bcut \in \mathbb{R}^{|K|}$: A one-hot vector of length $|K|$. The only positive value indicates the index of the keyphrase to be critiqued by the user $u$ at a given step $t$ of the user interaction with the recommender system.
	\item $\bzuct \in \mathbb{R}^{|H|}$: The latent representation of the critique $\bcut$.
	\item $\btzuct \in \mathbb{R}^{|H|}$: The updated latent representation of the user after the critique $\bcut$.
	\item $I^{+\bc} \in \{i|\bkIic=1, \forall i \in I\}$: The set of items that contain the critiqued keyphrase $\bc$.
	\item $I^{-\bc} \in \{i|\bkIic=0, \forall i \in I\}$: The set of items that do not contain the critiqued keyphrase $\bc$.
\end{itemize}

\subsection{Variational Autoencoder for Recommendation (VAE)}
\label{sec_vae}
A variational autoencoder (VAE)~\citep{kingmamw2014} is a generative model of the form $p_\theta(\bx, \bz) = p(\bz)p_\theta(\bx | \bz)$, where $p(\bz)$ is a prior and the likelihood $p_\theta(\bx | \bz)$ is parametrized by a neural network with parameters $\theta$. The model learns to maximize the marginal likelihood of the data $p_\theta(\bx)$ (i.e., the evidence that is intractable) by approximating the true unknown posterior $p_\theta(\bz | \bx)$ with a variational posterior $q_\phi(\bz | \bx)$. Applied to recommendation systems, the collaborative-filtering VAE~\citep{liang2018variational} considers as input data the sparse user preferences $\bru$ over $|I|$ items. More formally, the model optimizes a variational lower bound on the log likelihood of all observed user feedback $\sum_{u \in U} \log p(\bru)$ through stochastic gradient~descent:\begin{equation}
\begin{split}
	 \log p(\bru) &\ge \int_{\bzu} q_\phi(\bzu | \bru) \log \frac{p_\theta(\bru, \bzu)}{q_\phi(\bzu | \bru)} d\bzu \\ & \ge \mathbb{E}_{q_\phi(\bzu | \bru)} \bigl[\log p_\theta(\bru | \bzu)\bigr] - \beta \textrm{ D}_{\textrm{KL}} \bigl[ q_\phi(\bzu | \bru) ~||~ p(\bzu) \bigr],
\end{split}
\end{equation} where $\bzu$ is sampled\footnote{Using the reparametrization trick~\citep{kingmamw2014,10.5555/3044805.3045035}: $\bzu = \bmuu + \epsilon \bb{\sigma}_u$, where $\epsilon \sim \mathcal{N}(0, \mathbb{I}_H)$.} from the distribution $q_\phi(\bzu | \bru)$ with parameters $\bmuu$ and $\bSu$, and $\textrm{D}_{\textrm{KL}}[q, p]$ denotes the Kullback-Leibler divergence (KL) between the distributions $p$ and $q$. In practice, the prior $p(\bz)$ is usually a spherical Gaussian with parameters $\bmu$ and $\bS$. Finally, $\beta$ is a hyperparameter that controls the strength of the regularization relative to the reconstruction error, as motivated by the $\beta$-VAE of~\cite{higgins2016beta}, and is slowly annealed to 1, similarly to~\cite{bowman-etal-2016-generating}.

\subsection{Co-embedding of Language-based Feedback with the Variational Autoencoder (CE-VAE)} \label{sec_cevae}
Thus far, the variational autoencoder can only recommend items without generating any form of explanation. A recent study~\citep{luo2020} proposed the CE-VAE model, which integrates an explanation and critiquing module based on keyphrases. The authors support critiquing by first modeling the joint probability of a user's item preferences and keyphrase usage: \begin{equation}
 \begin{split}
\log p(\bru, \bku) &= \log p(\bku | \bru) + \log p(\bru) \\ &= \mathbb{E}_{q_{\Phi_r}(\bzu | \bru)} \bigl[\log p_{\Theta_k}(\bku | \bzu)\bigr] - \textrm{D}_{\textrm{KL}} \bigl[ q_{\Phi_r}(\bzu | \bru) ~||~ p(\bzu) \bigr]\\
&+  \mathbb{E}_{q_{\Psi_r}(\bzu | \bru)} \bigl[\log p_{\Theta_r}(\bru | \bzu)\bigr] + \mathcal{H}\bigl[q_{\Psi_r}(\bzu | \bru)\bigr] + \mathbb{E}_{q_{\Psi_r}(\bzu | \bru)} \bigl[p(\bzu) \bigr],
\end{split}
\end{equation}
where $\mathcal{H}$ is the entropy. Then, they incorporate an additional objective to learn a projection from the critiquing feedback into the latent space via another encoder (an inverse feedback loop). In other words, they reintroduce the user's keyphrase usage $\bku$ to approximate the variational lower bound of $p(\bzu)$ by marginalizing over $\bku$. More formally:\begin{equation}
\begin{split}
\log p(\bzu) &\ge \mathbb{E}_{q(\bku | \bzu)} \bigl[\log p(\bzu | \bku)\bigr] - \textrm{D}_{\textrm{KL}} \bigl[ q(\bku | \bzu) ~||~ p(\bku) \bigr]\\
&\approx \mathbb{E}_{p_{\Theta_k}(\bku | \bzu)} \bigl[\log p_{\Theta_k'}(\bzu | \bku)\bigr] - \textrm{D}_{\textrm{KL}} \bigl[ p_{\Theta_k}(\bku | \bzu) ~||~ p(\bku) \bigr],
\end{split}
\end{equation}
where $p(\bku)$ is a prior following a standard normal distribution and the weights of $q(\bku | \bzu)$ are shared with $p_{\Theta_k}(\bku | \bzu)$.

Finally, once the model is trained on the full objective function, the critiquing process for the critique $\bcu$ is performed as follows: 
\begin{enumerate}[topsep=0pt]
	\item compute the critique representation $\bzuc$ with $p_{\Theta_k'}(\bzu | \bku)$,
	\item average both the user latent representation~$\bzu$ and the critique representation $\bzuc$, and 
	\item predict the new feedback $\bhru$ with the generative network~$p_{\Theta_r}(\bru | \bzu)$.
\end{enumerate}

Overall, the CE-VAE framework is effective in practice for recommendation, keyphrase explanation, and single-step critiquing. However, it suffers from two key deficiencies that limit its performance (as we later show empirically):\begin{enumerate}
	\item The model learns a function to project the critiqued keyphrase into the user's latent space, from which the~feedback and the explanation are predicted. This mapping is learned via an autoencoder, which perturbs the training. Thus, there is an inherent trade-off between the performance of the recommendation and that of the~explanation.
	\item Although the joint objective also maximizes a latent representation likelihood with the Kullback-Leibler terms, it is unclear~whether the inverse function embeds the critique effectively and whether the mean reflects a critiquing mechanism.\footnote{The subsequent and concurrent work of \cite{10.1145/3404835.3463108} has addressed these issues using Bayesian approximation. Although the recommender system is trained to predict ratings, it could be adapted for implicit feedback. For future work, we will adapt the model and include it in our experiments.}
\end{enumerate}

\section{M\&Ms-VAE: A Mixture-of-Experts Multimodal Variational AutoEncoder}
Our goal is to build a more generalizable representation of users' preferences that is based on their observed interactions and keyphrase usage. 
Figure~\ref{fig_gm} depicts the graphical model of our proposed M\&Ms-VAE, and Figure~\ref{fig_training} shows the training scheme. 
Then, we leverage this representation to efficiently embed the user critiques and learn, in a self-supervised fashion, a blending module to re-rank recommended items for multi-step critiquing. Figure~\ref{fig_critiquing_general} illustrates the workflow.

\subsection{Model Overview}
\begin{figure}[!t]
\centering
  \begin{tikzpicture}
  	  \node[latent] (z) {$\bz$};
      \node[latent, below=of z, xshift=-0.75cm,path picture={\fill[gray!25] (path picture bounding box.south) rectangle (path picture bounding box.north west);}] (r) {$\br$} ;
      \node[latent, below=of z, xshift=0.75cm,path picture={\fill[gray!25] (path picture bounding box.south) rectangle (path picture bounding box.north west);}] (k) {$\bk$} ;
      
	  \node[xshift=1.9cm] (1) {$\Phi_k$};
	  \node[xshift=-1.9cm] (2) {$\Phi_r$};
	  \node[below=of z, yshift=-0.1cm, xshift=1.9cm] (3) {$\Theta_k$};
	  \node[below=of z, yshift=-0.1cm, xshift=-1.9cm] (4) {$\Theta_r$};
    \plate[inner sep=0.3cm, xshift=0cm, yshift=0cm] {plate} {(r) (k) (z)} {$u \in \{1 \dots U\}$};
	\path[->] (z) edge [bend left=20] node {} (r) ;
	\path[->] (z) edge [bend left=-20] node {} (k) ;
	\path[->, dashed] (r) edge [bend left=20] node {} (z) ;
	\path[->, dashed] (k) edge [bend left=-20] node {} (z) ;
	
	\path[->, dashed] (1) edge [] node {} (z) ;
	\path[->, dashed] (2) edge [] node {} (z) ;
	\path[->] (3) edge [] node {} (k) ;
	\path[->] (4) edge [] node {} (r) ;
  \end{tikzpicture}
    \caption{\label{fig_gm}Probabilistic-graphical-model view of our M\&Ms-VAE model. Both the implicit feedback $\bru$ and the keyphrase $\bku$ are generated from user $u$'s latent representation~$\bzu$. Solid lines denote the generative model $p_\Theta$, whereas dashed lines denote the variational approximation $q_\Phi$.}
\end{figure}
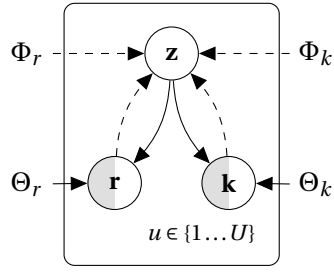

Like previously developed variational autoencoders for recommendation, we assume that the observed~user $u$'s interactions~$\bru$ and the keyphrase-usage preference $\bku$ are generated from a latent representation of the user preferences.

Differently from prior work, we seek to learn the joint distribution $p(\bru,\bku)$ under weak supervision. Our main~goal~is to learn a more generalizable representation of the user preferences. Therefore, we aim to design a generative model~that can recommend and generate keyphrase explanation \textbf{jointly but also independently} from each of the observed~variables (i.e., cross-modal generation). This condition is important as users are, most of the time, not required to write textual feedbacks of the consumed items.
It also allows us to apply the same technique to users who have not written~reviews or to cases in which keyphrases are unavailable.
 If this goal is achieved, we can then embed effectively the~user's observed interactions, the user's keyphrase preference, and the critique with the \textbf{same} inference network~$q_\Phi(\bzu | \bru, \bku)$.

Inspired by multimodal generative models~\citep{NEURIPS2018_1102a326,NEURIPS2019_0ae775a8,tsai2018learning,NEURIPS2020_43bb733c}, we treat $\bru$ and $\bku$ as different modalities, and we assume they are conditionally independent given the common latent variable $\bzu$. In other words, we assume a generative model of the form $p_\Theta(\bru, \bku, \bzu)=p(\bzu)p_{\Theta_r}(\bru|\bzu)p_{\Theta_k}(\bku|\bzu)$. An advantage of such a factorization is that if $\bru$ or~$\bku$ is unobserved, we can safely ignore it when evaluating the marginal likelihood~\citep{NEURIPS2018_1102a326}.

We start with the derivation of the joint log likelihood $\sum_{u \in U} \log p(\bru, \bku)$ over the observed interactions~$\bru$ and keyphrase-usage preference $\bku$ and all users $u$ as shown in Figure~\ref{fig_gm}:\begin{equation}
\label{eq_mm_1}
\begin{split}
\log~p(\bru, \bku) &= \log \int_{\bzu} p_\Theta(\bru, \bku, \bzu) d\bzu \\ &\ge \mathbb{E}_{q_\Phi(\bzu | \bru, \bku)} \bigl[\log p_\Theta(\bru, \bku | \bzu)\bigr] - \beta \textrm{ D}_{\textrm{KL}} \bigl[ q_\Phi(\bzu | \bru, \bku) ~||~ p(\bzu) \bigr],
\end{split}\end{equation} where we assume that the prior distribution $p(\bz)$ is a standard normal distribution and $\beta$ is a hyperparameter that controls the strength of the regularization relative to the reconstruction error.
Thanks to our assumption that $\bru$ and~$\bku$ are conditionally independent given the common latent variable $\bzu$, we can rewrite Equation~\ref{eq_mm_1} as follows (see Appendix~\ref{app_derivation} for the complete derivation):
\begin{equation}
\begin{split}
	\label{eq_mm_2}
	ELBO(\bru, \bku) = &\mathbb{E}_{q_\Phi(\bzu | \bru, \bku)} \bigl[\log p_{\Theta_r}(\bru | \bzu) + \log p_{\Theta_k}(\bku | \bzu)\bigr] \\ &- \beta \textrm{ D}_{\textrm{KL}} \bigl[ q_\Phi(\bzu | \bru, \bku) ~||~ p(\bzu) \bigr].
\end{split}
\end{equation}

Learning the variational joint posterior $q_\Phi(\bzu | \bru, \bku)$ of Equation~\ref{eq_mm_2} under its current form requires $\bru$ and $\bku$ to be presented at all times, thus making cross-modal recommendation difficult. Following our assumption, we can factorize the joint variational posterior as a function $\zeta(\cdot)$ of unimodal posteriors (or experts) $q_{\Phi_r}(\bzu | \bru)$ and $q_{\Phi_k}(\bzu | \bku)$: $q_\Phi(\bzu | \bru, \bku) = \zeta\bigl(q_{\Phi_r}(\bzu | \bru), q_{\Phi_k}(\bzu | \bku)\bigr)$, similarly to~\cite{NEURIPS2018_1102a326,NEURIPS2019_0ae775a8,tsai2018learning,NEURIPS2020_43bb733c}. In our case, the function $\zeta(\cdot)$ should be \begin{enumerate*}
 \item robust to overconfident experts if the marginal posterior $q_{\Phi_r}(\bzu | \bru)$ or $q_{\Phi_k}(\bzu | \bku)$ has low density, and
 \item robust to missing unobserved variable $\bru$ or $\bku$.
\end{enumerate*} Therefore, we propose to rely on a mixture of experts (MoE) with uniform weights: \begin{equation}
\label{eq_moe}
\begin{split}
	q_\Phi(\bzu | \bru, \bku) = \zeta\bigl(q_{\Phi_r}(\bzu | \bru), q_{\Phi_k}(\bzu | \bku)\bigr) =& \alpha \cdot q_{\Phi_r}(\bzu | \bru) + (1 - \alpha) \cdot q_{\Phi_k}(\bzu | \bku) \\ &\textrm{ with } \alpha = \begin{cases}
    \frac{1}{2}, & \text{if } \bru \textrm{ and } \bku \textrm{ are observed,}\\
    1, & \text{if only } \bru \textrm{ is observed,}\\
    0, & \text{if only } \bku \textrm{ is observed.}\\
\end{cases}
\end{split}
\end{equation}
We set the weights uniformly to explicitly enforce an equal contribution from each $\bru$ and $\bku$ when both are observed during training. In the case of an unobserved modality, we shift the importance distribution toward the presented one, which generalizes to weakly supervised learning (see Section~\ref{seq_training}). This is an important factor, because the inference network $q_{\Phi_k}(\bzu | \bku)$ will later induce the critique representation. Finally, one might be tempted to learn $\alpha$ jointly with the variational lower bound or dynamically. However, doing so might miscalibrate the precisions of the $q_{\Phi_r}(\bzu | \bru)$ or $q_{\Phi_k}(\bzu | \bku)$ and thus be detrimental to the whole model in terms of both prediction performance and generalization.

\subsection{Training Strategy}
\label{seq_training}
Combining Equations~\ref{eq_mm_2} and \ref{eq_moe} gives the full objective function, and M\&Ms-VAE can be trained on a complete dataset where all $\bru$ and $\bku$ are provided. However, in doing so, we never train the individual inference networks $q_{\Phi_r}(\bzu | \bru)$ and $q_{\Phi_k}(\bzu | \bku)$; only the relationship between the observed user interactions and keyphrase-usage preferences is captured. As a consequence, at inference, it is unclear how the model performs with a missing observation.

\begin{figure}[!t]
    \centering
    \includegraphics[width=.7\textwidth]{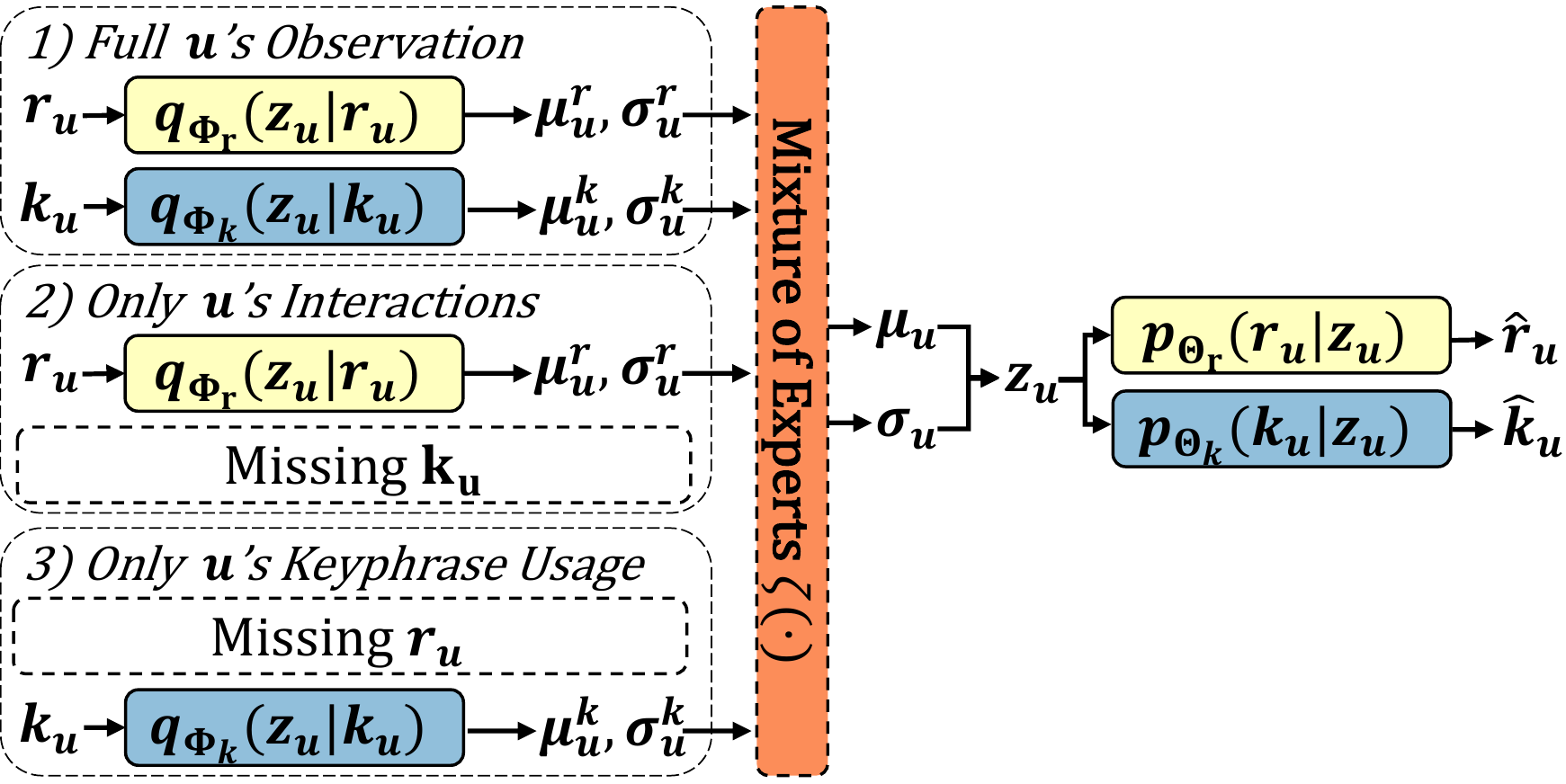}
    \caption{\label{fig_training}The proposed M\&Ms-VAE architecture and training scheme. Each pass infers the parameters $\bmuu$ and $\bb{\sigma}_u$ with the mixture of experts using either the joint inference network $q_\Phi(\bzu | \bru, \bku)$ or one of the individual networks ($q_{\Phi_r}(\bzu | \bru)$ or $q_{\Phi_k}(\bzu | \bku)$, respectively). The final gradient is computed on the sum of each $ELBO(\cdot)$ term.}
\end{figure}

To reach our goal of recommending given at least $\bru$ and embedding the critique effectively with the inference network $q_{\Phi_k}(\bzu | \bku)$, we propose a training strategy that mimics weakly supervised learning, similarly to~\cite{NEURIPS2018_1102a326}. Moreover, this allows us to handle incomplete datasets, where some samples are partially observed: data that contain only $\bru$~or $\bku$ \footnote{It also enables another way to solve the cold-start problem: new users can select a set of items and/or relevant keyphrases that reflect their preferences.}. The training strategy is shown in Figure~\ref{fig_training}. For each minibatch, we compute the gradient on the evidence lower bound of the joint observation and each single observation $\bru$ and~$\bku$. Our final training objective for all users $u$ is \begin{equation}
	\label{eq_mm_final}
	\begin{split}
	\mathcal{L}(\bR, \bK) =& \underbrace{\sum_{u \in U}\lambda \cdot \mathbb{E}_{q_\Phi(\bzu | \bru, \bku)} \bigl[\log p_{\Theta_r}(\bru | \bzu) + \log p_{\Theta_k}(\bku | \bzu)\bigr] - \beta \textrm{ D}_{\textrm{KL}} \bigl[ q_\Phi(\bzu | \bru, \bku) ~||~ p(\bzu) \bigr]}_{\text{$ELBO(\bru,\bku)$}} \\
	&+ \underbrace{\sum_{u \in U} \lambda \cdot \mathbb{E}_{q_{\Phi_r}(\bzu | \bru)} \bigl[\log p_{\Theta_r}(\bru | \bzu)\bigr] - \beta  \textrm{ D}_{\textrm{KL}} \bigl[ q_{\Phi_r}(\bzu | \bru) ~||~ p(\bzu) \bigr]}_{\text{$ELBO(\bru)$}} \\
	&+ \underbrace{\sum_{u \in U} \lambda \cdot \mathbb{E}_{q_{\Phi_k}(\bzu | \bku)} \bigl[\log p_{\Theta_k}(\bku | \bzu)\bigr] - \beta \textrm{ D}_{\textrm{KL}} \bigl[ q_{\Phi_k}(\bzu | \bku) ~||~ p(\bzu) \bigr]}_{\text{$ELBO(\bku)$}},
\end{split}
\end{equation}where $\lambda$ and $\beta$ control the strength of the reconstruction error and regularization, respectively.

\subsection{Self-Supervised Critiquing with M\&Ms-VAE}

\begin{figure}[!t]
  \centering
  \includegraphics[width=0.8\textwidth]{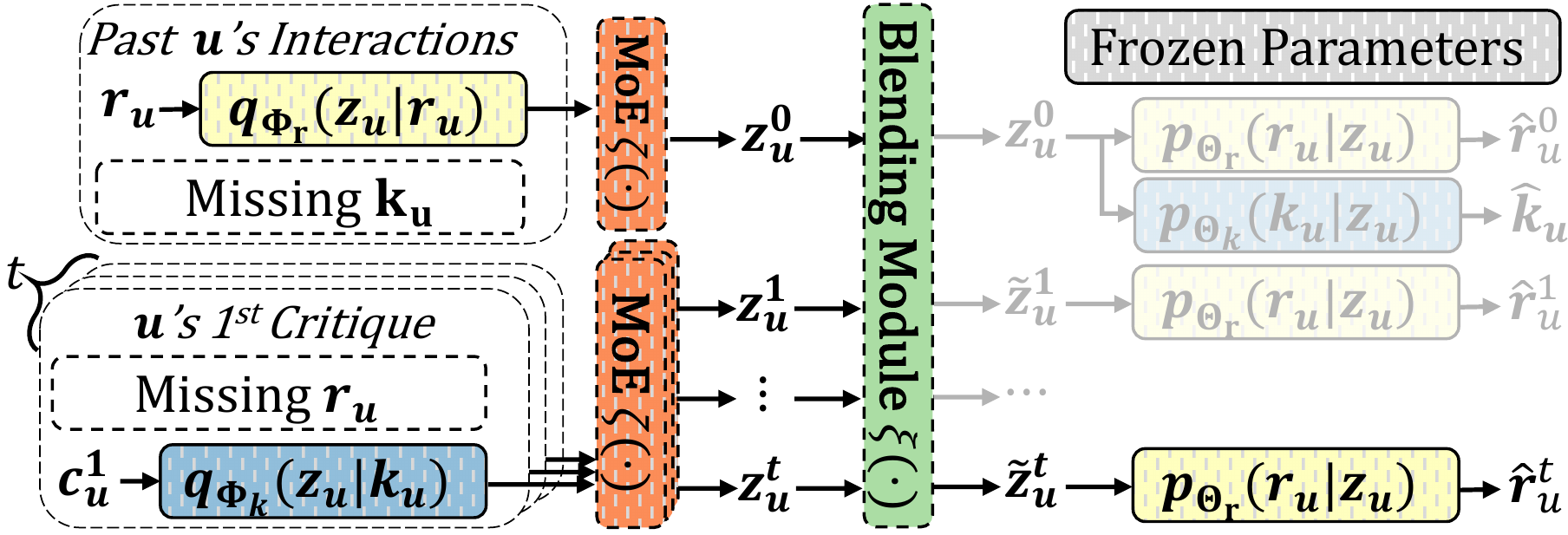}
  \caption{\label{fig_critiquing_general}Workflow of considering the recommendation of items~to~a user $u$ over $t$+1 time steps. First, M\&Ms-VAE produces the~initial set of recommended items $\bhruzero=\bhru$ using only the historical observed interactions $\bru$. Then, the user can provide a critique~$\bcut$ that is encoded into $\bzuct$ via the inference model~$q_{\Phi_k}(\bzu | \bku)$.~The blending module combines the previous representations~$\bz_{u}^{0}$,~$\bz_{u}^{1}$, $\dots, \bzuct$ into $\btzuct$, from which the subsequent recommendation~$\bhrut$ is computed. This process continues until the user $u$~accepts the recommendation and ceases to provide additional critiques.}
\end{figure}

The purpose of critiquing is to refine the recommendation $\bhru$ based on the user $u$'s interaction with the explanation $\bhku$, represented with a binary vector. The user can accept the recommended items, at which point the session terminates. In the other case, the user can provide a critique $\bc_{u}^{t}$ and obtain a new recommendation $\bhrut$. The process is repeated over $T$ iterations until the user $u$ is satisfied with the recommendation. Each critique $\bcut$ is encoded as a one-hot vector where the positive value indicates a keyphrase the user $u$ dislikes. The overall process is depicted in Figure~\ref{fig_critiquing_general}.

We leverage the generalization ability of the trained M\&Ms-VAE, especially the inference models $q_{\Phi_r}(\bzu | \bru)$ and $q_{\Phi_k}(\bzu | \bku)$. We use the former to represent the initial user preferences $\bru$ and the latter to embed the critique~$\bcut$. However, a crucial question remains: how should we blend the user representation $\bzu$ with the $t$\textsuperscript{th} critique representation ~$\bzuct$?

Prior work has implemented a blending function as a simple average~\citep{keyphraseExtractionDeep,luo2020} or as a linear programming task that looks for a convex combination of embeddings provided with a specific linear optimization objective~\citep{luo2020b,hanze2020}. As we demonstrate empirically later, the former yields poor performance when iterated for multi-step critiquing, whereas the latter is computationally slow because the optimization is performed for each critique and it cannot leverage GPUs.

\subsubsection{Blending Function Design.}

We consider the critiquing process as a short session. We propose to learn a blending function $\xi(\cdot)$ built upon a trained M\&Ms-VAE model whose weights are frozen. This two-step approach has several advantages:\begin{enumerate}
 	\item The original training of M\&Ms-VAE is not perturbed by the critiquing objective.
 	\item $\xi(\cdot)$ is decoupled from the model. It allows more flexibility in its architecture, objective function, and training.
 \end{enumerate}

We assume that each critique is independent, as in~\cite{luo2020b,keyphraseExtractionDeep,antognini2020interacting}. At time step $t$, we express the new user preferences~$\btzuct$ with a linear interpolation between the original latent representation $\bzu^0$, inferred from the observed interaction $\bru$ \textbf{only},
 and the critique $\bcut$'s representation $\bzuct$. More precisely, we use the gating mechanism of the gated recurrent unit~\citep{chung2014empirical} (we omit the biases to reduce notational clutter):\begin{equation}
 \begin{split}
 	\btzuct = \xi(\bzu^0,\bzuct) = h_2
 \end{split}
\quad
  \begin{split}
    h_2 &= (1-u_1) \cdot n_1 + u_1 \cdot h_1\\
    n_1 &= \textrm{tanh}(W_{in} \bzuct + W_{hn}(r_1 \odot h_1))\\
    u_1 &= \sigma(W_{iu} \bzuct + W_{hz} h_1)\\
    r_1 &= \sigma(W_{ir} \bzuct + W_{hr} h_1)
  \end{split}
\quad
  \begin{split}
    h_1 &= (1-u_0) \cdot n_0 + u_0 \cdot h_0\\
    n_0 &= \textrm{tanh}(W_{in} \bzu^0 + W_{hn}(r_0 \odot h_0))\\
    u_0 &= \sigma(W_{iu} \bzu^0 + W_{hz} h_0)\\
    r_0 &= \sigma(W_{ir} \bzu^0 + W_{hr} h_0)
  \end{split},
\end{equation} where $W_{ir},W_{iu},W_{in},W_{hr},W_{hz},W_{hn}$, and the bias vectors are the model parameters.

\subsubsection{Training.}

\begin{algorithm}[!t]
\caption{\label{alg_dataset}Synthetic Critiquing Dataset Creation}
\begin{algorithmic}[1]
\Function{Generate}{$\bR^{\textrm{val}}, \bKI$}
	\State Synthetic dataset $D \leftarrow \{\}$
    \For{each user $u$}
    	\For{each target item $i$, where $\brui^{\textrm{val}}=1$}
        \State Randomly sample a critique $\bc \in K \backslash\bkIi$
        \State Compute the item sets $I^{+\bc}$ and $I^{-\bc}$
        \State Update $D \leftarrow D \cup \{(u, i, \bc, I^{+\bc}, I^{-\bc})\}$
        \EndFor
    \EndFor
    \State \textbf{return} Synthetic dataset $D$
\EndFunction
\end{algorithmic}
\end{algorithm}

Thanks to our assumption and the generalization ability of M\&Ms-VAE, we can learn the weights of the blending module $\xi(\cdot)$ by creating a synthetic dataset based only on the validation set (see Algorithm~\ref{alg_dataset}).~For each user and observed interaction, we randomly sample a keyphrase $\bc$ that is inconsistent with the target item. Then, we calculate the item sets $I^{+\bc}$ that contain the critique and, symmetrically, the item sets $I^{-\bc}$ for those that do not~contain~it\footnote{In early experiments, we generalized $\xi(\bzu^0,\bzuct)$ to $\xi(\bzu^0,\bzu^1,\dots,\bz_{u}^{t})$ and updated Algorithm~\ref{alg_dataset} accordingly. However, our synthetic dataset cannot cover such a space due to the exponential number of combinations; session-based recommenders require millions of real sessions as training data~\citep{hidasi2015sessionbased,hidasi2018sessionbased}.}.

Our final objective is to re-rank items based on the user preferences and the provided critique~$\bc$. Recall that M\&Ms-VAE's weights are frozen. Let $\bhru^0$ be the user $u$'s initial predictions $\bhru$ and $\bhru^1$ those inferred from $\btzuct$ after the critique. We express this overall ranking-based objective via two differentiable max-margin objective functions~\citep{Rosasco_areloss}. We train $\xi(\cdot)$ and minimize the following loss function:
\begin{equation}
	\label{eq_ssc}
	\begin{split}
	\mathcal{L}(\bhRzero,\bhRone,u,\bc,I^{+\bc},I^{-\bc}) &= \mathcal{L}(p_{\Theta_r}(\bzu^0),p_{\Theta_r}\big(\xi(\bzu^0,\bzuct)\big),u,\bc,I^{+\bc},I^{-\bc })\\
	&= \sum_{i^+ \in I^{+\bc}} \biggl[ \max \Big\{0, h - (\bhr_{u,i^+}^{0} - \bhr_{u,i^+}^{1})\Big\} \biggr]\\
	 &+ \sum_{i^- \in I^{-\bc}} \biggl[ \max \Big\{0, h - (\bhr_{u,i^-}^{1} - \bhr_{u,i^-}^{0})\Big\} \biggr],
\end{split}
\end{equation} where $h$ is the margin. Intuitively, $\xi(\cdot)$ is encouraged to create a representation $\btzuct$ from which $p_{\Theta_r}(\cdot)$ gives a lower ranking to the items affected by the critique in the next iteration (i.e., $\bhr_{u,i^+}^{1} < \bhr_{u,i^+}^{0}$) and a higher ranking to the unaffected items (i.e., $\bhr_{u,i^-}^{1} > \bhr_{u,i^-}^{0}$).
Finally, Equation~\ref{eq_ssc} is efficiently parallelizable on both CPUs and GPUs. 

\section{Experiments}

In this section, we proceed to evaluate the proposed M\&Ms-VAE model in order to answer the following questions:
\begin{itemize}
	\item \textbf{RQ 1}: How does M\&Ms-VAE perform in terms of recommendation and explanation performance?
	\item \textbf{RQ 2}: Can M\&Ms-VAE with the self-supervised critiquing objective enable multi-step critiquing? 
	\item \textbf{RQ 3}: What is our proposed critiquing algorithm's computational time complexity compared to prior~work? 
	\item \textbf{RQ 4}: How does M\&Ms-VAE perform under weak supervision; how coherent is the joint and cross generation?
\end{itemize}

\begin{table}[!t]
    \centering
   \caption{\label{stats_datasets}Descriptive statistics of the datasets. Coverage shows the ratio of reviews having at least one of the selected keyphrases~(KPs)}
\hspace*{-1.0cm}
\begin{threeparttable}
\begin{tabular}{@{}l@{}c@{\hspace{2mm}}c@{\hspace{2mm}}c@{\hspace{2mm}}c@{\hspace{2mm}}c@{\hspace{2mm}}c@{\hspace{2mm}}c@{\hspace{3mm}}c@{}c@{}}
\textbf{Dataset} & \textbf{\#Users} & \textbf{\#Items} & \textbf{\#Interactions} & \textbf{Sparsity} & \textbf{\#KPs} & \textbf{KP Coverage} & \textbf{Avg. KPs/Review} & \textbf{AVG. KPs/User}\\
\toprule
Beer & 6,370 & 3,669 & 263,244 & 1.13\% & 75 & 99.27\% & 7.16 & 1,216\\
CDs\&Vinyl & 6,060 & 4,395 & 152,783 & 0.57\% & 40 & 74.59\% & 2.13 & 73\\
Yelp & 9,801 & 4,706 & 140,496 & 0.30\% & 234 & 96.65\% & 7.45 & 300\\
Hotel & 7,044 & 4,874 & 143,612 & 0.42\% & 141 & 99.99\% &17.42 & 419

\end{tabular}
\end{threeparttable}
\end{table}

\subsection{Datasets}
\label{sec_datasets}
We evaluate the quantitative performance of M\&Ms-VAE using four real-world, publicly available datasets: BeerAdvocate~\citep{beer}, Amazon CDs\&Vinyl~\citep{amazon1,amazon2}, Yelp~\citep{yelpdataset}, and HotelRec~\citep{antognini-faltings-2020-hotelrec}. Each contains more than 100k reviews with~five-star ratings. For the purpose of Top-N recommendation, we binarize the ratings with a threshold $t > 3.5$. Because~people tend to rate beers and restaurants positively, we set the threshold $t > 4$ and $t > 4.5$, respectively. We split each dataset into~60\%/20\%/20\% for the training, validation, and test sets. Table~\ref{stats_datasets} shows the data statistics. All contain complete~observations.
The datasets do not contain preselected keyphrases. Hence, we extract them for the explanations and critiquing with the frequency-based processing of~\cite{keyphraseExtractionDeep,hanze2020}. Some examples are shown in Table~\ref{tab_keyphrases}.

\begin{table}[!t]
    \centering
   \caption{\label{tab_keyphrases}Some keyphrases mined from the reviews. We manually grouped them by types for a better understanding.}
\begin{threeparttable}
\begin{tabular}{@{}lllc@{}}
\textbf{Dataset} & \textbf{Type} & \textbf{Keyphrases}\\
\toprule
\multirow{4}{*}{Beer} & Head & white, tan, offwhite, brown\\
& Malt & roasted, caramel, pale, wheat, rye\\
& Color & golden, copper, orange, black, yellow\\
& Taste & citrus, fruit, chocolate, cherry, plum\\
\bottomrule
\multirow{4}{*}{CDs\&Vinyl} & Genre & rock, pop, jazz, rap, hip hop, R\&B\\
& Instrument & orchestra, drum\\
& Style & concert, opera\\
& Religious & chorus, christian, gospel\\
\bottomrule
\multirow{4}{*}{Yelp} & Cuisine & chinese, thai, italian, mexican, french\\
& Drink & tea, coffee, bubble tea, wine, soft drinks\\
& Food & chicken, beef, fish, pork, seafood, cheese\\
& Price \& Service & cheap, pricy, expensive, busy, friendly\\
\bottomrule
\multirow{4}{*}{Hotel} & Service & bar, lobby, housekeeping, guest, shuttle\\
& Cleanliness & toilet, sink, tub, smoking, toiletry, bathroom\\
& Location & airport, downtown, city, shop, restaurant\\
& Room & bed, tv, balcony, terrace, kitchen, business\\
\bottomrule
\end{tabular}
\end{threeparttable}
\end{table}

\subsection{Experimental Settings}
Across experiments, we treat the prior and the likelihood as standard normal and multinomial distributions, respectively. The inference and generative networks consist of a two-layer neural network with a tanh nonlinearity as the activation function between the layers. We normalize the input and use dropout~\citep{srivastava2014dropout}. For learning, we employ the Adam optimizer~\citep{KingmaB14} with AMSGrad~\citep{47409} and a learning rate of $5\cdot10^{-5}$. We anneal linearly the regularization parameter $\beta$ of the Kullback-Leibler terms.
For the baselines, we reused the authors' code and tuning procedure.
We select hyperparameters and architectures for each model by evaluating NDCG on the validation set. We limit the search to a maximum of 100 trials. For critiquing, we tune our blending module on the synthetic dataset with the Falling MAP metric on the validation set, which measures the effect of a critique~\citep{keyphraseExtractionDeep}. 
For reproducibility purposes, we include additional details and the best hyperparameters in Appendix~\ref{app:add_repr}.

\subsection{RQ 1: How does M\&Ms-VAE perform in terms of recommendation and explanation performance?}
\label{sec_rq1}

In this section, we investigate whether the factorization and the training strategy of our proposed M\&Ms-VAE model benefit the recommendation and explanation performance.

\subsubsection{Baselines.} \label{sec_rq1_baselines}We compare our proposed M\&Ms-VAE model to the following baseline models. \textbf{POP} returns the most popular items without any kind of personalization. \textbf{AutoRec}~\citep{10.1145/2740908.2742726} is a neural autoencoder-based recommendation system. \textbf{BPR}~\citep{10.5555/1795114.1795167} is a Bayesian personalized ranking model that explicitly optimizes pairwise rankings. \textbf{CDAE}~\citep{10.1145/2835776.2835837} denotes a collaborative denoising autoencoder that is specifically optimized for implicit feedback recommendation tasks. \textbf{NCE-PLRec}~\citep{10.1145/3331184.3331201} represents the linear recommendation projected by noise-contrastive estimation; it augments PLRec with noise-contrasted item embeddings. \textbf{PLRec}~\citep{sedhain2016practical} is the ablation variant of NCE-PLRec without the noise-contrastive estimation. \textbf{PureSVD}~\citep{10.1145/1864708.1864721} denotes a similarity-based recommendation method that constructs a similarity matrix through SVD decomposition of the implicit rating matrix. \textbf{VAE-CF} is the variational autoencoder for collaborative filtering described in Section~\ref{sec_vae}. \textbf{CE-VNCF}~\citep{keyphraseExtractionDeep} is the extension of the neural collaborative filtering model~\citep{he2017neural} that is augmented with an explanation and a critiquing neural component. Finally, \textbf{CE-VAE}~\citep{luo2020} is a significant improvement over CE-VNCF, and it produces state-of-the-art performance (more details in Section~\ref{sec_cevae}).
For a fair comparison, we encode the user observations in M\&Ms-VAE using solely the inference network $q_{\Phi_r}(\bzu | \bru)$ at test time. We provide in Section~\ref{seq_rq4} the results with the other settings.

\subsubsection{Top-N Recommendation Performance.}
\label{sec_rq1_rec} We report the following metrics: R-Precision and NDCG; MAP,~Precision, and Recall at different Top-N. The main results are presented in Table~\ref{table_rec_perf}. We make the following key observations. 

Overall, M\&Ms-VAE shows the best recommendation performance for all metrics on three datasets and nearly all metrics on the CDs\&Vinyl dataset.
Compared to the original VAE recommender~(VAE-CF), M\&Ms-VAE achieves an improvement of 13\% on average. We conjecture that the extra loss terms ($ELBO(\bru, \bku)$ and $ELBO(\bku)$) help to generate better user~representations by leveraging both user preferences and keyphrase usage with the mixture of experts.

M\&Ms-VAE also significantly outperforms CE-VAE on the Yelp and Hotel datasets (by a factor of 1.9 and 1.7, respectively) and achieves an average improvement of 9\% on the Beer and CDs\&Vinyl datasets. We remark the same trend with CE-VNCF. These results emphasize the noise introduced in CE-VAE and CE-VNCF during training when learning the mapping between the keyphrases and the latent space. This is even more pronounced with a large number of keyphrases (i.e., over 100). In contrast, M\&Ms-VAE is more robust thanks to our factorization and training strategy, which mimics a weak supervision setting;

Interestingly, PureSVD exhibits the second-best performance on the CDs\&Vinyl and Yelp datasets. This shows that classic algorithms often remain competitive with state-of-the-art VAE-based recommender systems.

\begin{table}[H]
\footnotesize
    \centering
\caption{\label{table_rec_perf}Top-N recommendation results of all datasets. \textbf{Bold} and \underline{underline} denote the best and second-best results, respectively. We omit the error bars because the 95\% confidence interval is in 4\textsuperscript{th} digit.}
\hspace*{-1.25cm}
\begin{threeparttable}
\begin{tabular}{@{}cl@{}
c@{\hspace{1mm}}c@{\hspace{0mm}}c@{\hspace{4mm}}
c@{\hspace{2mm}}c@{\hspace{2mm}}c@{}c@{\hspace{4mm}}
c@{\hspace{2mm}}c@{\hspace{2mm}}c@{}c@{\hspace{4mm}}
c@{\hspace{2mm}}c@{\hspace{2mm}}c@{}}
& & & & & \multicolumn{3}{c}{\textbf{MAP@N}} & & \multicolumn{3}{c}{\textbf{Precision@N}} & & \multicolumn{3}{c}{\textbf{Recall@N}}\\
\cmidrule{6-8}\cmidrule{10-12}\cmidrule{14-16}
& \textbf{Model} & \textbf{R-Precision} & \textbf{NDCG} & & $N=5$ & $N=10$ & $N=20$ & & $N=5$ & $N=10$ & $N=20$ & & $N=5$ & $N=10$ & $N=20$\\
\toprule
\multirow{11.5}{*}{\rotatebox{90}{\textit{Beer}}}
& POP & $0.0307$ & $0.0777$ &  & $0.0388$ & $0.0350$ & $0.0319$ &  & $0.0346$ & $0.0298$ & $0.0279$ &  & $0.0241$ & $0.0408$ & $0.0737$ \\
& AutoRec & $0.0496$ & $0.1140$ &  & $0.0652$ & $0.0591$ & $0.0527$ &  & $0.0574$ & $0.0503$ & $0.0438$ &  & $0.0392$ & $0.0663$ & $0.1129$ \\
& BPR & $0.0520$ & $0.1214$ &  & $0.0646$ & $0.0597$ & $0.0538$ &  & $0.0596$ & $0.0525$ & $0.0449$ &  & $0.0451$ & $0.0744$ & $0.1214$ \\
& CDAE & $0.0414$ & $0.0982$ &  & $0.0504$ & $0.0477$ & $0.0434$ &  & $0.0482$ & $0.0432$ & $0.0368$ &  & $0.0330$ & $0.0576$ & $0.0969$ \\
& NCE-PLRec & $0.0501$ & $0.1151$ &  & $0.0643$ & $0.0594$ & $0.0532$ &  & $0.0589$ & $0.0518$ & $0.0440$ &  & $0.0418$ & $0.0714$ & $0.1177$ \\
& PLRec & $0.0497$ & $0.1113$ &  & $0.0655$ & $0.0599$ & $0.0532$ &  & $0.0590$ & $0.0515$ & $0.0431$ &  & $0.0421$ & $0.0704$ & $0.1127$ \\
& PureSVD & $0.0450$ & $0.1052$ &  & $0.0479$ & $0.0473$ & $0.0446$ &  & $0.0493$ & $0.0455$ & $0.0396$ &  & $0.0391$ & $0.0689$ & $0.1131$ \\
& VAE-CF & $\underline{0.0538}$ & $\underline{0.1275}$ &  & $0.0642$ & $0.0594$ & $0.0536$ &  & $0.0595$ & $0.0525$ & $0.0448$ &  & $\underline{0.0473}$ & $\underline{0.0808}$ & $\underline{0.1327}$ \\
& CE-VAE & $0.0520$ & $0.1215$ &  & $\underline{0.0675}$ & $\underline{0.0618}$ & $\underline{0.0555}$ &  & $\underline{0.0620}$ & $\underline{0.0536}$ & $\underline{0.0461}$ &  & $0.0442$ & $0.0737$ & $0.1255$ \\
& CE-VNCF & $0.0440$ & $0.1099$ &  & $0.0546$ & $0.0512$ & $0.0472$ &  & $0.0504$ & $0.0465$ & $0.0411$ &  & $0.0353$ & $0.0635$ & $0.1116$ \\
& M\&Ms-VAE (Ours) & $\mathbf{0.0545}$ & $\mathbf{0.1307}$ &  & $\mathbf{0.0706}$ & $\mathbf{0.0650}$ & $\mathbf{0.0580}$ &  & $\mathbf{0.0649}$ & $\mathbf{0.0563}$ & $\mathbf{0.0473}$ &  & $\mathbf{0.0492}$ & $\mathbf{0.0833}$ & $\mathbf{0.1349}$ \\
\bottomrule
\multirow{11.5}{*}{\rotatebox{90}{\textit{CDs\&Vinyl}}}
& POP & $0.0088$ & $0.0265$ &  & $0.0108$ & $0.0102$ & $0.0095$ &  & $0.0098$ & $0.0095$ & $0.0082$ &  & $0.0088$ & $0.0182$ & $0.0327$ \\
& AutoRec & $0.0227$ & $0.0537$ &  & $0.0284$ & $0.0257$ & $0.0220$ &  & $0.0255$ & $0.0213$ & $0.0165$ &  & $0.0254$ & $0.0418$ & $0.0627$ \\
& BPR & $0.0632$ & $0.1516$ &  & $0.0724$ & $0.0639$ & $0.0543$ &  & $0.0640$ & $0.0513$ & $0.0408$ &  & $0.0807$ & $0.1263$ & $0.1939$ \\
& CDAE & $0.0135$ & $0.0365$ &  & $0.0173$ & $0.0158$ & $0.0141$ &  & $0.0152$ & $0.0136$ & $0.0116$ &  & $0.0143$ & $0.0262$ & $0.0451$ \\
& NCE-PLRec & $0.0749$ & $\underline{0.1739}$ &  & $0.0728$ & $0.0678$ & $0.0586$ &  & $0.0698$ & $0.0584$ & $0.0441$ &  & $\mathbf{0.1010}$ & $\mathbf{0.1608}$ & $\underline{0.2308}$ \\
& PLRec & $\underline{0.0760}$ & $0.1626$ &  & $\mathbf{0.0889}$ & $\underline{0.0777}$ & $\underline{0.0642}$ &  & $\underline{0.0773}$ & $\underline{0.0608}$ & $\underline{0.0444}$ &  & $0.0960$ & $0.1461$ & $0.2025$ \\
& PureSVD & $0.0652$ & $0.1551$ &  & $0.0570$ & $0.0565$ & $0.0509$ &  & $0.0612$ & $0.0527$ & $0.0405$ &  & $0.0914$ & $0.1486$ & $0.2149$ \\
& VAE-CF & $0.0638$ & $0.1699$ &  & $0.0540$ & $0.0554$ & $0.0517$ &  & $0.0600$ & $0.0540$ & $0.0440$ &  & $0.0949$ & $\underline{0.1593}$ & $\mathbf{0.2381}$ \\
& CE-VAE & $0.0708$ & $0.1532$ &  & $0.0816$ & $0.0711$ & $0.0588$ &  & $0.0715$ & $0.0555$ & $0.0411$ &  & $0.0903$ & $0.1357$ & $0.1937$ \\
& CE-VNCF & $0.0654$ & $0.1524$ &  & $0.0746$ & $0.0662$ & $0.0560$ &  & $0.0663$ & $0.0534$ & $0.0411$ &  & $0.0829$ & $0.1299$ & $0.1931$ \\
& M\&Ms-VAE (Ours) & $\mathbf{0.0801}$ & $\mathbf{0.1765}$ &  & $\underline{0.0885}$ & $\mathbf{0.0784}$ & $\mathbf{0.0660}$ &  & $\mathbf{0.0779}$ & $\mathbf{0.0628}$ & $\mathbf{0.0482}$ &  & $\underline{0.0983}$ & $0.1529$ & $0.2263$ \\
\bottomrule
\multirow{11.5}{*}{\rotatebox{90}{\textit{Yelp}}}
& POP & $0.0026$ & $0.0129$ &  & $0.0024$ & $0.0026$ & $0.0026$ &  & $0.0028$ & $0.0028$ & $0.0025$ &  & $0.0042$ & $0.0087$ & $0.0151$ \\
& AutoRec & $0.0034$ & $0.0133$ &  & $0.0032$ & $0.0030$ & $0.0028$ &  & $0.0027$ & $0.0027$ & $0.0025$ &  & $0.0038$ & $0.0081$ & $0.0153$ \\
& BPR & $0.0160$ & $0.0609$ &  & $0.0168$ & $0.0156$ & $0.0143$ &  & $0.0156$ & $0.0140$ & $0.0122$ &  & $0.0236$ & $0.0435$ & $0.0748$ \\
& CDAE & $0.0028$ & $0.0135$ &  & $0.0027$ & $0.0028$ & $0.0027$ &  & $0.0030$ & $0.0027$ & $0.0026$ &  & $0.0044$ & $0.0084$ & $0.0161$ \\
& NCE-PLRec & $0.0197$ & $0.0739$ &  & $0.0220$ & $0.0200$ & $0.0177$ &  & $0.0198$ & $0.0169$ & $0.0143$ &  & $0.0300$ & $0.0505$ & $0.0864$ \\
& PLRec & $0.0191$ & $0.0703$ &  & $0.0207$ & $0.0189$ & $0.0171$ &  & $0.0185$ & $0.0166$ & $0.0143$ &  & $0.0291$ & $0.0513$ & $0.0866$ \\
& PureSVD & $\underline{0.0253}$ & $\underline{0.0825}$ &  & $\underline{0.0279}$ & $\underline{0.0249}$ & $\underline{0.0217}$ &  & $\underline{0.0240}$ & $\underline{0.0206}$ & $\underline{0.0173}$ &  & $\underline{0.0357}$ & $\underline{0.0597}$ & $\underline{0.1008}$ \\
& VAE-CF & $0.0214$ & $0.0801$ &  & $0.0232$ & $0.0216$ & $0.0195$ &  & $0.0214$ & $0.0192$ & $0.0163$ &  & $0.0319$ & $0.0589$ & $0.0995$ \\
& CE-VAE & $0.0136$ & $0.0533$ &  & $0.0143$ & $0.0132$ & $0.0121$ &  & $0.0125$ & $0.0119$ & $0.0104$ &  & $0.0197$ & $0.0367$ & $0.0636$ \\
& CE-VNCF & $0.0166$ & $0.0693$ &  & $0.0175$ & $0.0167$ & $0.0157$ &  & $0.0165$ & $0.0154$ & $0.0144$ &  & $0.0251$ & $0.0467$ & $0.0889$ \\
& M\&Ms-VAE (Ours) & $\mathbf{0.0264}$ & $\mathbf{0.0909}$ &  & $\mathbf{0.0284}$ & $\mathbf{0.0261}$ & $\mathbf{0.0231}$ &  & $\mathbf{0.0260}$ & $\mathbf{0.0223}$ & $\mathbf{0.0188}$ &  & $\mathbf{0.0395}$ & $\mathbf{0.0682}$ & $\mathbf{0.1154}$ \\
\bottomrule
\multirow{11.5}{*}{\rotatebox{90}{\textit{Hotel}}}
& POP & $0.0047$ & $0.0188$ &  & $0.0049$ & $0.0047$ & $0.0042$ &  & $0.0047$ & $0.0043$ & $0.0036$ &  & $0.0054$ & $0.0098$ & $0.0167$ \\
& AutoRec & $0.0051$ & $0.0193$ &  & $0.0053$ & $0.0050$ & $0.0044$ &  & $0.0052$ & $0.0042$ & $0.0037$ &  & $0.0061$ & $0.0097$ & $0.0169$ \\
& BPR & $0.0181$ & $0.0623$ &  & $0.0198$ & $0.0185$ & $0.0169$ &  & $0.0183$ & $0.0168$ & $0.0146$ &  & $0.0219$ & $0.0409$ & $0.0713$ \\
& CDAE & $0.0050$ & $0.0190$ &  & $0.0054$ & $0.0049$ & $0.0044$ &  & $0.0050$ & $0.0043$ & $0.0037$ &  & $0.0057$ & $0.0098$ & $0.0172$ \\
& NCE-PLRec & $0.0229$ & $0.0684$ &  & $0.0244$ & $0.0226$ & $0.0200$ &  & $0.0228$ & $0.0195$ & $0.0160$ &  & $0.0283$ & $0.0484$ & $0.0785$ \\
& PLRec & $0.0242$ & $0.0664$ &  & $0.0265$ & $0.0234$ & $0.0201$ &  & $0.0228$ & $0.0190$ & $0.0155$ &  & $0.0284$ & $0.0466$ & $0.0758$ \\
& PureSVD & $0.0179$ & $0.0541$ &  & $0.0193$ & $0.0173$ & $0.0152$ &  & $0.0169$ & $0.0145$ & $0.0121$ &  & $0.0206$ & $0.0357$ & $0.0594$ \\
& VAE-CF & $\underline{0.0243}$ & $\underline{0.0755}$ &  & $\underline{0.0267}$ & $\underline{0.0241}$ & $\underline{0.0213}$ &  & $\underline{0.0238}$ & $\underline{0.0206}$ & $\underline{0.0171}$ &  & $\underline{0.0295}$ & $\underline{0.0511}$ & $\underline{0.0848}$ \\
& CE-VAE & $0.0147$ & $0.0538$ &  & $0.0151$ & $0.0146$ & $0.0136$ &  & $0.0148$ & $0.0137$ & $0.0122$ &  & $0.0184$ & $0.0334$ & $0.0595$ \\
& CE-VNCF & $0.0165$ & $0.0575$ &  & $0.0180$ & $0.0166$ & $0.0152$ &  & $0.0159$ & $0.0149$ & $0.0129$ &  & $0.0200$ & $0.0370$ & $0.0635$ \\
& M\&Ms-VAE (Ours) & $\mathbf{0.0272}$ & $\mathbf{0.0804}$ &  & $\mathbf{0.0290}$ & $\mathbf{0.0265}$ & $\mathbf{0.0235}$ &  & $\mathbf{0.0260}$ & $\mathbf{0.0227}$ & $\mathbf{0.0189}$ &  & $\mathbf{0.0314}$ & $\mathbf{0.0555}$ & $\mathbf{0.0928}$
\end{tabular}
\end{threeparttable}
\end{table}

\subsubsection{Top-K Explanation Performance.}
\label{sec_rq1_exp} We also compare M\&Ms-VAE with the user and item-popularity baselines~\citep{keyphraseExtractionDeep} that predict the explanation through counting and ranking the frequency of keyphrases for the users (symmetrically, the items) in the training set. Among the recommender baselines, only CE-VAE and CE-VNCF produce an explanation alongside the recommendation. We report the following metrics: NDCG, MAP, Precision, and Recall at different~Top-K. 

\begin{table}[!t]
\small
    \centering
\caption{\label{table_exp_perf}Top-K keyphrase explanation quality of all datasets. \textbf{Bold} and \underline{underline} denote the best and second-best results, respectively. We omit the error bars because the 95\% confidence interval is in 4\textsuperscript{th} digit.}
\hspace*{-1.25cm}
\begin{threeparttable}
\begin{tabular}{@{}c@{\hspace{2mm}}l
@{\hspace{1mm}}c@{\hspace{2mm}}c@{\hspace{2mm}}c@{}c@{\hspace{3.5mm}}
c@{\hspace{2mm}}c@{\hspace{2mm}}c@{}c@{\hspace{3.5mm}}
c@{\hspace{2mm}}c@{\hspace{2mm}}c@{}c@{\hspace{3.5mm}}
c@{\hspace{2mm}}c@{\hspace{2mm}}c@{}}
& & \multicolumn{3}{c}{\textbf{NDCG@K}} & & \multicolumn{3}{c}{\textbf{MAP@K}} & & \multicolumn{3}{c}{\textbf{Precision@K}} & & \multicolumn{3}{c}{\textbf{Recall@K}}\\
\cmidrule{3-5}\cmidrule{7-9}\cmidrule{11-13}\cmidrule{15-17}
& \textbf{Model} & $K=5$ & $K=10$ & $K=20$ & & $K=5$ & $K=10$ & $K=20$ & & $K=5$ & $K=10$ & $K=20$ & & $K=5$ & $K=10$ & $K=20$\\
\toprule
\multirow{5.5}{*}{\rotatebox{90}{\textit{Beer}}}
& UserPop & $0.0550$ & $0.0852$ & $0.1484$ &  & $0.0812$ & $0.0795$ & $0.0824$ &  & $0.0782$ & $0.0799$ & $0.0933$ &  & $0.0446$ & $0.0913$ & $0.2186$ \\
& ItemPop & $0.0511$ & $0.0807$ & $0.1428$ &  & $0.0767$ & $0.0749$ & $0.0777$ &  & $0.0697$ & $0.0740$ & $0.0895$ &  & $0.0402$ & $0.0856$ & $0.2107$ \\
& CE-VAE & $\underline{0.2803}$ & $\underline{0.4104}$ & $\underline{0.5916}$ &  & $\underline{0.9418}$ & $\underline{0.9168}$ & $\underline{0.8820}$ &  & $\underline{0.9186}$ & $\underline{0.8760}$ & $\underline{0.8186}$ &  & $0.1448$ & $0.2683$ & $0.4829$ \\
& CE-VNCF & $0.2390$ & $0.3221$ & $0.4080$ &  & $0.3414$ & $0.3117$ & $0.2690$ &  & $0.3145$ & $0.2633$ & $0.2026$ &  & $\mathbf{0.1966}$ & $\mathbf{0.3263}$ & $\mathbf{0.4962}$ \\
& M\&Ms-VAE (Ours) & $\mathbf{0.2817}$ & $\mathbf{0.4147}$ & $\mathbf{0.5960}$ &  & $\mathbf{0.9463}$ & $\mathbf{0.9237}$ & $\mathbf{0.8885}$ &  & $\mathbf{0.9243}$ & $\mathbf{0.8861}$ & $\mathbf{0.8256}$ &  & $0.1457$ & $0.2722$ & $\underline{0.4869}$ \\
\bottomrule
\multirow{5.5}{*}{\rotatebox{90}{\textit{CDs\&Vinyl}}}
& UserPop & $0.1028$ & $0.1285$ & $0.1869$ &  & $0.0910$ & $0.0807$ & $0.0700$ &  & $0.0860$ & $0.0636$ & $0.0595$ &  & $0.1157$ & $0.1704$ & $0.3392$ \\
& ItemPop & $0.1109$ & $0.1357$ & $0.1935$ &  & $0.0928$ & $0.0833$ & $0.0716$ &  & $0.0929$ & $0.0657$ & $0.0606$ &  & $0.1288$ & $0.1825$ & $0.3499$ \\
& CE-VAE & $\underline{0.5243}$ & $\underline{0.6468}$ & $\underline{0.7454}$ &  & $\underline{0.6441}$ & $\underline{0.5609}$ & $\underline{0.4575}$ &  & $\underline{0.5564}$ & $\underline{0.4324}$ & $\underline{0.3040}$ &  & $0.4795$ & $\underline{0.6808}$ & $\underline{0.8757}$ \\
& CE-VNCF & $0.4590$ & $0.5338$ & $0.5860$ &  & $0.3657$ & $0.2981$ & $0.2258$ &  & $0.2893$ & $0.2010$ & $0.1260$ &  & $\mathbf{0.5081}$ & $0.6698$ & $0.8127$ \\
& M\&Ms-VAE (Ours) & $\mathbf{0.5447}$ & $\mathbf{0.6659}$ & $\mathbf{0.7628}$ &  & $\mathbf{0.6648}$ & $\mathbf{0.5779}$ & $\mathbf{0.4700}$ &  & $\mathbf{0.5777}$ & $\mathbf{0.4441}$ & $\mathbf{0.3091}$ &  & $\underline{0.5015}$ & $\mathbf{0.6996}$ & $\mathbf{0.8894}$ \\
\bottomrule
\multirow{5.5}{*}{\rotatebox{90}{\textit{Yelp}}}
& UserPop & $0.0007$ & $0.0009$ & $0.0066$ &  & $0.0009$ & $0.0010$ & $0.0016$ &  & $0.0013$ & $0.0009$ & $0.0061$ &  & $0.0007$ & $0.0011$ & $0.0129$ \\
& ItemPop & $0.0008$ & $0.0011$ & $0.0073$ &  & $0.0009$ & $0.0011$ & $0.0018$ &  & $0.0015$ & $0.0011$ & $0.0065$ &  & $0.0009$ & $0.0013$ & $0.0149$ \\
& CE-VAE & $\underline{0.1935}$ & $\underline{0.2763}$ & $\underline{0.3803}$ &  & $\underline{0.6653}$ & $\underline{0.6356}$ & $\underline{0.5916}$ &  & $\underline{0.6363}$ & $\underline{0.5876}$ & $\underline{0.5181}$ &  & $\underline{0.1017}$ & $\underline{0.1819}$ & $\underline{0.3080}$ \\
& CE-VNCF & $0.0883$ & $0.1164$ & $0.1505$ &  & $0.1195$ & $0.1052$ & $0.0901$ &  & $0.1023$ & $0.0848$ & $0.0690$ &  & $0.0779$ & $0.1270$ & $0.2027$ \\
& M\&Ms-VAE (Ours) & $\mathbf{0.1949}$ & $\mathbf{0.2787}$ & $\mathbf{0.3837}$ &  & $\mathbf{0.6738}$ & $\mathbf{0.6428}$ & $\mathbf{0.5976}$ &  & $\mathbf{0.6434}$ & $\mathbf{0.5935}$ & $\mathbf{0.5229}$ &  & $\mathbf{0.1019}$ & $\mathbf{0.1834}$ & $\mathbf{0.3107}$ \\
\bottomrule
\multirow{5.5}{*}{\rotatebox{90}{\textit{Hotel}}}
& UserPop & $0.0436$ & $0.0681$ & $0.1059$ &  & $0.1091$ & $0.1059$ & $0.1054$ &  & $0.1018$ & $0.1036$ & $0.1050$ &  & $0.0265$ & $0.0557$ & $0.1120$ \\
& ItemPop & $0.0483$ & $0.0756$ & $0.1152$ &  & $0.1159$ & $0.1137$ & $0.1124$ &  & $0.1108$ & $0.1131$ & $0.1111$ &  & $0.0303$ & $0.0633$ & $0.1237$ \\
& CE-VAE & $\underline{0.2425}$ & $\underline{0.3521}$ & $\underline{0.4984}$ &  & $\underline{0.9389}$ & $\underline{0.9113}$ & $\underline{0.8638}$ &  & $\underline{0.9209}$ & $\underline{0.8629}$ & $\underline{0.7831}$ &  & $0.1153$ & $\underline{0.2105}$ & $\underline{0.3704}$ \\
& CE-VNCF & $0.1873$ & $0.2527$ & $0.3280$ &  & $0.3975$ & $0.3671$ & $0.3230$ &  & $0.3732$ & $0.3154$ & $0.2558$ &  & $\mathbf{0.1252}$ & $0.2060$ & $0.3230$ \\
& M\&Ms-VAE (Ours) & $\mathbf{0.2500}$ & $\mathbf{0.3595}$ & $\mathbf{0.5054}$ &  & $\mathbf{0.9752}$ & $\mathbf{0.9393}$ & $\mathbf{0.8829}$ &  & $\mathbf{0.9498}$ & $\mathbf{0.8776}$ & $\mathbf{0.7895}$ &  & $\underline{0.1182}$ & $\mathbf{0.2131}$ & $\mathbf{0.3726}$
\end{tabular}
\end{threeparttable}
\end{table}

Table~\ref{table_exp_perf} contains the results. Both popularity baselines clearly underperform, showing that the task is not trivial (see Table~\ref{stats_datasets} for the number of keyphrases per dataset). Remarkably, the proposed M\&Ms-VAE model significantly outperforms the CE-VNCF baseline by a factor of 2.5 on average and by approximately by 3.5 on MAP and Precision. 
Moreover, M\&Ms-VAE retrieves 89\% percent of relevant keyphrases within the Top-20 explanations on the CDs\&Vinyl~dataset.

We observe that CE-VAE performs similarly to M\&Ms-VAE but still slightly underperforms by approximately 2\% on average. Finally, we note that CE-VNCF achieves the best results in terms of Recall for the Beer dataset and Recall@5 for the CDs\&Vinyl and Hotel datasets. Nevertheless, as seen in the recommendation performance, CE-VNCF significantly underperforms, highlighting the trade-off between recommendation and explanation.

\subsection{RQ 2: Can M\&Ms-VAE with the self-supervised critiquing objective enable multi-step critiquing?}\label{sec_rq2}

We now determine whether M\&Ms-VAE can enable multi-step critiquing and effectively re-rank recommended items.

\subsubsection{Baselines.}
\textbf{UAC}~\citep{luo2020b} denotes uniform average critiquing, in which the user embedding and all critique embeddings are averaged uniformly. \textbf{BAC}~\citep{luo2020b} is balanced average critiquing. It first averages the critique embeddings and then averages them again with the initial user embedding. \textbf{CE-VAE} and \textbf{CE-VNCF} were introduced in Section~\ref{sec_rq1_baselines}. During training, both learn an inverse feedback loop between a critique and the latent space. At inference, they average the original user embedding with the critique embedding. \textbf{LLC-Score}~\citep{luo2020b} and \textbf{LLC-Rank}~\citep{hanze2020} first extend the PLRec recommender system~\citep{sedhain2016practical} to co-embed keyphrases into the user embedding space with a linear regression. Afterwards, the models apply a weighted average between the initial user embedding and each critique embedding;~the weights are optimized in a linear programming formulation; LLC-Score uses a max-margin scoring-based objective and LLC-Rank a ranking-based objective. To limit computational complexity, the authors limit the number~of constraints to the top-100 rated items.
For a fair comparison, we also consider in M\&Ms-VAE the top-100 rated items~meeting the criteria for $I^{+\bc}$ and $I^{-\bc}$ for each critique $\bc$, although the computational time remains identical using the full sets.

\subsubsection{User Simulation.} Similarly to prior work~\citep{luo2020b,hanze2020}, we conduct a user simulation to asses each model's performance in a multi-step conversational recommendation scenario. Concretely, the simulation considers all users and follows Algorithm~\ref{alg_dataset} with the following differences: \begin{enumerate}[topsep=0pt]
 \item we track the conversational interaction session of simulated users by selecting all target items from their \textbf{test} set,
 \item the maximum allowed critiquing iterations is set to 10, and, 
 \item the conversation stops if the target item appears within the top-N recommendations on that iteration.
\end{enumerate}

As in~\cite{hanze2020}, we simulate a variety of user-critiquing styles. For each, the candidate keyphrases to critique are inconsistent (i.e., disjoint) with the target item's known keyphrase list. We experiment with the following three~methodologies: \begin{enumerate}
	\item \textbf{Random}: we assume the user randomly chooses a keyphrase to critique.
	\item \textbf{Pop}: we assume the user selects a keyphrase to critique according to the general keyphrase popularity. 
	\item \textbf{Diff}: we assume the user critiques a keyphrase that deviates the most from the known target item description. To do so, we compare the top recommended items' keyphrase frequency to the target item's keyphrases and select the keyphrase with the largest frequency differential.
\end{enumerate}

\subsubsection{Multi-Step Critiquing Performance.} Following~\cite{luo2020b,hanze2020}, we asses the models over all users and all items on~the test set using the average success rate and session length as metrics. The former is the percentage of target items that successfully reach a rank within the Top-N, and the latter is the average length of these sessions (with a limit of 10 iterations). 
In our~experiment framework, for each user and target, we sample alongside 299 unseen items. We believe that this setting reflects realistic scenarios, where users navigate through hundreds of items instead~thousands. 

\begin{figure}[!t]
\centering
\hspace*{-0.75cm}
\includegraphics[width=1.1\textwidth]{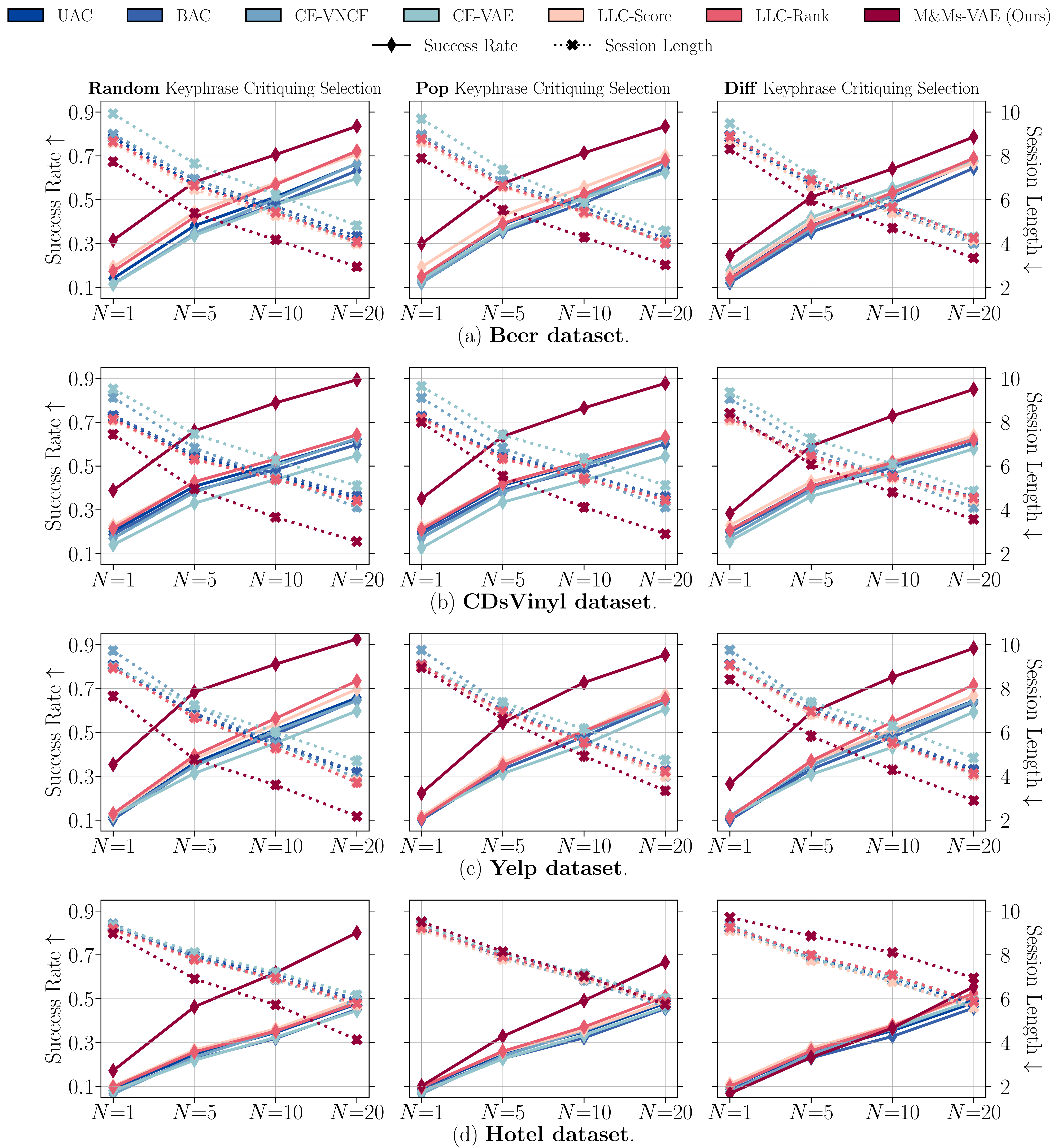}
\caption{\label{fig_multi_crit}Multi-step critiquing performance after 10 turns of conversation. For each dataset and keyphrase critiquing selection method, we report the average success rate (left y-axis) and session length (right y-axis) at different Top-N with 95\% confidence interval.}
\end{figure}

The results for each dataset and keyphrase critiquing selection method are depicted in~Figure~\ref{fig_multi_crit}.
Overall, all models' performance is generally better on the Beer and CDs\&Vinyl~datasets due to the higher density in terms of~the number of interactions.
Generally, all models tend to find the target item within the Top-20 in more than half the time and under six turns. This highlights that in practice, users are likely to find the desired item in a limited amount of~time.

Impressively, M\&Ms-VAE clearly outperforms all the baselines on both metrics on the Beer, CDs\&Vinyl, and Yelp datasets. On the Hotel dataset, the success rate is significantly higher for the Random and Pop cases and similar for the Diff case, whereas the session length is higher for the Pop and Diff selection methods. This is unsurprising, because the blending module is trained only once on the random keyphrase critiquing~selection (i.e., no assumption on the user's behavior).

Although the simple self-supervision objective in M\&Ms-VAE mimics only one-step random critiquing, we remark that the training strategy generalizes for multi-step critiquing and other keyphrase critiquing selection as well. This shows that M\&Ms-VAE efficiently embeds the critique, thanks to the multimodal modeling.

We observe that the simple UAC and BAC methods perform similarly or better than CE-VNCF and CE-VAE. However, they are outperformed by LLC-Score, LLC-Rank, and M\&Ms-VAE. These results confirm our observation~in Section~\ref{sec_cevae} that the critiquing objective introduces noise during training and does not accurately reflect the critiquing~mechanism.

Finally, we note that LLC-Score performs similarly to LLC-Rank in most cases.  When compared to M\&Ms-VAE, both significantly underperform on both metrics in 10 out of 12 cases. This highlights the effectiveness of our proposed critiquing algorithm compared to linear aggregation methods.


\begin{figure}[!t]
\centering
\hspace*{-0.75cm}
\includegraphics[width=1.1\textwidth,height=5.834in]{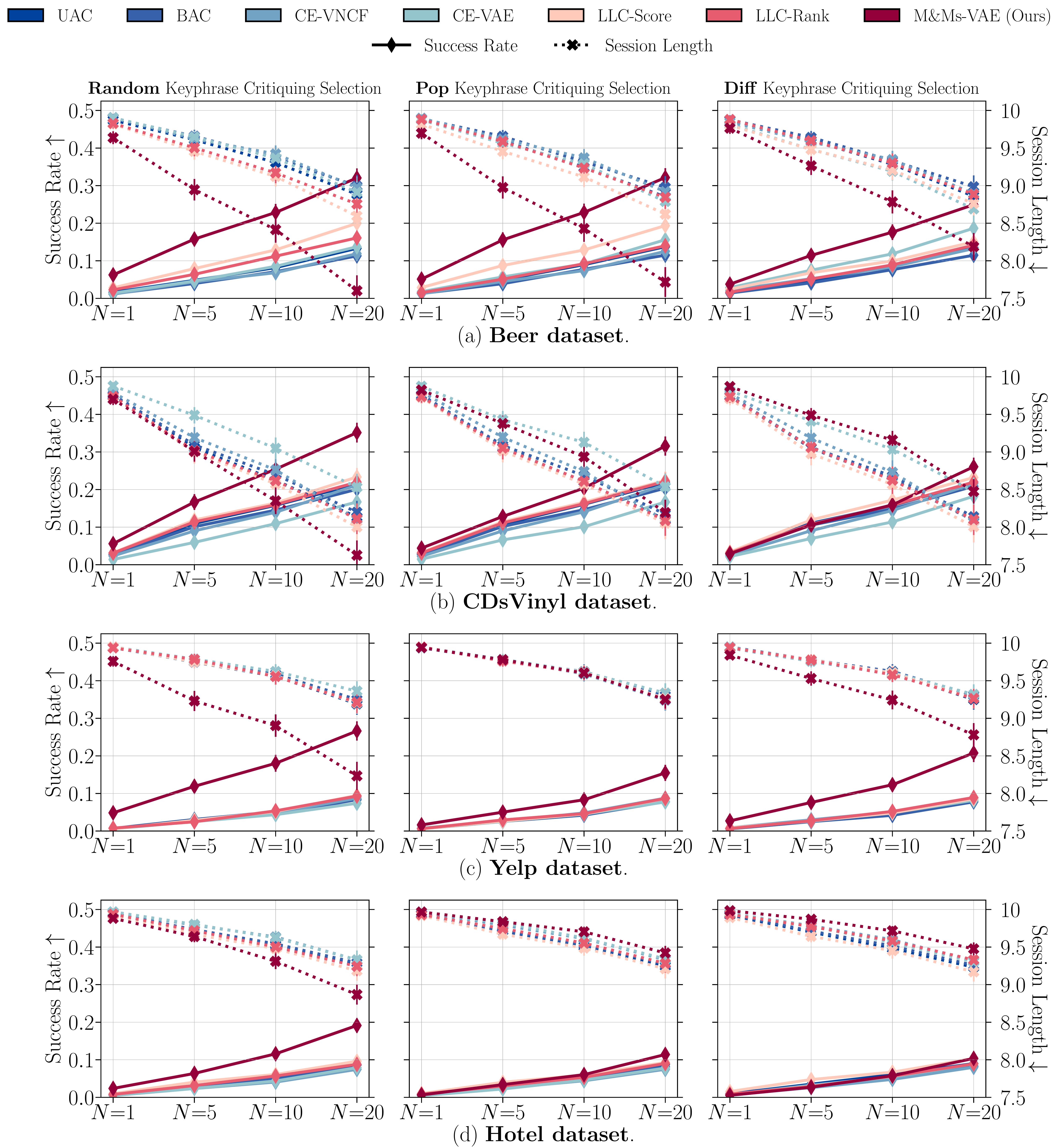}
\caption{\label{fig_multi_crit2}Multi-step critiquing performance after 10 turns of conversation \textbf{on all items}. For each dataset and keyphrase critiquing selection method, we report the average success rate (left y-axis) and session length (right y-axis) at different Top-N with 95\% confidence interval.}
\end{figure}

Finally, we replicate the experiment by using all the available items (see Table~\ref{stats_datasets} for the sizes). When the evaluation is conducted on 300 items (see Figure~\ref{fig_multi_crit}), we see that users indeed find a specific item using our technique with a high success rate (i.e., an average success rate of 80\%). However, in Figure~\ref{fig_multi_crit2} where thousands of items are available,~the results show that current methods are not yet good enough to achieve similar results for such a large number. Nevertheless, M\&Ms-VAE clearly outperforms other methods and still achieves an average success rate~of~20\%. 

\subsection{RQ 3: What is the critiquing computational time complexity for M\&Ms-VAE compared to prior~work?}\label{sec_rq3}
Now, we aim to empirically determine how the critiquing in M\&Ms-VAE compares to the best baselines in Section~\ref{sec_rq2}, LLC-Score and LLC-Rank, in terms of computational time. Because the baselines can not leverage the GPU due to their optimization framework, we also run M\&Ms-VAE on the CPU. We follow the same experiment settings as in Section~\ref{sec_rq2} and limit ourselves to 1,000 users and the Random keyphrase critiquing selection method. All models process exactly 10 critiques for each user-item pair. We employ a machine with a 2.5GHz 24-core CPU, Titan X GPU, and 256GB~memory (see Appendix~\ref{app:add_repr} for the detailed hardware).

\begin{figure}[!t]
\centering
\begin{subfigure}[t]{0.49\textwidth}
    \centering
    \includegraphics[width=0.925\textwidth,height=2.0396in]{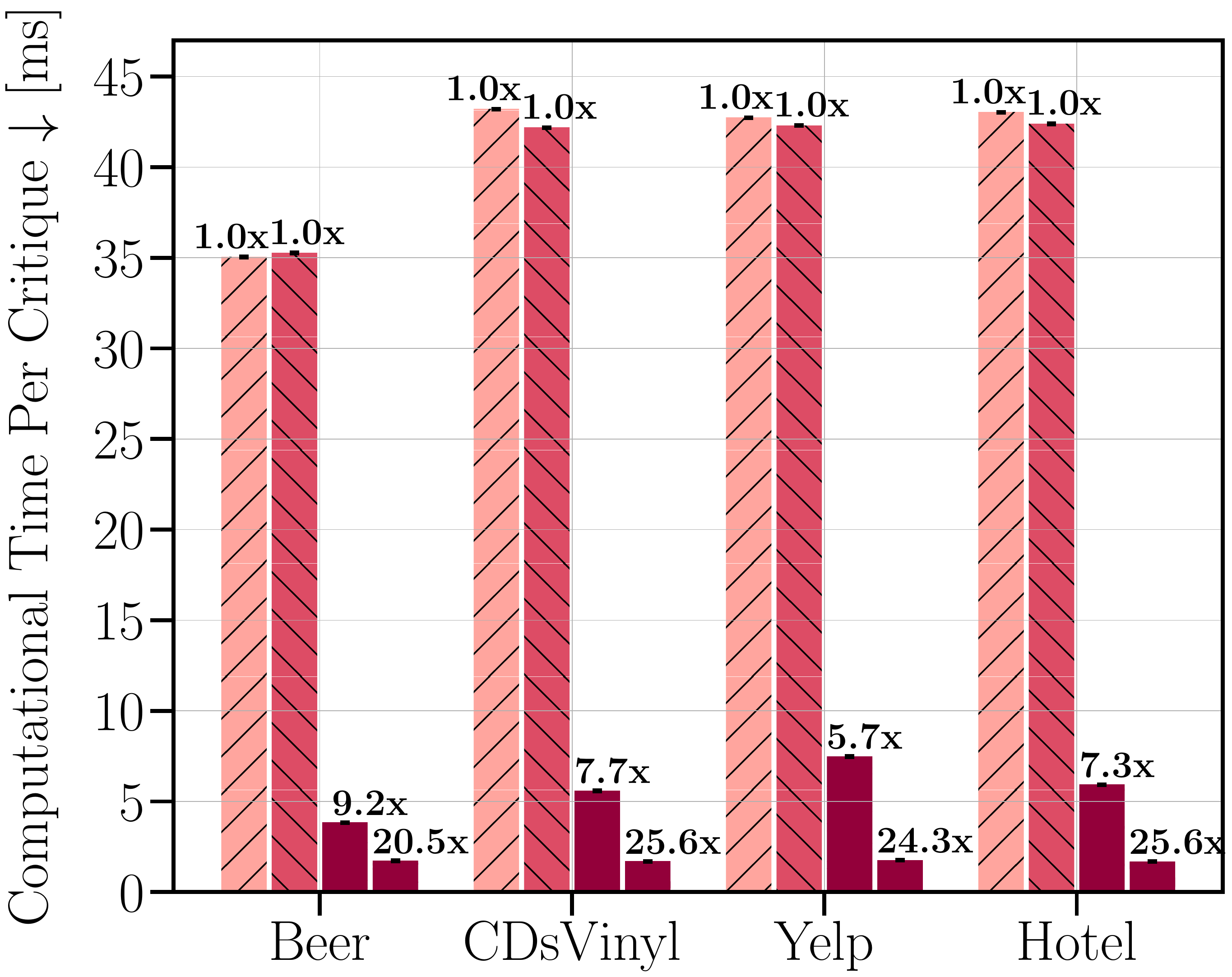}
    \caption{\label{fig_time_analysis_crit}Average computational time of a single critique.}
\end{subfigure}
~
\begin{subfigure}[t]{0.49\textwidth}
    \centering
    \includegraphics[width=0.95\textwidth,height=1.985in]{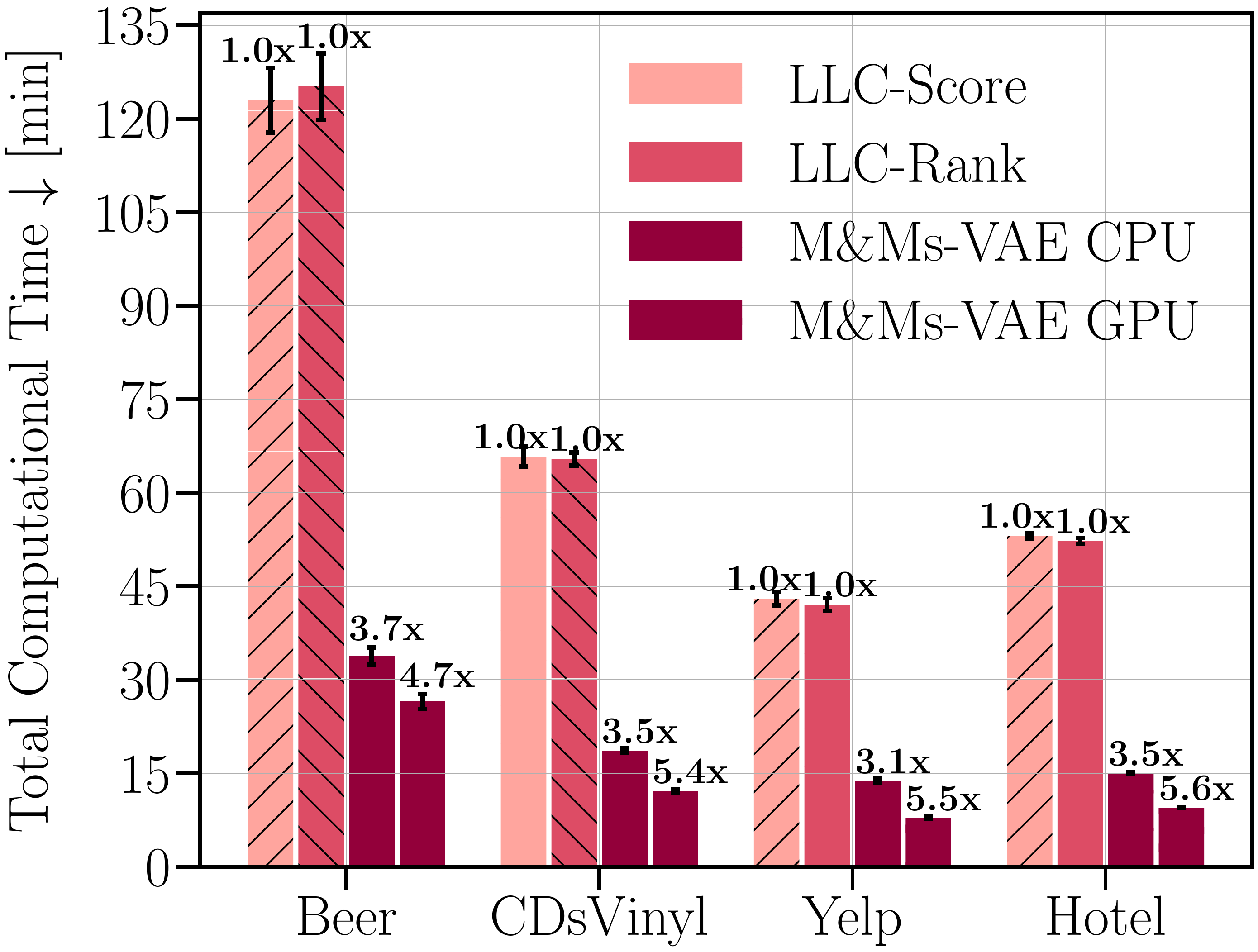}
    \caption{\label{fig_time_analysis_total}Average total computational time of the simulation.}
\end{subfigure}
\caption{\label{fig_time_analysis}Average time consumed for completing 10 runs for 1,000-user simulation after ten turns of conversation with 95\% confidence intervals. LLC-Score and LLC-Rank cannot leverage GPUs; we thus report the performance of M\&Ms-VAE on CPU and GPU. Additionally, we report the inference speedup compared to the slowest model.}
\end{figure}

Figure~\ref{fig_time_analysis} shows the average computational time over 10 runs. Particularly, the Figure~\ref{fig_time_analysis_crit} denotes the critiquing computational time in milliseconds, and Figure~\ref{fig_time_analysis_total} the overall simulation time in minutes. Impressively, we observe that the critiquing in M\&Ms-VAE's  is approximately 7.5x faster on CPU and up to 25.6x on the GPU than LLC-Score and LLC-Rank. This shows that once the critiquing module of M\&Ms-VAE is trained, which takes less than five minutes on the machine, the model achieves a lower latency (batch size of one). In Figure~\ref{fig_time_analysis_total}, the overall simulation in M\&Ms-VAE is at least 3.1x faster on CPU and approximately 5.3x faster on GPU. Finally, in real-life applications, we could leverage multiple users' critiques simultaneously and increase the throughput by considering a larger batch size.

\subsection{RQ4: How does M\&Ms-VAE perform under weak supervision; is the joint \& cross inference~coherent?}
\label{seq_rq4}
We first quantify the coherence of joint and cross generations of our M\&Ms-VAE model. We denote~three cases at test time: \begin{enumerate}[noitemsep,topsep=0pt]
 \item only the user's interactions are used, and the encoder is $q_{\Phi_r}(\bzu | \bru)$;
 \item only the user's keyphrase~preference is used, and the encoder is $q_{\Phi_k}(\bzu | \bku)$; and 
 \item both are used with the encoder $q_\Phi(\bzu | \bru, \bku)$.
 \end{enumerate}
 
 Second, we simulate~incomplete supervision by randomly selecting a subset of the training with fully observed samples. The other one~is split into two even parts: the first includes only observed $\bru$ and the second $\bku$. We retain the models and settings of~Section~\ref{sec_rq1}.

\begin{figure}[!t]
\centering
\hspace*{-1.4cm}
\includegraphics[width=1.2\textwidth]{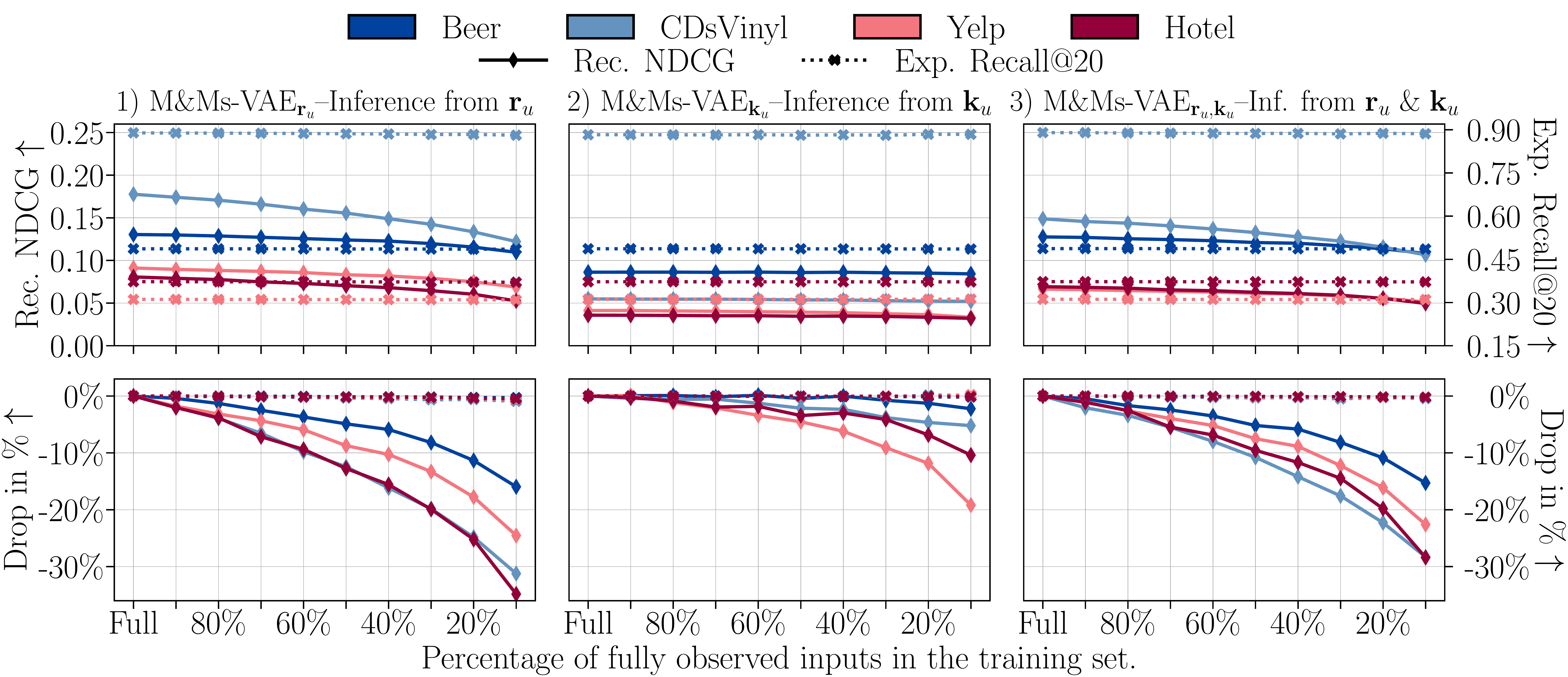}
\caption{\label{fig_rec_exp_mm_perf}Recommendation (left y-axis) and explanation (right y-axis) results averaged on five runs with different combinations of~modalities observed at inference. The x-axis denotes the percentage of fully observed inputs during training; the rest is partially observed.}
\end{figure}

Figure~\ref{fig_rec_exp_mm_perf} shows the results averaged on five runs for the three cases at different levels of supervision. The top row presents the explanation and recommendation performance in terms of NDCG and Recall@20 (the model behavior is consistent across the other metrics). On the full datasets, we note that the explanation performance is comparable across the three variants and higher than those of the baselines in Table~\ref{table_exp_perf}. 

Regarding the recommendation performance, M\&Ms-VAE\textsubscript{$\bru$} obtains the best results, followed closely by M\&Ms-VAE\textsubscript{$\bru,\bku$}, and M\&Ms-VAE\textsubscript{$\bku$} clearly underperforms. This aligns with the observations in~\cite{10.1145/3397271.3401281,10.1145/3178876.3186070}: recommender systems are limited if they use only review text as input, and not all reviews can be useful. Nevertheless, compared to  Table~\ref{table_rec_perf}, M\&Ms-VAE\textsubscript{$\bku$} always achieves better recommendation performance than the popularity baseline, and it performs better than AutoRec and CDAE on two~datasets.
\enlargethispage{\baselineskip}
The bottom row of Figure~\ref{fig_rec_exp_mm_perf} show the relative drop in performance on both metrics. The explanation performance seems unaffected by the sparsity, showing that the explanation task remains simple in comparison with the recommendation task. Remarkably, with only 50\% fully observed inputs and the rest partially observed, the recommendation performance of M\&Ms-VAE\textsubscript{$\bru$} and M\&Ms-VAE\textsubscript{$\bru,\bku$} is decreased by only 9\% on average. More so, with 90\% partial observations, the model can still achieve more than 70\% of its performance quality on the full datasets. Finally, these results emphasize that M\&Ms-VAE can effectively learn the joint distribution even in a weakly supervised~setting. 

\section{Conclusion}
Recommendations can have much more impact if they are supported by explanations that can be critiqued.
Previous research has developed methods that either perform poorly in multi-step critiquing or suffer from computational inefficiency at inference.
In this chapter, we presented M\&Ms-VAE, a novel variational autoencoder for recommendation~and explanation that treats the user preference and keyphrase usage as different observed variables. Additionally, we proposed a strategy that mimics weakly supervised learning and trains the inference networks jointly and~independently.

Our second contribution is a new critiquing module that leverages the generalizability of M\&Ms-VAE to embed the user preference and the critique. With a self-supervised objective and a synthetic dataset, we enable multi-step critiquing in M\&Ms-VAE.
Experiments on four datasets show that M\&Ms-VAE \begin{enumerate}
 	\item is the first model to obtain  substantially better recommendation, explanation, and multi-critiquing performance,
 	\item processes critiques up to 25.6x faster than previous state-of-the-art methods, and
 	\item produces coherent joint and cross generation, even under weak~supervision.
 \end{enumerate}

%% file: main/conclusion.tex
\chapter{Conclusion}

This dissertation's initial motivation comes from the observation that machine learning algorithms pervade our lives. Because more data are available than ever before, automated predictions accompanied by explanations can carry much more impact, can increase transparency, and may even be necessary. Nevertheless, it remains challenging to interact with such predictions when they are unsatisfying or wrong. Consequently, end users might feel frustrated with or, worse, experience distrust for or reject the model predictions. Although many interpretable techniques \citep[e.g.,][]{ribeiro2016should,serrano-smith-2019-attention,li2016understanding} and inherently interpretable models \citep[e.g.,][]{NEURIPS2018_3e9f0fc9,koh2020concept,lei-etal-2016-rationalizing} exist, the explanations generated are mostly informative and provided ``as it is''. However, it has been shown that users prefer the opportunity to provide feedback (i.e., tend to critique the model's outputs or its explanations) that will improve the model \citep{Amershi_Cakmak_Knox_Kulesza_2014,wall2019using,noexplainability2020}. This preference highlights the growing need for users to better understand the underlying computational processes and control their actions better.

Part~\ref{p1} of this dissertation examines the problem of \textbf{explanation generation}: how can we infer high-quality explanations from text documents? We base our solution on the paradigm of selective rationalization, that is, finding relevant subsets of the input text (called a rationale) that suffice on their own to yield the same prediction. Our key motivation arises from the limitations of rationales. Current rationalization models generate only one rationale: they strive for one hard (binary) overall selection to explain one outcome. This limitation imposes multiple constraints, such as assuming some priors in the data distribution, which do not reflect realistic settings, and makes it difficult to capture ambiguities in the text. We claim that useful rationales can be multi-faceted, thus involving support for different outcomes. For example, clinicians justify medical diagnoses and treatments of patients, based on multiple factors: symptoms, prior events, and preconditions. In the case of user reviews, users justify their overall feedback with multiple explanations - one for each aspect they deem important.

In Chapter~\ref{chapter_aaai2021}, we propose the notion of multi-dimensional rationales and a multi-task learning approach in which rationales are learned in an unsupervised manner. To the best of our knowledge, our study is the first to neither assume any prior on the data distribution nor train, tune, or maintain multiple models, one for each label to explain (e.g., one label per subrating). Therefore, our model can be deployed in realistic settings and learn soft multi-dimensional rationales in one training. We also empirically demonstrate that interpretability does not come at the cost of reduced performance and that the resulting explanations are more accurate and coherent than those generated by other methods. 

In Chapter~\ref{chapter_acl_2021}, we extend the multi-dimensional idea and generalize our model by assuming that only one label is observed (e.g., only the overall rating).  Inspired by the role of concept-based thinking in human reasoning \citep{ARMSTRONG1983263,tenenbaum1999bayesian}, we identify latent concepts and explain the outcome with a linear aggregation of concepts. Our study makes pioneering inroads into the extraction of multi-dimensional explanations and latent concepts for selective rationalization. It extends the application to cases in which multiple labels are not available at a large scale (e.g., on online platforms, subratings are optional whereas the overall rating is always mandatory). We empirically show that our model generates concepts that align with human rationalization while using only the overall label. Further, it outperforms state-of-the-art methods trained on each aspect label independently. Finally, both of the approaches we propose are trivially extendable for images as input data.

Part~\ref{p2} of this dissertation focuses on extending modern neural recommendation systems with explanations and interactive \textbf{critiquing}, which results in conversational explainable recommendation systems. Interactions are naturally present in recommendation systems, that constitute an impactful real-life application with hundreds of millions of predictions made daily.

Studies of explainable recommendation systems have shown that users prefer justifications alongside recommended items that are personalized and written in natural language (i.e., that could have been written by other humans) \citep{kunkel2019let,chang2016crowd,wilkinson2021}. However, there are no ground-truth datasets specifying what ``good'' justifications are. Prior studies have employed users' reviews to capture their preferences and writing styles \citep[e.g.,][]{ni2018personalized,li2019towards,dong2017learning}. From past reviews, they generate synthetic ones that serve as personalized explanations of ratings given by users. However, many reviews are noisy because they partly describe experiences or endorsements. \cite{ni-etal-2019-justifying} proposed a nonscalable pipeline for identifying justifications from reviews and asked humans to annotate them. \cite{chen2019co,chen2020towards} positioned the justification as the first sentence. However, the concepts of justification offered by these studies were ambiguous, and they assumed that a review contains only one justification. It is thus nontrivial to identify meaningful justifications within reviews.

Based on the models developed in Part~\ref{p1}, in Chapter~\ref{chapter_ijcai_2021}, we address the explanation generation challenge by introducing a scalable and data-driven justification extraction system, with no prior limits imposed on their number or structure. This approach relies on the faithfulness property of rationales, which is naturally compatible with recommendation: the data include user ratings and reviews. Our study is the first to introduce personalized explanations based on rationales and to show that they are significantly preferred by human users for recommendation in multiple domains over those produced by extant state-of-the-art techniques.

After nearly a decade in which critiquing approaches received little attention,~\cite{keyphraseExtractionDeep,luo2020,luo2020b,hanze2020} introduced a collaborative filtering recommender with keyword-based explanations and an embedding-based critiquing method. These methods allow users to critique the recommendation using arbitrary languages; a set of attributes is mined from reviews, and the users can interact~with them. However, isolated keywords are not the best type of explanation for users, and it is not trivial to co-embed item critiques with the latent user preferences. Furthermore, \cite{luo2020,luo2020b,hanze2020} showed that previous models suffer from computational inefficiency and an inherent trade-off between recommendation, explanation, and critiquing performance.

In the second part of Chapter~\ref{chapter_ijcai_2021}, we propose a transformer-based model \citep{vaswani2017attention} that generates abstractive natural language justifications as well as keyword-based explanations that accompany the recommendation. A crucial question remains: how can we critique the suggested items with their explanations made by the recommender system?  We incorporate critiquing via a novel unsupervised gradient-based technique that allows users to interact with the justifications. Keyword-based explanations are the key to connecting the justifications and the model's inner workings. We demonstrate empirically that our system successfully adapts to the preferences expressed in multi-step critiquing and generates consistent explanations without decreasing recommendation performance. Finally, to investigate the effects of different recommendation interfaces on users, we propose an online model-agnostic platform to foster research on recommendation with explanations and critiquing. Our interfaces allow to test such systems on a real use case and highlight some limitations of these approaches.

Finally, Chapter~\ref{chapter_recsys_2021} presents a novel variational autoencoder~\citep{kingmamw2014} for recommendation based on multimodal modeling assumptions and self-supervised training. It treats the user preference and keyphrase usage as different observed variables. Our critiquing module, which leverages the generalizability of our model to embed the user preference and the critique, is an important innovation. With a self-supervised ranking objective and an automatically generated synthetic dataset, we enable fast and efficient multi-step critiquing. Using four real-world, publicly available datasets , we show that compared to extant state-of-the-art models, our model not only exceeds or matches their performance in terms of recommendation, explanation, and multi-step critiquing, but also processes the critiques up to 25.6x faster. This may accelerate the adoption and deployment of conversational explainable recommendation systems into a variety of online platforms.

\section{Future Directions}

Our study points to several potential directions for future research. We outline the most promising ones below.

\begin{enumerate}
	\item \textbf{Decreasing neural hallucinations.}\hspace{\parindent} Due to the recent progress in natural language generation, models can now generate syntactically and grammatically correct sentences \citep{radford2019language,NEURIPS2020_1457c0d6}. As is often the case in neural text generation, however, the justifications generated may suffer from factual inconsistency \citep{rohrbach-etal-2018-object,holtzman2019curious}. For example, our system might recommend a hotel with a justification mentioning ``Lugano'' even though the hotel is located in Madrid. Although such justifications might be factually correct in some cases, they usually are not. A potential solution for this problem is to enrich the neural decoder with commonsense knowledge to verify the consistency of generated facts.
	\item \textbf{Treating critiques as harder constraints.}\hspace{\parindent} In Chapters~\ref{chapter_ijcai_2021} and \ref{chapter_recsys_2021}, we co-embed the language-based feedback (i.e., critique) and user preferences in the same space. Then, we combine each representation through a blending function to infer the final latent representation. This blending function plays a key role in satisfying the constraints based on the current and previous critiques. However, we have little control over it: there is no guarantee that the function will produce a latent representation that fully integrates and satisfies the critique. A promising direction is neural-symbolic reasoning over knowledge graphs. User behaviors and preferences, item attributes, and critiques can be represented as sets of symbolic rules derived from the knowledge graph. Nevertheless, the knowledge graph is domain dependent, which limits its generalization unless it is built automatically.
	\item \textbf{Choosing better keywords to critique and allowing categorical critiquing.}\hspace{\parindent} Another limitation of our system is the frequency-based approach with some postprocessing to mining keywords. Although our method is automatic and allows for a richer interaction space than does a set of fixed item attributes, meaningful keywords for the recommendation do not necessarily imply that the keywords are helpful for end users. We believe improvements can be made along two axes: the keywords' interactions and semantic representation. Regarding keywords' interactions, it is unclear, for example, what it means to critique the keyword ``breakfast'' positively or negatively. Currently, we assume only a binary constraint: the presence or absence of breakfast service in an establishment. It would be more useful to treat the constraint as a categorical variable~\citep{10.1145/3404835.3462893,louis-etal-2020-id}. Each possible value would also be mined from the context in which keywords appear (e.g., buffet or continental for the keyword  ``breakfast''). This could be achieved through aspect- and opinion-term-extraction techniques, and the critique encoding would be extended from a binary representation to an ordinal one. Regarding semantic representation, we could extract keywords based on their surface forms and their semantic representation \citep[e.g.,][]{bennani-smires-etal-2018-simple}. We could identify clusters with their representative keywords while controlling diversity. Thus, the keywords' latent representations would embed an actual meaning. As a result, the model would better understand the meaning of a critique, because the keywords and critiques would interact with the same latent space.
		
	\item \textbf{Relaxing the constraints of latent concepts.}\hspace{\parindent} In Chapter~\ref{chapter_acl_2021}, we generalize the approached described in Chapter~\ref{chapter_aaai2021} by observing only one variable. However, this generalization comes at the cost of two constraints specified beforehand: the number of concepts and their length. Early experiments without these constraints produced significantly worse results. We suspect that the predictor overfits the rationales found by the generator. Thus, both might be trapped in a suboptimal equilibrium. This phenomenon -- called model interlocking -- was recently confirmed in \cite{yu2021}. One means of alleviating this problem would be to pretrain the generator to extract more diverse and semantically meaningful text snippets via self-contrastive learning-based methods or topic-modeling-based approaches. In this way, the joint optimization might start with a better initialization and more successfully identify the text snippets that constitute potential rationales (i.e., meaningful for the prediction).
\end{enumerate}

The research directions proposed above can be combined to make even more significant improvements in the quality of justifications, recommendations, and user experience.

%% file: appendix.tex
\appendix

\chapter{Complementary Results}
\label{app_rq2}

\section{Inferred Concepts by ConRAT}
\label{app_rq2_conrad}
The following results complements Section~\ref{seq_rq2_concepts}.

\subsection{Objective Evaluation}
\label{app_rq2_obj}

The results for the concept length $\ell=5$ is shown in Table~\ref{exp_rq2_beer_objective_5}. 

\begin{table}[!ht]
    \centering
   \caption{\label{exp_rq2_beer_objective_5}Objective performance of rationales for the multi-aspect beer reviews with the concept length set to five. ConRAT only uses the overall rating and does not have access to the other aspect labels. All baselines are trained separately on each aspect label. \textbf{Bold} and \underline{underline} denote the best and second-best results, respectively.}
\hspace*{-1.5cm}
\begin{threeparttable}
\begin{tabular}{@{}c@{\hspace{1.5mm}}l@{\hspace{2.0mm}}
c@{\hspace{2.0mm}}
c@{\hspace{1mm}}c@{\hspace{1mm}}c@{}c@{\hspace{2.0mm}}
c@{\hspace{1mm}}c@{\hspace{1mm}}c@{}c@{\hspace{2.0mm}}
c@{\hspace{1mm}}c@{\hspace{1mm}}c@{}c@{\hspace{2.0mm}}
c@{\hspace{1mm}}c@{\hspace{1mm}}c@{}c@{\hspace{2.0mm}}
c@{\hspace{1mm}}c@{\hspace{1mm}}c@{}c@{\hspace{2.0mm}}
c@{\hspace{1mm}}c@{\hspace{1mm}}c@{}c
@{}}
& & & \multicolumn{3}{c}{\textit{Average}} & & \multicolumn{3}{c}{\textit{Appearance}} & & \multicolumn{3}{c}{\textit{Aroma}} & & \multicolumn{3}{c}{\textit{Palate}}& & \multicolumn{3}{c}{\textit{Taste}}& & \multicolumn{3}{c}{\textit{Overall}}\\
\cmidrule{4-6}\cmidrule{8-10}\cmidrule{12-14}\cmidrule{16-18}\cmidrule{20-22}\cmidrule{24-26}
& \textbf{Model} & \textbf{Acc.} & \textbf{P} & \textbf{R} &  \textbf{F} & & \textbf{P} & \textbf{R} & \textbf{F}  & & \textbf{P} & \textbf{R} & \textbf{F} & &  \textbf{P} & \textbf{R} & \textbf{F} & &  \textbf{P} & \textbf{R} & \textbf{F} & &  \textbf{P} & \textbf{R} & \textbf{F}\\
\toprule
\multirow{5}{*}{\rotatebox{90}{\textit{$\ell=5$}}}
& RNP & $80.8$ & $41.3$ & $10.4$ & $16.4$ & & $\underline{50.9}$ & $\mathbf{13.3}$ & $\mathbf{21.1}$ & & $\mathbf{43.2}$ & $\mathbf{12.7}$ & $\mathbf{19.7}$ & & $\underline{27.1}$ & $\underline{10.0}$ & $\underline{14.5}$ & & $5.5$ & $0.59$ & $1.07$ & & $\underline{80.0}$ & $\underline{15.3}$ & $\underline{25.7}$\\
& RNP-3P & $81.5$ & $32.9$ & $6.9$ & $11.2$ & & $35.1$ & $7.3$ & $12.1$ & & $25.6$ & $7.2$ & $11.3$ & & $17.0$ & $5.2$ & $8.0$ & & $28.6$ & $4.0$ & $7.1$ & & $58.2$ & $10.5$ & $17.8$\\
& Intro-3P & $\underline{84.6}$ & $29.8$ & $7.0$ & $11.3$ & & $47.3$ & $12.4$ & $19.7$ & & $35.4$ & $9.9$ & $15.5$ & & $9.7$ & $2.8$ & $4.3$ & & $24.3$ & $3.8$ & $6.6$ & & $32.4$ & $6.3$ & $10.6$\\
& InvRAT & $83.6$ & $\underline{46.4}$ & $\mathbf{11.4}$ & $\mathbf{18.1}$ & & $\mathbf{51.0}$ & $\underline{13.1}$ & $\underline{20.8}$ & & $\underline{40.6}$ & $\underline{11.9}$ & $\underline{18.4}$ & & $\mathbf{32.0}$ & $\mathbf{11.8}$ & $\mathbf{17.2}$ & & $\underline{36.1}$ & $\underline{5.6}$ & $\underline{9.6}$ & & $72.5$ & $14.7$ & $24.4$\\
& ConRAT\tnote{\textdagger} & $\mathbf{90.4}$ &$\mathbf{46.6}$ & $\underline{10.9}$ & $\underline{17.5}$ & & $47.2$ & $12.4$ & $19.6$ & & $26.9$ & $7.1$ & $11.3$ & & $26.6$ & $9.2$ & $13.7$ & & $\mathbf{39.2}$ & $\mathbf{6.2}$ & $\mathbf{10.8}$ & & $\mathbf{93.1}$ & $\mathbf{19.5}$ & $\mathbf{32.21}$\\
\end{tabular}
\begin{tablenotes}
     \item[*] The model is only trained on the overall label and does not have access to the other ground-truth labels.
  \end{tablenotes}
\end{threeparttable}
\end{table}

\subsection{Subjective Evaluation}
\label{app_rq2_sub}

\begin{figure}[!t]
\centering
\begin{subfigure}[t]{\textwidth}
\centering
\includegraphics[width=\textwidth]{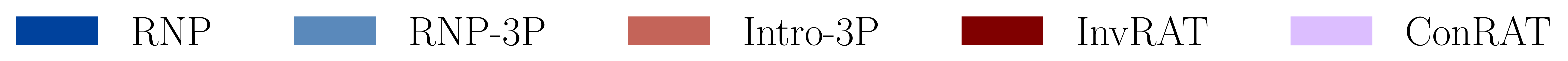}
\end{subfigure}

\begin{subfigure}[t]{0.33\textwidth}
\centering
\includegraphics[width=\textwidth,height=4.65cm]{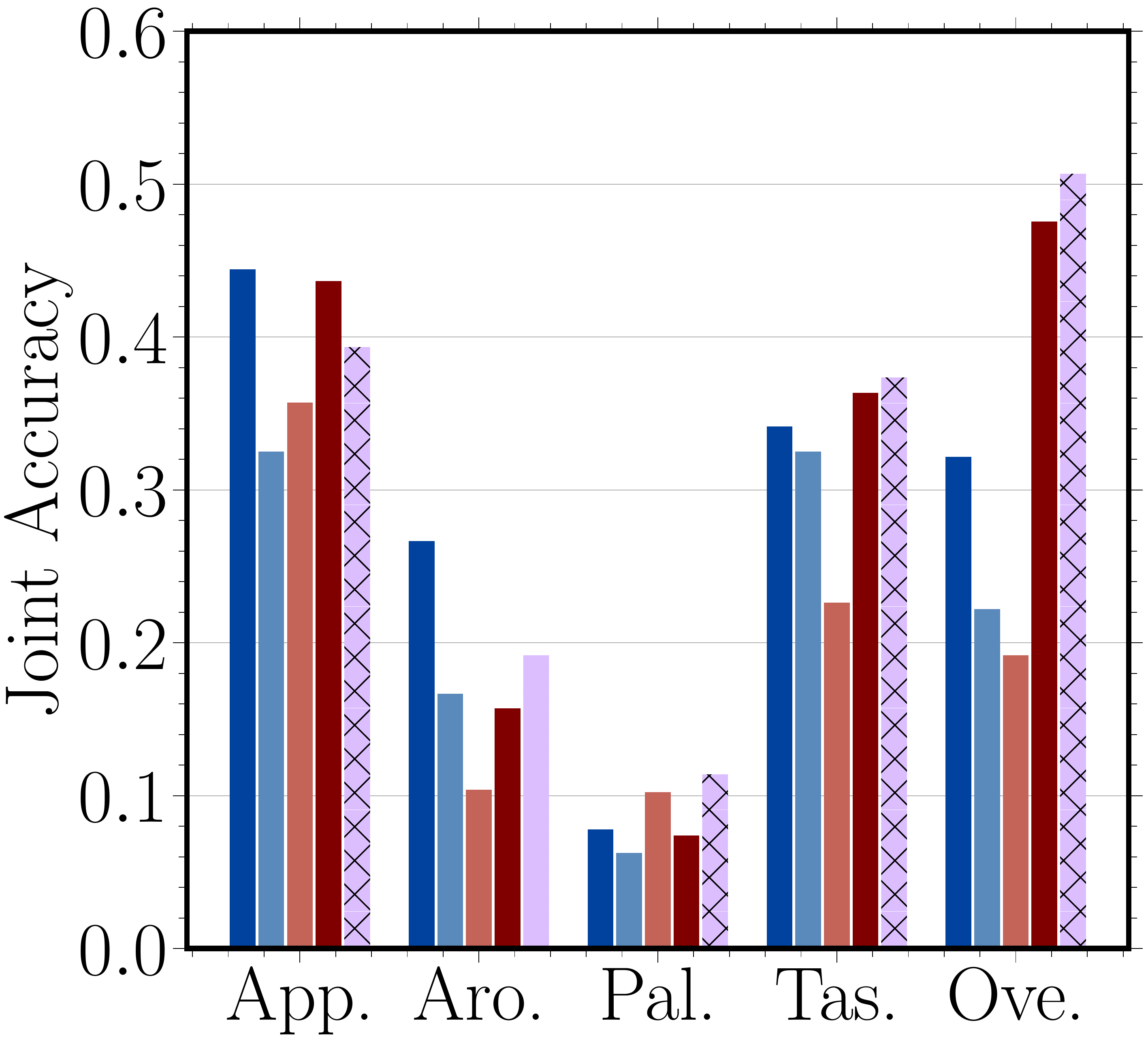}
\end{subfigure}
\begin{subfigure}[t]{0.33\textwidth}
\centering
\includegraphics[width=\textwidth,height=4.65cm]{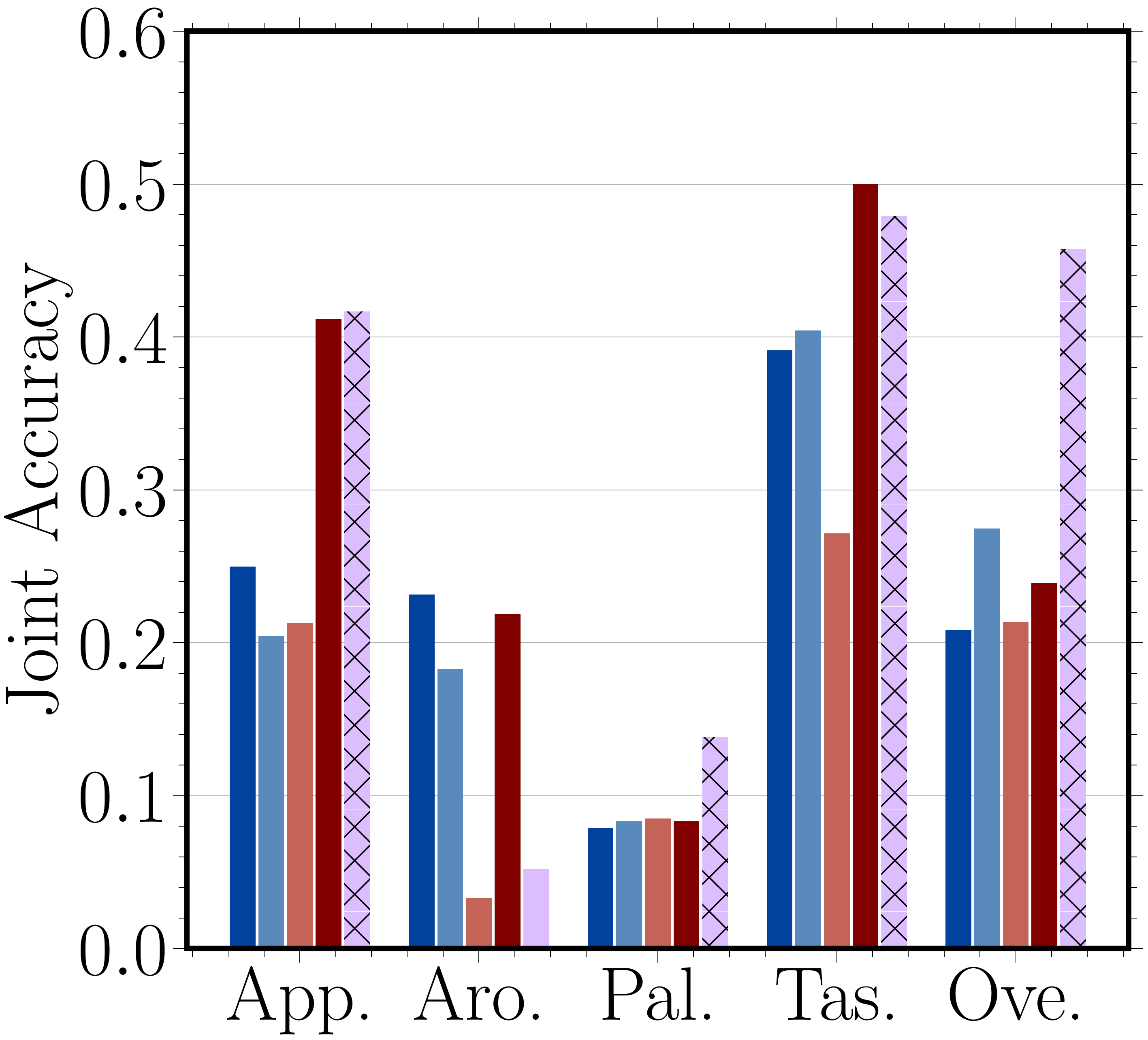}
\end{subfigure}
\begin{subfigure}[t]{0.33\textwidth}
\centering
\includegraphics[width=\textwidth,height=4.65cm]{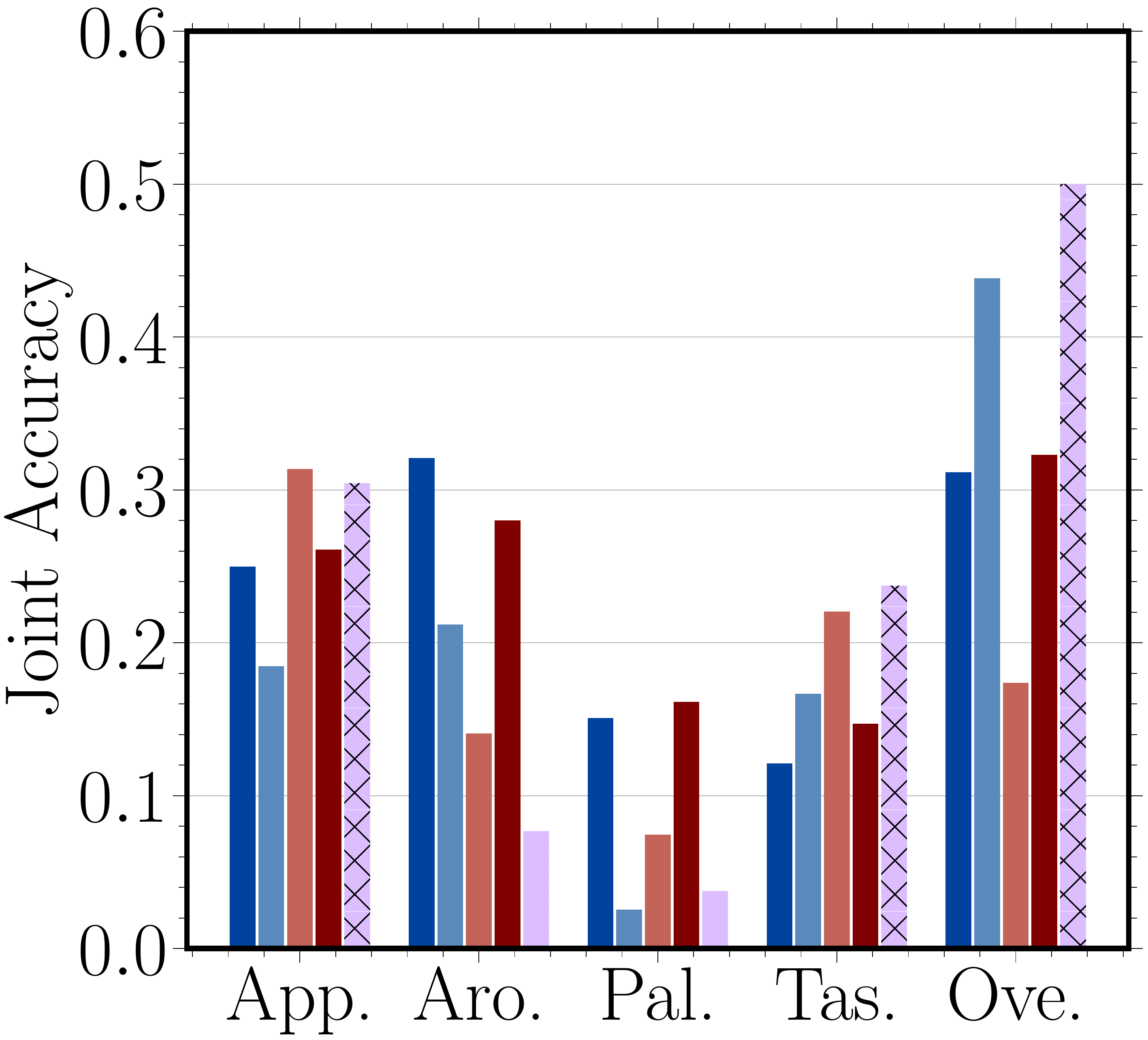}
\end{subfigure}
\par\bigskip
\par\bigskip

\begin{subfigure}[t]{0.33\textwidth}
\centering
\includegraphics[width=\textwidth,height=4.65cm]{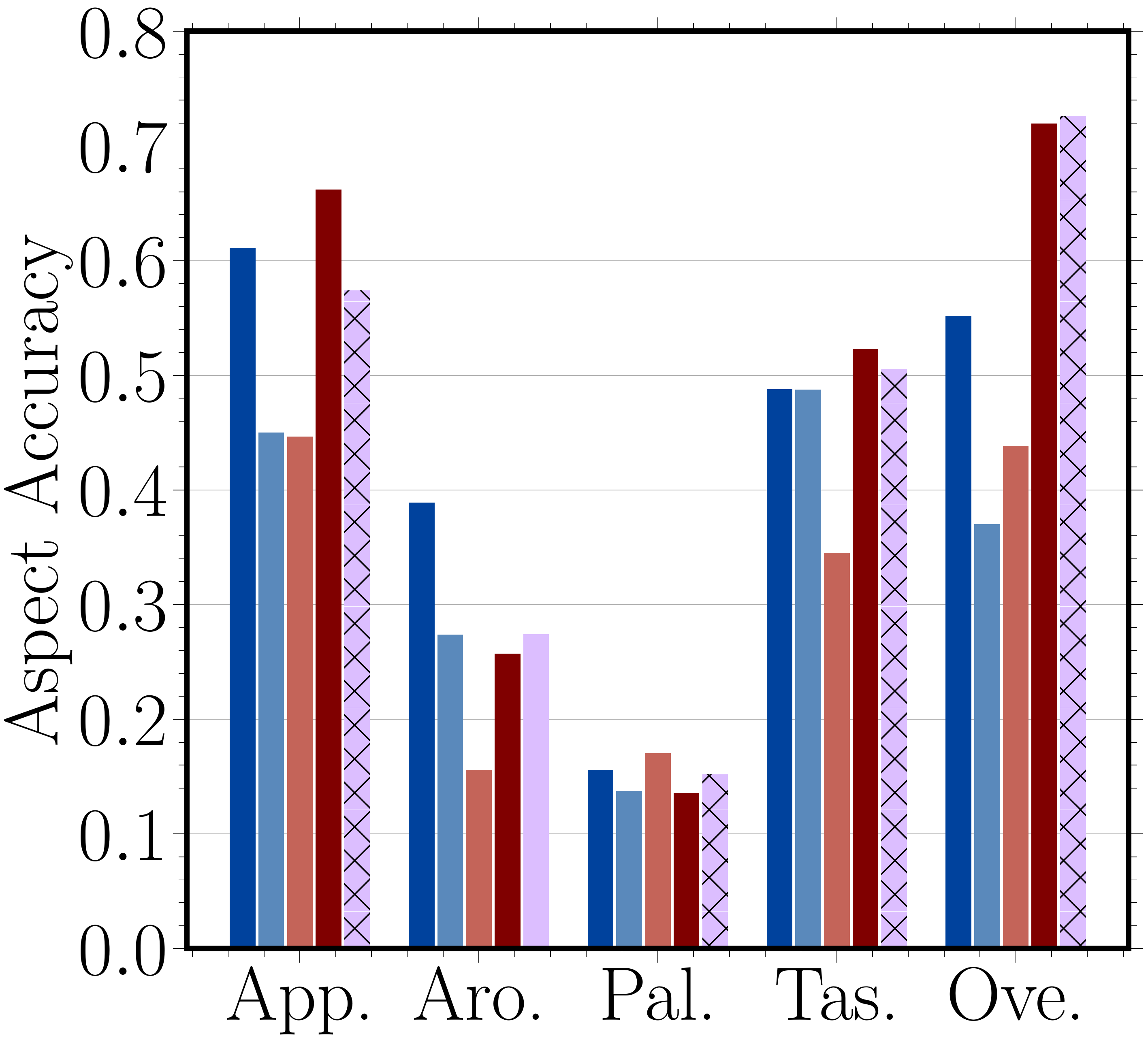}
\end{subfigure}
\begin{subfigure}[t]{0.33\textwidth}
\centering
\includegraphics[width=\textwidth,height=4.65cm]{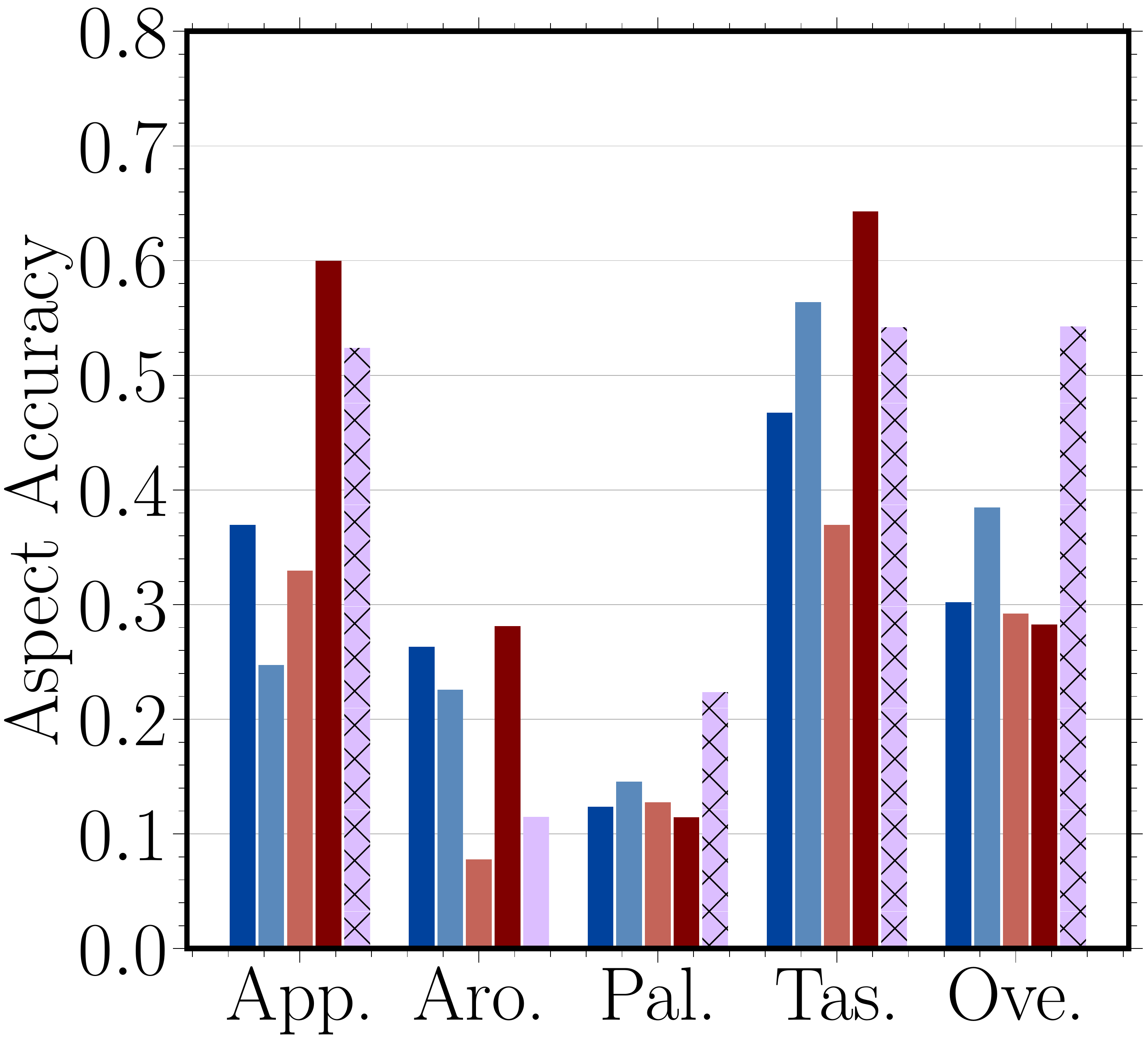}
\end{subfigure}
\begin{subfigure}[t]{0.33\textwidth}
\centering
\includegraphics[width=\textwidth,height=4.65cm]{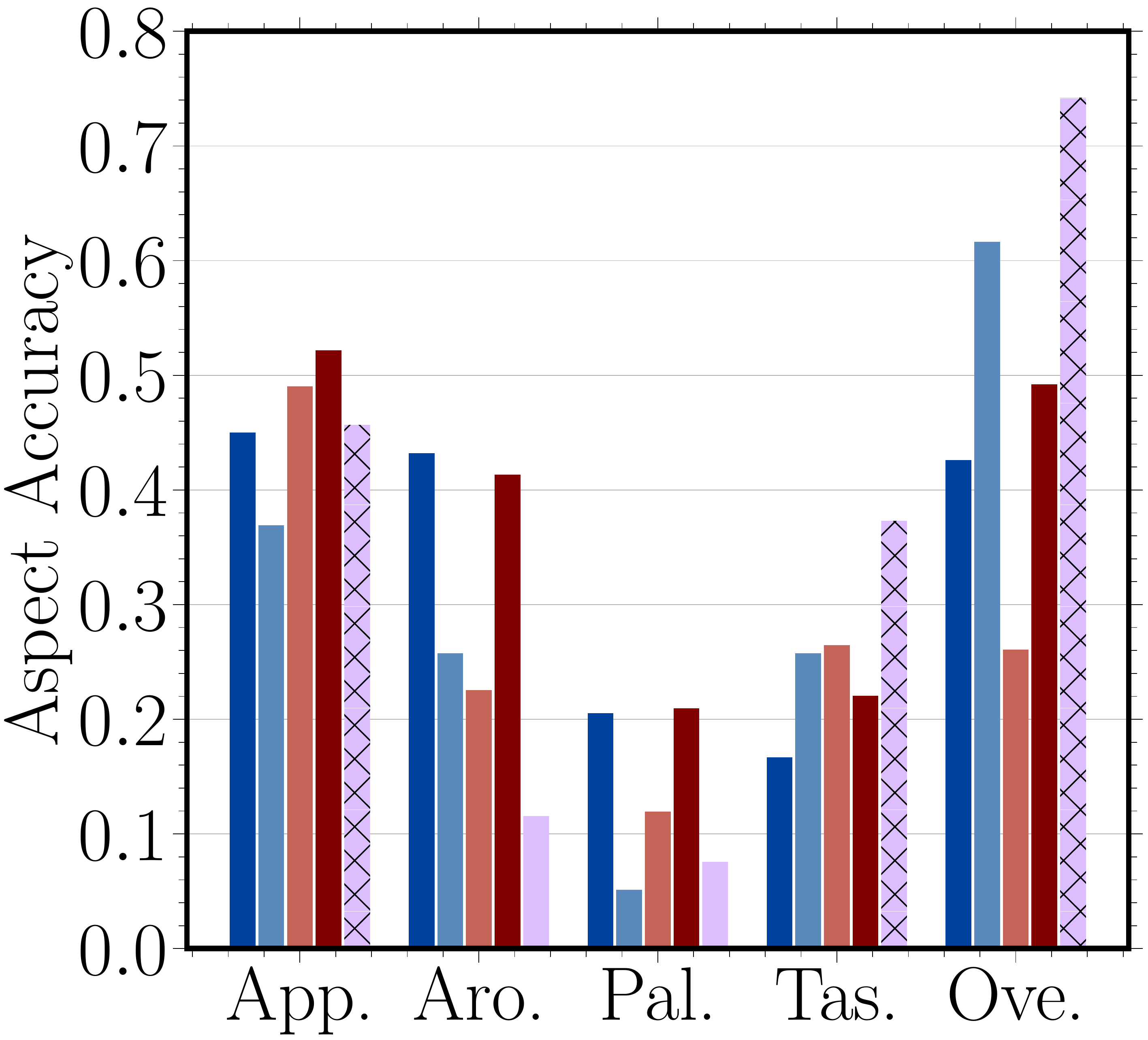}
\end{subfigure}
\par\bigskip
\par\bigskip

\begin{subfigure}[t]{0.33\textwidth}
\centering
\includegraphics[width=\textwidth,height=4.65cm]{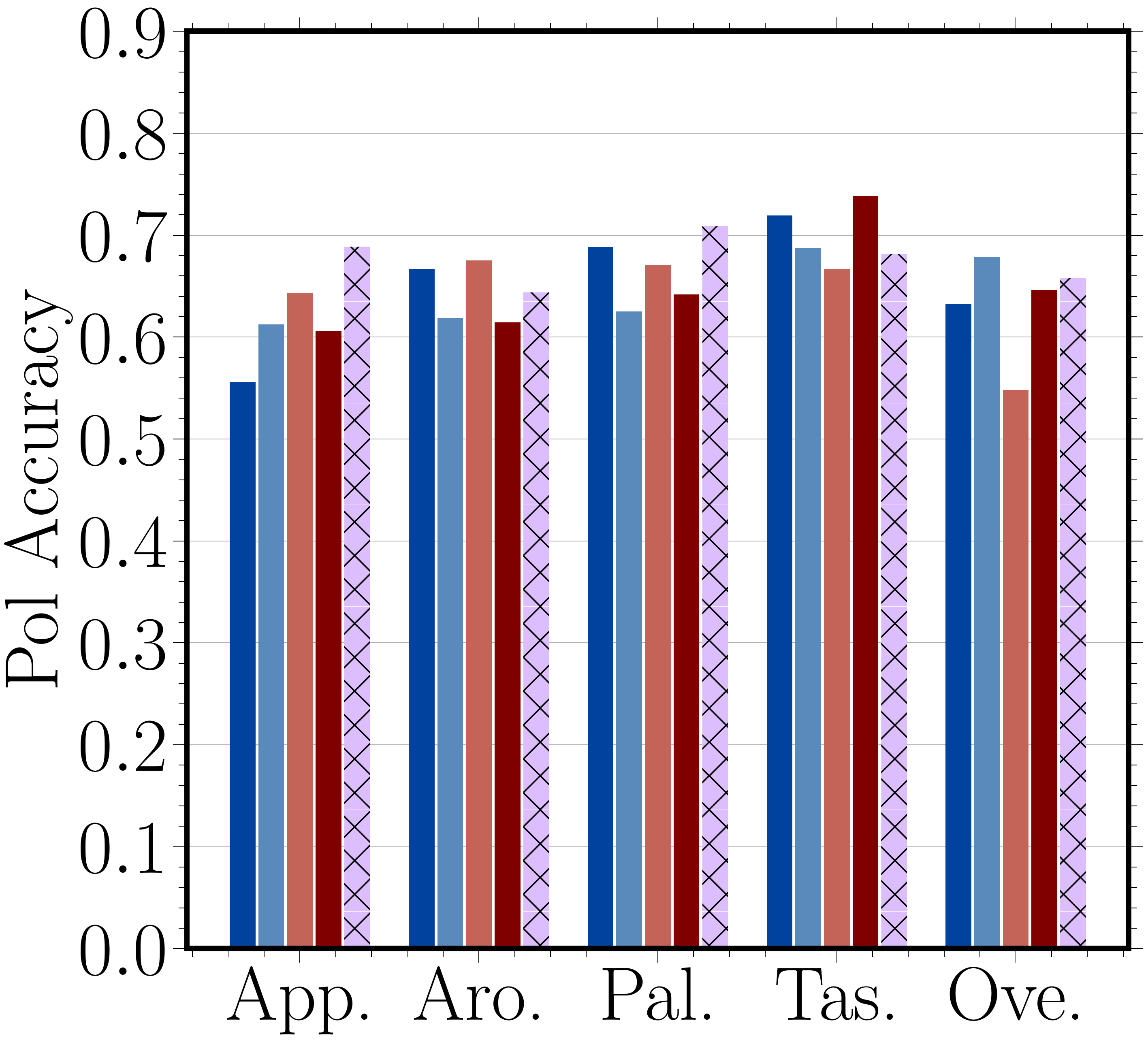}
\caption{Concept length $\ell=10$.}
\end{subfigure}
\begin{subfigure}[t]{0.33\textwidth}
\centering
\includegraphics[width=\textwidth,height=4.65cm]{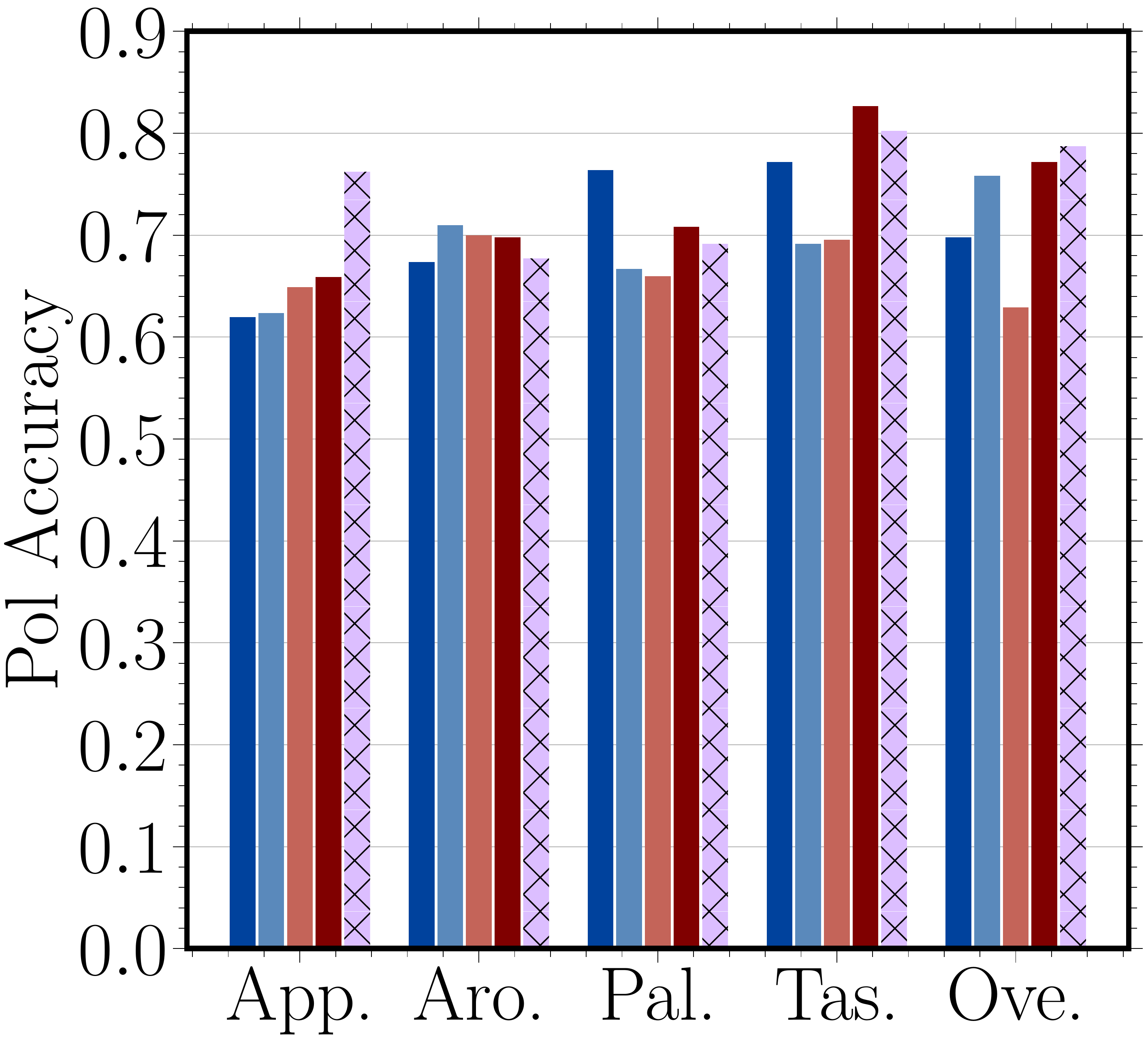}
\caption{Concept length $\ell=20$.}
\end{subfigure}
\begin{subfigure}[t]{0.33\textwidth}
\centering
\includegraphics[width=\textwidth,height=4.65cm]{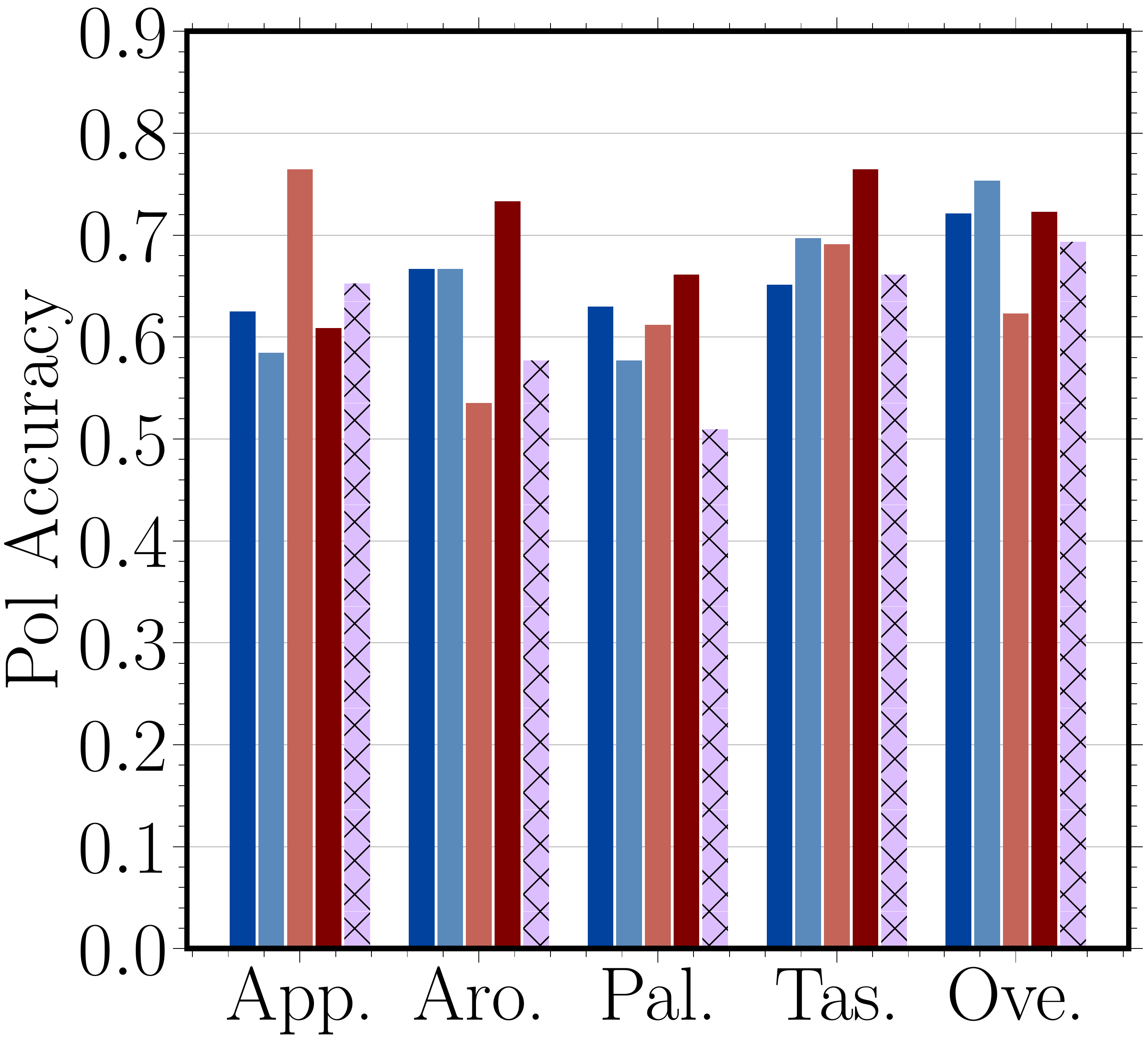}
\caption{Concept length $\ell=5$.}
\end{subfigure}

\caption{\label{fig_app_rq2_subj}Subjective performance per aspect of rationales for the multi-aspect beer reviews.}
\end{figure}

All results (for the joint, the aspect, and the polarity accuracy) are shown in Figure~\ref{fig_app_rq2_subj}. In total, we used 7,500 samples ($100 \times 5 \times 5 \times 3$).

We also studied the error rates on each aspect. The Aroma and Palate aspects cause the highest error for all models. One possible reason is that users confuse these with the aspect Taste, hence their high correlations in rating scores \citep{antognini2019multi}.

\section{Full Natural Language Explanations Results}
\label{app_all_results}

We also compare T-RECS with more models than these of Section~\ref{sec_rq2_ijcai}: LexRank~\citep{erkan2004lexrank}, NRT~\citep{li2017neural}, Item-Rand, Ref2Seq Top-k, and ACMLM \citep{ni-etal-2019-justifying}. LexRank is a unsupervised multi-document summarizer that selects an unpersonalized justification among all historical justifications of an item. NRT generates an explanation based on rating and the word distribution of the review. Item-Rand is an unpersonalized baseline which outputs a justification randomly from the justification history $J^i$ of item $i$. Ref2Seq Top-k is an extension of Ref2Seq, where we explore another decoding strategy called Top-k sampling \citep{radford2019language}, which should be more diverse and suitable on high-entropy tasks \citep{holtzman2019curious}. Finally, ACMLM is an aspect conditional masked language model that randomly chooses a justification from $J^i$ (similar to Item-Rand) and then iteratively edits it into new content by replacing random words. We also include more metrics and R\textsubscript{Sent}, which computes the percentages of generated justifications sharing the same polarity as the targets according to a sentiment classifier\footnote{We employ the sentiment classifiers trained jointly with Multi-Target Masker of Chapter~\ref{chapter_aaai2021}, used to infer the \textit{markers} from which the justifications are extracted.}.

The complete results are presented in Table~\ref{just_auto_perfs_add}. Interestingly, Item-Rand performs closely to LexRank: the best justification, according to LexRank, is slightly better than a random one. On the other hand, ACMLM edits the latter by randomly replacing tokens with the language model but produces poor quality justification, similarly to \cite{ni-etal-2019-justifying}. Finally, we also observe that the polarities of the generated justifications for beers match nearly perfectly, unlike in hotels where the positive and negative nuances are much harder to capture.
\begin{table}[!t]
\small
    \centering
\caption{\label{just_auto_perfs_add}Performance of the generated personalized justifications on automatic evaluation.}
\hspace*{-1.25cm}
\begin{threeparttable}
\begin{tabular}{@{}clcccc@{\hspace*{7mm}}ccc@{\hspace*{1mm}}c@{\hspace*{4mm}}c@{\hspace*{4mm}}cc@{}}
& \textbf{Model} & \textbf{B-1} & \textbf{B-2} & \textbf{B-3} & \textbf{B-4} & \textbf{R-1} & \textbf{R-2} & \textbf{R-L} & \textbf{BERT\textsubscript{Score}} & \textbf{PPL}$\downarrow$ & \textbf{R\textsubscript{KW}} & \textbf{R\textsubscript{Sent}}\\
\toprule
\multirow{11}{*}{\rotatebox{90}{\textit{Hotel}}} & Item-Rand & 11.50 & 2.88 & 0.91 & 0.32 & 12.65 & 0.87 & 9.75& 84.20 & - & 6.92 & 56.88\\
& LexRank & 12.12 & 3.31 & 1.10 & 0.41 & 14.74 & 1.16 & 10.61 & 83.91 & - & 10.32 & 58.51\\
& ExpansionNet & 4.03 & 1.95 & 1.01 & 0.53 & 34.22 & 9.65 & 6.91 & 74.81 & 28.87 & 60.09 & 61.38\\
\cdashlinelr{2-12}
& NRT & 17.00 & 6.51 & 3.06 & 1.51 & 18.21 & 2.88 & 16.08 & 86.31 & 29.75 & 11.44 & 64.46\\
& DualPC & 18.91 & 6.88 & 3.18 & 1.53 & 20.12 & 3.08 & 16.73 & 86.76 & 28.99 & 13.12 & 63.54\\
& CAML & 10.93 & 4.11 & 2.09 & 1.13 & 15.44 & 2.37 & 16.67 & 87.77 & 29.10 & 16.17 & 65.14\\
& Ref2Seq & 17.57 & 7.03 & 3.44 & 1.77 & 19.07 & 3.43 & 16.45 & 86.74 & \multirow{2}{*}{29.07} & 13.19 & 64.40\\
& Ref2Seq Top-k & 12.68 & 3.46 & 1.11 & 0.40 & 12.67 & 0.95 & 10.30 & 84.29 & & 6.38 & 58.11\\
& AP-Ref2Seq & 32.04 & 19.03 & 11.76 & 7.28 & 38.90 & 14.53 & 33.71 & 88.31 & 21.31 & 90.20 & 69.37\\
& ACMLM & 8.60 & 2.42 & 1.12 & 0.62 & 9.79 & 0.55 & 7.23 & 81.90 & - & 13.24 & 60.00\\
\cdashlinelr{2-12}
& T-RECS (Ours) & \textbf{33.53} & \textbf{19.76} & \textbf{12.14} & \textbf{7.47} & \textbf{40.29} & \textbf{14.74} & \textbf{34.10} & \textbf{90.23} & \textbf{17.80} & \textbf{93.57} & \textbf{70.12}\\
\bottomrule
\multirow{11}{*}{\rotatebox{90}{\textit{Beer}}} &  Item-Rand & 10.96 & 3.02 & 0.91 & 0.29 & 10.28 & 0.75 & 8.25 & 83.39 & - & 6.70 & 99.61\\
& LexRank & 12.23 & 3.58 & 1.12 & 0.38 & 13.81 & 1.16 & 9.90 & 83.42 & - & 10.79 & 99.88\\
& ExpansionNet & 6.48 & 3.59 & 2.06 & 1.22 & \textbf{54.53} & 18.24 & 9.68 & 72.32 & 22.28 & 82.49 & \textbf{99.99}\\
& NRT & 18.54 & 8.53 & 4.40 & 2.43 & 17.46 & 3.61 & 15.56 &  85.26 & 21.22 & 11.43 & \textbf{99.99}\\
& DualPC & 18.38 & 8.10 & 3.95 & 2.08 & 17.61 & 3.38 & 14.68 & 85.49 &21.15 & 10.60 & \textbf{99.99}\\
& CAML & 12.94 & 6.5 & 3.80 & 2.43 & 14.63 & 2.43 & 14.99 & 85.96 & 21.29 & 10.18 & \textbf{99.99}\\
\cdashlinelr{2-12}
& Ref2Seq & 18.75 & 9.47 & 5.51 & 3.51 & 18.25 & 4.52 & 15.96 & 85.27 & \multirow{2}{*}{22.34} & 12.10 & \textbf{99.99}\\
& Ref2Seq Top-k & 13.92 & 5.02 & 2.10 & 1.01 & 12.36 & 1.50 & 10.52 & 84.14 & & 8.51 & 99.83\\
& AP-Ref2Seq & 44.84 & 30.57 & 21.68 & 15.89 & 51.38 & 23.27 & 46.50 & 91.35 & 12.07 & 91.52 & \textbf{99.99}\\
& ACMLM & 7.76 & 2.54 & 0.91 & 0.34 & 8.33 & 0.87 & 6.17 & 80.94 & - & 10.33 & \textbf{99.99}\\
\cdashlinelr{2-12}
& T-RECS (Ours) & \textbf{46.50} & \textbf{31.56} & \textbf{22.42} & \textbf{16.54} & 53.12 & \textbf{23.86} & \textbf{47.20} & \textbf{91.50} & \textbf{10.24} & \textbf{94.96} & \textbf{99.99}\\
\bottomrule
\end{tabular}
\end{threeparttable}
\end{table}

\section{Full Keyphrase Explanation Results}
\label{app_kw_perfs2}
Table~\ref{kw_perfs2} contains more results to the keyphrase explanation quality experiment (Section~\ref{sec_keyphrases_explanations}).

\begin{table}[!h]
\small
    \centering
\caption{\label{kw_perfs2}Performance of personalized keyphrase explanation quality.}
\hspace*{-1.25cm}
\begin{threeparttable}
\begin{tabular}{@{}ll@{\hspace{1.5mm}}c@{\hspace{1mm}}c@{\hspace{1mm}}cc@{\hspace{1.5mm}}c@{\hspace{1mm}}c@{\hspace{1mm}}cc@{\hspace{1.5mm}}c@{\hspace{1mm}}c@{\hspace{1mm}}cc@{\hspace{1.5mm}}c@{\hspace{1mm}}c@{\hspace{1mm}}c@{}}
& & \multicolumn{3}{c}{\textbf{NDCG@N}} & & \multicolumn{3}{c}{\textbf{MAP@N}} & & \multicolumn{3}{c}{\textbf{Precision@N}} & & \multicolumn{3}{c}{\textbf{Recall@N}}\\
\cmidrule{3-5}\cmidrule{7-9}\cmidrule{11-13}\cmidrule{15-17}
& \textbf{Model} & N=5 & N=10 & N=20 & & N=5 & N=10 & N=20 & & N=5 & N=10 & N=20 & & N=5 & N=10 & N=20\\
\toprule
\multirow{8}{*}{\rotatebox{90}{\textit{Hotel}}} & UserPop & 0.2625 & 0.3128 & 0.3581 & & 0.2383 & 0.1950 & 0.1501 & & 0.1890 & 0.1332 & 0.0892 & & 0.2658 & 0.3694 & 0.4886\\
& ItemPop & 0.2801 & 0.3333 & 0.3822 & & 0.2533 & 0.2083 & 0.1608 & & 0.2041 & 0.1431 & 0.0959 & & 0.2866 & 0.3961 & 0.5245\\
\cdashlinelr{2-17}
& E-NCF & 0.2901 & 0.3410 & 0.3889 & & 0.2746 & 0.2146 & 0.1618 & & 0.1943 & 0.1366 & 0.0919 & & 0.2746 & 0.3802 & 0.5057\\
& CE-NCF & 0.1929 & 0.2286 & 0.2634 & & 0.1825 & 0.1432 & 0.1085 & & 0.1290 & 0.0918 & 0.0631 & & 0.1809 & 0.2548 & 0.3469\\
\cdashlinelr{2-17}
& E-VNCF & 0.2902 & 0.3441 & 0.3925 & & 0.2746 & 0.2158 & 0.1634 & & 0.1947 & 0.1391 & 0.0932 & & 0.2746 & 0.3860 & 0.5132\\
& CE-VNCF & 0.1727 & 0.2289 & 0.2761 & & 0.1530 & 0.1336 & 0.1115 & & 0.1275 & 0.1071 & 0.0767 & & 0.1795 & 0.2965 & 0.4200\\
\cdashlinelr{2-17}
& T-RECS (Ours) & \textbf{0.3158} & \textbf{0.3763} & \textbf{0.4319} & & \textbf{0.2919} & \textbf{0.2356} & \textbf{0.1807} & & \textbf{0.2223} & \textbf{0.1581} & \textbf{0.1068} & & \textbf{0.3109} & \textbf{0.4358} & \textbf{0.5812}\\
\cmidrule[0.125em]{1-17}
\multirow{8}{*}{\rotatebox{90}{\textit{Beer}}} & UserPop & 0.2049 & 0.2679 & 0.3357 & & 0.2749 & 0.2404 & 0.2014 & & 0.2366 & 0.1901 & 0.1445 & & 0.1716 & 0.2767 & 0.4207\\
& ItemPop & 0.1948 & 0.2495 & 0.3131 & & 0.2653 & 0.2291 & 0.1894 & & 0.2267 & 0.1759 & 0.1342 & & 0.1618 & 0.2529 & 0.3886\\
\cdashlinelr{2-17}
& E-NCF & 0.1860 & 0.2485 & 0.3158 & & 0.2488 & 0.2204 & 0.1877 & & 0.2162 & 0.1789 & 0.1389 & & 0.1571 & 0.2618 & 0.4040\\
& CE-NCF & 0.1471 & 0.1922 & 0.2422 & & 0.1967 & 0.1721 & 0.1446 & & 0.1687 & 0.1363 & 0.1050 & & 0.1227 & 0.1974 & 0.3033\\
\cdashlinelr{2-17}
& E-VNCF & 0.1763 & 0.2362 & 0.3055 & & 0.2389 & 0.2097 & 0.1797 & & 0.2031 & 0.1696 & 0.1356 & & 0.1471 & 0.2478 & 0.3943\\
& CE-VNCF & 0.1512 & 0.2025 & 0.2595 & & 0.1987 & 0.1784 & 0.1532 & & 0.1774 & 0.1475 & 0.1155 & & 0.1293 & 0.2146 & 0.3352\\
\cdashlinelr{2-17}
& T-RECS (Ours) & \textbf{0.2394} & \textbf{0.3163} & \textbf{0.3946} & & \textbf{0.3127} & \textbf{0.2799} & \textbf{0.2369} & & \textbf{0.2800} & \textbf{0.2284} & \textbf{0.1717} & & \textbf{0.2048} & \textbf{0.3320} & \textbf{0.4970}\\
\bottomrule
\end{tabular}
\end{threeparttable}
\end{table}

\section{Multi-Step Critiquing - Additional Metrics for T-RECS}
\label{app_multistep_add_metrics}
\begin{figure}[!h]
\centering
    \includegraphics[width=0.8\textwidth]{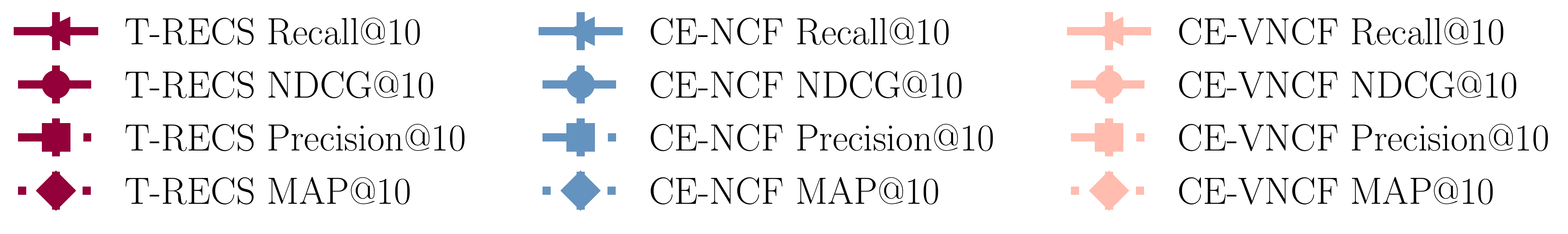}
\\
\begin{subfigure}[t]{0.4\textwidth}
    \centering
    \includegraphics[width=0.8\textwidth]{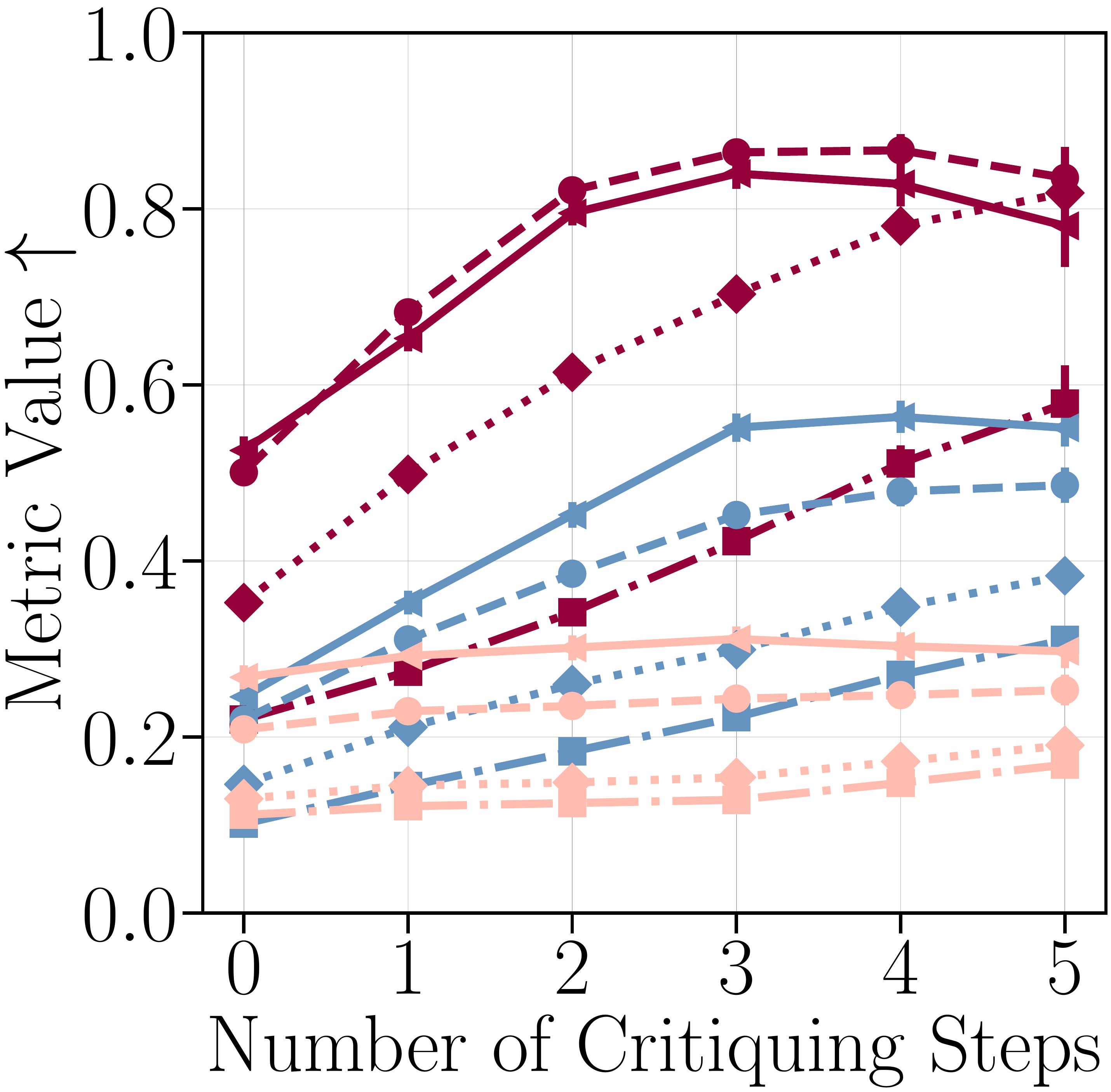}
    \caption{Results on the \textbf{hotel} dataset.}
\end{subfigure}
\hfill
\begin{subfigure}[t]{0.4\textwidth}
    \centering
    \includegraphics[width=0.8\textwidth]{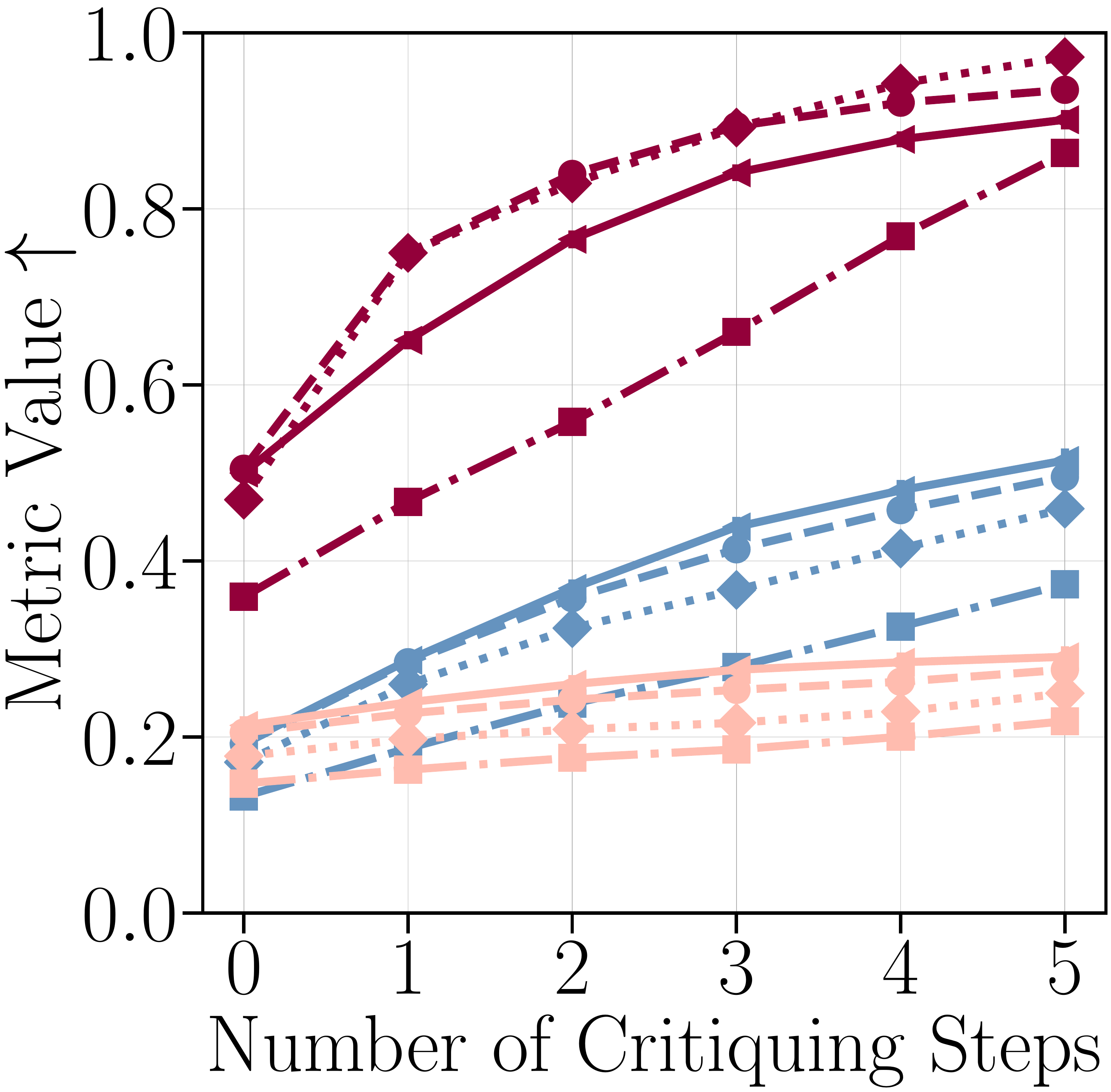}
    \caption{Results on the \textbf{beer} dataset.}
\end{subfigure}
\caption{\label{mul_mul_crit_both}Multi-step critiquing performance. Keyphrase prediction over multi-step critiquing in terms of Recall@10, NDCG@10, Precision@10, and MAP@10 with 95\% confidence interval. a)~Results on the hotel dataset, b)~on the beer dataset.}
\end{figure}

\chapter{Processing}

\section{Processing \& Filtering \textit{Markers}}
\label{app_process_filtering}

The method described in \cite{antognini2019multi} extracts, most of the time, \textit{markers} that consist of long, continuous spans of words. However, sometimes, the \textit{markers} are too short because some reviews do not include enough words to justify a certain aspect rating, or the \textit{markers} stop in the middle of a sentence. Although both are theoretically not wrong, we aim to create fluent and grammatically correct justifications. To this end, we exploit the constituency parse tree to ensure that \textit{markers} are noun/verb phrases. We apply the following steps to the entire set of reviews for each dataset:
\begin{enumerate}[topsep=0pt]
    \item Compute the constituency parse tree of each review;
    \item For each noun and adjective node in the constituency parse tree of a \textit{marker}, if the parent node is a verb or noun phrase, we add its children to the \textit{marker}. We follow the rules in \cite{giannakopoulos2017dataset};
    \item Filter out \textit{markers} having less than four tokens or including first and third-person pronouns.
\end{enumerate}
In cases where faceted ratings are not available at large cases, \cite{mukherjee2020uncertainty,niu-etal-2020-self} proposed elegant solutions to infer them from 20 or fewer samples. If faceted ratings are not present, one can use the concept rationalizer of \cite{antognini-2021-concept} to infer \textit{markers}.

\section{Extracting Keyphrase for T-RECS}
\label{app_kws_datasets}

None of our datasets contains initially preselected keyphrases. We extract $200$ keyphrases from the \textit{markers} used to model the user and item profiles. They serve~as~a basis for the explanation and the critiquing. Table~\ref{kws_datasets} shows some keyphrases for each dataset. We apply the following processing for each~dataset, similarly to \cite{keyphraseExtractionDeep}:
\begin{enumerate}[topsep=0pt]
    \item Group by aspect \textit{markers} from all reviews. The aspect sets come from faceted ratings\footnote{For the hotel reviews: service, cleanliness, value, location, and room. For beer reviews: appearance, smell, mouthfeel, and taste.};
    \item For each group of \textit{markers}:
    \begin{itemize}
        \item Tokenize and lemmatize the entire set of \textit{markers};
        \item Extract unigram lists of high-frequency noun and adjective phrases;
        \item Keep the top-$k$ most likely unigrams;
    \end{itemize}
    \item Represent each review as a binary vector indicating the presence of keyphrases.
\end{enumerate}

Another possibility to extract keywords from reviews is to leverage Microsoft Concept Graph as in \cite{chen2019co,chen2020towards}. However, the API is limited to single instance conceptualization.
\begin{table}
    \centering
   \caption{\label{kws_datasets}Some keyphrases mined from the inferred \textit{markers}. We grouped them by aspect for a better understanding.}
\begin{threeparttable}[t]
\begin{tabular}{@{}llc@{}}
\textbf{Dataset} & \textbf{Aspect} & \textbf{Keyphrases}\\
\cmidrule[0.08em]{1-3}
\multirow{5}{*}{Hotel}
& Service & bar, lobby, housekeeping, guest\\
& Cleanliness & carpet, toilet, bedding, cigarette\\
& Value & price, wifi, quality, motel, gym\\
& Location & airport, downtown, restaurant shop\\
& Room & bed, tv, balcony, fridge, microwave\\
\cmidrule[0.08em]{1-3}
\multirow{4}{*}{Beer} & Appearance & golden, dark, white, foamy\\
& Aroma & fruit, wheat, citrus, coffee\\
& Palate & creamy, chewy, syrupy, heavy\\
& Taste & bitter, sweet, balanced, nutty\\
\cmidrule[0.08em]{1-3}
\end{tabular}
\end{threeparttable}
\end{table}

\section{Addressing Users without Reviews in T-RECS}
\label{app_cold_start_trecs}
The cold-start problem is not particular to our method but a general problem in recommendation. However, our system could infer $\bb{\gamma^u}$ and $\bb{\beta^u}$ for users without reviews and then computes the initial latent representation $\bb{z}$ as in Section~\ref{sec_model}. We propose the following strategies:
\begin{enumerate}[topsep=0pt]
	\item \textbf{Users with ratings but no reviews}: we could leverage collaborative filtering techniques to identify users with similar ratings and build an initial representation.
	\item \textbf{New users}: following the observation of \cite{zhang2014explicit,musat2015personalizing}: ``users write opinions about the topics they care about'', we could ask new users to write about the different aspects they deem important. Their initial representation is an aggregation of other similar users. Another option is to ask users to select items they like based on a subset of items and build a profile from these preferences (see above).
\end{enumerate}

\chapter{Visualizations}

\section{Extracted Topic Words per Aspect from Rationales}
\label{sec_topic_words}

For each model, we computed the probability distribution of words per aspect by using the induced sub-masks $M_{a_1}, ..., M_{a_A}$ or attention values. Given an aspect $a_i$ and a set of top-$N$ words $\boldsymbol{w_{a_i}^N}$, the Normalized Pointwise Mutual Information~\citep{Bouma2009} coherence score is: \begin{equation}
    \text{NPMI}(\boldsymbol{w_{a_i}^N}) = \sum_{j=2}^{N} \sum_{k=1}^{j-1} \frac{\log \frac{P(w_{a_i}^k, w_{a_i}^j)}{P(w_{a_i}^k)P(w_{a_i}^j)}}{-\log P(w_{a_i}^k, w_{a_i}^j)}
\end{equation}
Top words of coherent topics (i.e., aspects) should share a similar semantic interpretation, and thus interpretability of a topic can be estimated by measuring how many words are not related. For each aspect $a_i$ and word $w$ having been highlighted at least once as belonging to aspect $a_i$, we computed the probability $P(w|a_i)$ on each dataset and sorted them in decreasing order of $P(w|a_i)$. Unsurprisingly, we found that the most common words are stop words such as ``a" and ``it", because masks are mostly word sequences instead of individual words. To gain a better interpretation of the aspect words, we followed the procedure in \cite{beer}: we first computed the averages across all aspect words for each word $w$ as follows: \begin{equation}
	b_w = \frac{1}{|A|}\sum_{i=1}^{|A|}P(w|a_i)
\end{equation} It represents a general distribution that includes words common to all aspects. The final word distribution per aspect is computed by removing the general distribution as follows:
\begin{equation}
	\hat{P}(w|a_i) = P(w|a_i) - b_w
\end{equation}
After generating the final word distribution per aspect, we picked the top ten words and asked two human annotators to identify intruder words (i.e., words not matching the corresponding aspect). We show in Table~\ref{topics_words_full_beer}, Table~\ref{topics_words_hotel}, and also Table~\ref{topics_words_dec_beer} the top ten words for each aspect, where \textcolor{redo}{\textbf{red}} denotes all words identified as unrelated to the aspect by the two annotators. Generally, our model finds better sets of words across the three datasets compared with other methods. Additionally, we observe that the aspects can be easily recovered, given its top words.

\begin{table}[!tb]
\centering
\caption{\label{topics_words_full_beer}Top ten words for each aspect from the \textbf{\textit{Beer}} dataset, learned by various models. \textcolor{redo}{\textbf{Red}} denotes intruders according to two annotators. Found words are generally noisier due to the high correlation between \textit{Taste} and other aspects. However, \textit{MTM} provides better results than other methods.}
\hspace*{-1.0cm}
\begin{tabular}{@{}c@{\hspace*{3mm}}l@{\hspace*{2mm}}l@{}}
\multicolumn{1}{c}{\textbf{}} & \multicolumn{1}{c}{\textbf{Model}} & \multicolumn{1}{c}{\textbf{Top-10 Words}}\\
\toprule

\multirow{4}{*}{\rotatebox{90}{\textit{Appearance}}}
& SAM & \textcolor{redo}{\textbf{nothing}}	beautiful	lager	nice	\textcolor{redo}{\textbf{average}}	macro	lagers	corn	\textcolor{redo}{\textbf{rich}}	gorgeous\\
& MASA & lacing	head	lace	\textcolor{redo}{\textbf{smell}}	amber	retention	beer	nice	carbonation	glass\\
& MAA & head lacing \textcolor{redo}{\textbf{smell}} \textcolor{redo}{\textbf{aroma}} color pours amber glass white retention\\
& MTM (Ours) & head	lacing	\textcolor{redo}{\textbf{smell}}	white	lace	retention	glass	\textcolor{redo}{\textbf{aroma}} tan	thin\\
\cmidrule{1-3}
\multirow{4}{*}{\rotatebox{90}{\textit{Smell}}}
& SAM & faint	\textcolor{redo}{\textbf{nice}}	\textcolor{redo}{\textbf{mild}}	light	slight	complex	good	wonderful	grainy	great\\
& MASA & aroma	hops	nose	chocolate	caramel	malt	citrus	fruit	smell	fruits\\
& MAA & \textcolor{redo}{\textbf{taste}} hints hint \textcolor{redo}{\textbf{lots}} \textcolor{redo}{\textbf{t-}} \textcolor{redo}{\textbf{starts}} blend mix \textcolor{redo}{\textbf{upfront}} malts\\
& MTM (Ours) & \textcolor{redo}{\textbf{taste}} 	malt	aroma	hops	sweet	citrus	caramel	nose	malts	chocolate \\
\cmidrule{1-3}
\multirow{4}{*}{\rotatebox{90}{\textit{Palate}}} 
& SAM & thin	\textcolor{redo}{\textbf{bad}}	light	watery	creamy	silky	medium	body	smooth	\textcolor{redo}{\textbf{perfect}}\\
& MASA & smooth	light	medium	thin	creamy	\textcolor{redo}{\textbf{bad}}	watery	\textcolor{redo}{\textbf{full}}	crisp	\textcolor{redo}{\textbf{clean}}\\
& MAA & good beer carbonation smooth \textcolor{redo}{\textbf{drinkable}} medium bodied \textcolor{redo}{\textbf{nice}} body \textcolor{redo}{\textbf{overall}}\\
& MTM (Ours) & carbonation	medium	mouthfeel	body	smooth	bodied \textcolor{redo}{\textbf{drinkability}}	creamy	light	\textcolor{redo}{\textbf{overall}} \\
\cmidrule{1-3}
\multirow{4}{*}{\rotatebox{90}{\textit{Taste}}}
& SAM & \textcolor{redo}{\textbf{decent}}	great	complex	delicious	tasty	favorite	\textcolor{redo}{\textbf{pretty}}	sweet	\textcolor{redo}{\textbf{well}}	\textcolor{redo}{\textbf{best}}\\
& MASA & good	\textcolor{redo}{\textbf{drinkable}}	\textcolor{redo}{\textbf{nice}}	tasty	great	enjoyable	\textcolor{redo}{\textbf{decent}}	\textcolor{redo}{\textbf{solid}}	balanced	\textcolor{redo}{\textbf{average}}\\
& MAA & malt hops flavor hop flavors caramel malts bitterness bit chocolate\\
& MTM (Ours) & malt	sweet	hops	flavor	bitterness	finish	chocolate	bitter	caramel	sweetness\\
\end{tabular}
\end{table}

\begin{table}[!htb]
\centering
\caption{\label{topics_words_hotel}Top ten words for each aspect from the \textbf{\textit{Hotel}} dataset, learned by various models. \textcolor{redo}{\textbf{Red}} denotes intruders according to human annotators. Besides \textit{SAM}, all methods find similar words for most aspects except the aspect \textit{Value}. The top words of \textit{MTM} do not contain any intruder.}
\begin{tabular}{@{}c@{\hspace*{3mm}}l@{\hspace*{2mm}}l@{}}
\multicolumn{1}{c}{\textbf{}} & \multicolumn{1}{c}{\textbf{Model}} & \multicolumn{1}{c}{\textbf{Top-10 Words}}\\
\toprule
\multirow{4}{*}{\rotatebox{90}{\textit{Service}}}
& SAM & staff service friendly nice told helpful good great lovely manager\\
& MASA & friendly helpful told rude nice good pleasant asked enjoyed worst\\
& MAA & staff service helpful friendly nice good rude excellent great desk\\
& MTM (Ours) & staff friendly service desk helpful manager reception told rude asked \\
\cmidrule{1-3}
\multirow{4}{*}{\rotatebox{90}{\textit{Cleanliness}}}
& SAM & clean cleaned dirty toilet smell cleaning sheets comfortable nice hair\\
& MASA & clean dirty cleaning spotless stains cleaned cleanliness mold filthy bugs \\
& MAA & clean dirty cleaned filthy stained well spotless carpet sheets stains\\
& MTM (Ours) & clean dirty bathroom room bed cleaned sheets smell carpet toilet\\
\cmidrule{1-3}
\multirow{4}{*}{\rotatebox{90}{\textit{Value}}}
& SAM & good stay great well \textcolor{redo}{\textbf{dirty}} recommend worth definitely \textcolor{redo}{\textbf{friendly}} charged\\
& MASA & great good poor excellent terrible awful \textcolor{redo}{\textbf{dirty}} horrible \textcolor{redo}{\textbf{disgusting}} \textcolor{redo}{\textbf{comfortable}}\\
& MAA &\textcolor{redo}{\textbf{night}} \textcolor{redo}{\textbf{stayed}} stay \textcolor{redo}{\textbf{nights}} \textcolor{redo}{\textbf{2}} \textcolor{redo}{\textbf{day}} price \textcolor{redo}{\textbf{water}} \textcolor{redo}{\textbf{4}} \textcolor{redo}{\textbf{3}}\\
& MTM (Ours) & good price expensive paid cheap worth better pay overall disappointed\\
\cmidrule{1-3}
\multirow{4}{*}{\rotatebox{90}{\textit{Location}}}
& SAM & location close far place walking \textcolor{redo}{\textbf{definitely}} located \textcolor{redo}{\textbf{stay}} short view\\
& MASA & location beach walk hotel town located restaurants walking close taxi\\
& MAA & location hotel place located close far area beach view situated\\
& MTM (Ours) & location great area walk beach hotel town close city street\\
\cmidrule{1-3}
\multirow{4}{*}{\rotatebox{90}{\textit{Room}}}
& SAM &\textcolor{redo}{\textbf{dirty} \textbf{clean}} small best comfortable large worst modern \textcolor{redo}{\textbf{smell}} spacious\\
& MASA & comfortable small spacious nice large dated well tiny modern basic \\
& MAA & room rooms bathroom bed spacious small beds large shower modern\\
& MTM (Ours)& comfortable room small spacious nice modern rooms large tiny walls\\
\end{tabular}
\end{table}

\begin{table}[!htb]
\centering
\caption{\label{topics_words_dec_beer}Top ten words for each aspect from the \textbf{\textit{decorrelated}} \textit{Beer} dataset, learned by various models. \textcolor{redo}{\textbf{Red}} denotes intruders according to two annotators. For the three aspects, MTM has only one word considered as an intruder, followed by MASA with SAM (two) and MAA (six).}
\hspace*{-0.75cm}
\begin{tabular}{@{}c@{\hspace*{3mm}}l@{\hspace*{2mm}}l@{}}
\multicolumn{1}{c}{\textbf{}} & \multicolumn{1}{c}{\textbf{Model}} & \multicolumn{1}{c}{\textbf{Top-10 Words}}\\
\toprule

\multirow{4}{*}{\rotatebox{90}{\textit{Appearance}}}
& SAM  & head color white brown dark lacing \textcolor{redo}{\textbf{pours}} amber clear black\\
& MASA & head	lacing	lace	retention	glass	foam	color	amber	yellow	cloudy\\
& MAA & nice dark amber \textcolor{redo}{\textbf{pours}} black hazy brown \textcolor{redo}{\textbf{great}} cloudy clear\\
& MTM (Ours) & head color lacing white brown clear amber glass black retention \\
\cmidrule{1-3}
\multirow{4}{*}{\rotatebox{90}{\textit{Smell}}}
& SAM & sweet malt hops coffee chocolate citrus hop strong smell aroma\\
& MASA & smell	aroma	nose	smells	sweet	aromas	scent	hops	malty	roasted\\
& MAA &\textcolor{redo}{\textbf{taste}} smell aroma sweet chocolate \textcolor{redo}{\textbf{lacing}} malt roasted hops nose\\
& MTM (Ours) & smell aroma nose smells sweet malt citrus chocolate caramel aromas\\
\cmidrule{1-3}
\multirow{4}{*}{\rotatebox{90}{\textit{Palate}}} 
& SAM & mouthfeel smooth medium carbonation bodied watery body thin creamy \textcolor{redo}{\textbf{full}}\\
& MASA & mouthfeel	medium	smooth	body	\textcolor{redo}{\textbf{nice}}	m-	feel	bodied	mouth	\textcolor{redo}{\textbf{beer}}\\
& MAA & carbonation mouthfeel medium \textcolor{redo}{\textbf{overall}} smooth finish body \textcolor{redo}{\textbf{drinkability}} bodied watery\\
& MTM (Ours)& mouthfeel carbonation medium smooth body bodied \textcolor{redo}{\textbf{drinkability}} good mouth thin\\
\end{tabular}
\end{table}

\clearpage

\section{Multi-Dimensional Rationales}
\label{samples_with_visualization}

\begin{figure}[!t]
\centering
\begin{tabular}{@{}c@{\makebox[0.25cm]{ }}c@{}}
    \includegraphics[width=0.475\textwidth]{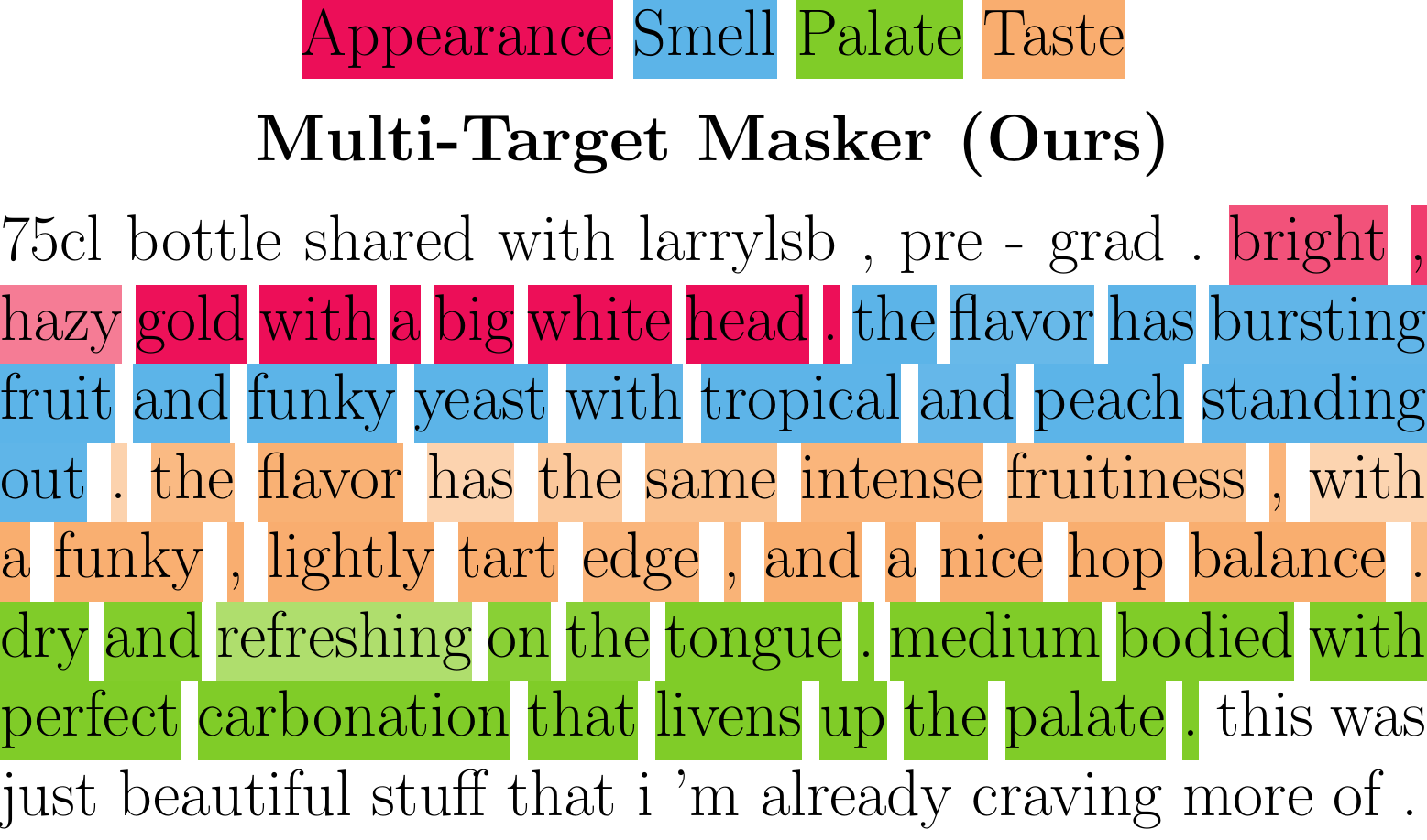} & 
    \includegraphics[width=0.475\textwidth]{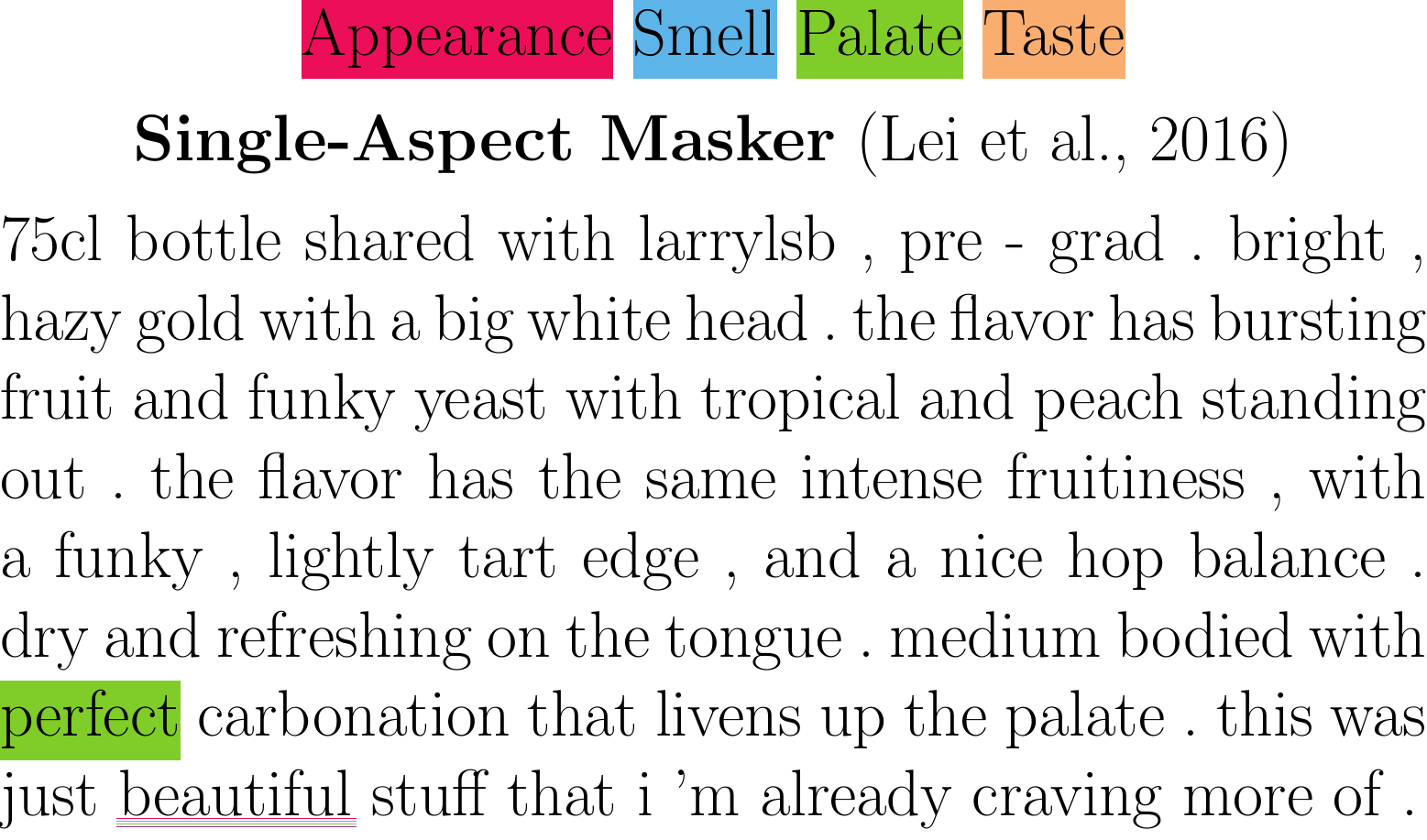} \\\\
    \includegraphics[width=0.475\textwidth]{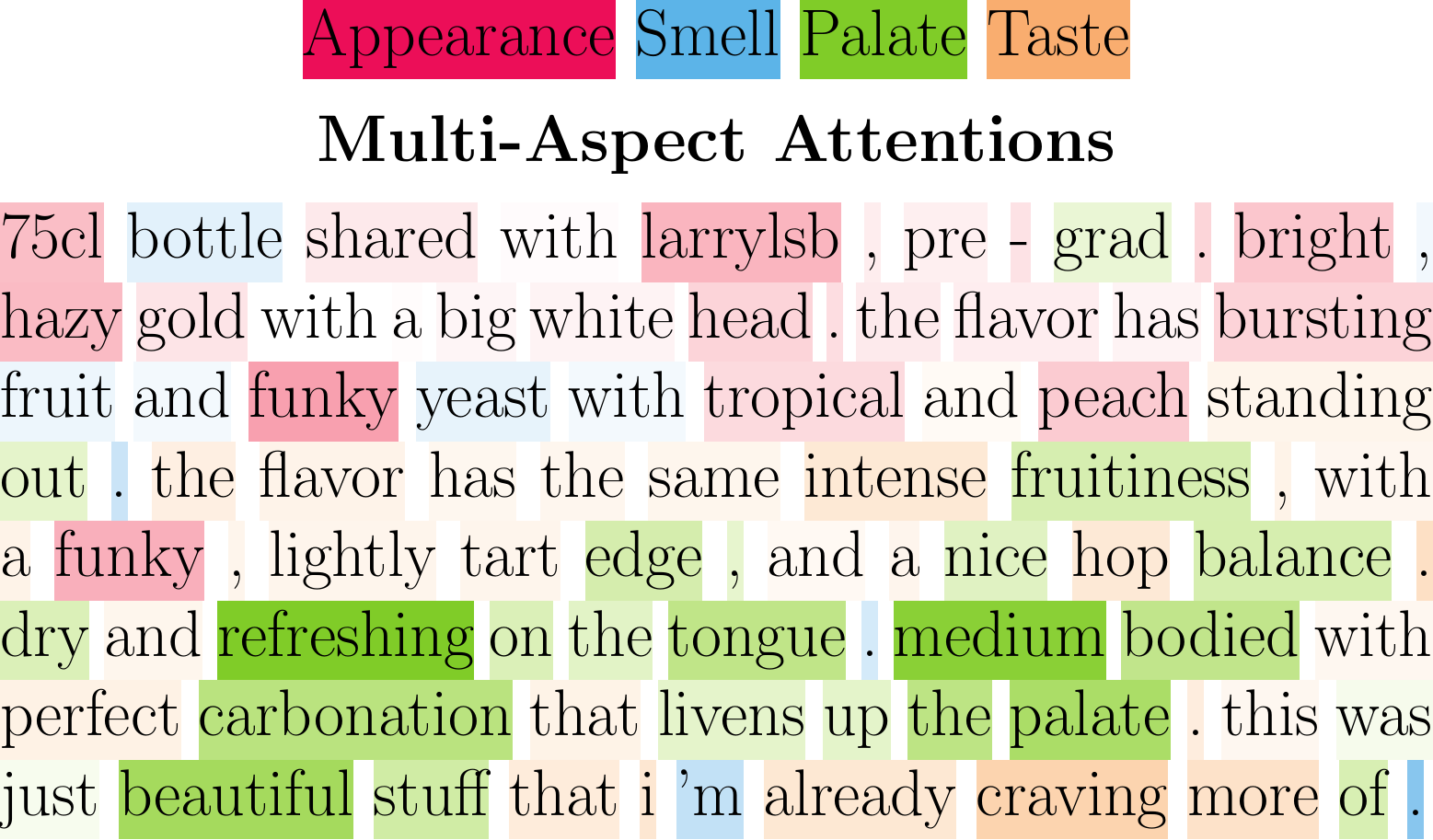} &
    \includegraphics[width=0.475\textwidth]{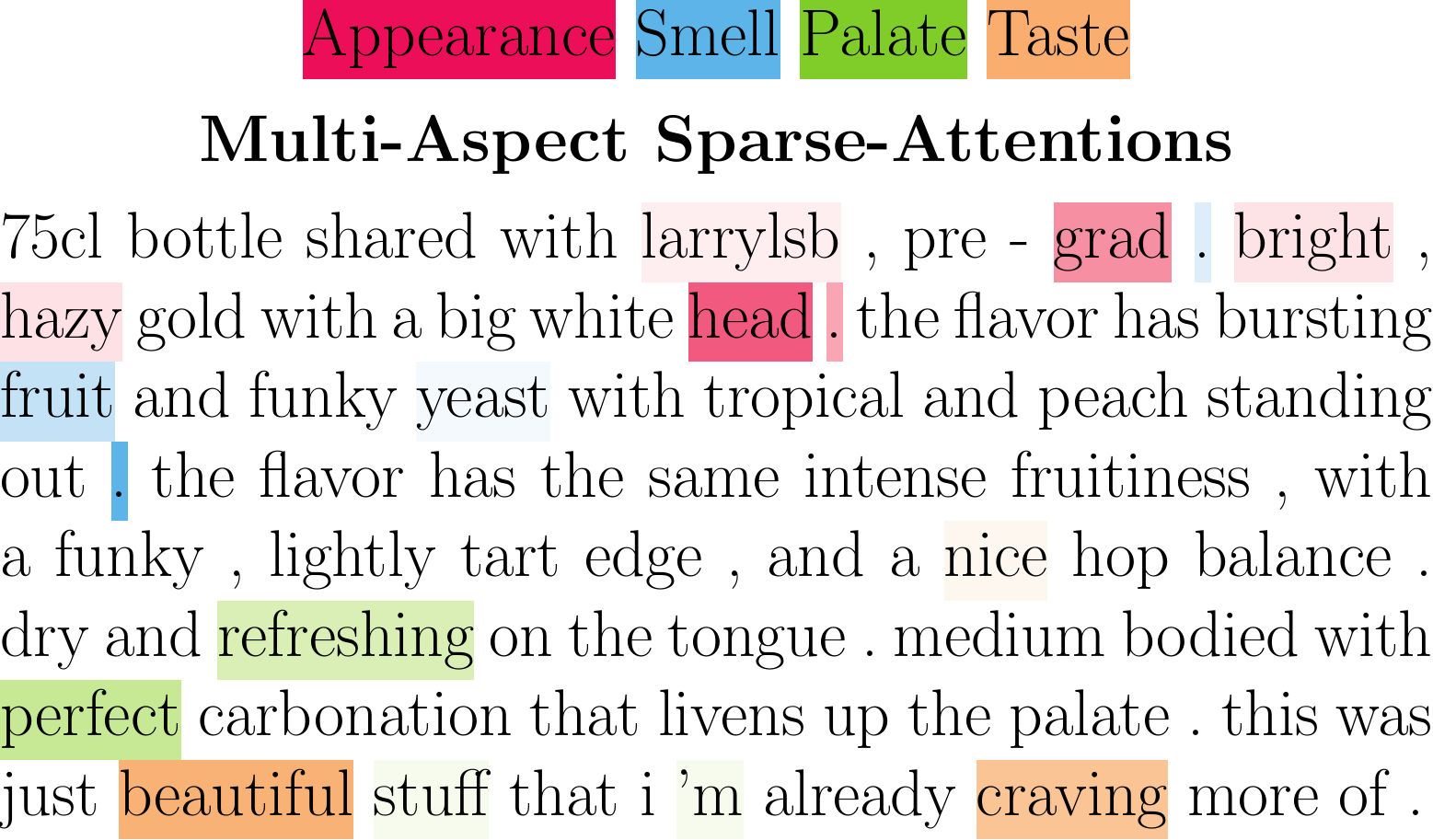}\\\\
\end{tabular}

\caption{\label{sample_full_beer_0}A sample review from the \textbf{\textit{Beer}} dataset, with computed masks from different methods. \textit{MTM} achieves near-perfect annotations, while \textit{SAM} highlights only two words where one is ambiguous with respect to the four aspects. \textit{MAA} mixes between the aspect \textit{Appearance} and \textit{Smell}. \textit{MASA} identifies some words but lacks coverage.}
\end{figure}

\begin{figure}[!htb]

\centering
\begin{tabular}{@{}c@{\makebox[0.25cm]{ }}c@{}}
     \includegraphics[width=0.475\textwidth,height=10cm]{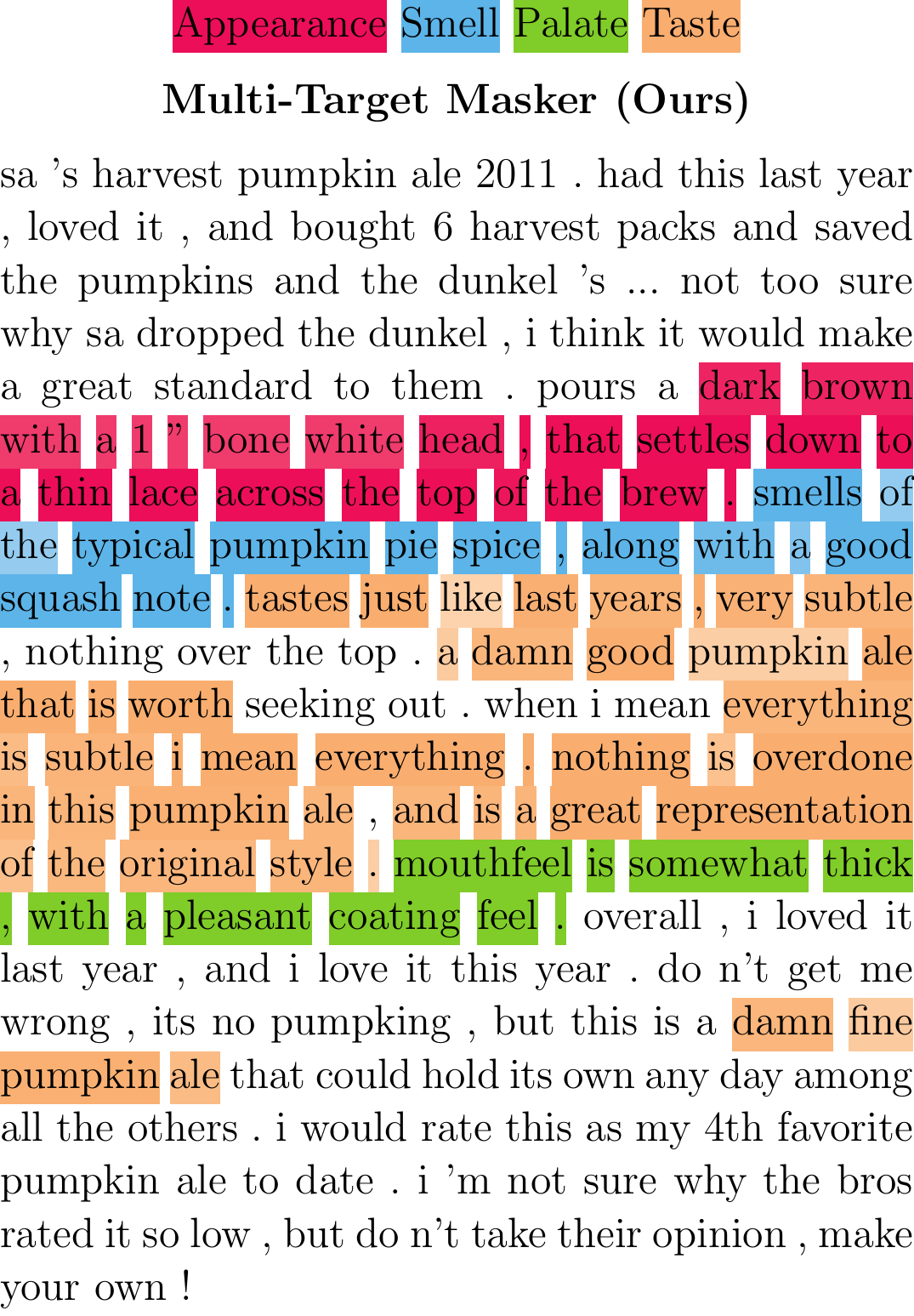} & 
     \includegraphics[width=0.475\textwidth,height=10cm]{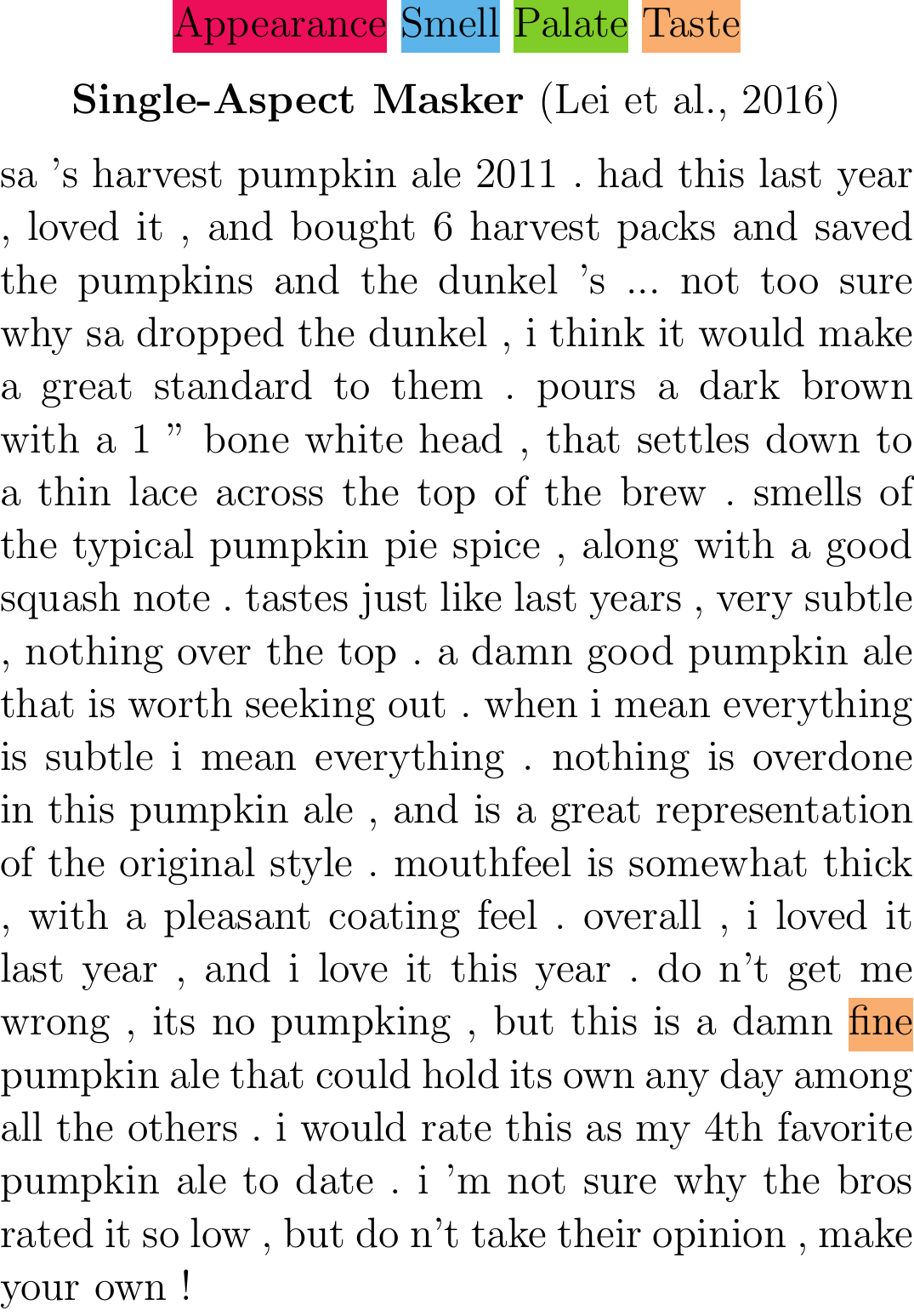} \\\\
     \includegraphics[width=0.475\textwidth,height=10cm]{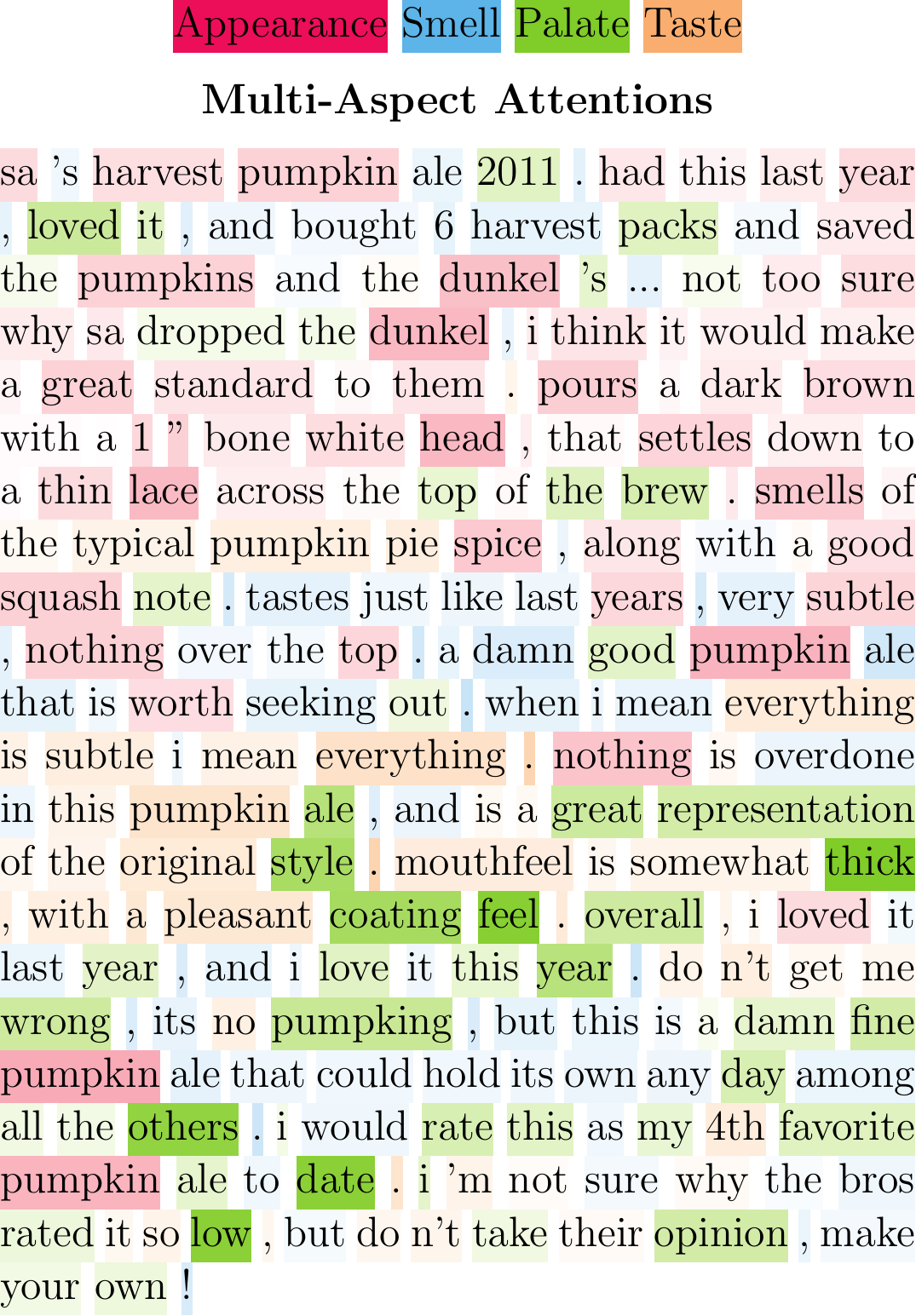} & 
     \includegraphics[width=0.475\textwidth,height=10cm]{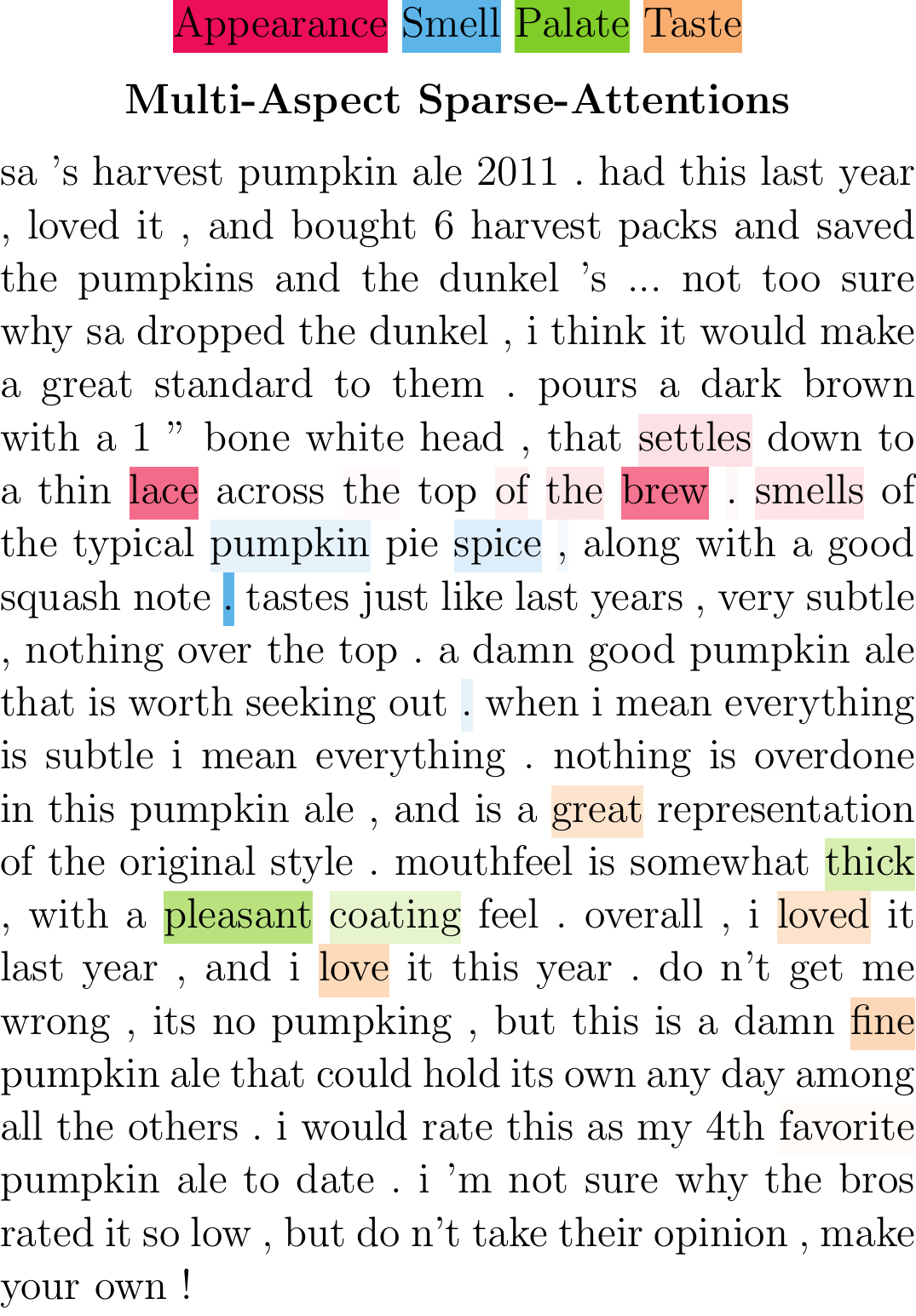}\\
\end{tabular}

\caption{\label{sample_full_beer_1}A sample review from the \textbf{\textit{Beer}} dataset, with computed masks from different methods. \textit{MTM} can accurately identify what parts of the review describe each aspect. \textit{MAA} provides very noisy labels due to the high imbalance and correlation between aspects, while \textit{MASA} highlights only a few important words. We can see that \textit{SAM} is confused and performs a poor selection.}
\end{figure}

\begin{figure}[!htb]
\centering
\begin{tabular}{@{}c@{\makebox[0.25cm]{ }}c@{}}
     \includegraphics[width=0.475\textwidth]{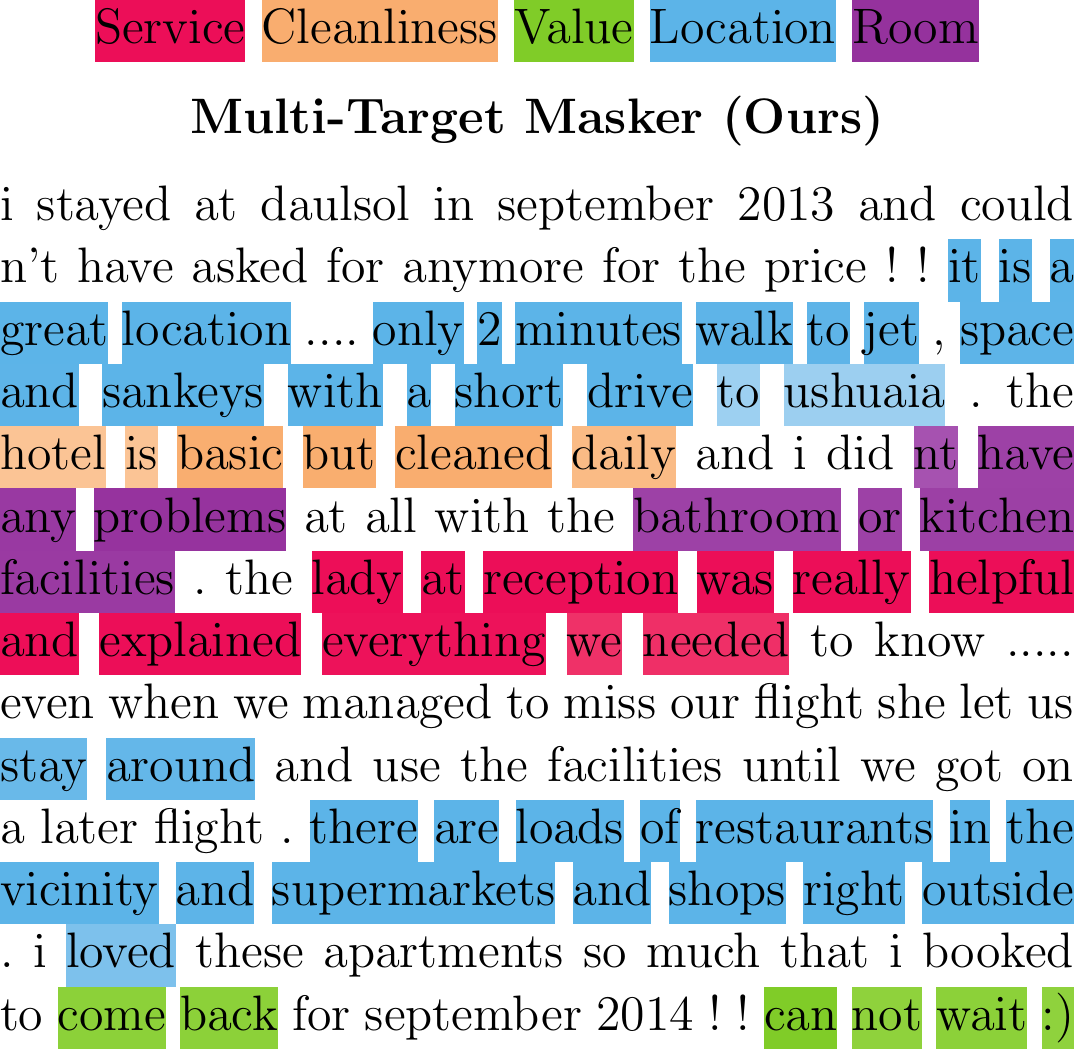} & 
     \includegraphics[width=0.475\textwidth]{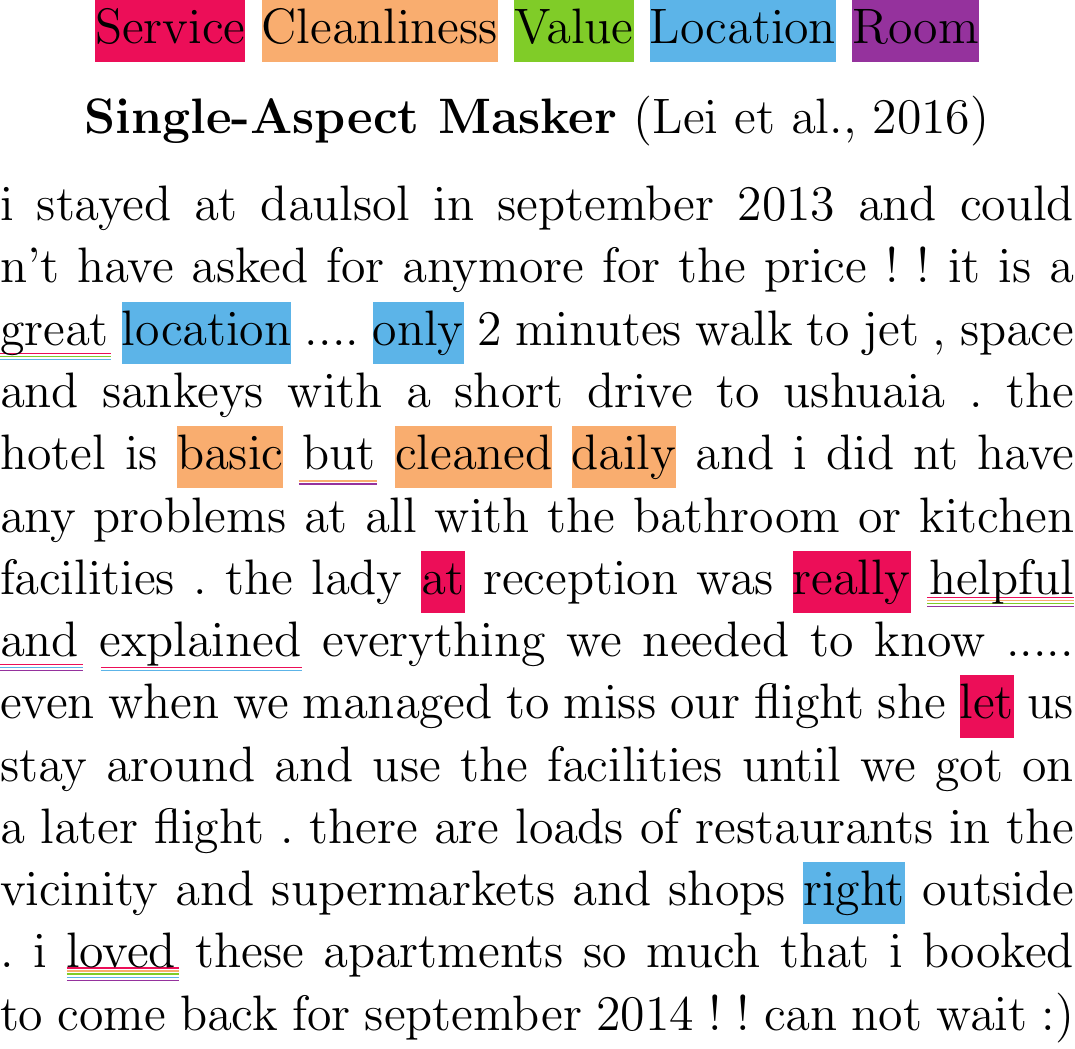} \\\\
     \includegraphics[width=0.475\textwidth]{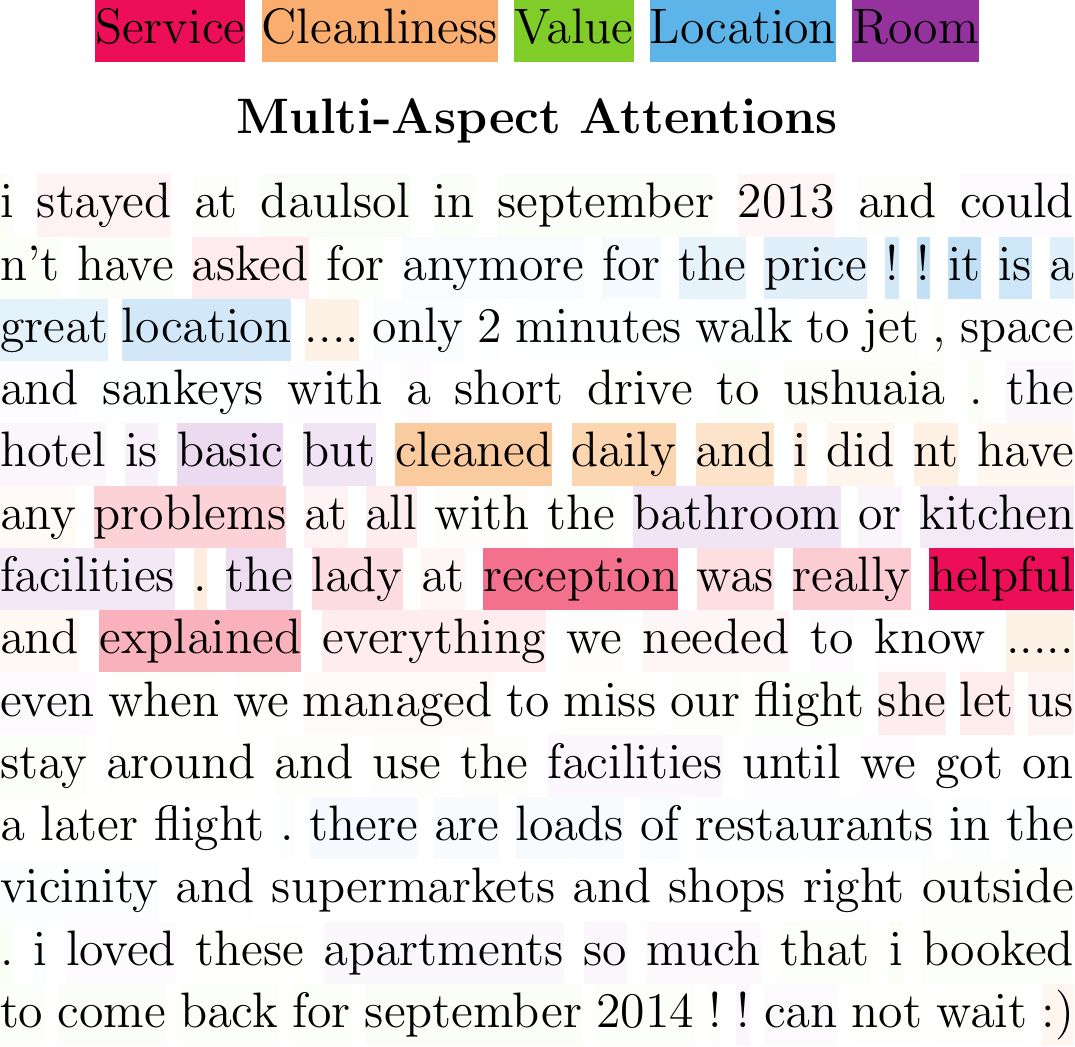} & 
     \includegraphics[width=0.475\textwidth]{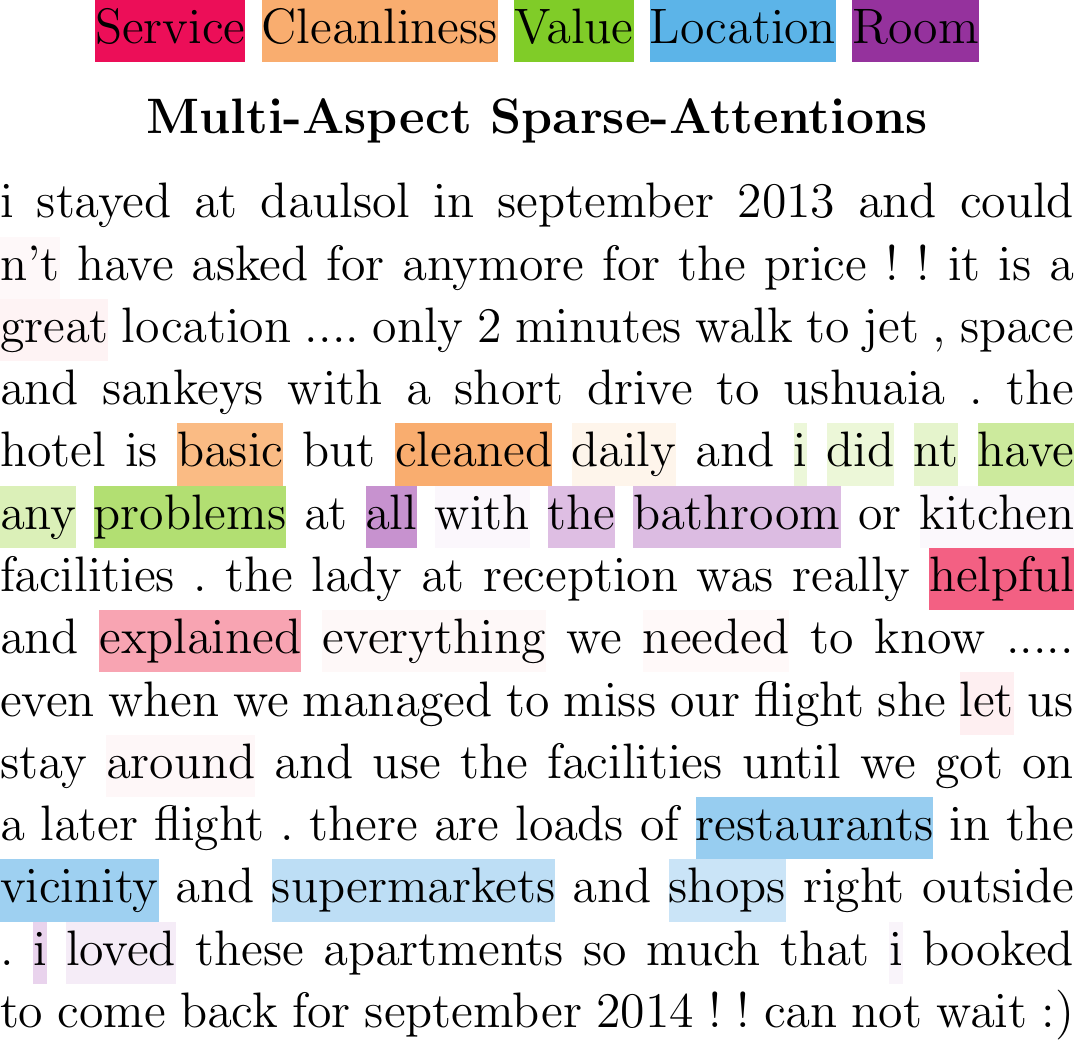}\\\\
\end{tabular}

\caption{\label{sample_hotel_0}A sample review from the \textbf{\textit{Hotel}} dataset, with computed masks from different methods. \textit{MTM} emphasizes consecutive words, identifies essential spans while having a small amount of noise. \textit{SAM} focuses on certain specific words and spans, but labels are ambiguous. The \textit{MAA} model highlights many words, ignores a few crucial keyphrases, but labels are noisy when the confidence is low. \textit{MASA} provides noisier tags than \textit{MAA}.}
\end{figure}

\begin{figure}[!htb]
\centering

\begin{tabular}{@{}c@{\makebox[0.25cm]{ }}c@{}}
     \includegraphics[width=0.475\textwidth,height=10cm]{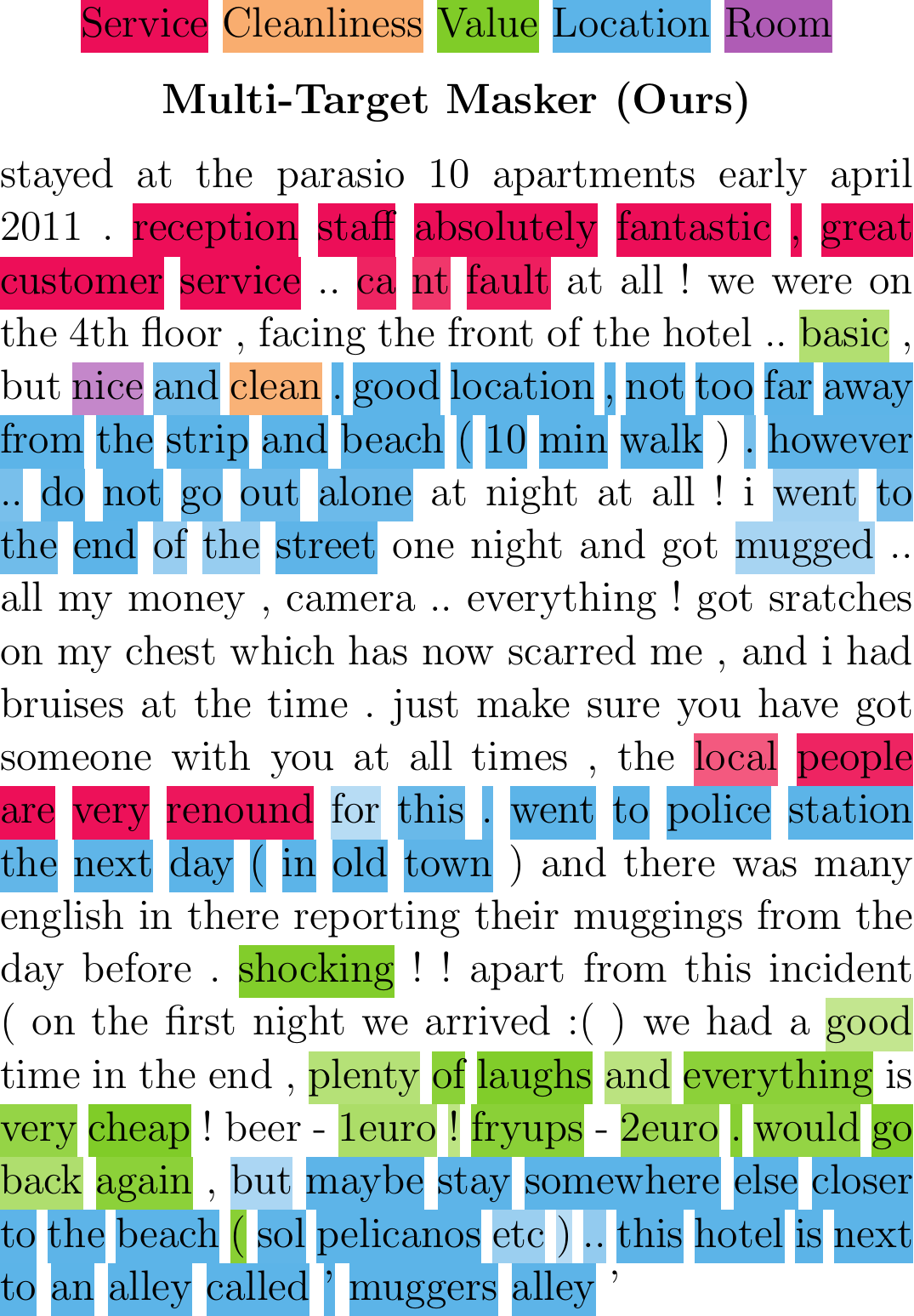} &
     \includegraphics[width=0.475\textwidth,height=10cm]{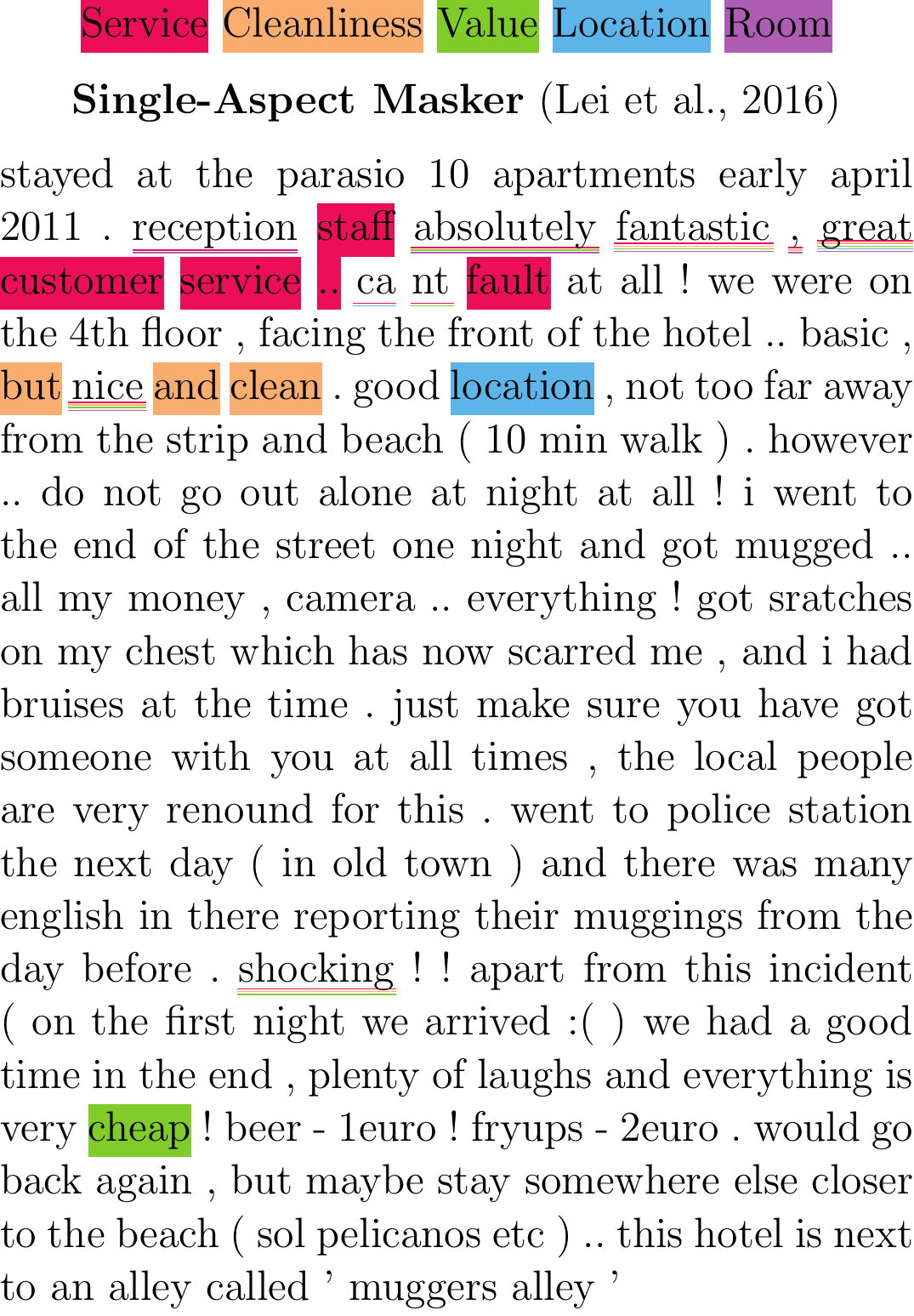} \\\\
     \includegraphics[width=0.475\textwidth,height=10cm]{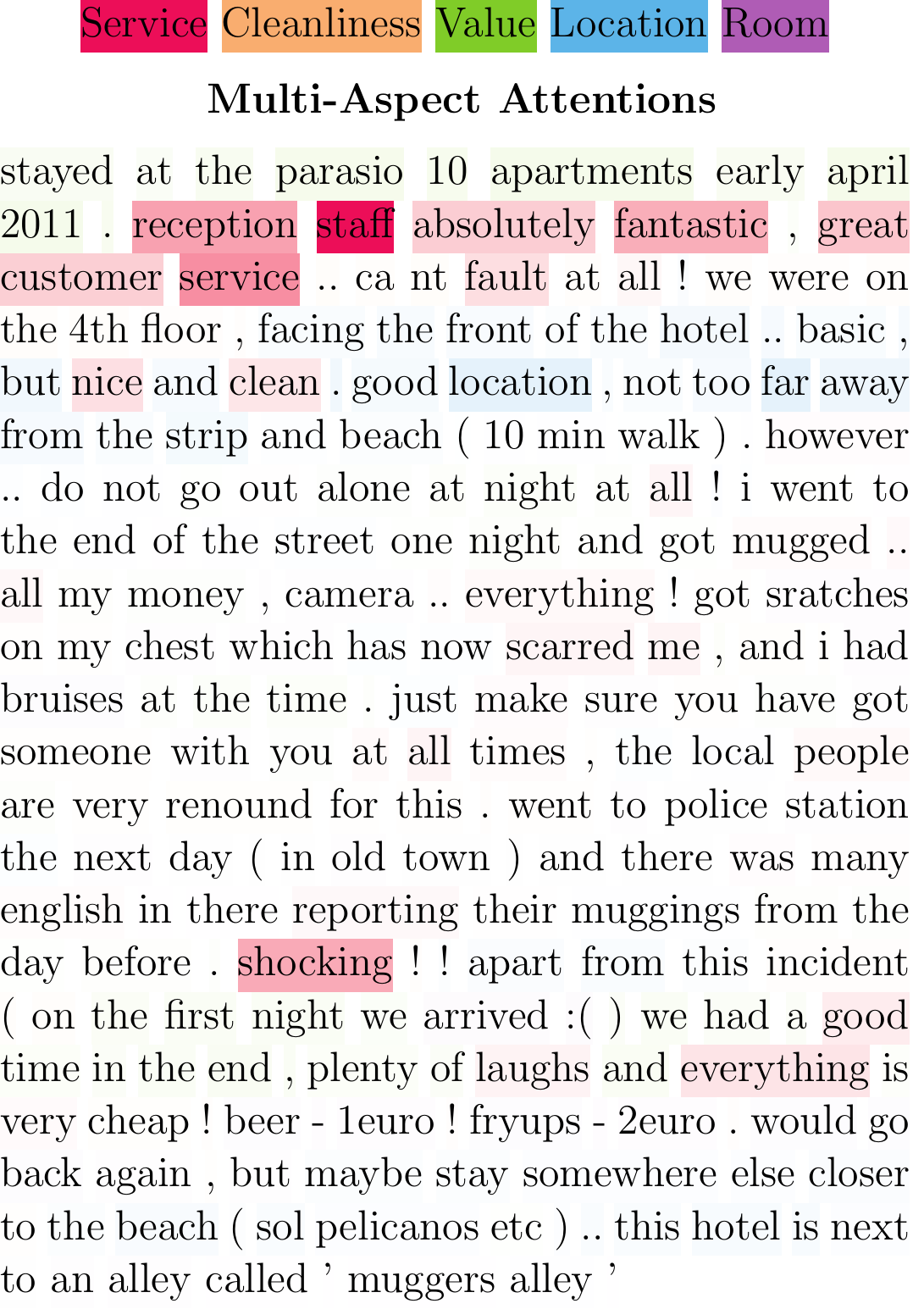} &
     \includegraphics[width=0.475\textwidth,height=10cm]{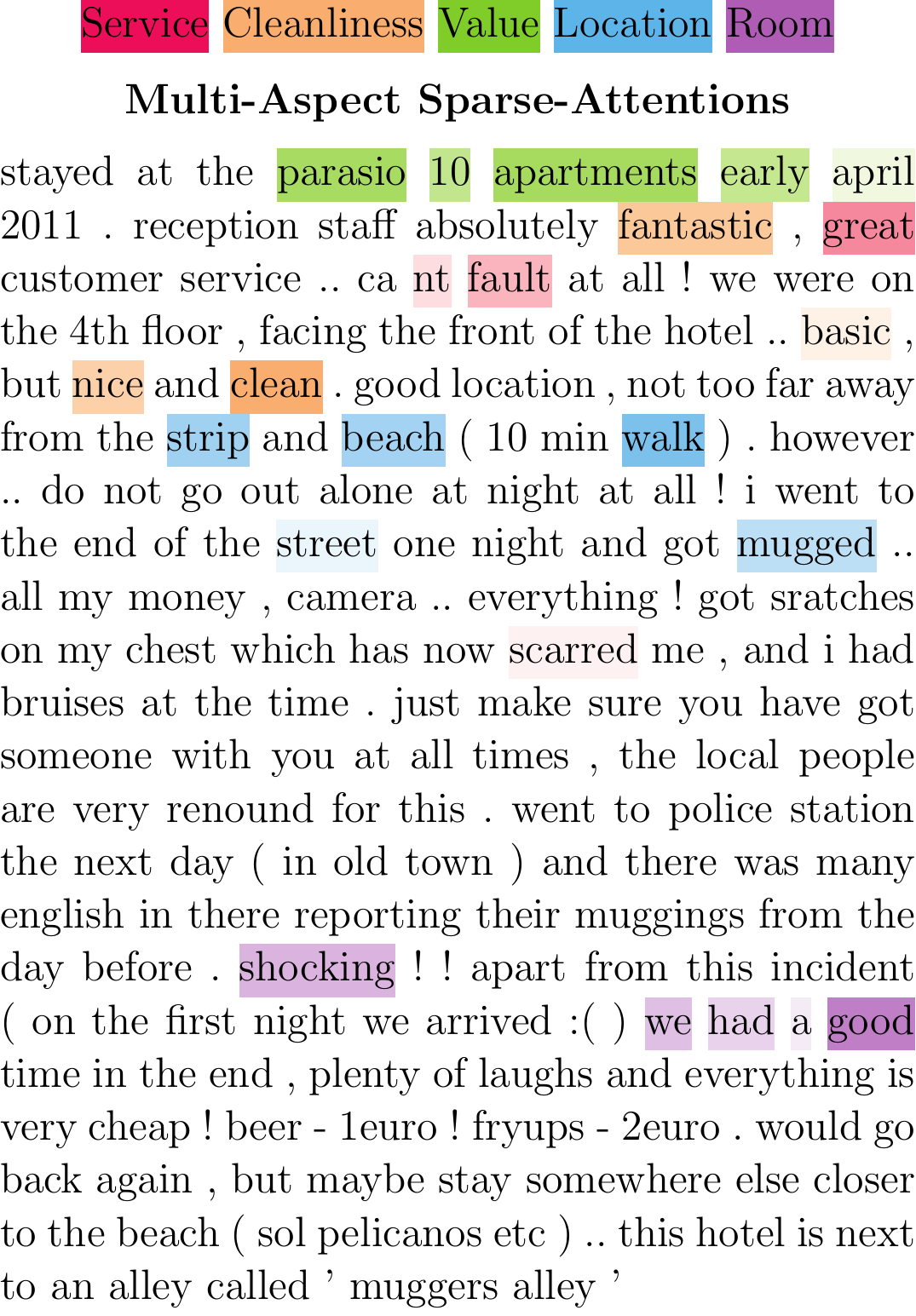}\\\\
\end{tabular}

\caption{\label{sample_hotel_1}A sample review from the \textbf{\textit{Hotel}} dataset, with computed masks from different methods. Our \textit{MTM} model finds most of the crucial span of words with a small amount of noise. \textit{SAM} lacks coverage but identifies words where half are correctly tags and the others ambiguous. \textit{MAA} partially correctly highlights words for the aspects \textit{Service}, \textit{Location}, and \textit{Value} while missing out on the aspect \textit{Cleanliness}. \textit{MASA} confidently finds a few important words.}
\end{figure}

\begin{figure}[!htb]
\centering
\begin{tabular}{@{}c@{\makebox[0.25cm]{ }}c@{}}
    \includegraphics[width=0.475\textwidth]{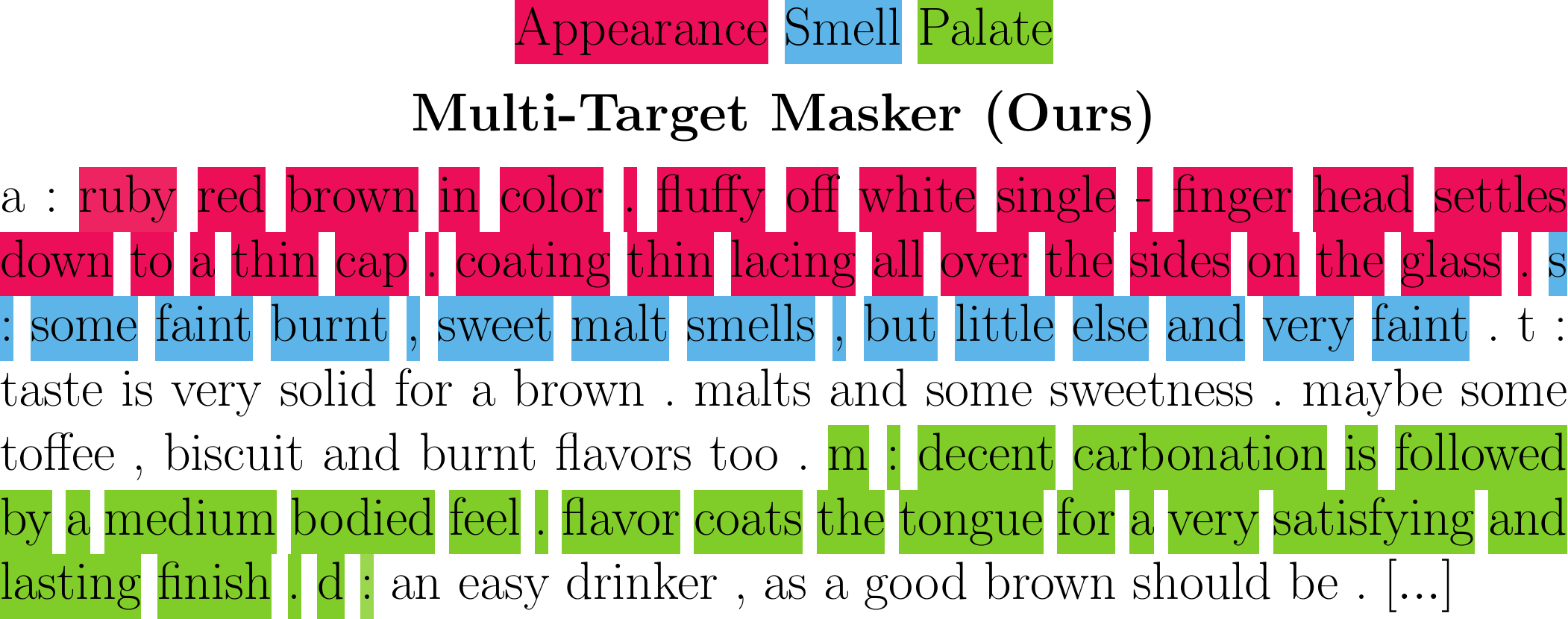} & 
    \includegraphics[width=0.475\textwidth]{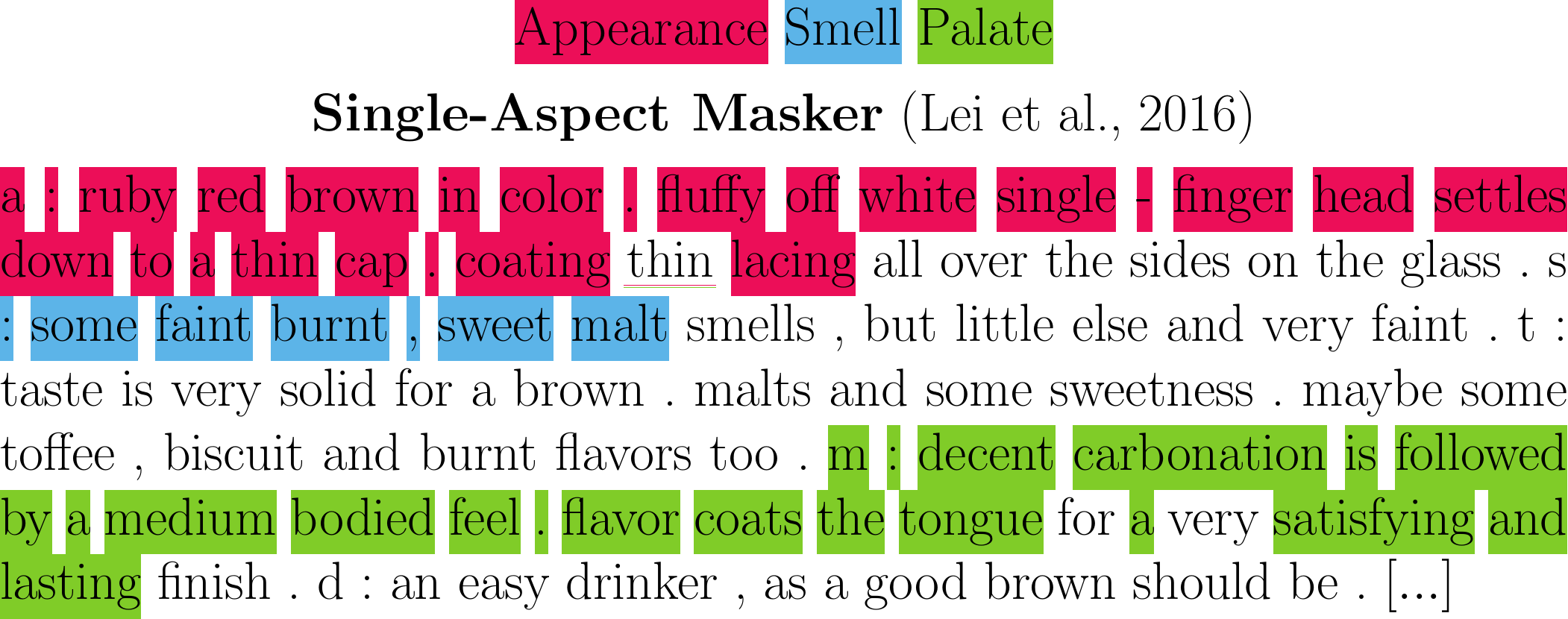} \\\\
    \includegraphics[width=0.475\textwidth]{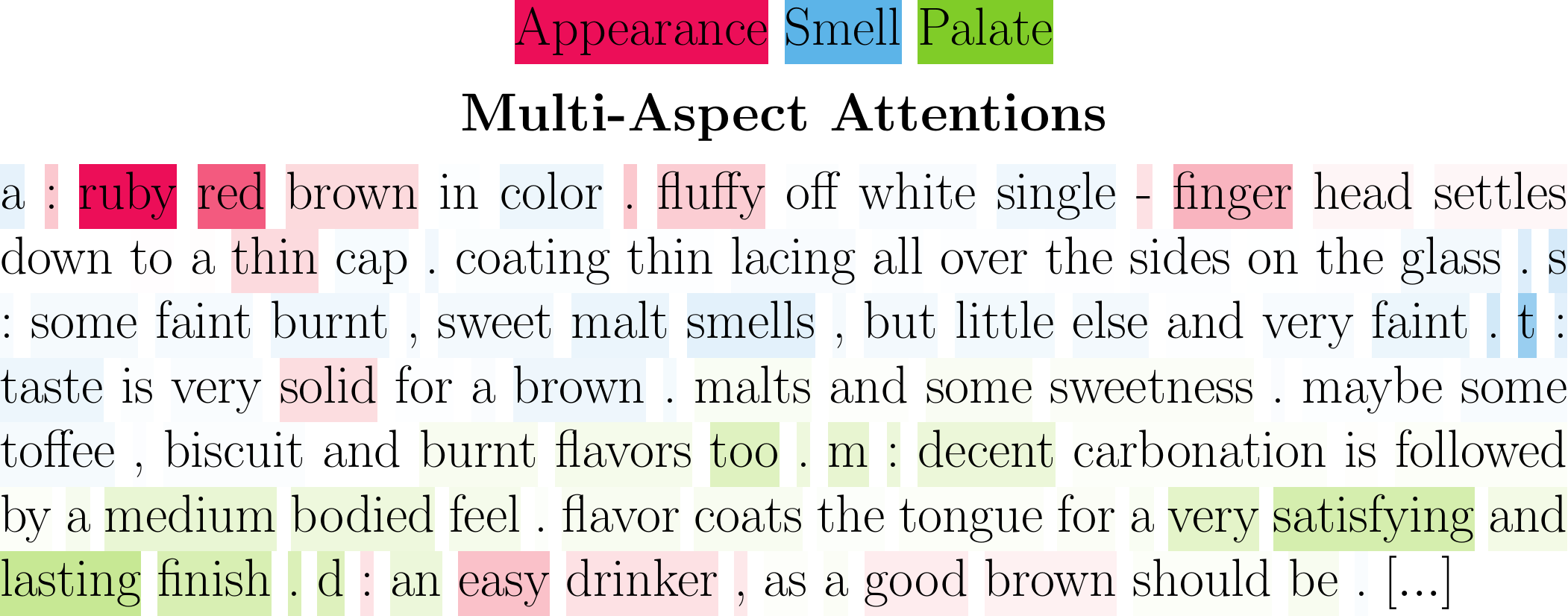} &
    \includegraphics[width=0.475\textwidth]{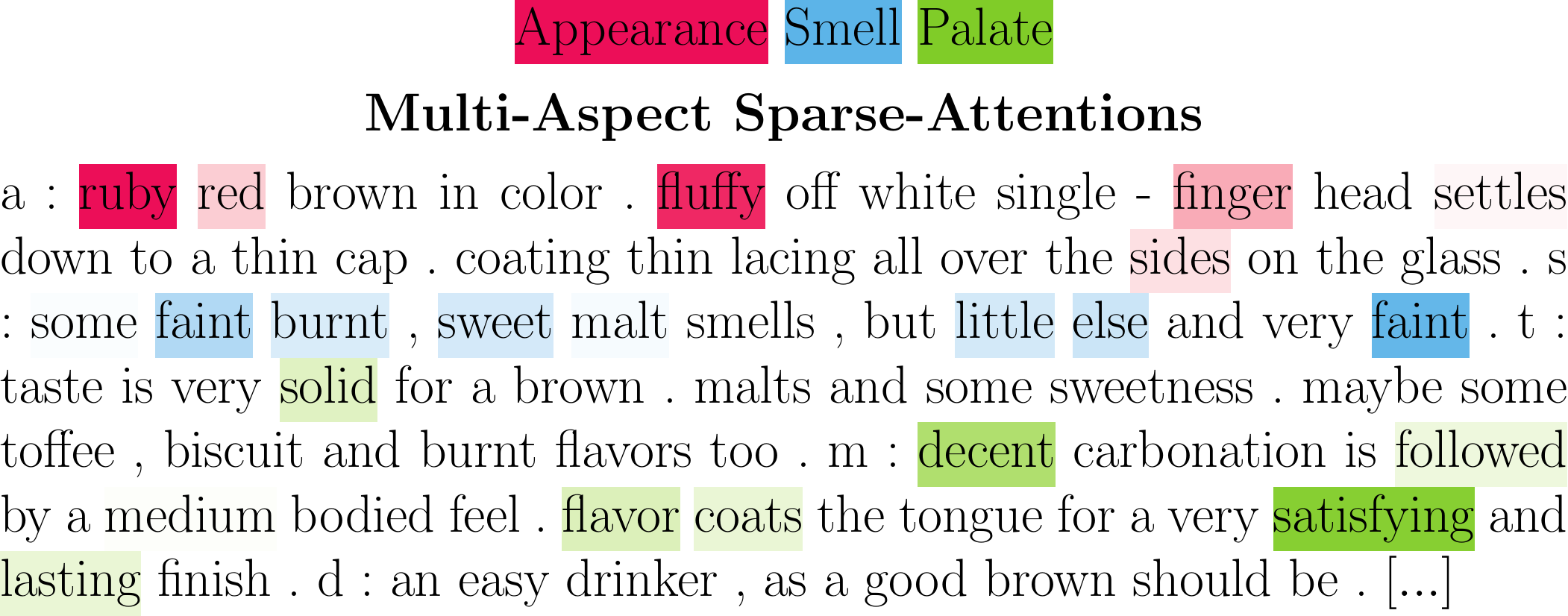}\\\\
\end{tabular}

\caption{\label{sample_beer_0}A sample review from the \textbf{\textit{decorrelated}} \textit{Beer} dataset, with computed masks from different methods. Our model \textit{MTM} highlights all the words corresponding to the aspects. \textit{SAM} only highlights the most crucial information, but some words are missing out and one is ambiguous. \textit{MAA} and \textit{MASA} fail to identify most of the words related to the aspect \textit{Appearance}, and only a few words have high confidence, resulting in noisy labeling. Additionally, \textit{MAA} considers words belonging to the aspect \textit{Taste} whereas this dataset does not include it in the aspect set (because it has a high correlation with other rating scores).}
\end{figure}

\begin{figure}[!htb]

\centering
\begin{tabular}{@{}c@{\makebox[0.25cm]{ }}c@{}}
    \includegraphics[width=0.475\textwidth]{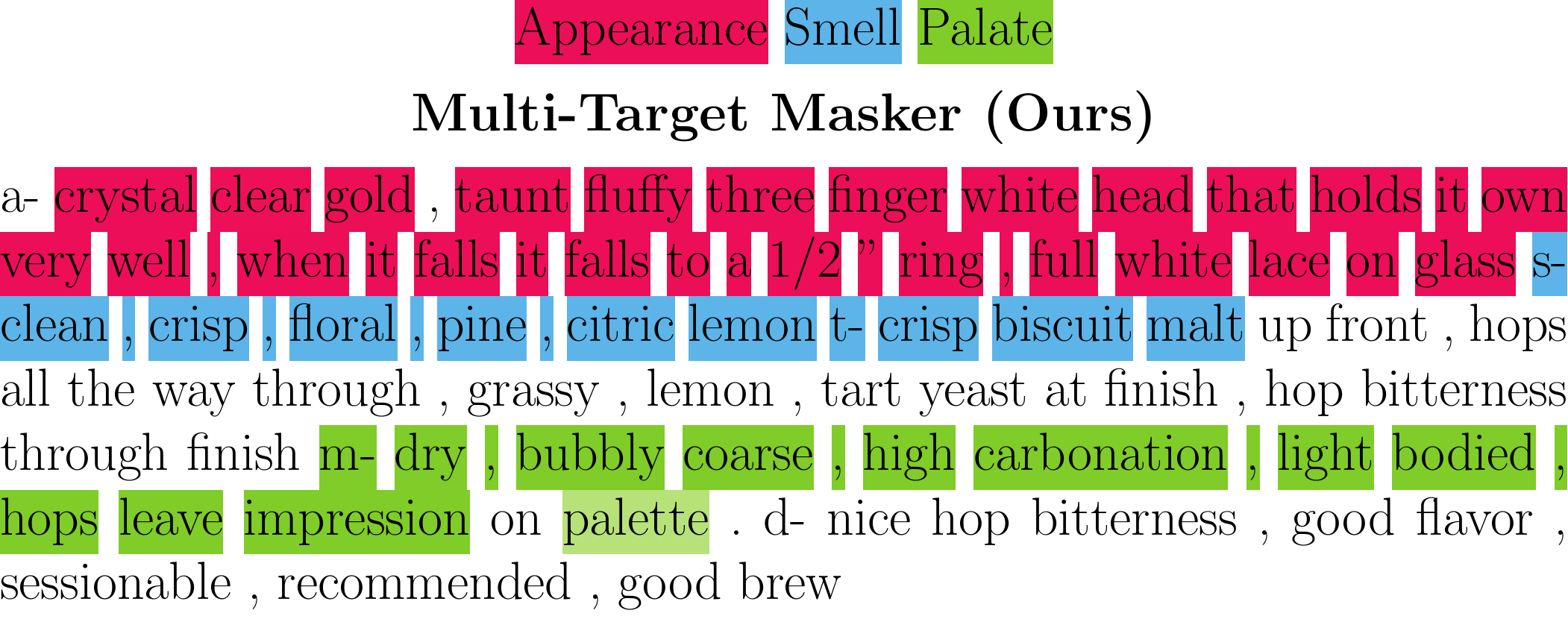} & 
    \includegraphics[width=0.475\textwidth]{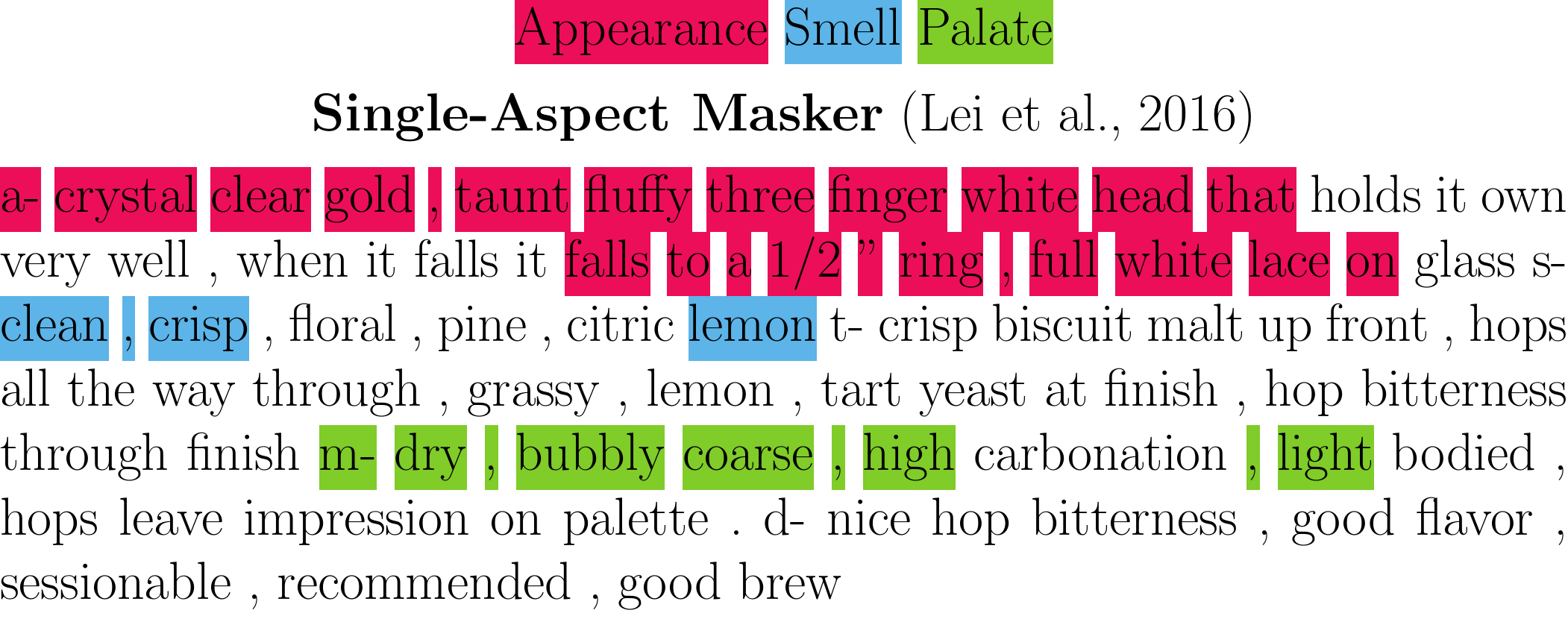} \\\\
    \includegraphics[width=0.475\textwidth]{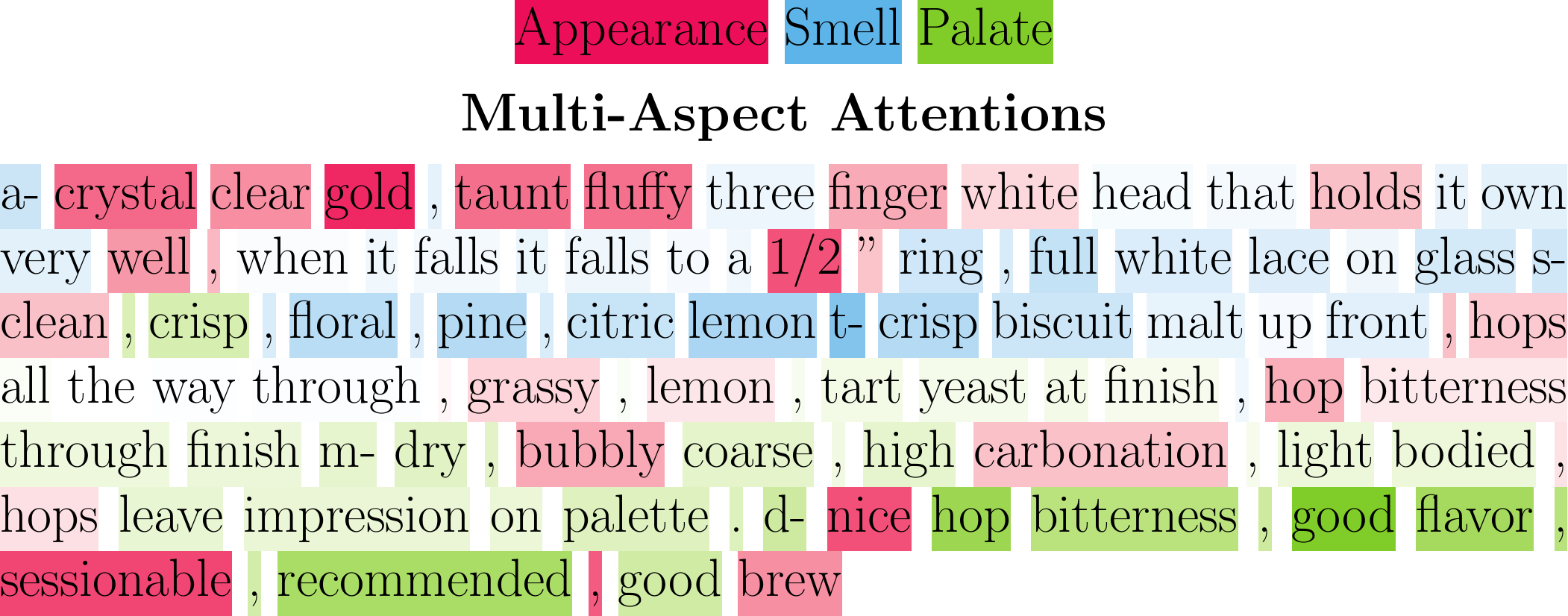} &
    \includegraphics[width=0.475\textwidth]{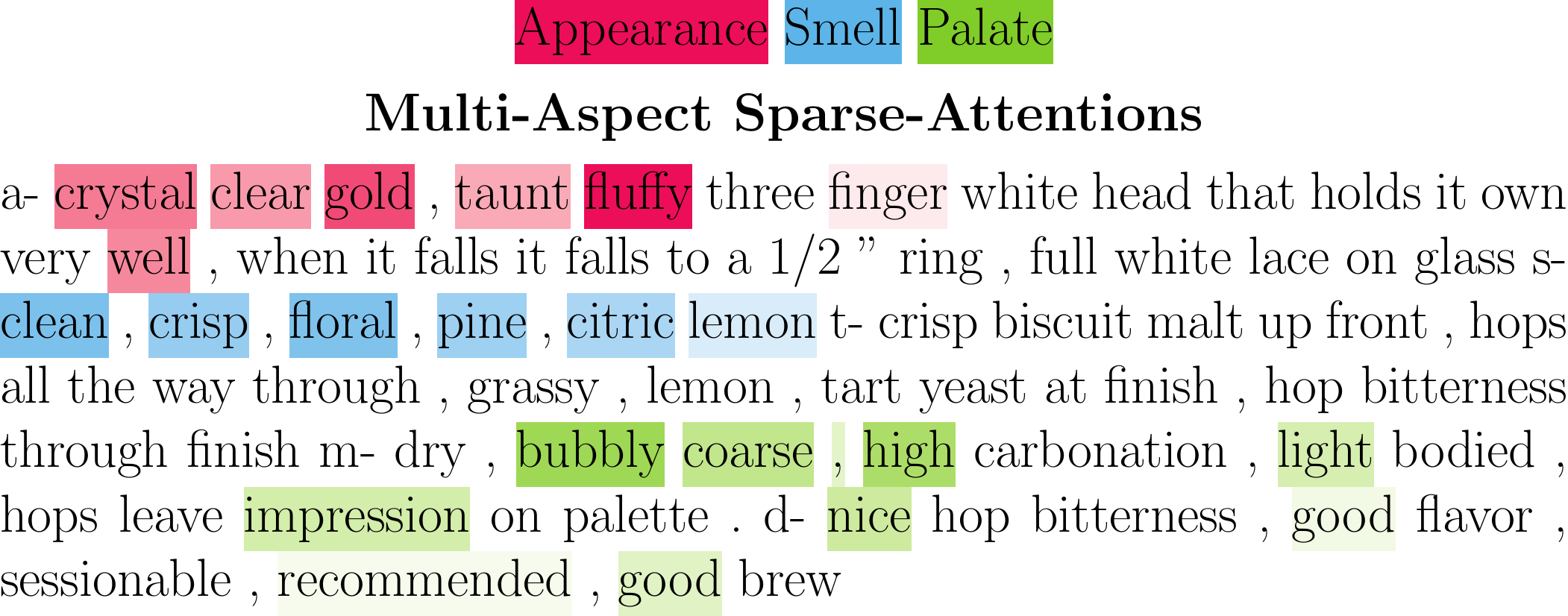}\\\\
\end{tabular}

\caption{\label{sample_beer_1}A sample review from the \textbf{\textit{decorrelated}} \textit{Beer} dataset, with computed masks from different methods. \textit{MTM} finds the exact parts corresponding to the aspect \textit{Appearance} and \textit{Palate} while covering most of the aspect \textit{Smell}. \textit{SAM} identifies key-information without any ambiguity, but lacks coverage. \textit{MAA} highlights confidently nearly all the words while having some noise for the aspect \textit{Appearance}. \textit{MASA} selects confidently only most predictive words.}
\end{figure}

We randomly sampled reviews from each dataset and computed the masks and attentions of four models: our Multi-Target Masker (\textit{MTM}), the Single-Aspect Masker (\textit{SAM}) \citep{lei-etal-2016-rationalizing}, and two attention models with additive and sparse attention, called Multi-Aspect Attentions (\textit{MAA}) and Multi-Aspect Sparse-Attentions (\textit{MASA}), respectively (see Section~\ref{sec_baselines}). Each color represents an aspect and the shade its confidence. All models generate soft attentions or masks besides SAM, which does hard masking. Samples for the \textit{Beer}, \textit{Hotel}, and \textit{Decorrelated Beer} datasets are shown in Figures~\ref{sample_full_beer_0}, \ref{sample_full_beer_1}, \ref{sample_hotel_0}, \ref{sample_hotel_1}, \ref{sample_beer_0} and \ref{sample_beer_1} respectively.

\label{full_beer_sample}

\label{hotel_samples}

\clearpage

\section{Concepts}
\label{app_samples}

\begin{figure}[!t]
\centering
\hspace*{-0.65cm}
\begin{tabular}{@{}c@{}c@{}c@{}}
  \multicolumn{3}{c}{\includegraphics[width=0.45\textwidth,height=.4cm]{main/ACL2021/Figures/samples/legend.pdf}}\\
   ConRAT (Ours) & InvRAT \cite{chang2020invariant} & RNP \cite{lei-etal-2016-rationalizing}\\
     \includegraphics[width=0.36\textwidth,height=5.1cm]{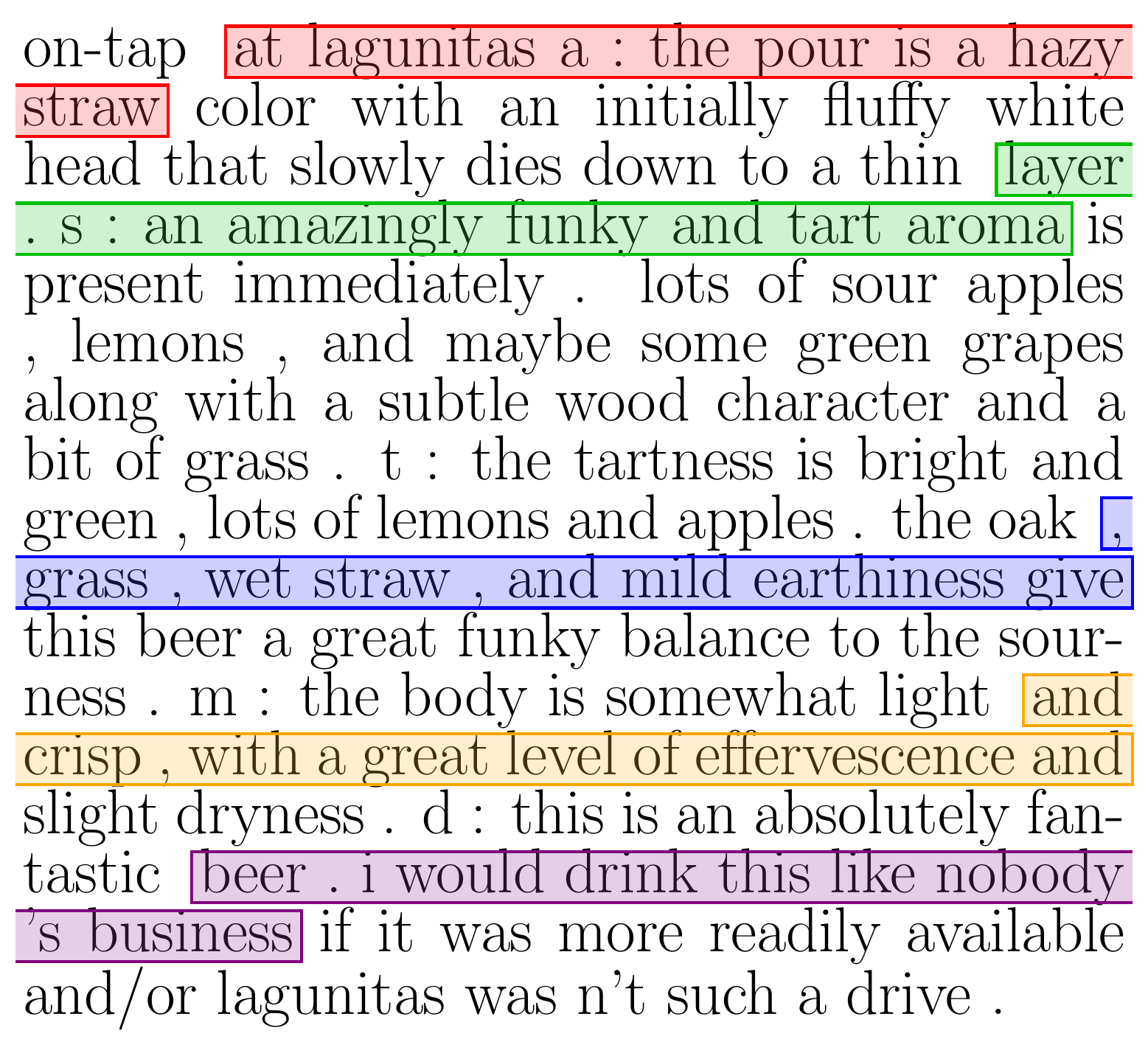}&     \includegraphics[width=0.36\textwidth,height=5.1cm]{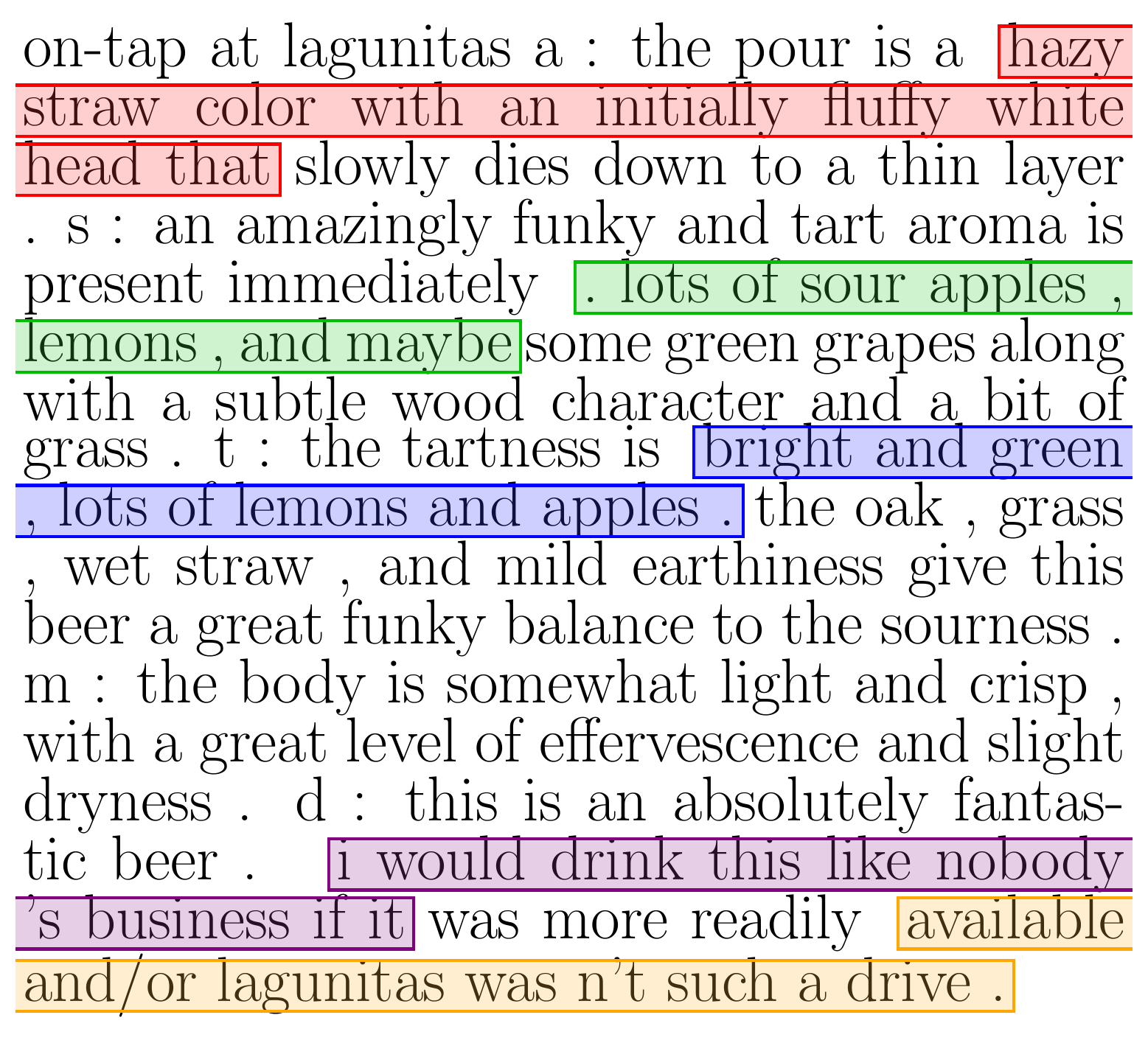}&     \includegraphics[width=0.36\textwidth,height=5.1cm]{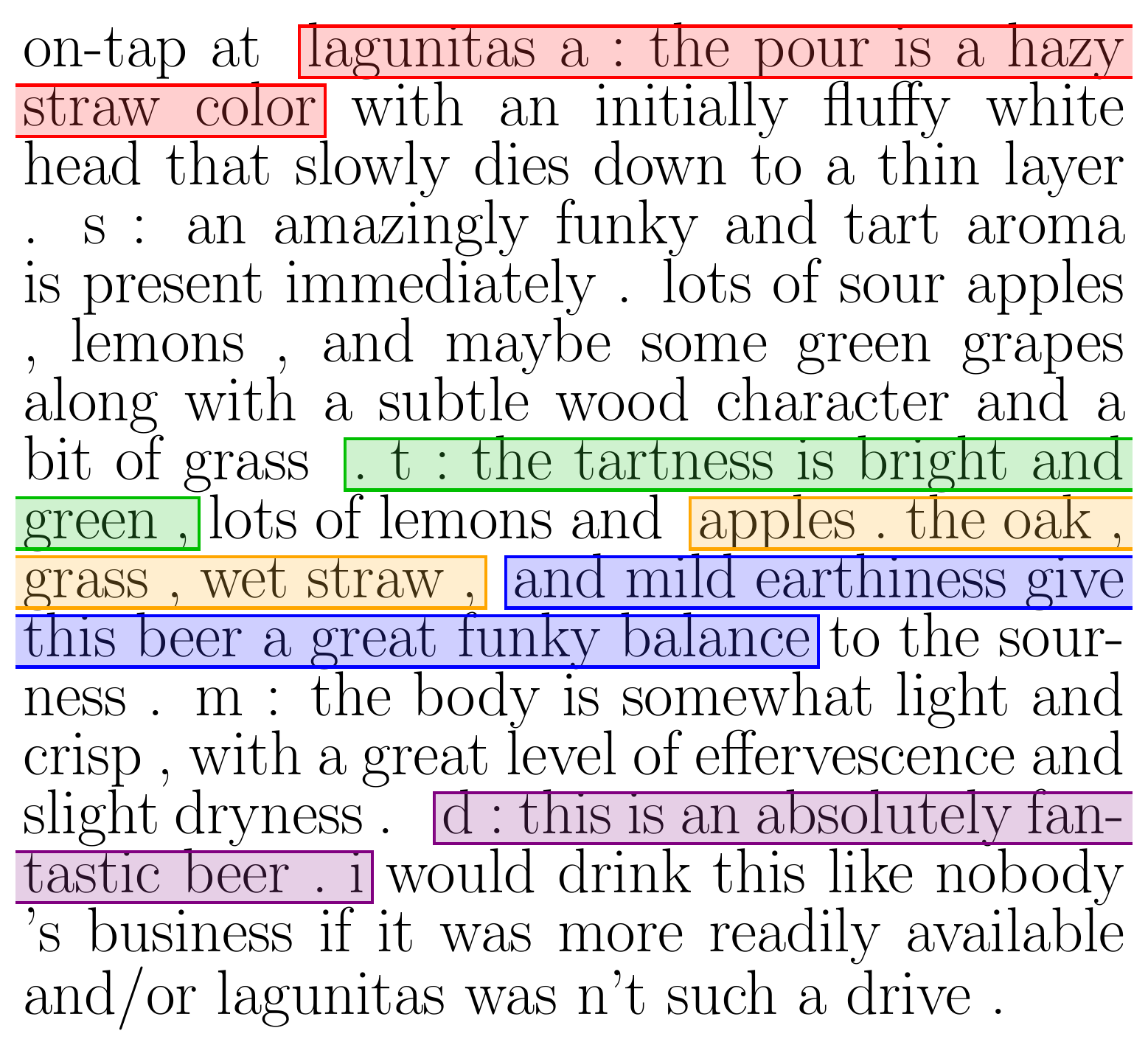}
  \\
   ConRAT (Ours) & InvRAT \cite{chang2020invariant} & RNP \cite{lei-etal-2016-rationalizing}\\
     \includegraphics[width=0.36\textwidth,height=3.5cm]{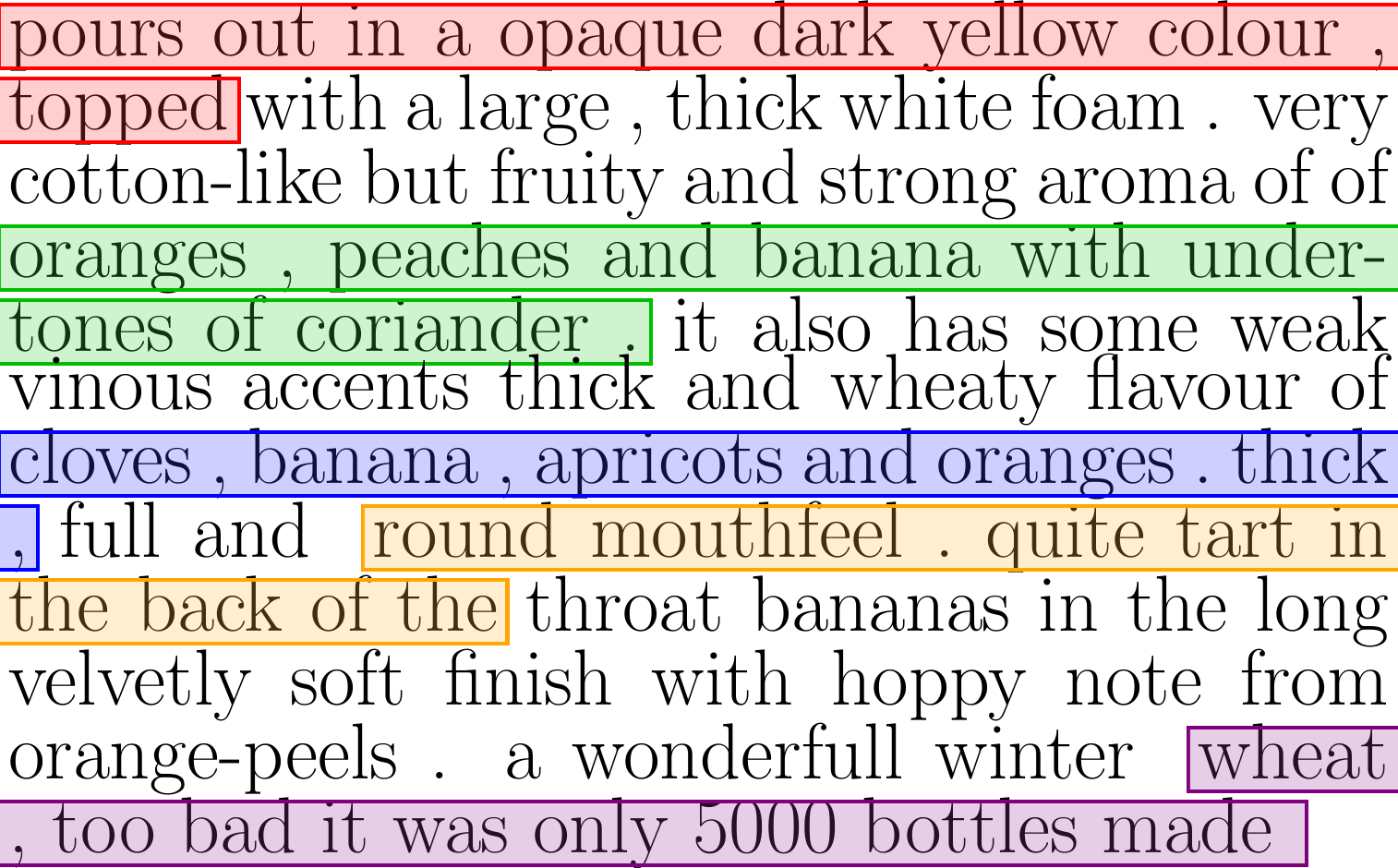}&     \includegraphics[width=0.36\textwidth,height=3.5cm]{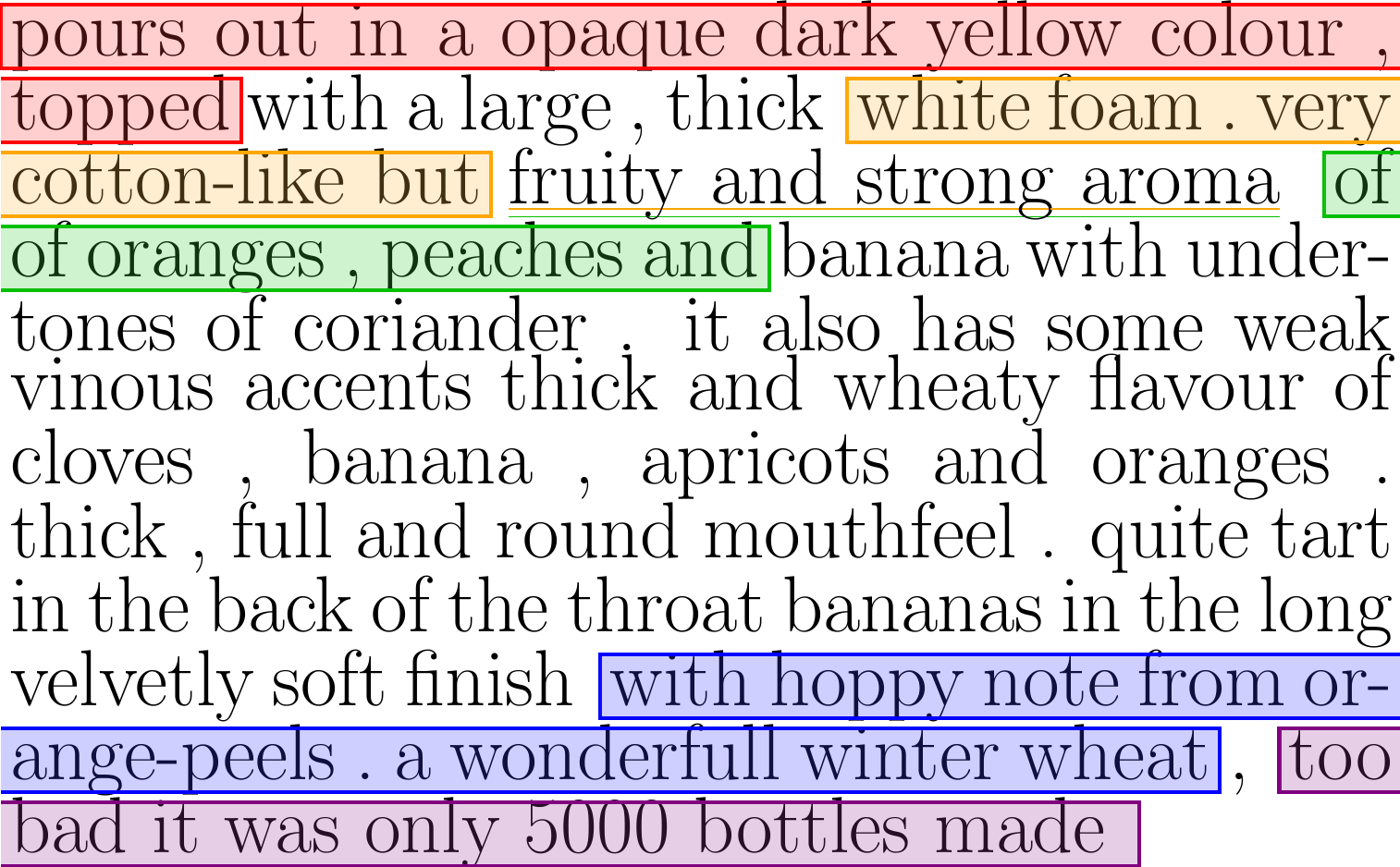}&     \includegraphics[width=0.36\textwidth,height=3.5cm]{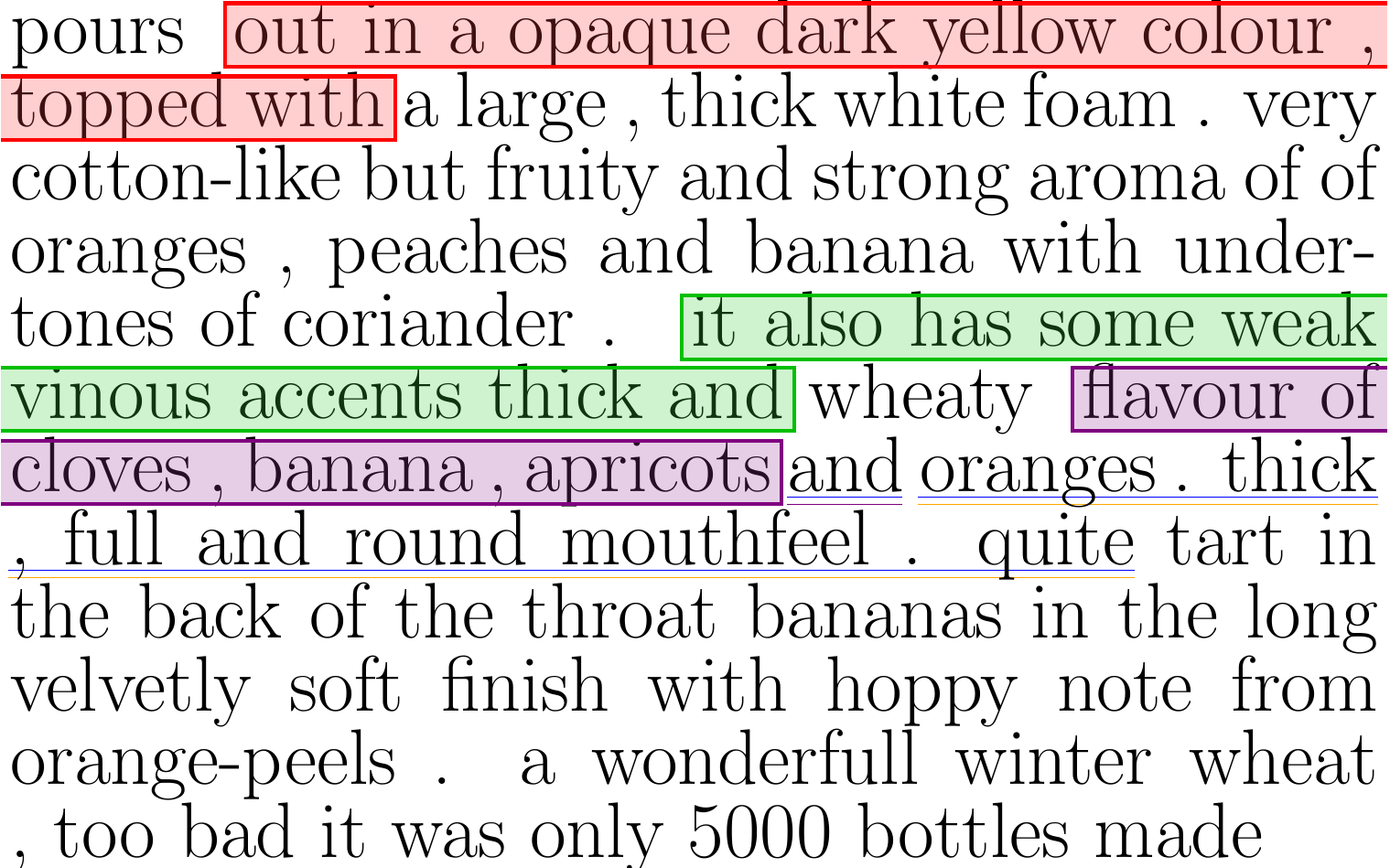}
\end{tabular}
\caption{\label{app_sample_10_2}Examples of generated rationales with $\ell=10$ for a beer review. \underline{Underline} highlights ambiguities.}
\end{figure}

\begin{figure}[!t]
\centering
\hspace*{-0.65cm}
\begin{tabular}{@{}c@{}c@{}c@{}}
\\
   ConRAT (Ours) & InvRAT \cite{chang2020invariant} & RNP \cite{lei-etal-2016-rationalizing}\\
     \includegraphics[width=0.36\textwidth,height=3.5cm]{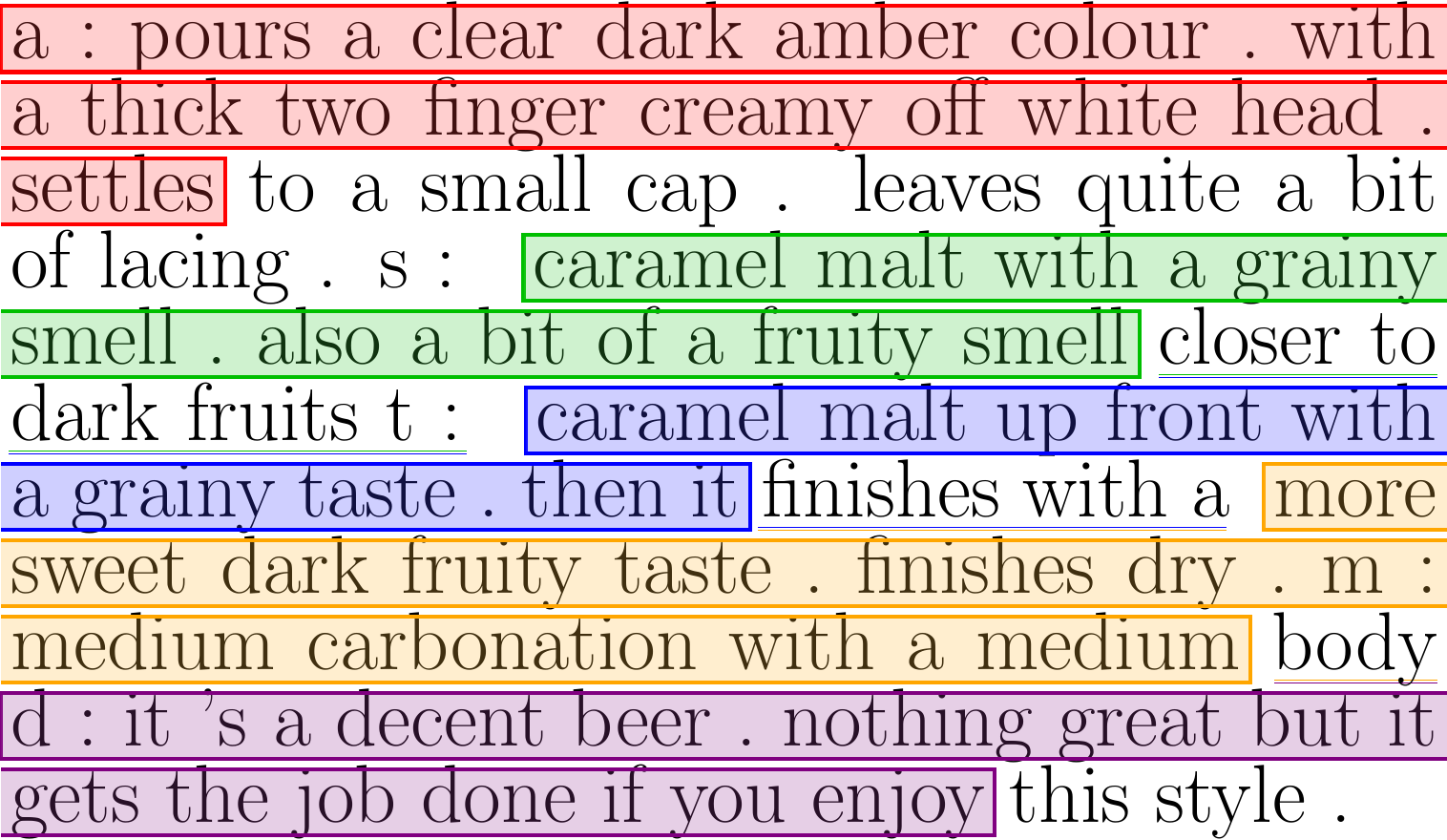}&     \includegraphics[width=0.36\textwidth,height=3.5cm]{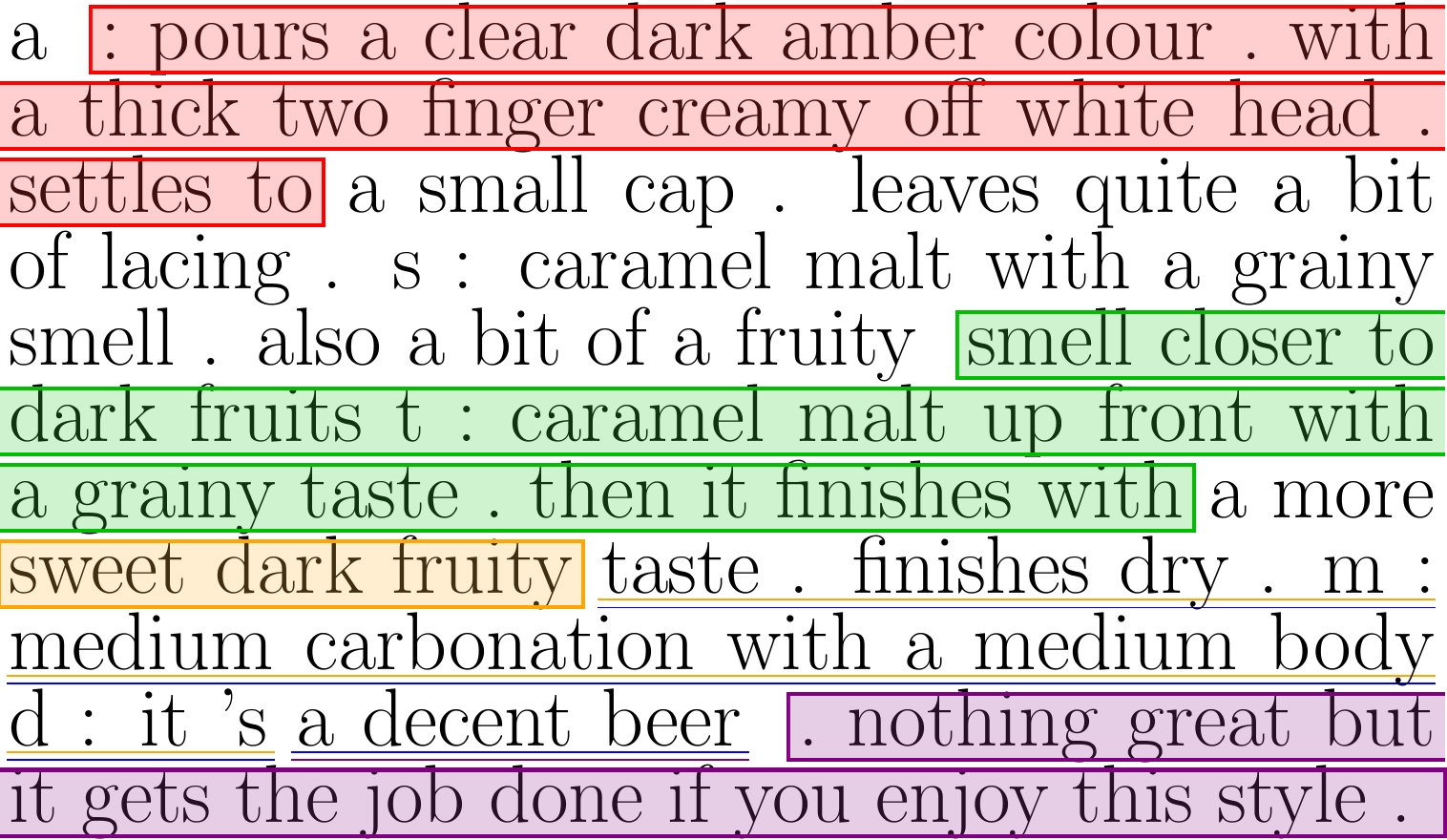}&     \includegraphics[width=0.36\textwidth,height=3.5cm]{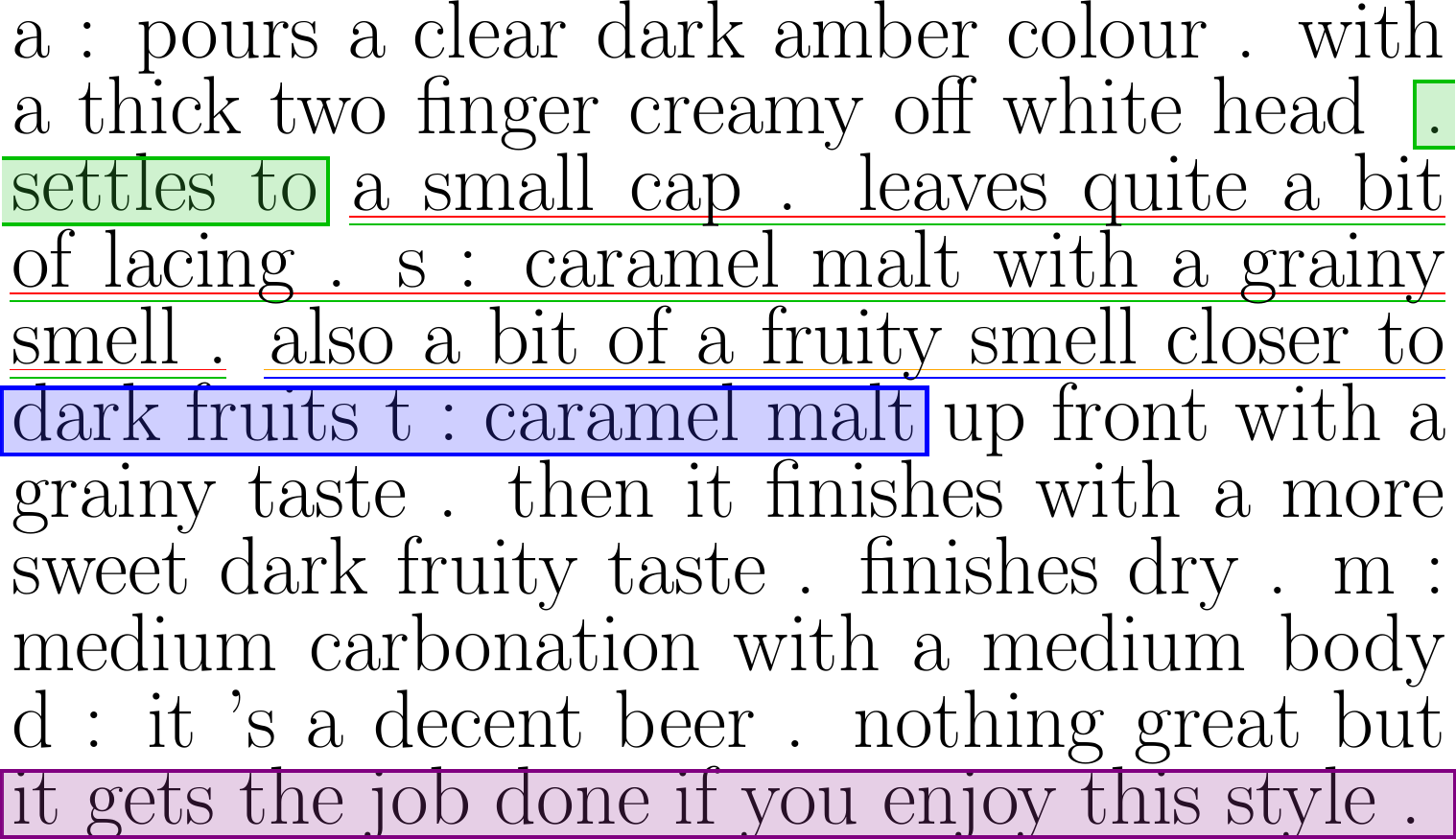}
\\

   ConRAT (Ours) & InvRAT \cite{chang2020invariant} & RNP \cite{lei-etal-2016-rationalizing}\\
     \includegraphics[width=0.36\textwidth,height=4.5cm]{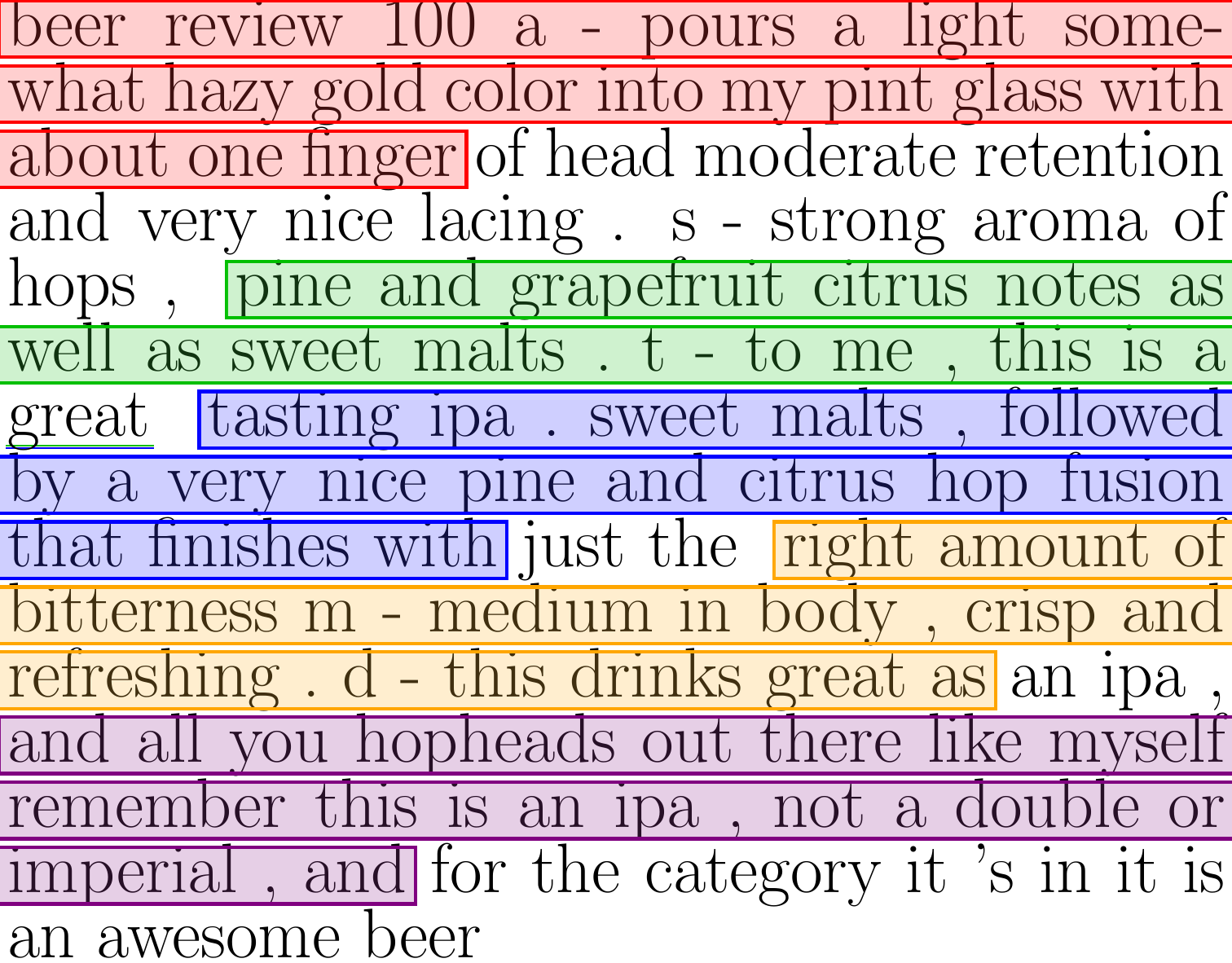}&     \includegraphics[width=0.36\textwidth,height=4.5cm]{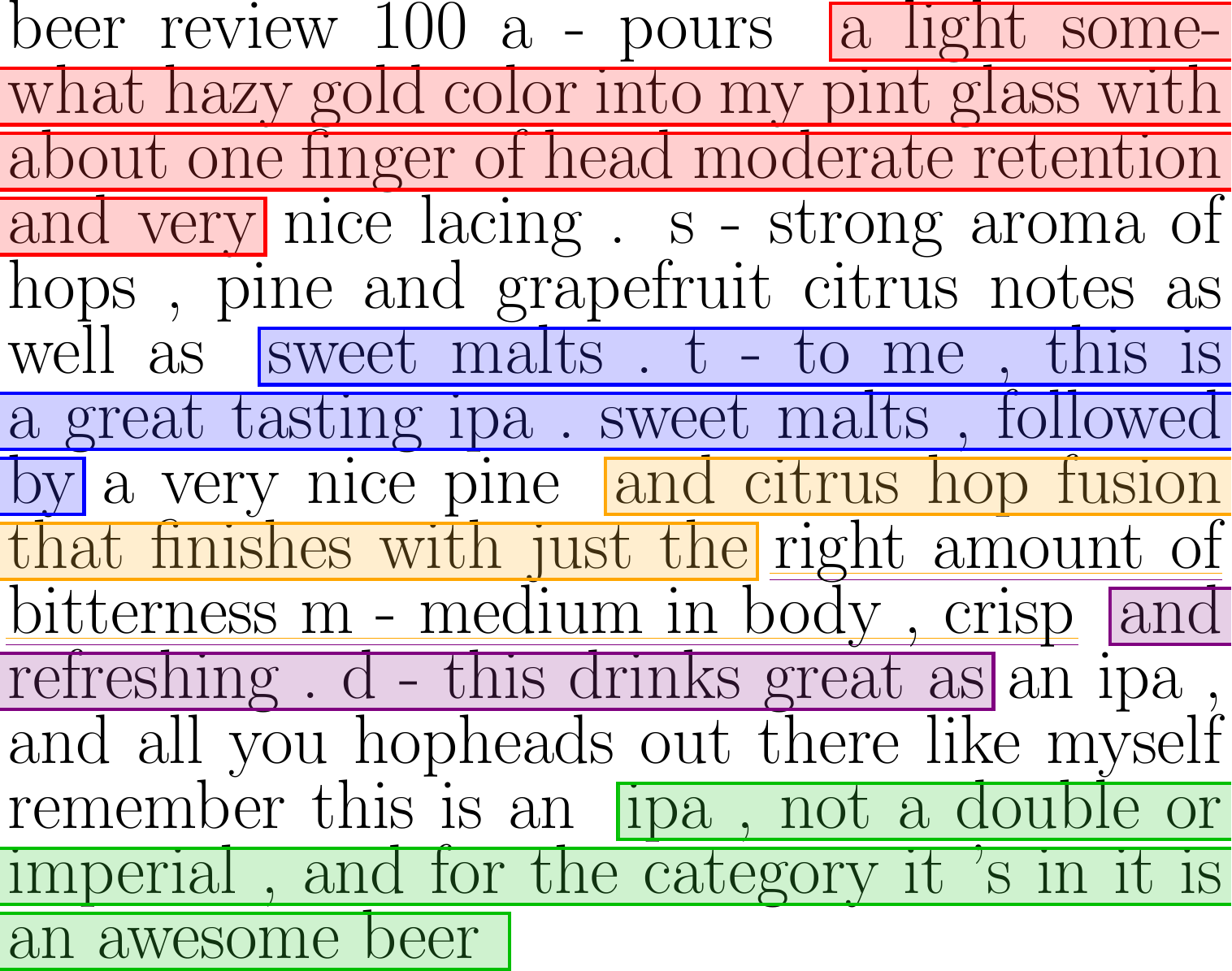}&     \includegraphics[width=0.36\textwidth,height=4.5cm]{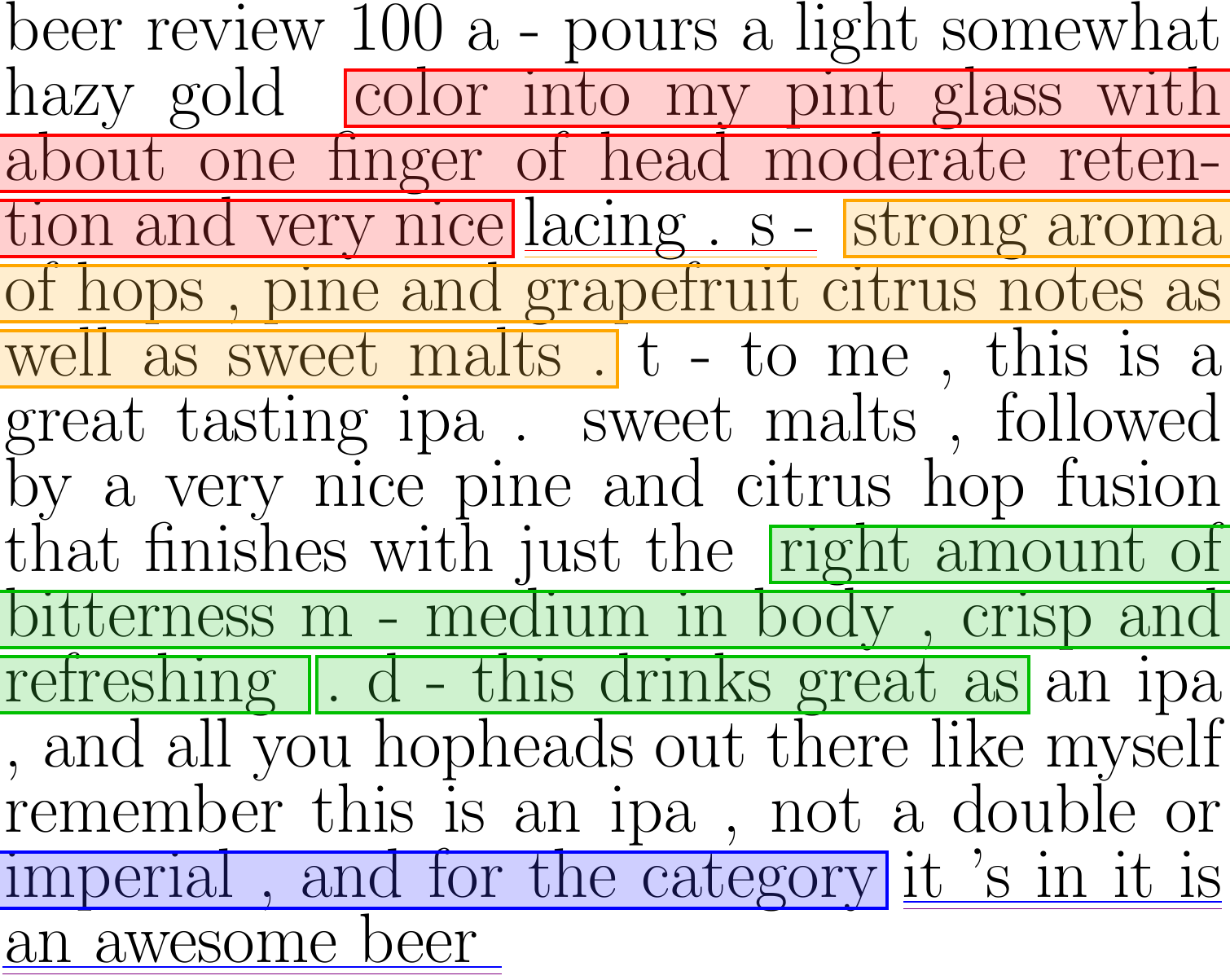}\\
\end{tabular}
\caption{\label{app_sample_20_2}Examples of generated rationales with $\ell=20$ for a beer review. \underline{Underline} highlights ambiguities.}
\end{figure}

Additional samples of generated rationales are shown in Figure~\ref{app_sample_10_2} and \ref{app_sample_20_2}. We can observe that baselines suffer from spurious correlations: the rationale for the aspect Aroma, Palate, and Taste are often exchanged, or several rationales pick the same text snippets. On the other hand, ConRAT finds better concepts while only trained on the overall aspect label. As it has been shown in prior work \citep{lei-etal-2016-rationalizing,chang2020invariant,antognini2019multi} rationale methods suffer from the high correlation between rating scores because each model is trained independently for each aspect. Therefore, they rely on the assumption that the data have low internal correlations, which does not reflect the real data distribution. By contrast, ConRAT alleviates this problem be finding all concepts in one training.

\chapter{Human Evaluation Details}

We use Amazon's Mechanical Turk crowdsourcing platform to recruit human annotators to evaluate the quality of extracted justifications and the generated justifications produced by each model. To ensure high-quality of the collected data, we restricted the pool to native English speakers from the U.S., U.K., Canada, or Australia. Additionally, we set the worker requirements at a 98\% approval rate and more than 1,000 HITS.

\section{Understandability of Concepts}
\label{app_rq2_hum}

\begin{figure}[t]
\centering
\hspace*{-1.35cm}
\includegraphics[width=1.2\textwidth]{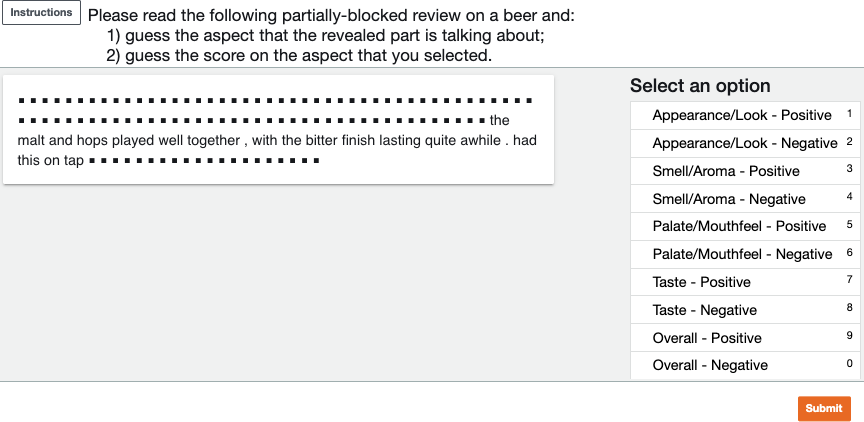}
\caption{\label{he_int_subj}Annotation platform for judging the quality of the concepts in the subjective evaluation on beer reviews.}
\end{figure}

The user interface used to judge the quality of the justifications extracted from different methods, in Section~\ref{rq2_exp_subj}, is shown in Figure~\ref{he_int_subj}.

\section{Justification Evaluation}
\label{app_hes}

\begin{figure}
\hspace*{-1.35cm}
\centering
\includegraphics[width=1.2\linewidth]{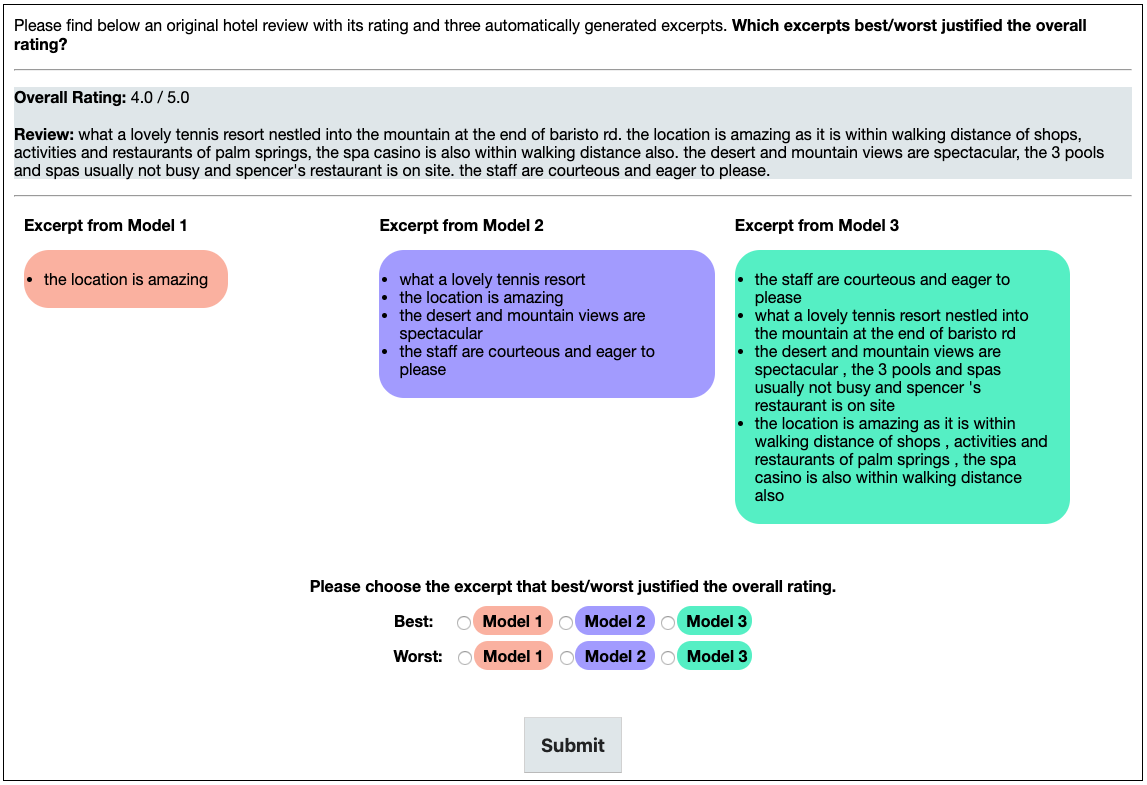}
\caption{\label{he_int_ex_ju}Annotation platform for judging the quality of extracted justifications from different methods. The justifications are shown in random order for each comparison. In this example, our method corresponds to the third model.}
\end{figure}

\begin{figure}
\hspace*{-1.35cm}
\centering
\includegraphics[width=1.2\linewidth]{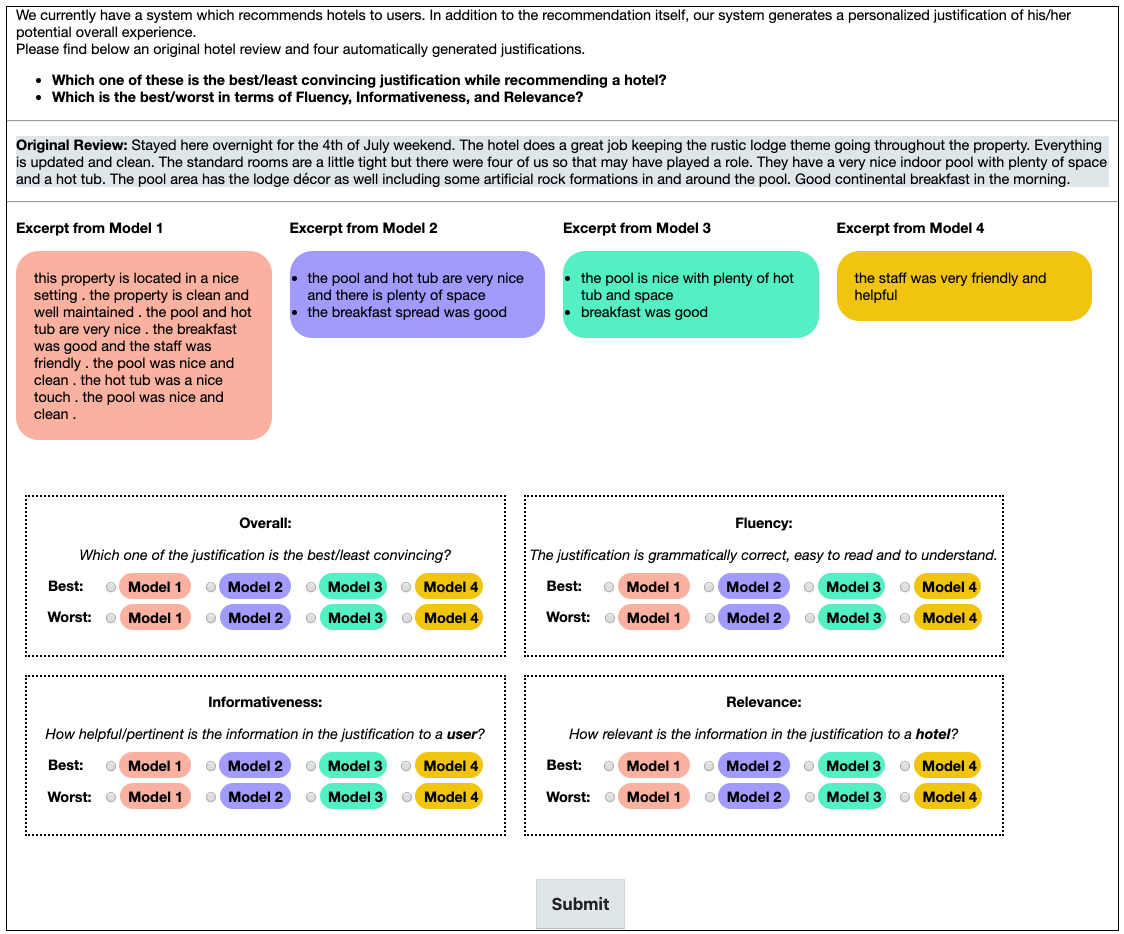}
\caption{\label{he_int_ge_ju}Annotation platform for judging the quality of generated justifications from different methods, on four dimensions. The justifications are shown in random order for each comparison. In this example, our method corresponds to the second~model.}
\end{figure}

The user interface used to judge the quality of the justifications extracted from different methods, in Section~\ref{sec_rq1_ijcai}, is shown in Figure~\ref{he_int_ex_ju}. Another human assessment evaluates the generated justifications (see Section~\ref{sec_rq2_ijcai}) on the four dimensions: \begin{enumerate}
    \item \textbf{Overall} measures the overall subjective quality;
    \item \textbf{Fluency} represents the readability;
    \item \textbf{Informativeness} indicates whether the justification contains information pertinent to the user;
    \item \textbf{Relevance} measures how relevant the information is to an item.
\end{enumerate} The interface is available in Figure~\ref{he_int_ge_ju}

\chapter{M\&Ms-VAE Derivation}
\label{app_derivation}
We provide the full derivation of the M\&Ms-VAE's evidence lower bound of Equation~\ref{eq_mm_2}.

\begin{align}
\log~p(\bru, \bku) &= \log \int_{\bzu} p_\Theta(\bru, \bku, \bzu) d\bzu\\
&= \log \int_{\bzu}  p_\Theta(\bru, \bku, \bzu) \frac{q_\Phi(\bzu | \bru, \bku)}{q_\Phi(\bzu | \bru, \bku)} d\bzu\\
&= \log \biggl( \mathbb{E}_{q_\Phi(\bzu | \bru, \bku)} \biggl[ \frac{p_\Theta(\bru, \bku, \bzu)}{q_\Phi(\bzu | \bru, \bku)} \biggr] \biggr)\\
&\begin{aligned}
\geq \mathbb{E}_{q_\Phi(\bzu | \bru, \bku)} \bigl[\log p_\Theta(\bru, \bku, \bzu)\bigr] &- \mathbb{E}_{q_\Phi(\bzu | \bru, \bku)} \bigl[\log q_\Phi(\bzu | \bru, \bku) \bigr]
\end{aligned}\\
&\begin{aligned}
\begin{split}
= \mathbb{E}_{q_\Phi(\bzu | \bru, \bku)} \bigl[\log p_\Theta(\bru, \bku | \bzu)\bigr] &+ \mathbb{E}_{q_\Phi(\bzu | \bru, \bku)} \bigl[\log p(\bzu) \bigr] \\ &- \mathbb{E}_{q_\Phi(\bzu | \bru, \bku)} \bigl[\log q_\Phi(\bzu | \bru, \bku) \bigr]\\
\end{split}
\end{aligned}\\
&\begin{aligned}
= \mathbb{E}_{q_\Phi(\bzu | \bru, \bku)} \bigl[\log p_\Theta(\bru, \bku | \bzu)\bigr] &- \textrm{D}_{\textrm{KL}} \bigl[ q_\Phi(\bzu | \bru, \bku) ~||~ p(\bzu) \bigr]\\
\end{aligned}\\
&\begin{aligned}
= \mathbb{E}_{q_\Phi(\bzu | \bru, \bku)} \bigl[\log p_{\Theta_r}(\bru | \bzu) + \log p_{\Theta_k}(\bku | \bzu)\bigr] &- \textrm{D}_{\textrm{KL}} \bigl[ q_\Phi(\bzu | \bru, \bku) ~||~ p(\bzu) \bigr]\\
\end{aligned}
\end{align}

\chapter{Reproducibility}

\section{Multi-Target Masker (MTM)}
\label{app:additional_training}
Most of the time, the model converges under $20$ epochs (maximum of $20$ and $3$ minutes per epoch for the \textit{Beer} and \textit{Hotel} dataset, respectively. The range of hyperparameters are the following for MTM (similar for other models).
\begin{itemize}
	\item Learning rate: $[0.001, 0.0005, 0.00075]$;
	\item Hidden size: $[50, 100, 200]$;
	\item Filter numbers (CNN): $[50, 100, 200]$;
	\item Bi-directional (LSTM): $[True, False]$;
	\item Dropout: $[0, 0.1, 0.2]$;
	\item Weight decay: $[0, 1e^{-6}, 1e^{-8}, 1e^{-10}]$;
	\item Gumbel Temperature $\tau$ :$[0.5, 0.8, 1.0, 1.2]$;
	\item $\lambda_{sel}$: $[0.01, 0.02, 0.03, 0.04, 0.05]$;
	\item $\lambda_p$: $[0.05, 0.06, 0.07, 0.08, 0.09, 0.1, 0.11, 0.12, 0.13, 0.14, 0.15]$;
	\item $\lambda_{cont}$: $[0.02, 0.04, 0.06, 0.08, 0.10]$;
\end{itemize}

\subsubsection{Hardware / Software}

\begin{itemize}
  \item \textbf{CPU}: 2x Intel Xeon E5-2680 v3, 2x 12 cores, 24 threads, 2.5 GHz, 30 MB cache;
  \item \textbf{RAM}: 16x16GB DDR4-2133;
  \item \textbf{GPU}: 1x Nvidia Titan X;
  \item \textbf{OS}: Ubuntu 18.04;
  \item \textbf{Software}: Python 3.6, PyTorch 1.3.1, CUDA 9.2.
\end{itemize}

\section{Concept-based Rationalizer (ConRAT)}
\label{app_training}

We tune all models on the dev set. We truncate all reviews to $320$ tokens for the beer dataset and $400$ tokens for Amazon reviews. We have operated a random search over $16$ trials. All baselines, except CAR, are tuned for each aspect ($80$ trials in total for the five aspects). We chose the models achieving the lowest validation accuracy. Most of the time, all models converged under $30$ epochs. The range of hyperparameters are the following for ConRAT (similar for other models):

\begin{itemize}
  \item Learning rate: $[0.0005, 0.00075, 0.001]$;
  \item Batch size: $[128]$;
  \item Hidden size: $[256]$;
  \item $\lambda_D$: $[0.01, 0.05, 0.1, 0.25, 0.5, 0.75, 1.0]$;
  \item $\lambda_O$: $[0.01, 0.05, 0.1, 0.25, 0.5, 0.75, 1.0]$;
  \item $\lambda_T$: $[0.5, 0.6]$;
  \item Dropout: $[0.0, 0.1]$;
  \item Weight decay: $[0.0, 10^{-8}, 10^{-10}]$;
  \item Gumbel temperature in $f_\theta(\cdot)$: $[1.0;1.5]$;
  \item Gumbel temperature in $s_\theta(\cdot)$: $[1.0;1.5]$;
\end{itemize}

\subsubsection{Hardware / Software}

\begin{itemize}
  \item \textbf{CPU}: 2x Intel Xeon E5-2680 v3, 2x 12 cores, 24 threads, 2.5 GHz, 30 MB cache;
  \item \textbf{RAM}: 16x16GB DDR4-2133;
  \item \textbf{GPU}: 2x Nvidia Titan X Maxwell;
  \item \textbf{OS}: Ubuntu 18.04;
  \item \textbf{Software}: Python 3.6, PyTorch 1.3, CUDA 10.
\end{itemize}

\section{A Multi-Task Transformer with Explanations and Critiquing (T-RECS)}
\label{app_reproducibility}

We build the justification history $J^u$,$J^i$, with $N_{just}=32$. In T-RECS, we set the embedding, latent, and self-attention dimension size to 256, and the dimension of the~feed-forward network to 1024. The encoder and decoder consist of two layers of Transformer with $4$~attention heads. We use a batch size of 128, dropout of 0.1, 4000 warm-up steps, smoothing parameter $\varepsilon=0.1$, and Adam with learning rate $0.001,\beta_1=0.9,\beta_2=0.98$, and $\epsilon=10^{-9}$. The~Rating Classifier and Keyphrase Explainer are two layers of 128 and 64 dimensions with LeakyReLU ($\alpha=0.2$).~For critiquing, we choose a threshold and decay coefficient $T=0.015,\zeta=0.9$ and $T=0.01,\zeta=0.975$~for hotel and beer reviews, respectively. We use the code from the authors for most models. We tune all models on the dev set. We have operated a random search over $10$ trials. We chose the models achieving the lowest validation loss. The range of hyperparameters are the following for T-RECS (similar for other models):

\begin{itemize}
	\item Learning rate: $[0.001, 0.0001]$;
	\item Max epochs: $[100, 200, 300]$;
	\item Batch size: $[128]$;
	\item Hidden size encoder/decoder: $[256]$;
	\item Attention heads: $[4]$;
	\item Number of layers: $[2]$;
	\item Dropout encoder: $[0.0, 0.1, 0.2, 0.3, 0.4, 0.5]$;
	\item Dropout decoder: $[0.0, 0.1, 0.2, 0.3]$;
	\item General dropout: $[0.0, 0.1, 0.2]$;
	\item Warmup: $[2000, 4000, 8000, 16000]$;
	\item $\lambda_{r}$, $\lambda_{kp}$, $\lambda_{just}$: $[1.0]$;
\end{itemize}

Most of the time, the model converges under $20$ epochs.
For critiquing, we employed:
\begin{itemize}
	\item Decay coefficient $\zeta$: $[0.5, 0.75, 0.8, 0.9, 0.95]$;
	\item Max iterations: $[25, 50, 75, 100, 200]$;
	\item Threshold: $[0.015, 0.01, 0.005]$;
\end{itemize}

\subsubsection{Hardware / Software}

\begin{itemize}
	\item \textbf{CPU}: 2x Intel Xeon E5-2680 v3 (Haswell), 2x 12 cores, 24 threads, 2.5 GHz, 30 MB cache;
	\item \textbf{RAM}: 16x 16GB DDR4-2133;
	\item \textbf{GPU}: 2x Nvidia Titan X Maxwell;
	\item \textbf{OS}: Ubuntu 18.04;
	\item \textbf{Software}: Python 3.6, PyTorch 1.3.0, CUDA 10.0.
\end{itemize}

\section{MultiModal Variational AutoEncoder (M\&Ms-VAE)}
\label{app:add_repr}
The official baselines' codes from the respective authors, including the tuning procedure, are available in \footnote{\url{https://github.com/wuga214/NCE_Projected_LRec}}\footnote{\url{https://github.com/k9luo/DeepCritiquingForVAEBasedRecSys}}\footnote{\url{https://github.com/wuga214/DeepCritiquingForRecSys}}\footnote{\url{https://github.com/litosly/RankingOptimizationApproachtoLLC}}. The~final hyperparameters for all models and datasets are shown in Table~\ref{table_parameters}. For all experiments, we used the following hardware:
\begin{table}
\small
    \centering
\caption{\label{table_parameters}Best hyperparameter setting for each model. The top table refers to Section~\ref{sec_rq1} and the bottom one to Section~\ref{sec_rq2}.}
\hspace*{-1.0cm}
\begin{threeparttable}
\begin{tabular}{@{}c@{\hspace{2mm}}lc@{\hspace{2mm}}c@{\hspace{2mm}}c@{\hspace{2mm}}c@{\hspace{2mm}}c@{\hspace{2mm}}c@{\hspace{2mm}}c@{\hspace{2mm}}c@{\hspace{2mm}}c@{\hspace{2mm}}c@{\hspace{2mm}}c@{\hspace{2mm}}c@{}}\\
\textbf{Dataset} & \textbf{Model} & $H$ & $LR$ & $\lambda_{L2}$ & $\lambda$ & $\lambda_{KP}$ & $\lambda_{C}$ & $\beta$ & Iteration & Epoch & Dropout & $\gamma$ & Neg. Samples\\
\toprule
\multirow{10}{*}{\rotatebox{90}{\textit{Beer}}}
& AutoRec & 200 & 0.0001 & 0.00001 & 1.0 & - & - &  - & - & 300 &-& - & -\\
& BPR & 200 & - & 0.0001 & 1.0 & - & - &  - & - & 30 &-& - & 1\\
& CDAE & 200 & 0.0001 & 0.00001 & 1.0 & - & - &  - & - & 300 & 0.2 & - & -\\
& NCE-PLRec & 50 & - & 10000.0 & 1.0 & - & - &  - & 10 & - &-& 1.1 & -\\
& PLRec & 400 & - & 10000.0 & 1.0 & - & - &  - & 10 & - &-& - & -\\
& PureSVD & 50 & - & - & - & - & - &  - & 10 & - &-& - & -\\
& VAE-CF & 50 & 0.0001 & 0.0001 & 1.0 & - & - &  0.2 & - & - & 0.4 & - & -\\
& CE-VAE & 100 & 0.0001 & 0.0001 & 1.0 & 0.01 & 0.01 &  0.001 & - & 300 & 0.5 & - & -\\
& CE-VNCF & 100 & 0.0005 & 0.00005 & 1.0 & 1.0 & 1.0 &  0.1 & - & 100 & 0.1 & - & 5\\
& M\&Ms-VAE & 300 & 0.00005 & 1e-10 & 3.0 & - & - & 0.7 & - & 300 & 0.4 & - & -\\
\bottomrule
\multirow{10}{*}{\rotatebox{90}{\textit{CDs\&Vinyl}}}
& AutoRec & 200 & 0.0001 & 0.00001 & 1.0 & - & - &  - & - & 300 &-& - & -\\
& BPR & 200 & - & 0.0001 & 1.0 & - & - &  - & - & 30 &-& - & 1\\
& CDAE & 200 & 0.0001 & 0.00001 & 1.0 & - & - &  - & - & 300 & 0.2 & - & -\\
& NCE-PLRec & 200 & - & 1000.0 & 1.0 & - & - &  - & 10 & - &-& 1.3 & -\\
& PLRec & 400 & - & 1000.0 & 1.0 & - & - &  - & 10 & - &-& - & -\\
& PureSVD & 200 & - & - & - & - &  - & 10 & - &-& - & -\\
& VAE-CF & 200 & 0.0001 & 0.00001 & 1.0 & - & - &  0.2 & - & - & 0.3 & - & -\\
& CE-VAE & 200 & 0.0001 & 0.0001 & 1.0 & 0.001 & 0.001 &  0.0001 & - & 600 & 0.5 & - & -\\
& CE-VNCF & 100 & 0.0001 & 0.0001 & 1.0 & 1.0 & 1.0 &  0.1 & - & 100 & 0.1 & - & 5\\
& M\&Ms-VAE & 400 & 0.00005 & 1e-12 & 1.0 & - & - & 0.4 & - & 400 & 0.4 & - & -\\
\bottomrule
\multirow{10}{*}{\rotatebox{90}{\textit{Yelp}}}
& AutoRec & 50 & 0.0001 & 0.001 & 1.0 & - & - &  - & - & 300 &-& - & -\\
& BPR & 100 & - & 0.0001 & 1.0 & - & - &  - & - & 30 &-& - & 1\\
& CDAE & 50 & 0.0001 & 0.001 & 1.0 & - & - &  - & - & 300 & 0.4 & - & -\\
& NCE-PLRec & 50 & - & 10000.0 & 1.0 & - & - &  - & 10 & - &-& 1.3 & -\\
& PLRec & 400 & - & 10000.0 & 1.0 & - & - &  - & 10 & - &-& - & -\\
& PureSVD & 50 & - & - & - & - & - &  - & 10 & - &-& - & -\\
& VAE-CF & 50 & 0.0001 & 0.001 & 1.0 & - & - &  0.2 & - & - & 0.2 & - & -\\
& CE-VAE & 200 & 0.0001 & 0.0001 & 1.0 & 0.01 & 0.01 &  0.001 & - & 600 & 0.4 & - & -\\
& CE-VNCF & 100 & 0.0005 & 0.0001 & 1.0 & 1.0 & 1.0 &  0.1 & - & 100 & 0.1 & - & 5\\
& M\&Ms-VAE & 500 & 0.00005 & 1e-10 & 10.0 & - & - & 0.8 & - & 300 & 0.7 & - & -\\
\bottomrule
\multirow{10}{*}{\rotatebox{90}{\textit{Hotel}}}
& AutoRec & 50 & 0.0001 & 1e-05 & 1.0 & - & - &  - & - & 300 &-& - & -\\
& BPR & 200 & - & 0.0001 & 1.0 & - & - &  - & - & 30 &-& - & 1\\
& CDAE & 200 & 0.0001 & 0.001 & 1.0 & - & - &  - & - & 300 & 0.2 & - & -\\
& NCE-PLRec & 50 & - & 10000.0 & 1.0 & - & - &  - & 10 & - &-& 1.3 & -\\
& PLRec & 400 & - & 10000.0 & 1.0 & - & - &  - & 10 & - &-& - & -\\
& PureSVD & 50 & - & - & - & - & - & - & 10 & - &-& - & -\\
& VAE-CF & 50 & 0.0001 & 1e-05 & 1.0 & - & - &  0.2 & - & - & 0.5 & - & -\\
& CE-VAE & 200 & 0.0001 & 0.0001 & 1.0 & 0.01 & 0.01 &  0.001 & - & 600 & 0.2 & - & -\\
& CE-VNCF & 100 & 0.0005 & 0.0001 & 1.0 & 1.0 & 1.0 &  0.1 & - & 100 & 0.1 & - & 5\\
& M\&Ms-VAE & 400 & 0.00005 & 1e-12 & 2.0 & - & - & 0.8 & - & 300 & - & - & -\\
\bottomrule
\end{tabular}
\end{threeparttable}
\begin{threeparttable}
\begin{tabular}{@{}lcccc@{}}\\
\textbf{Dataset} & \textbf{Model} & $h$ & $LR$ & $\lambda_{L2}$\\
\toprule
Beer & \multirow{4}{*}{\parbox{3cm}{\centering M\&Ms-VAE $\xi(\cdot)$\\(Critiquing)}} & 0.75 & 0.001 & 0\\
CDsVinyl &  & 3.0 & 0.001 & 1e-10\\
Yelp & & 2.0 & 0.001 & 0\\
Hotel &  & 5.0 & 0.001 & 1e-10\\
\bottomrule
\end{tabular}
\end{threeparttable}
\end{table}

\subsubsection{Hardware / Software}

\begin{itemize}
  \item \textbf{CPU}: 2x Intel Xeon E5-2680 v3, 2x 12 cores, 24 threads, 2.5 GHz, 30 MB cache;
  \item \textbf{RAM}: 16x16GB DDR4-2133;
  \item \textbf{GPU}: 1x Nvidia Titan X Maxwell;
  \item \textbf{OS}: Ubuntu 18.04;
  \item \textbf{Software}: Python 3.6, PyTorch 1.6.1, CUDA 10.2.
\end{itemize}

%% file: biblio.tex
\cleardoublepage
\phantomsection
\addcontentsline{toc}{chapter}{References}
\bibliographystyle{apalike}

%% file: my_thesis.bbl
\begin{thebibliography}{}

\bibitem[Alaniz and Akata, 2019]{alaniz2019explainable}
Alaniz, S. and Akata, Z. (2019).
\newblock Explainable observer-classifier for explainable binary decisions.
\newblock {\em arXiv preprint arXiv:1902.01780}.

\bibitem[Aletras and Stevenson, 2013]{aletras2013evaluating}
Aletras, N. and Stevenson, M. (2013).
\newblock Evaluating topic coherence using distributional semantics.
\newblock In {\em Proceedings of the 10th International Conference on
  Computational Semantics ({IWCS} 2013) {--} Long Papers}, pages 13--22,
  Potsdam, Germany. Association for Computational Linguistics.

\bibitem[Alvarez-Melis and Jaakkola, 2017]{alvarez-melis-jaakkola-2017-causal}
Alvarez-Melis, D. and Jaakkola, T. (2017).
\newblock A causal framework for explaining the predictions of black-box
  sequence-to-sequence models.
\newblock In {\em Proceedings of the 2017 Conference on Empirical Methods in
  Natural Language Processing}, pages 412--421, Copenhagen, Denmark.
  Association for Computational Linguistics.

\bibitem[Alvarez{-}Melis and Jaakkola, 2018]{NEURIPS2018_3e9f0fc9}
Alvarez{-}Melis, D. and Jaakkola, T.~S. (2018).
\newblock Towards robust interpretability with self-explaining neural networks.
\newblock In Bengio, S., Wallach, H.~M., Larochelle, H., Grauman, K.,
  Cesa{-}Bianchi, N., and Garnett, R., editors, {\em Advances in Neural
  Information Processing Systems 31: Annual Conference on Neural Information
  Processing Systems 2018, NeurIPS 2018, December 3-8, 2018, Montr{\'{e}}al,
  Canada}, pages 7786--7795.

\bibitem[Amershi et~al., 2014]{Amershi_Cakmak_Knox_Kulesza_2014}
Amershi, S., Cakmak, M., Knox, W.~B., and Kulesza, T. (2014).
\newblock Power to the people: The role of humans in interactive machine
  learning.
\newblock volume~35, pages 105--120.

\bibitem[Angwin et~al., 2016]{propublica16}
Angwin, J., Larson, J., Mattu, S., and Kirchner, L. (2016).
\newblock Machine bias.
\newblock {\em ProPublica, May}, 23:2016.

\bibitem[Antognini and Faltings, 2019]{antognini-faltings-2019-learning}
Antognini, D. and Faltings, B. (2019).
\newblock Learning to create sentence semantic relation graphs for
  multi-document summarization.
\newblock In {\em Proceedings of the 2nd Workshop on New Frontiers in
  Summarization}, pages 32--41, Hong Kong, China. Association for Computational
  Linguistics.

\bibitem[Antognini and Faltings, 2020a]{antognini-faltings-2020-gamewikisum}
Antognini, D. and Faltings, B. (2020a).
\newblock {G}ame{W}iki{S}um: a novel large multi-document summarization
  dataset.
\newblock In {\em Proceedings of the 12th Language Resources and Evaluation
  Conference}, pages 6645--6650, Marseille, France. European Language Resources
  Association.

\bibitem[Antognini and Faltings, 2020b]{antognini-faltings-2020-hotelrec}
Antognini, D. and Faltings, B. (2020b).
\newblock {H}otel{R}ec: a novel very large-scale hotel recommendation dataset.
\newblock In {\em Proceedings of the 12th Language Resources and Evaluation
  Conference}, pages 4917--4923, Marseille, France. European Language Resources
  Association.

\bibitem[Antognini and Faltings, 2021a]{fast_critiquing}
Antognini, D. and Faltings, B. (2021a).
\newblock Fast multi-step critiquing for vae-based recommender systems.
\newblock In {\em Fifteenth ACM Conference on Recommender Systems}, RecSys '21,
  page 209–219, New York, NY, USA. Association for Computing Machinery.

\bibitem[Antognini and Faltings, 2021b]{antognini-2021-concept}
Antognini, D. and Faltings, B. (2021b).
\newblock Rationalization through concepts.
\newblock In {\em Findings of the Association for Computational Linguistics:
  ACL 2021}, pages 761--775, Online. Association for Computational Linguistics.

\bibitem[Antognini and Faltings, 2022]{fast_critiquing_pos}
Antognini, D. and Faltings, B. (2022).
\newblock Positive \& negative critiquing for vae-based recommenders.
\newblock {\em arXiv preprint arXiv:2204.02162}.

\bibitem[Antognini et~al., 2023]{editing_cooking}
Antognini, D., Li, S., Faltings, B., and McAuley, J. (2023).
\newblock Assistive recipe editing through critiquing.
\newblock In {\em Proceedings of the 17th Conference of the European Chapter of
  the Association for Computational Linguistics: Main Volume}, Online.
  Association for Computational Linguistics.

\bibitem[Antognini et~al., 2021a]{antognini2020interacting}
Antognini, D., Musat, C., and Faltings, B. (2021a).
\newblock Interacting with explanations through critiquing.
\newblock In Zhou, Z.-H., editor, {\em Proceedings of the Thirtieth
  International Joint Conference on Artificial Intelligence, {IJCAI-21}}, pages
  515--521. International Joint Conferences on Artificial Intelligence
  Organization.
\newblock Main Track.

\bibitem[Antognini et~al., 2021b]{antognini2019multi}
Antognini, D., Musat, C., and Faltings, B. (2021b).
\newblock Multi-dimensional explanation of target variables from documents.
\newblock In {\em Proceedings of the AAAI Conference on Artificial
  Intelligence, (AAAI 2021)}, volume~35, pages 12507--12515.

\bibitem[Armstrong et~al., 1983]{ARMSTRONG1983263}
Armstrong, S.~L., Gleitman, L.~R., and Gleitman, H. (1983).
\newblock What some concepts might not be.
\newblock {\em Cognition}, 13(3):263--308.

\bibitem[Bahdanau et~al., 2015]{iclr2015}
Bahdanau, D., Cho, K., and Bengio, Y. (2015).
\newblock Neural machine translation by jointly learning to align and
  translate.
\newblock In {\em 3rd International Conference on Learning Representations,
  {ICLR} 2015}.

\bibitem[Balog et~al., 2021]{10.1145/3404835.3462893}
Balog, K., Radlinski, F., and Karatzoglou, A. (2021).
\newblock {\em On Interpretation and Measurement of Soft Attributes for
  Recommendation}, page 890–899.
\newblock Association for Computing Machinery, New York, NY, USA.

\bibitem[Bao et~al., 2018]{bao-etal-2018-deriving}
Bao, Y., Chang, S., Yu, M., and Barzilay, R. (2018).
\newblock Deriving machine attention from human rationales.
\newblock In {\em Proceedings of the Conference on Empirical Methods in Natural
  Language Processing}, pages 1903--1913, Brussels, Belgium.

\bibitem[Barocas et~al., 2019]{barocas-hardt-narayanan}
Barocas, S., Hardt, M., and Narayanan, A. (2019).
\newblock {\em Fairness and Machine Learning}.
\newblock fairmlbook.org.
\newblock \url{http://www.fairmlbook.org}.

\bibitem[Bastings et~al., 2019]{bastings-etal-2019-interpretable}
Bastings, J., Aziz, W., and Titov, I. (2019).
\newblock Interpretable neural predictions with differentiable binary
  variables.
\newblock In {\em Proceedings of the 57th Annual Meeting of the Association for
  Computational Linguistics}, pages 2963--2977, Florence, Italy. Association
  for Computational Linguistics.

\bibitem[Bau et~al., 2017]{bau2017network}
Bau, D., Zhou, B., Khosla, A., Oliva, A., and Torralba, A. (2017).
\newblock Network dissection: Quantifying interpretability of deep visual
  representations.
\newblock In {\em Proceedings of the IEEE conference on computer vision and
  pattern recognition}, pages 6541--6549.

\bibitem[Bellini et~al., 2018]{bellini2018knowledge}
Bellini, V., Schiavone, A., Di~Noia, T., Ragone, A., and Di~Sciascio, E.
  (2018).
\newblock Knowledge-aware autoencoders for explainable recommender systems.
\newblock In {\em Proceedings of the 3rd Workshop on Deep Learning for
  Recommender Systems}.

\bibitem[Bengio et~al., 2013]{bengio2013estimating}
Bengio, Y., L{\'e}onard, N., and Courville, A. (2013).
\newblock Estimating or propagating gradients through stochastic neurons for
  conditional computation.
\newblock {\em arXiv preprint arXiv:1308.3432}.

\bibitem[Bennani-Smires et~al., 2018]{bennani-smires-etal-2018-simple}
Bennani-Smires, K., Musat, C., Hossmann, A., Baeriswyl, M., and Jaggi, M.
  (2018).
\newblock Simple unsupervised keyphrase extraction using sentence embeddings.
\newblock In {\em Proceedings of the 22nd Conference on Computational Natural
  Language Learning}, pages 221--229, Brussels, Belgium. Association for
  Computational Linguistics.

\bibitem[Blei et~al., 2003]{blei2003latent}
Blei, D.~M., Ng, A.~Y., and Jordan, M.~I. (2003).
\newblock Latent dirichlet allocation.
\newblock {\em Journal of machine Learning research}, 3(Jan):993--1022.

\bibitem[Bouchacourt and Denoyer, 2019]{bouchacourt2019educe}
Bouchacourt, D. and Denoyer, L. (2019).
\newblock Educe: Explaining model decisions through unsupervised concepts
  extraction.
\newblock {\em arXiv preprint arXiv:1905.11852}.

\bibitem[Bouma, 2009]{Bouma2009}
Bouma, G. (2009).
\newblock Normalized (pointwise) mutual information in collocation extraction.
\newblock In Chiarcos, C., de~Castilho, E., and Stede, M., editors, {\em Von
  der Form zur Bedeutung: Texte automatisch verarbeiten / From Form to Meaning:
  Processing Texts Automatically, Proceedings of the Biennial GSCL Conference
  2009}, pages 31--40, T{\"u}bingen. Gunter Narr Verlag.

\bibitem[Bowman et~al., 2016]{bowman-etal-2016-generating}
Bowman, S.~R., Vilnis, L., Vinyals, O., Dai, A., Jozefowicz, R., and Bengio, S.
  (2016).
\newblock Generating sentences from a continuous space.
\newblock In {\em Proceedings of The 20th {SIGNLL} Conference on Computational
  Natural Language Learning}, pages 10--21, Berlin, Germany. Association for
  Computational Linguistics.

\bibitem[Brand et~al., 2015]{brand2015beyond}
Brand, A., Allen, L., Altman, M., Hlava, M., and Scott, J. (2015).
\newblock Beyond authorship: attribution, contribution, collaboration, and
  credit.
\newblock {\em Learned Publishing}, 28(2):151--155.

\bibitem[Brown et~al., 2020]{NEURIPS2020_1457c0d6}
Brown, T., Mann, B., Ryder, N., Subbiah, M., Kaplan, J.~D., Dhariwal, P.,
  Neelakantan, A., Shyam, P., Sastry, G., Askell, A., Agarwal, S.,
  Herbert-Voss, A., Krueger, G., Henighan, T., Child, R., Ramesh, A., Ziegler,
  D., Wu, J., Winter, C., Hesse, C., Chen, M., Sigler, E., Litwin, M., Gray,
  S., Chess, B., Clark, J., Berner, C., McCandlish, S., Radford, A., Sutskever,
  I., and Amodei, D. (2020).
\newblock Language models are few-shot learners.
\newblock In Larochelle, H., Ranzato, M., Hadsell, R., Balcan, M.~F., and Lin,
  H., editors, {\em Advances in Neural Information Processing Systems},
  volume~33, pages 1877--1901. Curran Associates, Inc.

\bibitem[Buolamwini and Gebru, 2018]{pmlr-v81-buolamwini18a}
Buolamwini, J. and Gebru, T. (2018).
\newblock Gender shades: Intersectional accuracy disparities in commercial
  gender classification.
\newblock In Friedler, S.~A. and Wilson, C., editors, {\em Proceedings of the
  1st Conference on Fairness, Accountability and Transparency}, volume~81 of
  {\em Proceedings of Machine Learning Research}, pages 77--91. PMLR.

\bibitem[Burke et~al., 1996]{unitcritiquing}
Burke, R.~D., Hammond, K.~J., and Young, B.~C. (1996).
\newblock Knowledge-based navigation of complex information spaces.
\newblock In {\em Proceedings of the Thirteenth National Conference on
  Artificial Intelligence - Volume 1}, AAAI'96, pages 462--468. AAAI Press.

\bibitem[Caruana et~al., 2015]{10.1145/2783258.2788613}
Caruana, R., Lou, Y., Gehrke, J., Koch, P., Sturm, M., and Elhadad, N. (2015).
\newblock Intelligible models for healthcare: Predicting pneumonia risk and
  hospital 30-day readmission.
\newblock In {\em Proceedings of the 21th ACM SIGKDD International Conference
  on Knowledge Discovery and Data Mining}, KDD '15, page 1721–1730, New York,
  NY, USA. Association for Computing Machinery.

\bibitem[Castelo et~al., 2019]{castelo2019task}
Castelo, N., Bos, M.~W., and Lehmann, D.~R. (2019).
\newblock Task-dependent algorithm aversion.
\newblock {\em Journal of Marketing Research}, 56(5):809--825.

\bibitem[Catherine and Cohen, 2017]{catherine2017transnets}
Catherine, R. and Cohen, W. (2017).
\newblock Transnets: Learning to transform for recommendation.
\newblock In {\em Proceedings of the Eleventh ACM Conference on Recommender
  Systems}, RecSys '17, page 288–296, New York, NY, USA. Association for
  Computing Machinery.

\bibitem[Chang et~al., 2016]{chang2016crowd}
Chang, S., Harper, F.~M., and Terveen, L.~G. (2016).
\newblock Crowd-based personalized natural language explanations for
  recommendations.
\newblock In {\em Proceedings of the 10th ACM Conference on Recommender
  Systems}, RecSys '16, page 175–182, New York, NY, USA. Association for
  Computing Machinery.

\bibitem[Chang et~al., 2019]{chang2019game}
Chang, S., Zhang, Y., Yu, M., and Jaakkola, T. (2019).
\newblock A game theoretic approach to class-wise selective rationalization.
\newblock In Wallach, H., Larochelle, H., Beygelzimer, A., d\textquotesingle
  Alch\'{e}-Buc, F., Fox, E., and Garnett, R., editors, {\em Advances in Neural
  Information Processing Systems}, volume~32. Curran Associates, Inc.

\bibitem[Chang et~al., 2020]{chang2020invariant}
Chang, S., Zhang, Y., Yu, M., and Jaakkola, T. (2020).
\newblock Invariant rationalization.
\newblock In III, H.~D. and Singh, A., editors, {\em Proceedings of the 37th
  International Conference on Machine Learning}, volume 119 of {\em Proceedings
  of Machine Learning Research}, pages 1448--1458. PMLR.

\bibitem[Chen et~al., 2018a]{10.1145/3178876.3186070}
Chen, C., Zhang, M., Liu, Y., and Ma, S. (2018a).
\newblock Neural attentional rating regression with review-level explanations.
\newblock In {\em Proceedings of the 2018 World Wide Web Conference}, WWW '18,
  pages 1583--1592. International World Wide Web Conferences Steering
  Committee.

\bibitem[Chen et~al., 2018b]{chen2018learning}
Chen, J., Song, L., Wainwright, M., and Jordan, M. (2018b).
\newblock Learning to explain: An information-theoretic perspective on model
  interpretation.
\newblock In {\em International Conference on Machine Learning}, pages
  883--892.

\bibitem[Chen et~al., 2019a]{chen2018shapley}
Chen, J., Song, L., Wainwright, M.~J., and Jordan, M.~I. (2019a).
\newblock L-shapley and c-shapley: Efficient model interpretation for
  structured data.
\newblock In {\em International Conference on Learning Representations}.

\bibitem[Chen and Pu, 2007a]{HybridChen}
Chen, L. and Pu, P. (2007a).
\newblock Hybrid critiquing-based recommender systems.
\newblock In {\em Proceedings of the 12th International Conference on
  Intelligent User Interfaces}, IUI '07, page 22–31, New York, NY, USA.
  Association for Computing Machinery.

\bibitem[Chen and Pu, 2007b]{10.1007/978-3-540-73078-1_11}
Chen, L. and Pu, P. (2007b).
\newblock Preference-based organization interfaces: Aiding user critiques in
  recommender systems.
\newblock In {\em Proceedings of the 11th International Conference on User
  Modeling}, UM '07, page 77–86, Berlin, Heidelberg. Springer-Verlag.

\bibitem[Chen and Pu, 2012]{chen2012critiquing}
Chen, L. and Pu, P. (2012).
\newblock Critiquing-based recommenders: survey and emerging trends.
\newblock {\em User Modeling and User-Adapted Interaction}, 22(1-2).

\bibitem[Chen et~al., 2020]{chen2020towards}
Chen, Z., Wang, X., Xie, X., Parsana, M., Soni, A., Ao, X., and Chen, E.
  (2020).
\newblock Towards explainable conversational recommendation.
\newblock In Bessiere, C., editor, {\em Proceedings of the Twenty-Ninth
  International Joint Conference on Artificial Intelligence, {IJCAI-20}}, pages
  2994--3000. International Joint Conferences on Artificial Intelligence
  Organization.
\newblock Main track.

\bibitem[Chen et~al., 2019b]{chen2019co}
Chen, Z., Wang, X., Xie, X., Wu, T., Bu, G., Wang, Y., and Chen, E. (2019b).
\newblock Co-attentive multi-task learning for explainable recommendation.
\newblock In {\em Proceedings of the Twenty-Eighth International Joint
  Conference on Artificial Intelligence, {IJCAI-19}}, pages 2137--2143.
  International Joint Conferences on Artificial Intelligence Organization.

\bibitem[Chung et~al., 2014]{chung2014empirical}
Chung, J., Gulcehre, C., Cho, K., and Bengio, Y. (2014).
\newblock Empirical evaluation of gated recurrent neural networks on sequence
  modeling.
\newblock {\em arXiv preprint arXiv:1412.3555}.

\bibitem[Citron and Pasquale, 2014]{citron2014scored}
Citron, D.~K. and Pasquale, F. (2014).
\newblock The scored society: Due process for automated predictions.
\newblock {\em Wash. L. Rev.}, 89:1.

\bibitem[Clark et~al., 2019]{clark-etal-2019-bert}
Clark, K., Khandelwal, U., Levy, O., and Manning, C.~D. (2019).
\newblock What does {BERT} look at? an analysis of {BERT}{'}s attention.
\newblock In {\em Proceedings of the 2019 ACL Workshop BlackboxNLP: Analyzing
  and Interpreting Neural Networks for NLP}, Florence, Italy. Association for
  Computational Linguistics.

\bibitem[Costa et~al., 2018]{costa2018automatic}
Costa, F., Ouyang, S., Dolog, P., and Lawlor, A. (2018).
\newblock Automatic generation of natural language explanations.
\newblock In {\em Proceedings of the 23rd International Conference on
  Intelligent User Interfaces Companion}, IUI '18 Companion, New York, NY, USA.
  Association for Computing Machinery.

\bibitem[Cremonesi et~al., 2010]{10.1145/1864708.1864721}
Cremonesi, P., Koren, Y., and Turrin, R. (2010).
\newblock Performance of recommender algorithms on top-n recommendation tasks.
\newblock In {\em Proceedings of the Fourth ACM Conference on Recommender
  Systems}, RecSys '10, pages 39--46, New York, NY, USA. Association for
  Computing Machinery.

\bibitem[Dathathri et~al., 2020]{abs-1912-02164}
Dathathri, S., Madotto, A., Lan, J., Hung, J., Frank, E., Molino, P., Yosinski,
  J., and Liu, R. (2020).
\newblock Plug and play language models: {A} simple approach to controlled text
  generation.
\newblock In {\em International Conference on Learning Representations (ICLR)}.

\bibitem[Devlin et~al., 2019]{devlin-etal-2019-bert}
Devlin, J., Chang, M.-W., Lee, K., and Toutanova, K. (2019).
\newblock {BERT}: Pre-training of deep bidirectional transformers for language
  understanding.
\newblock In {\em Proceedings of NAACL-HLT}, pages 4171--4186.

\bibitem[DeYoung et~al., 2020]{deyoung-etal-2020-eraser}
DeYoung, J., Jain, S., Rajani, N.~F., Lehman, E., Xiong, C., Socher, R., and
  Wallace, B.~C. (2020).
\newblock {ERASER}: {A} benchmark to evaluate rationalized {NLP} models.
\newblock In {\em Proceedings of the 58th Annual Meeting of the Association for
  Computational Linguistics}, pages 4443--4458, Online. Association for
  Computational Linguistics.

\bibitem[Dieng et~al., 2017]{dieng2016topicrnn}
Dieng, A.~B., Wang, C., Gao, J., and Paisley, J.~W. (2017).
\newblock Topicrnn: {A} recurrent neural network with long-range semantic
  dependency.
\newblock In {\em 5th International Conference on Learning Representations,
  {ICLR} 2017, Toulon, France, April 24-26, 2017, Conference Track
  Proceedings}. OpenReview.net.

\bibitem[Dietvorst et~al., 2015]{dietvorst2015algorithm}
Dietvorst, B.~J., Simmons, J.~P., and Massey, C. (2015).
\newblock Algorithm aversion: People erroneously avoid algorithms after seeing
  them err.
\newblock {\em Journal of Experimental Psychology: General}, 144(1):114.

\bibitem[Dietvorst et~al., 2018]{dietvorst2018algorithm}
Dietvorst, B.~J., Simmons, J.~P., and Massey, C. (2018).
\newblock Overcoming algorithm aversion: People will use imperfect algorithms
  if they can even slightly modify them.
\newblock {\em Manage. Sci.}, 64(3):1155–1170.

\bibitem[Dong et~al., 2017]{dong2017learning}
Dong, L., Huang, S., Wei, F., Lapata, M., Zhou, M., and Xu, K. (2017).
\newblock Learning to generate product reviews from attributes.
\newblock In {\em Proceedings of the 15th Conference of the {E}uropean Chapter
  of the Association for Computational Linguistics: Volume 1, Long Papers},
  pages 623--632, Valencia, Spain. Association for Computational Linguistics.

\bibitem[Doshi-Velez and Kim, 2017]{doshi2017towards}
Doshi-Velez, F. and Kim, B. (2017).
\newblock Towards a rigorous science of interpretable machine learning.
\newblock {\em arXiv preprint arXiv:1702.08608}.

\bibitem[Erkan and Radev, 2004]{erkan2004lexrank}
Erkan, G. and Radev, D.~R. (2004).
\newblock Lexrank: Graph-based lexical centrality as salience in text
  summarization.
\newblock {\em Journal of AI research}, 22:457--479.

\bibitem[Faltings et~al., 2004a]{Faltings2004b}
Faltings, B., Pu, P., Torrens, M., and Viappiani, P. (2004a).
\newblock Designing example-critiquing interaction.
\newblock In {\em Proceedings of the 9th International Conference on
  Intelligent User Interfaces}, IUI '04, pages 22--29, Island of Madeira,
  Portugal. ACM Press.

\bibitem[Faltings et~al., 2004b]{faltings2004solution}
Faltings, B., Torrens, M., and Pu, P. (2004b).
\newblock Solution generation with qualitative models of preferences.
\newblock {\em Computational Intelligence}, 20(2).

\bibitem[Faruqui et~al., 2015a]{faruqui-etal-2015-retrofitting}
Faruqui, M., Dodge, J., Jauhar, S.~K., Dyer, C., Hovy, E., and Smith, N.~A.
  (2015a).
\newblock Retrofitting word vectors to semantic lexicons.
\newblock In {\em Proceedings of the 2015 Conference of the North {A}merican
  Chapter of the Association for Computational Linguistics: Human Language
  Technologies}, pages 1606--1615, Denver, Colorado. Association for
  Computational Linguistics.

\bibitem[Faruqui et~al., 2015b]{faruqui2015sparse}
Faruqui, M., Tsvetkov, Y., Yogatama, D., Dyer, C., and Smith, N.~A. (2015b).
\newblock Sparse overcomplete word vector representations.
\newblock In {\em Proceedings of the 53rd Annual Meeting of the Association for
  Computational Linguistics and the 7th International Joint Conference on
  Natural Language Processing (Volume 1: Long Papers)}, pages 1491--1500,
  Beijing, China. Association for Computational Linguistics.

\bibitem[Figurnov et~al., 2018]{figurnov2018implicit}
Figurnov, M., Mohamed, S., and Mnih, A. (2018).
\newblock Implicit reparameterization gradients.
\newblock In Bengio, S., Wallach, H., Larochelle, H., Grauman, K.,
  Cesa-Bianchi, N., and Garnett, R., editors, {\em Advances in Neural
  Information Processing Systems}, volume~31. Curran Associates, Inc.

\bibitem[Filipovic et~al., 2021]{model_evaluation}
Filipovic, M., Mitrevski, B., Antognini, D., Lejal~Glaude, E., Faltings, B.,
  and Musat, C. (2021).
\newblock Modeling online behavior in recommender systems: The importance of
  temporal context.
\newblock In {\em Proceedings of the Perspectives on the Evaluation of
  Recommender Systems Workshop at RecSys 2021 (PERSPECTIVE 2021)}.

\bibitem[Ghorbani et~al., 2019]{ghorbani2019towards}
Ghorbani, A., Wexler, J., Zou, J.~Y., and Kim, B. (2019).
\newblock Towards automatic concept-based explanations.
\newblock In Wallach, H., Larochelle, H., Beygelzimer, A., d\textquotesingle
  Alch\'{e}-Buc, F., Fox, E., and Garnett, R., editors, {\em Advances in Neural
  Information Processing Systems}, volume~32. Curran Associates, Inc.

\bibitem[Giannakopoulos et~al., 2017]{giannakopoulos2017dataset}
Giannakopoulos, A., Antognini, D., Musat, C., Hossmann, A., and Baeriswyl, M.
  (2017).
\newblock Dataset construction via attention for aspect term extraction with
  distant supervision.
\newblock In {\em 2017 IEEE International Conference on Data Mining Workshops
  (ICDMW)}, pages 373--380. IEEE.

\bibitem[Goodman and Flaxman, 2017]{Goodman_Flaxman_2017}
Goodman, B. and Flaxman, S. (2017).
\newblock European union regulations on algorithmic decision-making and a
  “right to explanation”.
\newblock {\em AI Magazine}, 38(3):50--57.

\bibitem[Goyal et~al., 2019]{goyal2019explaining}
Goyal, Y., Feder, A., Shalit, U., and Kim, B. (2019).
\newblock Explaining classifiers with causal concept effect (cace).
\newblock {\em arXiv preprint arXiv:1907.07165}.

\bibitem[Gunning, 2016]{Gunning2016}
Gunning, D. (2016).
\newblock {\em Explainable artificial intelligence (XAI): Technical Report}.
\newblock Defense Advanced Research Projects Agency DARPA-BAA-16-53, DARPA,
  Arlington, USA.

\bibitem[He and McAuley, 2016]{amazon2}
He, R. and McAuley, J. (2016).
\newblock Ups and downs: Modeling the visual evolution of fashion trends with
  one-class collaborative filtering.
\newblock In {\em Proceedings of the 25th International Conference on World
  Wide Web}, WWW '16, pages 507--517, Republic and Canton of Geneva, CHE.
  International World Wide Web Conferences Steering Committee.

\bibitem[He et~al., 2017]{he2017neural}
He, X., Liao, L., Zhang, H., Nie, L., Hu, X., and Chua, T.-S. (2017).
\newblock Neural collaborative filtering.
\newblock In {\em Proceedings of the 26th International Conference on World
  Wide Web}, WWW '17, page 173–182, Republic and Canton of Geneva, CHE.
  International World Wide Web Conferences Steering Committee.

\bibitem[Herbelot and Vecchi, 2015]{herbelot2015building}
Herbelot, A. and Vecchi, E.~M. (2015).
\newblock Building a shared world: mapping distributional to model-theoretic
  semantic spaces.
\newblock In {\em Proceedings of the 2015 Conference on Empirical Methods in
  Natural Language Processing}, pages 22--32, Lisbon, Portugal. Association for
  Computational Linguistics.

\bibitem[Herlocker et~al., 2000]{herlocker2000explaining}
Herlocker, J.~L., Konstan, J.~A., and Riedl, J. (2000).
\newblock Explaining collaborative filtering recommendations.
\newblock In {\em Proceedings of the 2000 ACM Conference on Computer Supported
  Cooperative Work}, CSCW '00, page 241–250, New York, NY, USA. Association
  for Computing Machinery.

\bibitem[Hidasi and Karatzoglou, 2018]{hidasi2018sessionbased}
Hidasi, B. and Karatzoglou, A. (2018).
\newblock Recurrent neural networks with top-k gains for session-based
  recommendations.
\newblock In {\em Proceedings of the 27th ACM International Conference on
  Information and Knowledge Management}, CIKM '18, pages 843--852, New York,
  NY, USA. Association for Computing Machinery.

\bibitem[Hidasi et~al., 2016]{hidasi2015sessionbased}
Hidasi, B., Karatzoglou, A., Baltrunas, L., and Tikk, D. (2016).
\newblock Session-based recommendations with recurrent neural networks.
\newblock In {\em the International Conference on Learning Representations
  (ICLR)}.

\bibitem[Higgins et~al., 2017]{higgins2016beta}
Higgins, I., Matthey, L., Pal, A., Burgess, C., Glorot, X., Botvinick, M.,
  Mohamed, S., and Lerchner, A. (2017).
\newblock beta-vae: Learning basic visual concepts with a constrained
  variational framework.
\newblock In {\em the International Conference on Learning Representations
  (ICLR)}.

\bibitem[Hinton et~al., 2015]{44873}
Hinton, G., Vinyals, O., and Dean, J. (2015).
\newblock Distilling the knowledge in a neural network.
\newblock In {\em NIPS Deep Learning and Representation Learning Workshop}.

\bibitem[Hochreiter and Schmidhuber, 1997]{hochreiter1997long}
Hochreiter, S. and Schmidhuber, J. (1997).
\newblock Long short-term memory.
\newblock {\em Neural computation}, 9(8):1735--1780.

\bibitem[Holtzman et~al., 2020]{holtzman2019curious}
Holtzman, A., Buys, J., Du, L., Forbes, M., and Choi, Y. (2020).
\newblock The curious case of neural text degeneration.
\newblock In {\em International Conference on Learning Representations}.

\bibitem[Hoyer, 2004]{hoyer2004non}
Hoyer, P.~O. (2004).
\newblock Non-negative matrix factorization with sparseness constraints.
\newblock {\em Journal of machine learning research}, 5(Nov):1457--1469.

\bibitem[Jacovi and Goldberg, 2020]{jacovi-goldberg-2020-towards}
Jacovi, A. and Goldberg, Y. (2020).
\newblock Towards faithfully interpretable {NLP} systems: How should we define
  and evaluate faithfulness?
\newblock In {\em Proceedings of the 58th Annual Meeting of the Association for
  Computational Linguistics}, pages 4198--4205, Online. Association for
  Computational Linguistics.

\bibitem[Jain and Wallace, 2019]{jain2019attention}
Jain, S. and Wallace, B.~C. (2019).
\newblock {A}ttention is not {E}xplanation.
\newblock In {\em Proceedings of the 2019 Conference of the North {A}merican
  Chapter of the Association for Computational Linguistics: Human Language
  Technologies, Volume 1 (Long and Short Papers)}, pages 3543--3556,
  Minneapolis, Minnesota. Association for Computational Linguistics.

\bibitem[Jain et~al., 2020]{jain-etal-2020-learning}
Jain, S., Wiegreffe, S., Pinter, Y., and Wallace, B.~C. (2020).
\newblock {L}earning to faithfully rationalize by construction.
\newblock In {\em Proceedings of the 58th Annual Meeting of the Association for
  Computational Linguistics}, pages 4459--4473, Online. Association for
  Computational Linguistics.

\bibitem[Jang et~al., 2017]{JangGP17}
Jang, E., Gu, S., and Poole, B. (2017).
\newblock Categorical reparameterization with gumbel-softmax.
\newblock In {\em 5th International Conference on Learning Representations,
  {ICLR} 2017, Toulon, France, April 24-26}.

\bibitem[Jannach and Jugovac, 2019]{10.1145/3370082}
Jannach, D. and Jugovac, M. (2019).
\newblock Measuring the business value of recommender systems.
\newblock {\em ACM Trans. Manage. Inf. Syst.}, 10(4).

\bibitem[Jussupow et~al., 2020]{madoc56152}
Jussupow, E., Benbasat, I., and Heinzl, A. (2020).
\newblock Why are we averse towards algorithms? a comprehensive literature
  review on algorithm aversion.
\newblock In Rowe, F., editor, {\em 28th European Conference on Information
  Systems - Liberty, Equality, and Fraternity in a Digitizing World, ECIS 2020,
  Marrakech, Morocco, June 15-17, 2020 : Proceedings}, page RP 168, Atlanta,
  GA. AISeL.

\bibitem[Karpathy et~al., 2015]{KarpathyJL15}
Karpathy, A., Johnson, J., and Fei-Fei, L. (2015).
\newblock Visualizing and understanding recurrent networks.
\newblock {\em arXiv preprint arXiv:1506.02078}.

\bibitem[Kendall, 1938]{kendall1938measure}
Kendall, M.~G. (1938).
\newblock A new measure of rank correlation.
\newblock {\em Biometrika}, 30(1/2):81--93.

\bibitem[Kim et~al., 2015]{kim2015mind}
Kim, B., Shah, J.~A., and Doshi-Velez, F. (2015).
\newblock Mind the gap: A generative approach to interpretable feature
  selection and extraction.
\newblock In Cortes, C., Lawrence, N., Lee, D., Sugiyama, M., and Garnett, R.,
  editors, {\em Advances in Neural Information Processing Systems}, volume~28.
  Curran Associates, Inc.

\bibitem[Kim et~al., 2018]{conf/icml/KimWGCWVS18}
Kim, B., Wattenberg, M., Gilmer, J., Cai, C.~J., Wexler, J., Vi\'{e}gas, F.~B.,
  and Sayres, R. (2018).
\newblock Interpretability beyond feature attribution: Quantitative testing
  with concept activation vectors (tcav).
\newblock In Dy, J.~G. and Krause, A., editors, {\em ICML}, volume~80 of {\em
  Proceedings of Machine Learning Research}, pages 2673--2682. PMLR.

\bibitem[Kingma and Ba, 2015]{KingmaB14}
Kingma, D.~P. and Ba, J. (2015).
\newblock Adam: {A} method for stochastic optimization.
\newblock In {\em 3rd International Conference on Learning Representations,
  {ICLR} 2015}.

\bibitem[Kingma and Welling, 2014]{kingmamw2014}
Kingma, D.~P. and Welling, M. (2014).
\newblock {Auto-Encoding Variational Bayes}.
\newblock In {\em 2nd International Conference on Learning Representations,
  {ICLR} 2014, Banff, AB, Canada, April 14-16, 2014, Conference Track
  Proceedings}.

\bibitem[Kiritchenko and Mohammad, 2016]{kiritchenko2016capturing}
Kiritchenko, S. and Mohammad, S.~M. (2016).
\newblock Capturing reliable fine-grained sentiment associations by
  crowdsourcing and best{--}worst scaling.
\newblock In {\em Proceedings of the 2016 Conference of the North {A}merican
  Chapter of the Association for Computational Linguistics: Human Language
  Technologies}, pages 811--817, San Diego, California. Association for
  Computational Linguistics.

\bibitem[Knijnenburg et~al., 2012]{knijnenburg2012explaining}
Knijnenburg, B.~P., Willemsen, M.~C., Gantner, Z., Soncu, H., and Newell, C.
  (2012).
\newblock Explaining the user experience of recommender systems.
\newblock {\em User Modeling and User-Adapted Interaction}, 22(4):441--504.

\bibitem[Koh et~al., 2020]{koh2020concept}
Koh, P.~W., Nguyen, T., Tang, Y.~S., Mussmann, S., Pierson, E., Kim, B., and
  Liang, P. (2020).
\newblock Concept bottleneck models.
\newblock In {\em Proceedings of the 37th International Conference on Machine
  Learning}.

\bibitem[Konstan and Riedl, 2012]{KonstanRiedl12umuai}
Konstan, J.~A. and Riedl, J. (2012).
\newblock Recommender systems: From algorithms to user experience.
\newblock {\em User Modeling and User-Adapted Interaction}, 22(1--2):101--123.

\bibitem[Kovaleva et~al., 2019]{kovaleva-etal-2019-revealing}
Kovaleva, O., Romanov, A., Rogers, A., and Rumshisky, A. (2019).
\newblock Revealing the dark secrets of {BERT}.
\newblock In {\em Proceedings of the 2019 Conference on Empirical Methods in
  Natural Language Processing and the 9th International Joint Conference on
  Natural Language Processing (EMNLP-IJCNLP)}, pages 4365--4374, Hong Kong,
  China. Association for Computational Linguistics.

\bibitem[Kunkel et~al., 2018]{Kunkel2018TrustrelatedEO}
Kunkel, J., Donkers, T., Barbu, C.-M., and Ziegler, J. (2018).
\newblock Trust-related effects of expertise and similarity cues in
  human-generated recommendations.
\newblock {\em Companion Proceedings of the 23rd International on Intelligent
  User Interfaces: 2nd Workshop on Theory-Informed User Modeling for Tailoring
  and Personalizing Interfaces (HUMANIZE).}

\bibitem[Kunkel et~al., 2019]{kunkel2019let}
Kunkel, J., Donkers, T., Michael, L., Barbu, C.-M., and Ziegler, J. (2019).
\newblock Let me explain: Impact of personal and impersonal explanations on
  trust in recommender systems.
\newblock In {\em Proceedings of the 2019 CHI Conference on Human Factors in
  Computing Systems}, page 1–12, New York, NY, USA. Association for Computing
  Machinery.

\bibitem[Lau and Baldwin, 2016]{lau2016sensitivity}
Lau, J.~H. and Baldwin, T. (2016).
\newblock The sensitivity of topic coherence evaluation to topic cardinality.
\newblock In {\em Proceedings of the 2016 Conference of the North American
  Chapter of the Association for Computational Linguistics}, pages 483--487.

\bibitem[Lau et~al., 2014]{lau2014machine}
Lau, J.~H., Newman, D., and Baldwin, T. (2014).
\newblock Machine reading tea leaves: Automatically evaluating topic coherence
  and topic model quality.
\newblock In {\em Proceedings of the 14th Conference of the {E}uropean Chapter
  of the Association for Computational Linguistics}, pages 530--539,
  Gothenburg, Sweden. Association for Computational Linguistics.

\bibitem[Lei et~al., 2016]{lei-etal-2016-rationalizing}
Lei, T., Barzilay, R., and Jaakkola, T. (2016).
\newblock Rationalizing neural predictions.
\newblock In {\em Proceedings of the 2016 Conference on Empirical Methods in
  Natural Language Processing}, pages 107--117, Austin, Texas. Association for
  Computational Linguistics.

\bibitem[Li et~al., 2020]{hanze2020}
Li, H., Sanner, S., Luo, K., and Wu, G. (2020).
\newblock A ranking optimization approach to latent linear critiquing for
  conversational recommender systems.
\newblock In {\em Fourteenth ACM Conference on Recommender Systems}, RecSys
  '20, pages 13--22, New York, NY, USA. Association for Computing Machinery.

\bibitem[Li et~al., 2016a]{li2016visualizing}
Li, J., Chen, X., Hovy, E., and Jurafsky, D. (2016a).
\newblock Visualizing and understanding neural models in nlp.
\newblock In {\em Proceedings of the 2016 Conference of the North American
  Chapter of the Association for Computational Linguistics: Human Language
  Technologies}, pages 681--691.

\bibitem[Li et~al., 2016b]{li2016understanding}
Li, J., Monroe, W., and Jurafsky, D. (2016b).
\newblock Understanding neural networks through representation erasure.
\newblock {\em arXiv preprint arXiv:1612.08220}.

\bibitem[Li et~al., 2018a]{li2018document}
Li, J., Yang, H., and Zong, C. (2018a).
\newblock Document-level multi-aspect sentiment classification by jointly
  modeling users, aspects, and overall ratings.
\newblock In {\em Proceedings of the 27th International Conference on
  Computational Linguistics}, pages 925--936, Santa Fe, New Mexico, USA.
  Association for Computational Linguistics.

\bibitem[Li et~al., 2018b]{li2018deep}
Li, O., Liu, H., Chen, C., and Rudin, C. (2018b).
\newblock Deep learning for case-based reasoning through prototypes: A neural
  network that explains its predictions.
\newblock In {\em Proceedings of the AAAI Conference on Artificial
  Intelligence}, volume~32.

\bibitem[Li and Tuzhilin, 2019]{li2019towards}
Li, P. and Tuzhilin, A. (2019).
\newblock Towards controllable and personalized review generation.
\newblock In {\em Proceedings of the 2019 Conference on Empirical Methods in
  Natural Language Processing and the 9th International Joint Conference on
  Natural Language Processing (EMNLP-IJCNLP)}, pages 3237--3245, Hong Kong,
  China. Association for Computational Linguistics.

\bibitem[Li et~al., 2019]{li2019persona}
Li, P., Wang, Z., Bing, L., and Lam, W. (2019).
\newblock Persona-aware tips generation?
\newblock In {\em The World Wide Web Conference}, WWW '19, page 1006–1016,
  New York, NY, USA. Association for Computing Machinery.

\bibitem[Li et~al., 2017]{li2017neural}
Li, P., Wang, Z., Ren, Z., Bing, L., and Lam, W. (2017).
\newblock Neural rating regression with abstractive tips generation for
  recommendation.
\newblock In {\em the ACM SIGIR conference on Research and Development in
  Information Retrieval}.

\bibitem[Liang et~al., 2018]{liang2018variational}
Liang, D., Krishnan, R.~G., Hoffman, M.~D., and Jebara, T. (2018).
\newblock Variational autoencoders for collaborative filtering.
\newblock In {\em Proceedings of the 2018 World Wide Web Conference}, WWW '18,
  page 689–698, Republic and Canton of Geneva, CHE. International World Wide
  Web Conferences Steering Committee.

\bibitem[Lin and Hovy, 2002]{lin2002manual}
Lin, C.-Y. and Hovy, E. (2002).
\newblock Manual and automatic evaluation of summaries.
\newblock In {\em Proceedings of the ACL-02 Workshop on Automatic
  Summarization-Volume 4}, pages 45--51. Association for Computational
  Linguistics.

\bibitem[Lin et~al., 2017]{LinFSYXZB17}
Lin, Z., Feng, M., dos Santos, C.~N., Yu, M., Xiang, B., Zhou, B., and Bengio,
  Y. (2017).
\newblock A structured self-attentive sentence embedding.
\newblock In {\em 5th International Conference on Learning Representations,
  {ICLR} 2017, Toulon, France, April 24-26, 2017, Conference Track
  Proceedings}. OpenReview.net.

\bibitem[Linden et~al., 1997]{multistepcritiquingNeal}
Linden, G., Hanks, S., and Lesh, N. (1997).
\newblock Interactive assessment of user preference models: The automated
  travel assistant.
\newblock In Jameson, A., Paris, C., and Tasso, C., editors, {\em User
  Modeling}, pages 67--78, Vienna. Springer Vienna.

\bibitem[Louis et~al., 2020]{louis-etal-2020-id}
Louis, A., Roth, D., and Radlinski, F. (2020).
\newblock {``}{I}{'}d rather just go to bed{''}: Understanding indirect
  answers.
\newblock In {\em Proceedings of the 2020 Conference on Empirical Methods in
  Natural Language Processing (EMNLP)}, pages 7411--7425, Online. Association
  for Computational Linguistics.

\bibitem[Louviere et~al., 2015]{louviere_flynn_marley_2015}
Louviere, J.~J., Flynn, T.~N., and Marley, A. A.~J. (2015).
\newblock {\em Best-Worst Scaling: Theory, Methods and Applications}.
\newblock Cambridge University Press.

\bibitem[Lu et~al., 2018]{lu2018like}
Lu, Y., Dong, R., and Smyth, B. (2018).
\newblock Why i like it: Multi-task learning for recommendation and
  explanation.
\newblock In {\em Proceedings of the 12th ACM Conference on Recommender
  Systems}, RecSys '18, page 4–12, New York, NY, USA. Association for
  Computing Machinery.

\bibitem[Lundberg and Lee, 2017]{NIPS2017_7062}
Lundberg, S.~M. and Lee, S.-I. (2017).
\newblock A unified approach to interpreting model predictions.
\newblock In Guyon, I., Luxburg, U.~V., Bengio, S., Wallach, H., Fergus, R.,
  Vishwanathan, S., and Garnett, R., editors, {\em Advances in Neural
  Information Processing Systems 30}, volume~30, pages 4765--4774. Curran
  Associates, Inc.

\bibitem[Luo et~al., 2020a]{luo2020b}
Luo, K., Sanner, S., Wu, G., Li, H., and Yang, H. (2020a).
\newblock Latent linear critiquing for conversational recommender systems.
\newblock In {\em Proceedings of The Web Conference 2020}, WWW '20, pages
  2535--2541, New York, NY, USA. Association for Computing Machinery.

\bibitem[Luo et~al., 2020b]{luo2020}
Luo, K., Yang, H., Wu, G., and Sanner, S. (2020b).
\newblock Deep critiquing for vae-based recommender systems.
\newblock In {\em Proceedings of the 43rd International ACM SIGIR Conference on
  Research and Development in Information Retrieval}, SIGIR '20, pages
  1269--1278, New York, NY, USA. Association for Computing Machinery.

\bibitem[Luong et~al., 2015]{luong-etal-2015-effective}
Luong, T., Pham, H., and Manning, C.~D. (2015).
\newblock Effective approaches to attention-based neural machine translation.
\newblock In {\em Proceedings of the 2015 Conference on Empirical Methods in
  Natural Language Processing}, pages 1412--1421, Lisbon, Portugal.

\bibitem[Lyu et~al., 2021]{10.1145/3442381.3450123}
Lyu, S., Rana, A., Sanner, S., and Bouadjenek, M.~R. (2021).
\newblock A workflow analysis of context-driven conversational recommendation.
\newblock In {\em Proceedings of the Web Conference 2021}, WWW '21, page
  866–877, New York, NY, USA. Association for Computing Machinery.

\bibitem[MacKenzie et~al., 2013]{mackenzie2013retailers}
MacKenzie, I., Meyer, C., and Noble, S. (2013).
\newblock How retailers can keep up with consumers.
\newblock {\em McKinsey \& Company}, 18:1.

\bibitem[Maddison et~al., 2017]{MaddisonMT17}
Maddison, C.~J., Mnih, A., and Teh, Y.~W. (2017).
\newblock The concrete distribution: {A} continuous relaxation of discrete
  random variables.
\newblock In {\em 5th International Conference on Learning Representations,
  {ICLR} 2017, Toulon, France, April 24-26}.

\bibitem[Mann and Thompson, 1988]{mann1988rhetorical}
Mann, W.~C. and Thompson, S.~A. (1988).
\newblock Rhetorical structure theory: Toward a functional theory of text
  organization.
\newblock {\em Text-interdisciplinary Journal for the Study of Discourse},
  8(3):243--281.

\bibitem[Martins and Astudillo, 2016]{martins2016softmax}
Martins, A. F.~T. and Astudillo, R.~F. (2016).
\newblock From softmax to sparsemax: A sparse model of attention and
  multi-label classification.
\newblock In {\em Proceedings of the 33rd International Conference on
  International Conference on Machine Learning - Volume 48}, ICML'16, page
  1614–1623. JMLR.org.

\bibitem[McAuley and Leskovec, 2013]{McAuley:2013:HFH:2507157.2507163}
McAuley, J. and Leskovec, J. (2013).
\newblock Hidden factors and hidden topics: Understanding rating dimensions
  with review text.
\newblock In {\em Proceedings of the 7th ACM Conference on Recommender
  Systems}, RecSys '13, pages 165--172, New York, NY, USA. ACM, ACM.

\bibitem[{McAuley} et~al., 2012]{beer}
{McAuley}, J., {Leskovec}, J., and {Jurafsky}, D. (2012).
\newblock Learning attitudes and attributes from multi-aspect reviews.
\newblock In {\em 2012 IEEE 12th International Conference on Data Mining}, ICDM
  '12, pages 1020--1025, Washington, DC, USA.

\bibitem[McAuley et~al., 2015]{amazon1}
McAuley, J., Targett, C., Shi, Q., and van~den Hengel, A. (2015).
\newblock Image-based recommendations on styles and substitutes.
\newblock In {\em Proceedings of the 38th International ACM SIGIR Conference on
  Research and Development in Information Retrieval}, SIGIR '15, pages 43--52,
  New York, NY, USA. Association for Computing Machinery.

\bibitem[McCarthy et~al., 2010]{mccarthy2010experience}
McCarthy, K., Salem, Y., and Smyth, B. (2010).
\newblock Experience-based critiquing : reusing critiquing experiences to
  improve conversational recommendation.
\newblock {\em Bichindaritz, I. and Montani, S. (eds.). Case-based reasoning
  research and development : 18th International Conference on Case-based
  Reasoning, ICCBR 2010, Alessandria, Italy, July 2010 : proceedings}.

\bibitem[Mikolov et~al., 2013]{mikolov2013distributed}
Mikolov, T., Sutskever, I., Chen, K., Corrado, G.~S., and Dean, J. (2013).
\newblock Distributed representations of words and phrases and their
  compositionality.
\newblock In Burges, C. J.~C., Bottou, L., Welling, M., Ghahramani, Z., and
  Weinberger, K.~Q., editors, {\em Advances in Neural Information Processing
  Systems}, volume~26. Curran Associates, Inc.

\bibitem[Milenkoski et~al., 2021]{cross_recommendation}
Milenkoski, M., Antognini, D., and Musat, C. (2021).
\newblock Recommending burgers based on pizza preferences: Addressing data
  sparsity with a product of experts.
\newblock In {\em Proceedings of the First Workshop of Cross-Market
  Recommendation at RecSys 2021 (XMRec 2021)}.

\bibitem[Miller, 2019]{MILLER20191}
Miller, T. (2019).
\newblock Explanation in artificial intelligence: Insights from the social
  sciences.
\newblock {\em Artificial Intelligence}, 267:1--38.

\bibitem[Milojkovic et~al., 2020]{milojkovic2019multi}
Milojkovic, N., Antognini, D., Bergamin, G., Faltings, B., and Musat, C.
  (2020).
\newblock Multi-gradient descent for multi-objective recommender systems.
\newblock In {\em Proceedings of the AAAI (2020) - Workshop on Interactive and
  Conversational Recommendation Systems (WICRS)}.

\bibitem[Mitrevski et~al., 2021]{momentum_gradient}
Mitrevski, B., Filipovic, M., Antognini, D., Lejal~Glaude, E., Faltings, B.,
  and Musat, C. (2021).
\newblock Momentum-based gradient methods in multi-objective recommendation.
\newblock In {\em Proceedings of the First Workshop on Multi-Objective
  Recommender Systems at RecSys 2021 (MORS 2021)}.

\bibitem[Mnih and Salakhutdinov, 2008]{mnih2008probabilistic}
Mnih, A. and Salakhutdinov, R.~R. (2008).
\newblock Probabilistic matrix factorization.
\newblock In Platt, J., Koller, D., Singer, Y., and Roweis, S., editors, {\em
  Advances in Neural Information Processing Systems}, volume~20. Curran
  Associates, Inc.

\bibitem[Montavon et~al., 2018]{montavon2018methods}
Montavon, G., Samek, W., and M{\"u}ller, K.-R. (2018).
\newblock Methods for interpreting and understanding deep neural networks.
\newblock {\em Digital Signal Processing}, 73:1--15.

\bibitem[Mukherjee and Awadallah, 2020]{mukherjee2020uncertainty}
Mukherjee, S. and Awadallah, A. (2020).
\newblock Uncertainty-aware self-training for few-shot text classification.
\newblock In Larochelle, H., Ranzato, M., Hadsell, R., Balcan, M.~F., and Lin,
  H., editors, {\em Advances in Neural Information Processing Systems},
  volume~33, pages 21199--21212. Curran Associates, Inc.

\bibitem[Musat and Faltings, 2015]{musat2015personalizing}
Musat, C.-C. and Faltings, B. (2015).
\newblock Personalizing product rankings using collaborative filtering on
  opinion-derived topic profiles.
\newblock In {\em Proceedings of the 24th International Conference on
  Artificial Intelligence}, IJCAI'15, page 830–836. AAAI Press.

\bibitem[Musat et~al., 2013]{musat2013}
Musat, C.-C., Liang, Y., and Faltings, B. (2013).
\newblock Recommendation using textual opinions.
\newblock In {\em Proceedings of the Twenty-Third International Joint
  Conference on Artificial Intelligence}, IJCAI '13, page 2684–2690. AAAI
  Press.

\bibitem[Nair and Hinton, 2010]{nair2010rectified}
Nair, V. and Hinton, G.~E. (2010).
\newblock Rectified linear units improve restricted boltzmann machines.
\newblock In {\em Proceedings of the 27th international conference on machine
  learning (ICML-10)}, pages 807--814.

\bibitem[Ni et~al., 2019]{ni-etal-2019-justifying}
Ni, J., Li, J., and McAuley, J. (2019).
\newblock Justifying recommendations using distantly-labeled reviews and
  fine-grained aspects.
\newblock In {\em the 2019 Conference on Empirical Methods in Natural Language
  Processing (EMNLP-IJCNLP)}, pages 188--197, Hong Kong, China. Association for
  Computational Linguistics.

\bibitem[Ni and McAuley, 2018]{ni2018personalized}
Ni, J. and McAuley, J. (2018).
\newblock Personalized review generation by expanding phrases and attending on
  aspect-aware representations.
\newblock In {\em Proceedings of the 56th Annual Meeting of the Association for
  Computational Linguistics (Volume 2: Short Papers)}, pages 706--711,
  Melbourne, Australia. Association for Computational Linguistics.

\bibitem[Nikolakopoulos et~al., 2019]{PERDIF2019}
Nikolakopoulos, A.~N., Berberidis, D., Karypis, G., and Giannakis, G.~B.
  (2019).
\newblock Personalized diffusions for top-n recommendation.
\newblock In {\em the 13th ACM Conference on Recommender Systems}, pages
  260--268.

\bibitem[Niu et~al., 2020]{niu-etal-2020-self}
Niu, Y., Jiao, F., Zhou, M., Yao, T., Xu, J., and Huang, M. (2020).
\newblock A self-training method for machine reading comprehension with soft
  evidence extraction.
\newblock In {\em Proceedings of the 58th Annual Meeting of the Association for
  Computational Linguistics}, pages 3916--3927, Online. Association for
  Computational Linguistics.

\bibitem[Padh et~al., 2021]{fairness_kirtan}
Padh, K., Antognini, D., Lejal~Glaude, E., Faltings, B., and Musat, C. (2021).
\newblock Addressing fairness in classification with a model-agnostic
  multi-objective algorithm.
\newblock In Peters, J. and Sontag, D., editors, {\em Proceedings of the 37th
  Conference on Uncertainty in Artificial Intelligence (UAI)}, Proceedings of
  Machine Learning Research. PMLR.

\bibitem[Papineni et~al., 2002]{papineni2002bleu}
Papineni, K., Roukos, S., Ward, T., and Zhu, W.-J. (2002).
\newblock {B}leu: a method for automatic evaluation of machine translation.
\newblock In {\em Proceedings of the 40th Annual Meeting of the Association for
  Computational Linguistics}, pages 311--318, Philadelphia, Pennsylvania, USA.
  Association for Computational Linguistics.

\bibitem[Pappas and Popescu-Belis, 2014]{pappas2014explaining}
Pappas, N. and Popescu-Belis, A. (2014).
\newblock Explaining the stars: Weighted multiple-instance learning for
  aspect-based sentiment analysis.
\newblock In {\em Proceedings of the 2014 Conference on Empirical Methods in
  Natural Language Processing ({EMNLP})}, pages 455--466, Doha, Qatar.
  Association for Computational Linguistics.

\bibitem[Paranjape et~al., 2020]{paranjape-etal-2020-information}
Paranjape, B., Joshi, M., Thickstun, J., Hajishirzi, H., and Zettlemoyer, L.
  (2020).
\newblock An information bottleneck approach for controlling conciseness in
  rationale extraction.
\newblock In {\em Proceedings of the 2020 Conference on Empirical Methods in
  Natural Language Processing (EMNLP)}, pages 1938--1952, Online. Association
  for Computational Linguistics.

\bibitem[Pariser, 2011]{pariser11}
Pariser, E. (2011).
\newblock {\em The filter bubble: What the Internet is hiding from you}.
\newblock Penguin UK.

\bibitem[Pennington et~al., 2014]{pennington2014glove}
Pennington, J., Socher, R., and Manning, C. (2014).
\newblock {G}lo{V}e: Global vectors for word representation.
\newblock In {\em Proceedings of the 2014 Conference on Empirical Methods in
  Natural Language Processing ({EMNLP})}, pages 1532--1543, Doha, Qatar.
  Association for Computational Linguistics.

\bibitem[Peters et~al., 2018]{peters-etal-2018-deep}
Peters, M., Neumann, M., Iyyer, M., Gardner, M., Clark, C., Lee, K., and
  Zettlemoyer, L. (2018).
\newblock Deep contextualized word representations.
\newblock In {\em Proceedings of the 2018 Conference of the North {A}merican
  Chapter of the Association for Computational Linguistics}, pages 2227--2237,
  New Orleans.

\bibitem[Petrescu et~al., 2021]{multi_step_demo}
Petrescu, D.~A., Antognini, D., and Faltings, B. (2021).
\newblock Multi-step critiquing user interface for recommender systems.
\newblock In {\em Fifteenth ACM Conference on Recommender Systems}, RecSys '21,
  page 760–763, New York, NY, USA. Association for Computing Machinery.

\bibitem[Pruthi et~al., 2020]{pruthi-etal-2020-learning}
Pruthi, D., Gupta, M., Dhingra, B., Neubig, G., and Lipton, Z.~C. (2020).
\newblock Learning to deceive with attention-based explanations.
\newblock In {\em Proceedings of the 58th Annual Meeting of the Association for
  Computational Linguistics}, pages 4782--4793, Online. Association for
  Computational Linguistics.

\bibitem[Pu and Chen, 2005]{pu2005integrating}
Pu, P. and Chen, L. (2005).
\newblock Integrating tradeoff support in product search tools for e-commerce
  sites.
\newblock In {\em Proceedings of the 6th ACM conference on Electronic
  commerce}.

\bibitem[Pu and Chen, 2006]{putrust2006}
Pu, P. and Chen, L. (2006).
\newblock Trust building with explanation interfaces.
\newblock In {\em Proceedings of the 11th International Conference on
  Intelligent User Interfaces}, IUI '06, page 93–100, New York, NY, USA.
  Association for Computing Machinery.

\bibitem[Pu and Faltings, 2000]{10.1145/332040.332446}
Pu, P. and Faltings, B. (2000).
\newblock Enriching buyers' experiences: The smartclient approach.
\newblock In {\em Proceedings of the SIGCHI Conference on Human Factors in
  Computing Systems}, CHI '00, pages 289--296, New York, NY, USA. Association
  for Computing Machinery.

\bibitem[Pu et~al., 2006]{pu2006increasing}
Pu, P., Viappiani, P., and Faltings, B. (2006).
\newblock Increasing user decision accuracy using suggestions.
\newblock In {\em Proceedings of the SIGCHI conference on Human Factors in
  computing systems}, pages 121--130.

\bibitem[Quint et~al., 2018]{quint2018interpretable}
Quint, E., Wirka, G., Williams, J., Scott, S., and Vinodchandran, N. (2018).
\newblock Interpretable classification via supervised variational autoencoders
  and differentiable decision trees.

\bibitem[Radford et~al., 2019]{radford2019language}
Radford, A., Wu, J., Child, R., Luan, D., Amodei, D., and Sutskever, I. (2019).
\newblock Language models are unsupervised multitask learners.
\newblock {\em OpenAI blog}.

\bibitem[Reddi et~al., 2018]{47409}
Reddi, S., Kale, S., and Kumar, S. (2018).
\newblock On the convergence of adam and beyond.
\newblock In {\em International Conference on Learning Representations (ICLR)}.

\bibitem[Reilly et~al., 2005]{reilly2005explaining}
Reilly, J., McCarthy, K., McGinty, L., and Smyth, B. (2005).
\newblock Explaining compound critiques.
\newblock {\em Artificial Intelligence Review}.

\bibitem[Reilly et~al., 2007]{10.1145/1250910.1250929}
Reilly, J., Zhang, J., McGinty, L., Pu, P., and Smyth, B. (2007).
\newblock Evaluating compound critiquing recommenders: A real-user study.
\newblock In {\em Proceedings of the 8th ACM Conference on Electronic
  Commerce}, EC '07, pages 114--123, New York, NY, USA. Association for
  Computing Machinery.

\bibitem[Rendle et~al., 2009]{10.5555/1795114.1795167}
Rendle, S., Freudenthaler, C., Gantner, Z., and Schmidt-Thieme, L. (2009).
\newblock Bpr: Bayesian personalized ranking from implicit feedback.
\newblock In {\em Proceedings of the Twenty-Fifth Conference on Uncertainty in
  Artificial Intelligence}, UAI '09, pages 452--461, Arlington, Virginia, USA.
  AUAI Press.

\bibitem[Rezende et~al., 2014]{10.5555/3044805.3045035}
Rezende, D.~J., Mohamed, S., and Wierstra, D. (2014).
\newblock Stochastic backpropagation and approximate inference in deep
  generative models.
\newblock In {\em Proceedings of the 31st International Conference on
  International Conference on Machine Learning - Volume 32}, ICML'14, pages
  II--1278--II--1286. JMLR.org.

\bibitem[Ribeiro et~al., 2016]{ribeiro2016should}
Ribeiro, M.~T., Singh, S., and Guestrin, C. (2016).
\newblock " why should i trust you?" explaining the predictions of any
  classifier.
\newblock In {\em Proceedings of the 22nd ACM SIGKDD international conference
  on knowledge discovery and data mining}, KDD '16, pages 1135--1144, New York,
  NY, USA. Association for Computing Machinery.

\bibitem[Ricci et~al., 2011]{ricci2011introduction}
Ricci, F., Rokach, L., and Shapira, B. (2011).
\newblock Introduction to recommender systems handbook.
\newblock In {\em Recommender systems handbook}, pages 1--35. Springer.

\bibitem[Rohrbach et~al., 2018]{rohrbach-etal-2018-object}
Rohrbach, A., Hendricks, L.~A., Burns, K., Darrell, T., and Saenko, K. (2018).
\newblock Object hallucination in image captioning.
\newblock In {\em Proceedings of the 2018 Conference on Empirical Methods in
  Natural Language Processing}, pages 4035--4045, Brussels, Belgium.
  Association for Computational Linguistics.

\bibitem[Rosasco et~al., 2004]{Rosasco_areloss}
Rosasco, L., Vito, E.~D., Caponnetto, A., Piana, M., and Verri, A. (2004).
\newblock Are loss functions all the same?
\newblock {\em Neural Computation}, 16(5):1063--1076.

\bibitem[Sachdeva and McAuley, 2020]{10.1145/3397271.3401281}
Sachdeva, N. and McAuley, J. (2020).
\newblock How useful are reviews for recommendation? a critical review and
  potential improvements.
\newblock In {\em Proceedings of the 43rd International ACM SIGIR Conference on
  Research and Development in Information Retrieval}, SIGIR '20, pages
  1845--1848, New York, NY, USA. Association for Computing Machinery.

\bibitem[Sedhain et~al., 2016]{sedhain2016practical}
Sedhain, S., Bui, H., Kawale, J., Vlassis, N., Kveton, B., Menon, A.~K., Bui,
  T., and Sanner, S. (2016).
\newblock Practical linear models for large-scale one-class collaborative
  filtering.
\newblock In {\em Proceedings of the Twenty-Fifth International Joint
  Conference on Artificial Intelligence}, pages 3854--3860.

\bibitem[Sedhain et~al., 2015]{10.1145/2740908.2742726}
Sedhain, S., Menon, A.~K., Sanner, S., and Xie, L. (2015).
\newblock Autorec: Autoencoders meet collaborative filtering.
\newblock In {\em Proceedings of the 24th International Conference on World
  Wide Web}, WWW '15 Companion, pages 111--112, New York, NY, USA. Association
  for Computing Machinery.

\bibitem[Serrano and Smith, 2019]{serrano-smith-2019-attention}
Serrano, S. and Smith, N.~A. (2019).
\newblock Is attention interpretable?
\newblock In {\em Proceedings of the 57th Annual Meeting of the Association for
  Computational Linguistics}, pages 2931--2951, Florence, Italy.

\bibitem[Shi et~al., 2019]{NEURIPS2019_0ae775a8}
Shi, Y., N, S., Paige, B., and Torr, P. (2019).
\newblock Variational mixture-of-experts autoencoders for multi-modal deep
  generative models.
\newblock In Wallach, H., Larochelle, H., Beygelzimer, A., d\textquotesingle
  Alch\'{e}-Buc, F., Fox, E., and Garnett, R., editors, {\em Advances in Neural
  Information Processing Systems}, volume~32. Curran Associates, Inc.

\bibitem[Shneiderman et~al., 2016]{shneiderman2016}
Shneiderman, B., Plaisant, C., Cohen, M., Jacobs, S., Elmqvist, N., and
  Diakopoulos, N. (2016).
\newblock Grand challenges for hci researchers.
\newblock {\em Interactions}, 23(5):24–25.

\bibitem[Shrikumar et~al., 2017]{pmlr-v70-shrikumar17a}
Shrikumar, A., Greenside, P., and Kundaje, A. (2017).
\newblock Learning important features through propagating activation
  differences.
\newblock In Precup, D. and Teh, Y.~W., editors, {\em Proceedings of the 34th
  International Conference on Machine Learning}, volume~70 of {\em Proceedings
  of Machine Learning Research}, pages 3145--3153, International Convention
  Centre, Sydney, Australia. PMLR.

\bibitem[Sinha and Swearingen, 2002]{sinha2002role}
Sinha, R. and Swearingen, K. (2002).
\newblock The role of transparency in recommender systems.
\newblock In {\em CHI'02 extended abstracts on Human factors in computing
  systems}, CHI EA '02, pages 830--831, New York, NY, USA. Association for
  Computing Machinery.

\bibitem[Smith-Renner et~al., 2020]{noexplainability2020}
Smith-Renner, A., Fan, R., Birchfield, M., Wu, T., Boyd-Graber, J., Weld,
  D.~S., and Findlater, L. (2020).
\newblock No explainability without accountability: An empirical study of
  explanations and feedback in interactive ml.
\newblock In {\em Proceedings of the 2020 CHI Conference on Human Factors in
  Computing Systems}, CHI '20, page 1–13, New York, NY, USA. Association for
  Computing Machinery.

\bibitem[Srivastava et~al., 2014]{srivastava2014dropout}
Srivastava, N., Hinton, G., Krizhevsky, A., Sutskever, I., and Salakhutdinov,
  R. (2014).
\newblock Dropout: a simple way to prevent neural networks from overfitting.
\newblock {\em The journal of machine learning research}, 15(1):1929--1958.

\bibitem[Sun et~al., 2020]{dual2020}
Sun, P., Wu, L., Zhang, K., Fu, Y., Hong, R., and Wang, M. (2020).
\newblock Dual learning for explainable recommendation: Towards unifying user
  preference prediction and review generation.
\newblock In {\em Proceedings of The Web Conference 2020}, WWW '20, page
  837–847, New York, NY, USA. Association for Computing Machinery.

\bibitem[Sundararajan et~al., 2017]{pmlr-v70-sundararajan17a}
Sundararajan, M., Taly, A., and Yan, Q. (2017).
\newblock Axiomatic attribution for deep networks.
\newblock In Precup, D. and Teh, Y.~W., editors, {\em Proceedings of the 34th
  International Conference on Machine Learning}, volume~70 of {\em Proceedings
  of Machine Learning Research}, pages 3319--3328, International Convention
  Centre, Sydney, Australia. PMLR.

\bibitem[Sutter et~al., 2020]{NEURIPS2020_43bb733c}
Sutter, T., Daunhawer, I., and Vogt, J. (2020).
\newblock Multimodal generative learning utilizing jensen-shannon-divergence.
\newblock In Larochelle, H., Ranzato, M., Hadsell, R., Balcan, M.~F., and Lin,
  H., editors, {\em Advances in Neural Information Processing Systems},
  volume~33, pages 6100--6110. Curran Associates, Inc.

\bibitem[Szegedy et~al., 2016]{szegedy2016rethinking}
Szegedy, C., Vanhoucke, V., Ioffe, S., Shlens, J., and Wojna, Z. (2016).
\newblock Rethinking the inception architecture for computer vision.
\newblock In {\em the IEEE conference on computer vision and pattern
  recognition}.

\bibitem[Tenenbaum, 1999]{tenenbaum1999bayesian}
Tenenbaum, J.~B. (1999).
\newblock {\em A Bayesian framework for concept learning}.
\newblock PhD thesis, Massachusetts Institute of Technology.

\bibitem[Tintarev and Masthoff, 2015]{ExplainingRecommendation}
Tintarev, N. and Masthoff, J. (2015).
\newblock {\em Explaining Recommendations: Design and Evaluation}, pages
  353--382.
\newblock Springer.

\bibitem[Torrens et~al., 2002]{Torrens2002abcd}
Torrens, M., Faltings, B., and Pu, P. (2002).
\newblock Smart clients: Constraint satisfaction as a paradigm for scaleable
  intelligent information systems.
\newblock {\em Constraints. Kluwer Academic Publishers, Hingham, MA, USA.},
  7(1):49--69.

\bibitem[Tsai et~al., 2019]{tsai2018learning}
Tsai, Y.-H.~H., Liang, P.~P., Zadeh, A., Morency, L.-P., and Salakhutdinov, R.
  (2019).
\newblock Learning factorized multimodal representations.
\newblock In {\em International Conference on Learning Representations (ICLR)}.

\bibitem[van~der Lee et~al., 2019]{van-der-lee-etal-2019-best}
van~der Lee, C., Gatt, A., van Miltenburg, E., Wubben, S., and Krahmer, E.
  (2019).
\newblock Best practices for the human evaluation of automatically generated
  text.
\newblock In {\em International Conference on Learning Representations}.

\bibitem[Vaswani et~al., 2017]{vaswani2017attention}
Vaswani, A., Shazeer, N., Parmar, N., Uszkoreit, J., Jones, L., Gomez, A.~N.,
  Kaiser, L.~u., and Polosukhin, I. (2017).
\newblock Attention is all you need.
\newblock In Guyon, I., Luxburg, U.~V., Bengio, S., Wallach, H., Fergus, R.,
  Vishwanathan, S., and Garnett, R., editors, {\em Advances in Neural
  Information Processing Systems}, volume~30. Curran Associates, Inc.

\bibitem[Viappiani et~al., 2006]{Viappiani06preference-basedsearch}
Viappiani, P., Faltings, B., and Pu, P. (2006).
\newblock Preference-based search using example-critiquing with suggestions.
\newblock {\em Journal of Artificial Intelligence Research (JAIR}, pages
  465--503.

\bibitem[Viappiani et~al., 2008]{viappiani2008preference}
Viappiani, P., Pu, P., and Faltings, B. (2008).
\newblock Preference-based search with adaptive recommendations.
\newblock {\em Ai Communications}, 21(2-3):155--175.

\bibitem[Voita et~al., 2019]{voita-etal-2019-analyzing}
Voita, E., Talbot, D., Moiseev, F., Sennrich, R., and Titov, I. (2019).
\newblock Analyzing multi-head self-attention: Specialized heads do the heavy
  lifting, the rest can be pruned.
\newblock In {\em Proceedings of the 57th Annual Meeting of the Association for
  Computational Linguistics}, pages 5797--5808, Florence, Italy. Association
  for Computational Linguistics.

\bibitem[Vyas et~al., 2021]{vyas2021hidden}
Vyas, D.~A., Eisenstein, L.~G., and Jones, D.~S. (2021).
\newblock Hidden in plain sight—reconsidering the use of race correction in
  clinical algorithms.
\newblock {\em Obstetrical \& Gynecological Survey}, 76(1):5--7.

\bibitem[Wall et~al., 2019]{wall2019using}
Wall, E., Ghorashi, S., and Ramos, G. (2019).
\newblock Using expert patterns in assisted interactive machine learning: A
  study in machine teaching.
\newblock In {\em Interact 2019}.

\bibitem[Wang et~al., 2019]{abs-1905-12926}
Wang, K., Hua, H., and Wan, X. (2019).
\newblock Controllable unsupervised text attribute transfer via editing
  entangled latent representation.
\newblock In Wallach, H., Larochelle, H., Beygelzimer, A., d\textquotesingle
  Alch\'{e}-Buc, F., Fox, E., and Garnett, R., editors, {\em Advances in Neural
  Information Processing Systems}, volume~32. Curran Associates, Inc.

\bibitem[Wang and Manning, 2012]{wang-manning-2012-baselines}
Wang, S. and Manning, C. (2012).
\newblock Baselines and bigrams: Simple, good sentiment and topic
  classification.
\newblock In {\em Proceedings of the 50th Annual Meeting of the Association for
  Computational Linguistics}, pages 90--94, Jeju Island, Korea.

\bibitem[Wiegreffe and Marasovic, 2021]{wiegreffe2021teach}
Wiegreffe, S. and Marasovic, A. (2021).
\newblock Teach me to explain: A review of datasets for explainable natural
  language processing.
\newblock In {\em Thirty-fifth Conference on Neural Information Processing
  Systems Datasets and Benchmarks Track (Round 1)}.

\bibitem[Wilkinson et~al., 2021]{wilkinson2021}
Wilkinson, D., Alkan, O., Liao, Q.~V., Mattetti, M., Vejsbjerg, I.,
  Knijnenburg, B.~P., and Daly, E. (2021).
\newblock Why or why not? the effect of justification styles on chatbot
  recommendations.
\newblock volume~39, New York, NY, USA. Association for Computing Machinery.

\bibitem[Williams and Tou, 1982]{williams1982rabbit}
Williams, M.~D. and Tou, F.~N. (1982).
\newblock Rabbit: an interface for database access.
\newblock In {\em Proceedings of the ACM Conference}, pages 83--87.

\bibitem[Williams, 1992]{williams1992simple}
Williams, R.~J. (1992).
\newblock Simple statistical gradient-following algorithms for connectionist
  reinforcement learning.
\newblock {\em Machine learning}, 8(3-4):229--256.

\bibitem[Williams and Zipser, 1989]{teacherforcing}
Williams, R.~J. and Zipser, D. (1989).
\newblock A learning algorithm for continually running fully recurrent neural
  networks.
\newblock {\em Neural Comput.}, 1(2):270–280.

\bibitem[Wu et~al., 2019a]{keyphraseExtractionDeep}
Wu, G., Luo, K., Sanner, S., and Soh, H. (2019a).
\newblock Deep language-based critiquing for recommender systems.
\newblock In {\em Proceedings of the 13th ACM Conference on Recommender
  Systems}, RecSys '19, page 137–145, New York, NY, USA. Association for
  Computing Machinery.

\bibitem[Wu et~al., 2019b]{10.1145/3331184.3331201}
Wu, G., Volkovs, M., Soon, C.~L., Sanner, S., and Rai, H. (2019b).
\newblock Noise contrastive estimation for one-class collaborative filtering.
\newblock In {\em Proceedings of the 42nd International ACM SIGIR Conference on
  Research and Development in Information Retrieval}, SIGIR'19, pages 135--144,
  New York, NY, USA. Association for Computing Machinery.

\bibitem[Wu and Goodman, 2018]{NEURIPS2018_1102a326}
Wu, M. and Goodman, N. (2018).
\newblock Multimodal generative models for scalable weakly-supervised learning.
\newblock In Bengio, S., Wallach, H., Larochelle, H., Grauman, K.,
  Cesa-Bianchi, N., and Garnett, R., editors, {\em Advances in Neural
  Information Processing Systems}, volume~31. Curran Associates, Inc.

\bibitem[Wu et~al., 2016]{10.1145/2835776.2835837}
Wu, Y., DuBois, C., Zheng, A.~X., and Ester, M. (2016).
\newblock Collaborative denoising auto-encoders for top-n recommender systems.
\newblock In {\em Proceedings of the Ninth ACM International Conference on Web
  Search and Data Mining}, WSDM '16, pages 153--162, New York, NY, USA.
  Association for Computing Machinery.

\bibitem[Xu et~al., 2015a]{xu2015empirical}
Xu, B., Wang, N., Chen, T., and Li, M. (2015a).
\newblock Empirical evaluation of rectified activations in convolutional
  network.
\newblock {\em arXiv preprint arXiv:1505.00853}.

\bibitem[Xu et~al., 2015b]{pmlr-v37-xuc15}
Xu, K., Ba, J., Kiros, R., Cho, K., Courville, A., Salakhudinov, R., Zemel, R.,
  and Bengio, Y. (2015b).
\newblock Show, attend and tell: Neural image caption generation with visual
  attention.
\newblock In Bach, F. and Blei, D., editors, {\em Proceedings of the 32nd
  International Conference on Machine Learning}, volume~37 of {\em Proceedings
  of Machine Learning Research}, pages 2048--2057, Lille, France. PMLR.

\bibitem[Yala et~al., 2019]{yala2019deep}
Yala, A., Lehman, C., Schuster, T., Portnoi, T., and Barzilay, R. (2019).
\newblock A deep learning mammography-based model for improved breast cancer
  risk prediction.
\newblock {\em Radiology}, 292(1):60--66.

\bibitem[Yang et~al., 2021]{10.1145/3404835.3463108}
Yang, H., Shen, T., and Sanner, S. (2021).
\newblock {\em Bayesian Critiquing with Keyphrase Activation Vectors for
  VAE-Based Recommender Systems}, page 2111–2115.
\newblock Association for Computing Machinery, New York, NY, USA.

\bibitem[Yang et~al., 2016]{yang2016hierarchical}
Yang, Z., Yang, D., Dyer, C., He, X., Smola, A., and Hovy, E. (2016).
\newblock Hierarchical attention networks for document classification.
\newblock In {\em Proceedings of the 2016 Conference of the North {A}merican
  Chapter of the Association for Computational Linguistics: Human Language
  Technologies}, pages 1480--1489, San Diego, California. Association for
  Computational Linguistics.

\bibitem[Yao et~al., 2019]{yao2019plan}
Yao, L., Peng, N., Weischedel, R., Knight, K., Zhao, D., and Yan, R. (2019).
\newblock Plan-and-write: Towards better automatic storytelling.
\newblock In {\em Proceedings of the AAAI Conference on Artificial
  Intelligence}, pages 7378--7385.

\bibitem[Yelp, 2016]{yelpdataset}
Yelp (2016).
\newblock Yelp open dataset 2016 - https://www.yelp.com/dataset.

\bibitem[Yeomans et~al., 2019]{yeomans2019making}
Yeomans, M., Shah, A., Mullainathan, S., and Kleinberg, J. (2019).
\newblock Making sense of recommendations.
\newblock {\em Journal of Behavioral Decision Making}, 32(4):403--414.

\bibitem[Yin et~al., 2017]{yin2017document}
Yin, Y., Song, Y., and Zhang, M. (2017).
\newblock Document-level multi-aspect sentiment classification as machine
  comprehension.
\newblock In {\em Proceedings of the 2017 Conference on Empirical Methods in
  Natural Language Processing}, pages 2044--2054, Copenhagen, Denmark.
  Association for Computational Linguistics.

\bibitem[Yu et~al., 2019]{yu-etal-2019-rethinking}
Yu, M., Chang, S., Zhang, Y., and Jaakkola, T. (2019).
\newblock Rethinking cooperative rationalization: Introspective extraction and
  complement control.
\newblock In {\em Proceedings of the 2019 Conference on Empirical Methods in
  Natural Language Processing and the 9th International Joint Conference on
  Natural Language Processing (EMNLP-IJCNLP)}, pages 4094--4103, Hong Kong,
  China. Association for Computational Linguistics.

\bibitem[Yu et~al., 2021]{yu2021}
Yu, M., Zhang, Y., Chang, S., and Jaakkola, T. (2021).
\newblock Understanding interlocking dynamics of cooperative rationalization.
\newblock In Wallach, H., Larochelle, H., Beygelzimer, A., d\textquotesingle
  Alch\'{e}-Buc, F., Fox, E., and Garnett, R., editors, {\em Advances in Neural
  Information Processing Systems}, volume~34. Curran Associates, Inc.

\bibitem[Zaheer et~al., 2017]{zaheer2017latent}
Zaheer, M., Ahmed, A., and Smola, A.~J. (2017).
\newblock Latent lstm allocation: Joint clustering and non-linear dynamic
  modeling of sequence data.
\newblock In {\em International Conference on Machine Learning}, pages
  3967--3976.

\bibitem[Zhang and Curley, 2018]{zhang2018exploring}
Zhang, J. and Curley, S.~P. (2018).
\newblock Exploring explanation effects on consumers' trust in online
  recommender agents.
\newblock {\em International Journal of Human--Computer Interaction},
  34(5):421--432.

\bibitem[Zhang et~al., 2020]{bert-score}
Zhang, T., Kishore, V., Wu, F., Weinberger, K.~Q., and Artzi, Y. (2020).
\newblock Bertscore: Evaluating text generation with bert.
\newblock In {\em International Conference on Learning Representations}.

\bibitem[Zhang et~al., 2014]{zhang2014explicit}
Zhang, Y., Lai, G., Zhang, M., Zhang, Y., Liu, Y., and Ma, S. (2014).
\newblock Explicit factor models for explainable recommendation based on
  phrase-level sentiment analysis.
\newblock In {\em Proceedings of the 37th International ACM SIGIR Conference on
  Research \& Development in Information Retrieval}, SIGIR '14, page 83–92,
  New York, NY, USA. Association for Computing Machinery.

\bibitem[Zhang et~al., 2016]{zhang-etal-2016-rationale}
Zhang, Y., Marshall, I., and Wallace, B.~C. (2016).
\newblock Rationale-augmented convolutional neural networks for text
  classification.
\newblock In {\em Proceedings of the 2016 Conference on Empirical Methods in
  Natural Language Processing}, pages 795--804, Austin, Texas. Association for
  Computational Linguistics.

\bibitem[Zhao et~al., 2018]{zhao2018categorical}
Zhao, Q., Chen, J., Chen, M., Jain, S., Beutel, A., Belletti, F., and Chi,
  E.~H. (2018).
\newblock Categorical-attributes-based item classification for recommender
  systems.
\newblock In {\em Proceedings of the 12th ACM Conference on Recommender
  Systems}, RecSys '18, page 320–328, New York, NY, USA. Association for
  Computing Machinery.

\bibitem[Zhou et~al., 2018]{10.1007/978-3-030-01237-3_8}
Zhou, B., Sun, Y., Bau, D., and Torralba, A. (2018).
\newblock Interpretable basis decomposition for visual explanation.
\newblock In Ferrari, V., Hebert, M., Sminchisescu, C., and Weiss, Y., editors,
  {\em Computer Vision -- ECCV 2018}, pages 122--138, Cham. Springer
  International Publishing.

\end{thebibliography}
